\documentclass[10pt,twocolumn,letterpaper]{article}

\usepackage[pagenumbers]{cvpr} % To force page numbers, e.g. for an arXiv version
\usepackage{amsmath,amsfonts,bm}

\def\eqref#1{equation~\ref{#1}}
\def\1{\bm{1}}

\DeclareMathAlphabet{\mathsfit}{\encodingdefault}{\sfdefault}{m}{sl}
\SetMathAlphabet{\mathsfit}{bold}{\encodingdefault}{\sfdefault}{bx}{n}

\DeclareMathOperator*{\argmin}{arg\,min}

\DeclareMathOperator{\sign}{sign}

\usepackage{amssymb,amsmath,amsthm}
\DeclareMathOperator{\diag}{diag}
\newtheorem{theorem}{Theorem}
\newtheorem{lemma}{Lemma}
\newtheorem{corollary}{Corollary}

\newtheorem{assumption}{Assumption}

\usepackage[dvipsnames]{xcolor}

\usepackage{color}

\definecolor{cvprblue}{rgb}{0.21,0.49,0.74}
\usepackage[pagebackref,breaklinks,colorlinks,citecolor=cvprblue]{hyperref}

\def\paperID{15252} % *** Enter the Paper ID here
\def\confName{CVPR}
\def\confYear{2024}

\title{Towards Robust Learning to Optimize with Theoretical Guarantees}

\author{Qingyu Song\\
CUHK
\and
Wei Lin\\
CUHK
\and
Juncheng Wang\\
HKBU
\and
Hong Xu\\
CUHK
}
\usepackage{tocloft}

\begin{document}
\maketitle
\begin{abstract}
    Learning to optimize (L2O) is an emerging technique to solve mathematical optimization problems with learning-based methods. Although with great success in many real-world scenarios such as wireless communications, computer networks, and electronic design, existing L2O works lack theoretical demonstration of their performance and robustness in out-of-distribution (OOD) scenarios. We address this gap by providing comprehensive proofs. First, we prove a sufficient condition for a robust L2O model with homogeneous convergence rates over all In-Distribution (InD) instances. We assume an L2O model achieves robustness for an InD scenario. Based on our proposed methodology of aligning OOD problems to InD problems, we also demonstrate that the L2O model's convergence rate in OOD scenarios will deteriorate by an equation of the L2O model's input features. Moreover, we propose an L2O model with a concise gradient-only feature construction and a novel gradient-based history modeling method. Numerical simulation demonstrates that our proposed model outperforms the state-of-the-art baseline in both InD and OOD scenarios and achieves up to 10 $\times$ convergence speedup. The code of our method can be found from \url{https://github.com/NetX-lab/GoMathL2O-Official}.
\end{abstract}
%!TEX root = main.tex
% \vspace{-6mm}
\section{Introduction}
\label{sec:intro}

% \begin{enumerate}
%   \item What is L2O? A brief introduction of L2O and how it works: META training.
%   \item What is Math-Inspired L2O? State why is math-inspired L2O SOTA? Provide an explicit manipulation of L2O workflow and achieve SOTA results.
%   \item What is the generalization problem in L2O? Optimizer generalization, optimizee generalization. Two cases, ref.
%   \item How other works deal with generalization problem in L2O? One ref: generalization in black-box case, add a further regularization term. 
%   \item White-box case: Can we analyze the generalization ability under a certain assumption? 
%   \begin{enumerate}
%     \item Why is removing parameters essential? Theoretically prove OOD problem is unavoidable w.r.t. each feature.
%     \item How much gain if a parameter is removed? Compare the bounds between with and without a parameter.
%   \end{enumerate}
%   \item Workflow:
%   \begin{enumerate}
%     \item Inference only, asssume training is stable and different training configurations will not influence L2O parameters.
%   \end{enumerate}
%   \item How to ensure the optimality still holds after removing $x$, in-distribution?
%   \begin{enumerate}
%     \item Theoretically compare the convergence rate between with and without $x$.
%     \item Empirically show that the learned coefficient w.r.t. $x$ is close to zero.
%   \end{enumerate}
% \end{enumerate}

% \paragraph{Introduction of L2O}
Learning to Optimize (L2O) is a promising new approach in applying learning-based methods to tackle optimization problems. In particular, L2O concentrates on problems with well-defined objective functions and constraints \cite{Chen2022learning}. Thus, black-box optimization strategies, such as Bayesian Optimization \cite{Snoek2012practical}, typically fall outside its scope. L2O has shown benefits in problems from various domains, including LASSO regression in sparse coding using multilayer perceptrons \cite{Gregor2010learning}, and utility maximization in resource allocation wherein neural networks (NN) serve to approximate the expensive matrix inversion \cite{Hu2020iterative}.

% \paragraph{Different kinds of L2O}
L2O can be categorized into three main types: black-box \cite{Sun2018, Shen2021, Zhao2021, Cao2021}, algorithm-unrolling \cite{Hu2020iterative, Schynol2022, Zhu2022learning}, and math-inspired \cite{Liu2023, Heaton2023safeguarded}. 
Black-box L2O approaches the optimization problem as a traditional pattern recognition task, approximating a mapping function from manually constructed features to the solutions \cite{Sun2018}. Algorithm-unrolling L2O leverages well-defined algorithms, such as gradient descent \cite{Ruder16}, to approximate the solutions of complex calculations. 
Besides, much research has gone into explainable and trustworthy L2O. For example, \citet{Heaton2023safeguarded} employ an existing algorithm to prevent the L2O model from entering irrecoverable areas. 
\citet{Liu2023} introduce a mathematics-driven L2O (Math-L2O) framework for convex optimization, offering a general workflow for formulating an L2O model. 
Despite empirical results, a theoretical analysis on the robustness of L2O models under out-of-distribution (OOD) conditions is still missing in \cite{Liu2023}.

OOD generalization for L2O has emerged as a vital issue, often considered more critical in L2O than in other deep learning applications \cite{Shen2108towards}. For L2O, OOD's challenge involves resolving previously unseen problems, potentially involving novel optimization problems with unique objectives \cite{Yang2023learning}. 
Guaranteeing convergence in OOD scenarios remains elusive. 
For instance, a model's output in an OOD scenario could potentially veer into unpredictable areas when the domain changes significantly to an InD scenario.

Numerous efforts have been made to enhance the robustness of L2O models in training. 
\citet{Lv2017learning} employ data augmentation to prevent L2O models from overfitting to specific tasks. 
\citet{Almeida2021generalizable} transform the L2O model into a hyperparameter tuner for existing optimization algorithms. 
\citet{Wichrowska2017learned} focus on minimizing parameters in NNs and assembling heterogeneous optimization tasks. 
\citet{Liu2023} try to regularize L2O models with inspirations from existing algorithms. 
However, these studies predominantly aim to mitigate the limitations inherent in existing L2O methods, with no comprehensive analysis conducted on the impact of OOD on the deterioration of convergence. 
This gap in the literature motivates us to quantify this deterioration with rigorous analysis.

% In this framework, the authors derive a white-box L2O model from the sufficient and necessary conditions for optimality. 
% However, they do not apply any initiative strategies to solving the out-of-distribution (OOD) problems. Although the proposed method outperforms other learning and non-learning competitors in several OOD scenarios, they fail to explain how mathematical inspiration improve the generalization ability when OOD happens. 

% First, we hierarchically define the OOD problem in 

%  we demonstrate that a definitive formula exists to quantify the deterioration caused by OOD. 

The central thesis of this paper is to propose a general and robust L2O model for both InD and OOD scenarios. 
Chiefly, we first investigate L2O's convergence behavior in InD contexts and derive the criteria for a uniformly robust model applicable to all InD instances. 
Then, we characterize L2O's degradation of convergence under OOD conditions, presenting our findings as a series of corollaries. The main contributions of this paper are as follows.
% We utilize the Math-L2O approach from \citet{Liu2023} to define an L2O model. When utilizing an L2O model to solve an optimization problem iteratively, 
% \item We propose 
% The design of our L2O model is based on insights from both InD and OOD situations.
\begin{enumerate}
  \item We propose a methodology to link the L2O model's performances in InD and OOD situations based on the Math-L2O approach from \citet{Liu2023}. First, we construct a virtual feature by subtracting the L2O model's input feature in InD from that in OOD. 
  We then compute the corresponding difference in the model's outputs by applying this virtual feature. To depict a comprehensive deviation of OOD from InD, we align the variable sequence from an OOD situation with that from InD and construct a trajectory of virtual features. We use this trajectory to illustrate OOD's divergence from InD and then conduct theoretical analyses.
  \item We establish the criteria for a robust L2O model in an InD setting and examine its response to OOD. 
  First, we present a sufficient condition to guarantee a homogeneous convergence improvement in each iteration, confirming robustness in InD scenarios. Then, we derive the equations describing convergence gain in a single iteration and the overall convergence rate of the entire sequence relative to our proposed virtual feature. 
  A collection of theorems and observations underscore that the magnitude of virtual features inherently exacerbates the deterioration of convergence in OOD situations.
  \item Based on our theoretical insights, we propose a robust L2O model, GO-Math-L2O, that exclusively employs gradients as input features. This gradient-only approach enables a more concise virtual feature in OOD settings. We introduce a new gradient-only history modeling technique to model the optimization process's historical sequence. This method employs gradient (and subgradient) values as status indicators to modulate updates provided by the L2O model. We propose to recover the historical subgradient from an inversible model definition, thus eliminating the ambiguity of subgradient selection.
  \item Through numerical experiments, we show that GO-Math-L2O outperforms state-of-the-art (SOTA) L2O models on convergence and optimality across both InD and OOD scenarios. Following training with a synthetic dataset, we deploy various OOD test cases with identical optimal values. Our proposed model's convergence speed is up to 10$\times$ faster than SOTA L2O models in OOD scenarios.
  % Empirical results from both scenarios substantiate that our proposed model achieves superior convergence and stability compared to all existing SOTA models in InD and OOD contexts.
\end{enumerate}

% We comprehensively define L2O's OOD problem and propose a virtual feature methodology to analyze the solution sequence in OOD scenarios. First, We propose a hierarchical methodology to construct the definition. We define the L2O problem by optimization problem: In each iteration, the L2O model takes features from variable and objective functions to generate updates for the solution. Then, we utilize the OOD on the inner problem to define the L2O OOD problem: Using an L2O model trained with InD problems to solve OOD problems. 
% Moreover, we align each OOD variable to an InD variable and utilize their difference to construct a relationship between the variable updates in OOD and InD scenarios. 

The rest of this paper is organized as follows. In \cref{sec:definition}, we define OOD problems for L2O. In \cref{sec:vv_traj}, we propose a method to quantify the solutions given by an L2O model in OOD scenarios. Then, in \cref{sec:design_theorem}, we derive the convergence rate of an L2O model in OOD scenarios. Based on this, we propose our robust GO-Math-L2O model in \cref{sec:design_model}. 
We empirically verify the proposed model with simulations in \cref{sec:eval}, and conclude the work in \cref{sec:conclusion}.

\textit{Notations:} 
A smooth convex function and a non-smooth convex function are denoted by $f$ and $r$, respectively. NNs' input vectors are denoted by $z$ and $z^\prime$. Variables of an optimization problem are denoted by $x$ and $x^\prime$. The optimal solution is denoted by $x^*$. An iteration and stopping iteration are denoted by $k$ and $K$, respectively. A smooth gradient at $x_k$ and a set of subgradients at $x_k$ are denoted by $\nabla f(x_k)$ and $\partial r(x_k)$, respectively. A subgradient value of $\partial r(x_k)$ is denoted by $g_k$. Frobenius norm for a matrix and $L2$-norm for a vector is denoted by $\|\cdot\|_\mathrm{F}$ and $\|\cdot\|$ respectively. Transpose is defined by $^\top$. The maximum length of history modeling is denoted by $T$. The Jacobian matrix of a vector-to-vector function is denoted by $\mathbf{J}$. An L2O model is denoted by $d$. A NN is denoted by operator $\mathbf{N}$.
%!TEX root = main.tex
\section{Definitions}
\label{sec:definition}

In this section, we first introduce the objective of the L2O problem. We then introduce the Math-L2O model in \cite{Liu2023}, whose iterative updates are defined by NNs. 
Last, we define the domains for both InD and OOD scenarios, which leads to the definitions of InD L2O and OOD L2O problems.

% The abbreviations such as OOD InD, l2O, NN are already given in the Intro, you do not need to define them again in the rest of the paper. Also, I feel keep using Definition in the subsection tiles are a bit redundant.

% We delineate the Out-of-Distritbution (OOD) problem in the L2O framework step-by-step. First, we define the space of the L2O problem based on the space definition of objective optimization problems. We select the Math-L2O model \cite{Liu2023} whose update formulations are derived from rigorous mathematical derivations and define the neural networks (NN) as the functions of specific spaces. 

% Subsequently, we define the domains of optimization problems for in-distribution (InD) and OOD scenarios. This process leads us to distinguish between the \textit{InD L2O Problem}, which involves solving problems within the InD domain, and the \textit{OOD L2O Problem}, which is concerned with solving problems within the OOD domain. We assess the generalization capability of L2O through the convergence rate when solving problems across the OOD domain.

\subsection{Optimizee (Optimization Objective)} \label{sec:def_optimizee}
Consider function $F(x) = f(x) + r(x)$. Here, $f(x)$ is a $L$-smooth function, and $r(x)$ is a non-smooth function. They are defined within the following function spaces:
\begin{equation*} \label{eq:function_space}
    \begin{aligned}
        \mathcal{F}_L\left(\mathbb{R}^n\right)= & \{f: \mathbb{R}^n \rightarrow \mathbb{R} | f \text { is convex, differentiable, and } \\
        & \|\nabla f(x)-\nabla f(y)\| \leq L\|x-y\|, \forall x, y \in \mathbb{R}^n\}, \\
        \mathcal{F}\left(\mathbb{R}^n\right)= & \left\{r: \mathbb{R}^n \rightarrow \mathbb{R} | r \text { is proper, closed, and convex}\right\}.
    \end{aligned}
\end{equation*}
We assume $r(x)$ is sub-differentiable, with its subgradient set at any point $x$ defined below:
\begin{equation*}
    \partial r(x)=\{g \in \mathbb{R}^n \mid r(y)-r(x) \geq g^{\top}(y-x), \forall x,y \in \mathbb{R}^n \}.
\end{equation*}
We note here that the above optimization objective applies to both the InD and the OOD scenarios.

\subsection{Optimizor (L2O Model)} \label{sec:def_l2o}
Denote the L2O model as $d(z)$, where the input vector space is designated as $\mathcal{Z}$ such that $z \in \mathcal{Z} \subseteq \mathbb{R}^m$. 
We define $d(z)$ as a function mapping within the given function space \cite{Liu2023}:
\begin{equation} \label{eq:d_bound}
    \begin{aligned} 
        \mathcal{D}_C(\mathcal{Z})=\{d: \mathcal{Z} & \rightarrow  \mathbb{R}^n \mid d \text { is differentiable, } \\ 
        & \|\mathbf{J}_{d(z)}\|_\mathrm{F} \leq C, \forall z \in \mathcal{Z}, C \in \mathbb{R}^+ \}.
    \end{aligned}
\end{equation} 
We choose features from $x$ and $F(x)$ to define $z$, offering a wide range of feasible options. For instance, $z$ could be defined with the optimization variable and its gradient as $\begin{bmatrix} {x}^\top, \nabla f(x)^\top \end{bmatrix}^\top$ in \cite{Liu2023}. Different from \cite{Liu2023}, we propose to define $z$ solely as $\nabla f(x)$ to improve convergence in OOD scenarios. From our experimental results, our approach achieves near-optimal solutions in some OOD cases and more robust performance than SOTA baselines in all OOD scenarios. 
Moreover, Corollaries~\ref{lm:lemma2_smooth_converge_gain_ub} and~\ref{lm:lemma3_smooth_converge_rate_ub} theoretically demonstrate the outperformance over the method in \cite{Liu2023}.

$d(z)$ iteratively updates the optimization variable. At each iteration $k$, 
given the previous variable $x_{k-1} \in \mathbb{R}^n$ and the input vector $z_{k-1}$ for the L2O model,  $d(z_{k-1})$ updates $x_k$ as follows:
\begin{equation} \label{eq:l2o_ind}
    x_k=x_{k-1} - d (z_{k-1}).
\end{equation}

% Moreover, we inductively define the compositions in $d(z)$ that $d(z) := \diag(\mathbf{N}_{1})^\top\nabla f(x) + \mathbf{N}_{2}$, where 
% $\mathbf{N}_{1} \in \mathcal{D}_{C_1}(\mathcal{Z})$ and $\mathbf{N}_{2} \in \mathcal{D}_{C_2}(\mathcal{Z})$  are two NNs for some constants $C_1, C_1 \in \mathcal{R}^+$.

\subsection{InD and OOD Problems}
The InD and OOD problems share the same space of optimization objective defined in \cref{sec:def_optimizee} but with different optimization objectives or variable domains. Consider a convex and compact set, $\mathcal{S}_P \subseteq \mathbb{R}^n$. The complementary set of $\mathcal{S}_P$ is denoted as $\mathcal{S}_O$, such that $\mathcal{S}_O := \mathbb{R}^n \setminus \mathcal{S}_P$. We also suppose the existence of two function sets: $\mathcal{F}_{L,P} \subseteq \mathcal{F}_L\left(\mathbb{R}^n\right)$ and $\mathcal{F}_P \subseteq \mathcal{F}\left(\mathbb{R}^n\right)$. We define InD optimization problems as follows:
\begin{equation} \label{obj:ind}
    \min_{x} F(x), \tag{P}
\end{equation}
where $x \in \mathcal{S}_P$,  $F(x) = f(x) + r(x)$, $f \in \mathcal{F}_{L,P}$, and $r \in \mathcal{F}_P$. The dataset employed for training an L2O model is derived from a specific domain of $x$, $f$, and $r$. Consider an L2O model $d(z)$ that has undergone training with a domain of $x$, $f$, and $r$, sampled from Problem~\ref{obj:ind}. We then define the \textit{InD L2O Problem} as: Given any initial point $x_0 \in \mathcal{S}_P$, using $d(z)$ to iteratively update $x_0$ in order to find a solution for any arbitrary InD problem as depicted in Problem~\ref{obj:ind}. 

Note that instances outside this domain potentially yield more erroneous $d(z)$ outputs. Furthermore, non-learning algorithms, such as gradient descent, have demonstrated robustness across all domains \cite{Ruder16}. One of the main goals of this paper is to propose an L2O model that is robust to OOD.

We characterize OOD in the context of L2O in the optimization objective's domain. We define the \textit{OOD L2O Problem} as: 
Consider an L2O model $d(z)$ that has undergone training with a domain of $x$, $f$, and $r$, sampled from Problem~\ref{obj:ind}, using $d(z)$ to iteratively update $x_0^\prime \in \mathcal{S}_O$ in order to a solution for any following problem:
\begin{equation} \label{obj:ood}
    \min_{x^\prime} F^\prime(x^\prime). \tag{O}
\end{equation}
where $F^\prime(x^\prime) = f^\prime(x^\prime) + r^\prime(x^\prime)$, $f^\prime \notin \mathcal{F}_{L,P}$, and $r^\prime \notin \mathcal{F}_{P}$.

We delineate the InD and OOD input vector spaces of $d(z)$. We denote the input vector spaces for an L2O model in the context of \textit{InD L2O Problem} and \textit{OOD L2O Problem} as $\mathcal{Z}_P$ and $\mathcal{Z}_O$, respectively. Then, we choose features of the variables and the objective functions to construct the input feature of $d(z)$. Specifically, we define $\mathcal{Z}_P$ and $\mathcal{Z}_O$ as the ensuing sets:
\begin{equation*}
    \begin{aligned}
        \mathcal{Z}_P = &\big\{ \big[ x\text{-feature}^\top,  f(x)\text{-feature}^\top, r(x)\text{-feature}^\top, \dots \big]^\top \\
        & \qquad \mid \forall x \in \mathcal{S}_P, \forall f^\prime \in \mathcal{F}_{L,P},  \forall r^\prime \in \mathcal{F}_{P} \big \},\\
        \!  \mathcal{Z}_O = &\big\{ \big[ x^\prime\text{-feature}^\top, f^\prime(x^\prime)\text{-feature}^\top, r^\prime(x^\prime)\text{-feature}^\top, \dots \big]^\top \\
        & \qquad \mid \exists x^\prime \in \mathcal{S}_O \text{ or } \exists  f^\prime \notin \mathcal{F}_{L,P} \text{ or }  \exists r^\prime \notin \mathcal{F}_{P} \big\},
    \end{aligned}
\end{equation*}
where ``$\dots$'' represents other feasible features such as the history of $x$. Some feasible feature constructions for $x$, $f(x)$, and $r(x)$ include $x$ itself, $\nabla f(x)$, and $\partial r(x)$. Later in \cref{sec:design_model}, we show how to contruct the input features of the L2O model $d(z)$ based only on $\nabla{f}(x)$ and $\partial{r}(x)$.

%!TEX root = main.tex
\section{Virtual Feature and Trajectory}
\label{sec:vv_traj}

In this section, we introduce a virtual feature methodology to correlate any arbitrary variable yielded by the L2O model in the OOD scenario ($x_k^\prime$) to a corresponding variable $x_k$ in the InD scenario. The virtual features are generated as a linear combination of the OOD and InD features and serve as a bridge to connect each L2O model's OOD outcome to its InD outcome. We then leverage the virtual-feature method to connect OOD and InD variable trajectories generated by the L2O model. Since the convergence of InD trajectories is deterministic, such a method facilitates the convergence and robustness analysis for OOD scenarios in \cref{sec:design_theorem}.

% \qy{evaluate the robustness of the \textit{OOD L2O Problem} by correlating an arbitrary OOD scenario to a deterministic one. We select the InD scenario for this purpose. The solution for each OOD variable is constructed as the summation of two hypothetical variables' solutions: an InD hypothetical variable derived from the \textit{InD L2O Problem}, and an OOD hypothetical variable generated by subtracting the InD hypothetical variable from the original OOD variable. This method offers a resilient framework for scrutinizing the OOD trajectory, contingent on the \textit{InD L2O Problem} maintaining its invariance post-training.}

\subsection{Virtual Feature} \label{sec:virtual_feature}
Consider an arbitrary OOD variable $x^\prime \in \mathcal{S}_O$ and a InD variable $x \in \mathcal{S}_P$ yielded by the L2O model. Let $s \in \mathbb{R}^n$ such that $s = x^\prime - x$. In that case, we define the difference $s^\prime$ between the L2O model's features in the OOD scenario $z^\prime$ and the features in the InD scenario $z$. From the Mean Value Theorem \cite{Rudin1976}, there exists a virtual Jacobian matrix $\mathbf{J}_{d}, \|\mathbf{J}_{d}\| \leq C \sqrt{n}$ such that the following equality holds:
\begin{equation} \label{eq:traj_mean_value_th}
    d(z^\prime) = d(z) + \mathbf{J}_{d}(z^\prime - z) = d(z) + \mathbf{J}_{d}s^\prime.
\end{equation}

% at $\tilde{z}$, where $\tilde{z} = \alpha z^\prime + (1-\alpha)z$ with $\alpha \in [0,1]$ is a virtual feature 
The demonstrations are in \cref{sec:thpf_smooth_pre}. 
From \eqref{eq:traj_mean_value_th}, we can relate any variable of the L2O model in the OOD scenario to the InD scenario. Although the virtual Jacobian matrix $\mathbf{J}_{d}$ is non-deterministic, it is upper bounded from the definition of $d(z)$ in \eqref{eq:d_bound}. This suffices for a quantitative analysis of the impact of the "shift" $s^\prime$ on convergence. For instance, our proposed Theorem~\ref{th:theorem1_smooth_converge_gain} in \cref{sec:design_theorem} provides an upper bound on the convergence gain for a single iteration.

% Nonetheless, identifying an appropriate $x$ is of paramount importance. We need a suitable $x$ that we can trust, one where the optimization process executed by the L2O model is deterministic and straightforward to scrutinize. The absence of such alignment can exacerbate the complexity of analysis.

\subsection{Trajectory} \label{sec:trajcetory}

% We propose selecting the variables of the trajectory generated by the L2O model, originating from a stable InD initial point. Denote an InD initial point as $x_0 \in \mathcal{S}_P$. For an InD problem, defined by Problem~\ref{obj:ind}, we establish two trajectories, the objective variable $x$ and the corresponding input vector $z$ of the L2O model, as below:

For the OOD Problem~\ref{obj:ood}, denote the initial variable as $x_0^\prime \in \mathcal{S}_O$. In the optimization process, we have two trajectories for the variable $x^\prime$ and the features of the L2O model $z^\prime$:
\begin{equation*}
    \{x_0^\prime, x_1^\prime, x_2^\prime, \dots, x_K^\prime\}, \{z_0^\prime, z_1^\prime, z_2^\prime, \dots, z_K^\prime\}. 
\end{equation*}
where $x_k^\prime \in \mathcal{S}_O, z_k^\prime \in \mathcal{Z}_O, k = {0, 1, 2, \dots, K}$. Similarly, for the InD Problem~\ref{obj:ind}, denote the initial variable as $x_0 \in \mathcal{S}_P$. We have also have two trajectories for the variables $x$ and the features of the L2O model z:
\begin{equation*}
    \{x_0, x_1, x_2, \dots, x_K\}, \{z_0, z_1, z_2, \dots, z_K\}, 
\end{equation*}
where $x_k \in \mathcal{S}_P, z_k \in \mathcal{Z}_P, k = {0, 1, 2, \dots, K}$. Utilizing the definitions in \cref{sec:virtual_feature}, we compute the differences between the variables and the features of the OOD trajectories and the InD trajectories as follows:
\begin{equation*}
    \{s_0, s_1, s_2, \dots, s_K\}, \{s_0^\prime, s_1^\prime, s_2^\prime, \dots, s_K^\prime\}, 
\end{equation*}
where $s_k := x_k^\prime - x_k$ and $s_k^\prime := z_k^\prime - z_k$. 
Thus, we can represent the OOD trajectory by $\{x_k + s_k\}$ and $\{z_k + s_k^\prime\}$. Furthermore, utilizing the virtual-feature method in \cref{sec:virtual_feature}, we have:
\begin{equation} \label{eq:s_update}
    d(z_{k-1}^\prime) = d(z_{k-1}) + \mathbf{J}_{d, k-1} s_{k-1}^\prime,
\end{equation}
where $\mathbf{J}_{d, k-1}$ is a virtual Jacobian matrix of $d(\tilde{z}_{k-1})$. Due to \eqref{eq:l2o_ind} in \cref{sec:def_l2o}, $x_k^\prime$ is updated by $x_{k-1}^\prime - d(z_{k-1}^\prime)$ and $x_k$ is updated by $x_{k-1} - d(z_{k-1})$. Based on \eqref{eq:s_update}, we have:
\begin{equation} \label{eq:shifting_trajectory}
    s_k = s_{k-1} - \mathbf{J}_{d, k-1}s_{k-1}^\prime.
\end{equation}

% with $\tilde{z}_{k-1} = \alpha z_{k-1}^\prime + (1-\alpha)z_{k-1}$ for a $\alpha_{k-1} \in [0,1]$
%!TEX root = main.tex
\section{White-Box OOD Generalization Analysis}
\label{sec:design_theorem}
% We propose \qy{four} theorems in this section, which 
In this section, we rigorously demonstrate that the robustness of the L2O model is limited by its input features of NNs. We prove that increased features adversely impact the L2O model's generalization ability in OOD scenarios.

\subsection{The Smooth Case} \label{sec:smooth}
Building upon the state-of-the-art Math-L2O \cite{Liu2023}, we systematically detail our conclusions through a series of theorems and lemmas. 

We analyze the convergence rate of the OOD scenario when the objective function $F(x)$ is smooth, i.e., $r(x)=0$ and $F(x)=f(x)$. Leveraging Theorem 1 from \cite{Liu2023}, the update of the variable at the $k$-th iteration can be expressed as $x_k=x_{k-1}-\mathbf{P}_{k-1} \nabla f(x_{k-1})-b_{k-1}$, where $\mathbf{P}_{k-1} \in \mathbb{R}^{n \times n}$ and $b_{k-1} \in \mathbb{R}^n$ are parameters learned by NNs.

Let $\mathbf{P}_{k-1}$ and $b_{k-1}$ be $\mathbf{N}_1(\mathcal{Z}) \in \mathcal{D}_{C_1}(\mathcal{Z})$ and $\mathbf{N}_2(\mathcal{Z}) \in \mathcal{D}_{C_2}(\mathcal{Z})$ respectively, for some positive constants $C_1, C_2 \in \mathbb{R}^+$. As suggested in \cite{Liu2023}, we assign $\mathbf{P}_k$ as a diagonal matrix. Without loss of generality, for any given variable $x_{k-1}$, where $x_{k-1} \in \mathbb{R}^n$, and any given function $f \in \mathcal{F}_L(\mathbb{R}^n)$, we define $z_{k-1}=[x_{k-1}^\top, {\nabla f(x_{k-1})}^\top]^\top$ \cite{Liu2023}. The update of variable $x_k$ at each iteration $k$ can then be expressed as:
\begin{equation} \label{eq:l2o_smooth}
    \! x_k = x_{k-1} - \diag(\mathbf{N}_1(z_{k-1})) \nabla f(x_{k-1}) - \mathbf{N}_2(z_{k-1}).
\end{equation} 
The OOD shift applied to the variable and its gradient yields the definition of virtual feature (\cref{sec:vv_traj}):
\begin{equation} \label{eq:ood_vector}
    s^\prime_{k-1} := [s_{k-1}^\top, (\nabla f^\prime(x^\prime_{k-1}) - \nabla f(x_{k-1}))^\top]^\top.
\end{equation}

We present the following lemma for $\mathbf{N}_1(z)$ and $\mathbf{N}_2(z)$ to yield a variable $x_k$ that is no worse than the previous variable $x_{k-1}$ at each iteration $k$. 
\begin{lemma} \label{lm:lemma1_ind_homogeneous_converge_gain}
    Denote the angle between $\mathbf{N}_2(z_{k-1})$ and corresponding $\nabla f(x_{k-1})$ as $\theta_{k-1}$. 
    For $\forall z_{k-1} \in \mathcal{Z}_P, \forall x_{k-1} \in \mathcal{S}_P$,
    if $\mathbf{N}_1(z_{k-1})$ and $\mathbf{N}_2(z_{k-1})$ are respectively bounded by following compact sets:
    \begin{equation*}
        \begin{aligned}
            &\mathbf{N}_1(z_{k-1}) := \lambda_{k-1} \mathbf{1}, \lambda_{k-1} \in \left[0, \frac{1}{L} \right], \\
            &\mathbf{N}_2(z_{k-1}) \in \left[\mathbf{0}, \frac{\|\nabla f(x_{k-1})\|\cos(\theta_{k-1})}{L}\mathbf{1} \right], \theta \in \left[0, \frac{\pi}{2}\right],
        \end{aligned}
    \end{equation*}
    then, for $x_k$ generated by L2O model in \eqref{eq:l2o_smooth}, we have:
    \begin{equation*}
        F(x_k) - F(x_{k-1}) \leq 0.
    \end{equation*}
\end{lemma}
\begin{proof}
    See \cref{sec:lemma1_proof} in Appendix.
\end{proof}

As stated in Lemma~\ref{lm:lemma1_ind_homogeneous_converge_gain}, to maintain homogeneous improvement on the convergence, it is sufficient to set $\mathbf{N}_1(z)$ as an input-invariant constant, and limit $\mathbf{N}_2(z)$ according to the gradient $\nabla{F}(x_{k-1})$. 
Moreover, we can utilize some bounded activation functions in training an L2O model to fulfill the conditions to ensure convergence, such as Sigmoid \cite{Narayan199769} and Tanh \cite{Kalman1992}.

The proof for Lemma~\ref{lm:lemma1_ind_homogeneous_converge_gain} establishes that improvement is characterized by a quadratic relation to each element in $\mathbf{N}_1(z_{k-1})$ and $\|\mathbf{N}_2(z_{k-1})\|$. We can identify the optimal upper bound for convergence improvement in the InD L2O model by optimizing this quadratic relation, leading us to Corollary~\ref{ob:obs_best_converge_gain}.
\begin{corollary} \label{ob:obs_best_converge_gain}
    For any $z_{k-1} \in \mathcal{Z}_P$, we let: 
    \begin{equation*}
        \mathbf{N}_1(z_{k-1}) := \frac{1}{2L}\mathbf{1}, \mathbf{N}_2(z_{k-1}) := \frac{\nabla f(x_{k-1})}{2L}, 
    \end{equation*}
    the Math-L2O model in \eqref{eq:l2o_smooth} is exactly gradient descent update with convergence rate:
    \begin{equation*}
        F(x_K) - F(x^*) \leq \frac{L}{2K} \|x_0 -x^*\|^2.
    \end{equation*}
\end{corollary}
\begin{proof}
    See \cref{sec:obs1_proof} in Appendix.
\end{proof}
Corollary~\ref{ob:obs_best_converge_gain} implies that the L2O model can achieve gradient descent's convergence rate by particular settings. The $\mathbf{N}_1(z_{k-1})$ is set to be a homogeneous constant across all elements. The $\mathbf{N}_2(z_{k-1})$ is set to in correspondence with the gradient $\nabla f(x_{k-1})$. Moreover, Corollary~\ref{ob:obs_best_converge_gain} also provides the most robust L2O model with an identical per iteration convergence gain among all InD instances.

% the convergence gain upper bound (CGUB) of Math-L2O \cite{Liu2023} is lower-bounded by gradient descent. This observation illustrates that, in the worst-case scenario, the convergence gain of a single iteration of the L2O model cannot outperform that of gradient descent. However, it is essential to note that the lower bound of the convergence gain is non-deterministic and is influenced by the training process.
% On the other hand, if we let a Math-L2O model achieve the best convergence improvement (upper bound) in every iteration, the L2O model must emulate a gradient-descent algorithm. 
% The lower bound of convergence is the best convergence contingent upon the training process. 

\subsubsection*{Per-Iteration Convergence Gain}
To ascertain the convergence rate of OOD, following Corollary~\ref{ob:obs_best_converge_gain}, we suppose that after training, the following assumption holds for the \textit{InD L2O Problem} (not for the \textit{OOD L2O Problem}) to ensure best robustness for the InD scenario:
\begin{assumption} \label{as:assump1_ind_robust}
    After training, $\forall x_{k-1} \in \mathcal{S}_P, \forall z_{k-1} \in \mathcal{Z}_P$, $\mathbf{N}_1(z_{k-1}):= \frac{1}{2L}\mathbf{1}$ and $\mathbf{N}_2(z_{k-1}) := \frac{\nabla f(x_{k-1})}{2L}$.
\end{assumption}

Based on the Lemma~\ref{lm:lemma1_ind_homogeneous_converge_gain} and Corollary~\ref{ob:obs_best_converge_gain}, Assumption \ref{as:assump1_ind_robust} leads to an L2O model with best robustness on all InD instances. In the following theorem, we quantify the diminution in convergence rate instigated by the virtual feature $s^\prime$ defined in \cref{sec:vv_traj}. 
\begin{theorem}\label{th:theorem1_smooth_converge_gain}
    Under Assumption \ref{as:assump1_ind_robust}, there exists virtual Jacobian matrices $\mathbf{J}_{1,k-1}, \mathbf{J}_{2,k-1}, k= 1,2, \dots, K$ that the per iteration convergence improvement in the OOD scenario is upper bounded by:
    \begin{equation*}
        \begin{aligned}
            & F^\prime(x_k + s_k) - F^\prime(x_{k-1} + s_{k-1}) \\
             \leq &-\frac{\|\nabla f^\prime(x_{k-1} + s_{k-1})\|^2}{2L} \\
            &  + L\|\diag(\mathbf{J}_{1,k-1} s^\prime) \nabla f^\prime(x_{k-1} + s_{k-1})\|^2 \\
            &  + L \|\frac{\nabla f^\prime(x_{k-1} + s_{k-1})-\nabla f(x_{k-1})}{2L} - \mathbf{J}_{2,k-1} s^\prime \|^2.
        \end{aligned}
    \end{equation*}
\end{theorem}
\begin{proof}
    See \cref{sec:th1_proof} in Appendix.
\end{proof}

Theorem~\ref{th:theorem1_smooth_converge_gain} discloses that for a single iteration, the convergence improvement of OOD is bounded by the gradient descent with a step size of $1/L$, resulting in $-|\nabla f|^2/2L$ convergence improvement. Hence, when Math-L2O is adequately trained, any OOD will dampen convergence. 
Additionally, given that the expression on the right-hand side is not strictly non-positive, we cannot unequivocally affirm that convergence will transpire within a single iteration. Further investigation also intimates that, even in the context of convex optimization problems, scenarios may arise where the value of the objective function deteriorates. 

While the existence of virtual Jacobian matrices in Theorem~\ref{th:theorem1_smooth_converge_gain} is assured, their specific values remain unknown. Given that boundedness is a defined characteristic of these matrices, we relax this constraint in Theorem~\ref{th:theorem1_smooth_converge_gain} and introduce Corollary~\ref{lm:lemma2_smooth_converge_gain_ub}.

\begin{corollary} \label{lm:lemma2_smooth_converge_gain_ub}
    Under Assumption \ref{as:assump1_ind_robust}, the per iteration convergence improvement in the OOD scenario can be upper bounded w.r.t. $\| s_{k-1}^\prime \|$ by:
    \begin{equation*} 
        \begin{aligned}
            & F^\prime(x_k + s_k) - F^\prime(x_{k-1} + s_{k-1})  \\
            \leq & -\frac{\|\nabla f^\prime(x_{k-1} + s_{k-1})\|^2}{2L} \\
            & + \frac{\|\nabla f^\prime(x_{k-1} + s_{k-1})-\nabla f(x)\|^2}{2L} \\
            &  + \left(L C_1^2 n \|\nabla f^\prime(x_{k-1} + s_{k-1})\|^2  + 2L C_2^2 n\right) \|s^\prime \|^2.
        \end{aligned}
    \end{equation*}
\end{corollary}
\begin{proof}
    See \cref{sec:lm2_proof} in Appendix. 
\end{proof}

Corollary~\ref{lm:lemma2_smooth_converge_gain_ub} further elucidates that the decline in the convergence improvement of OOD is determined by the magnitude of the input (virtual) feature $s^\prime$ of the L2O model, as outlined in \eqref{eq:ood_vector}. This magnitude is intrinsically related to the vector's dimensionality, which relies on the feature construction of the L2O model. For example, to reduce its magnitude, we can eliminate $s_{k-1}$ in \eqref{eq:ood_vector}. We achieve this feature shrinking and propose a novel gradient-only L2O model in \cref{sec:design_model}.

\subsection*{Multi-Iteration Convergence Rate}
Building upon Theorem~\ref{th:theorem1_smooth_converge_gain}, we extrapolate the convergence rate across numerous iterations, as delineated in Theorem~\ref{th:theorem2_smooth_converge_rate}. 

\begin{theorem} \label{th:theorem2_smooth_converge_rate}
    Under Assumption \ref{as:assump1_ind_robust}, the $K$ iterations' convergence rate in the OOD scenario is upper bounded by:
    \begin{equation*}
        \begin{aligned}
            & \min_{k=1,\dots, K}F^\prime(x_k+s_k) - F^\prime(x^*+s^*) \\
            \leq & \frac{L}{2}\|x_0 -x^* + s_0- s^*\|^2  - \frac{L}{2}\|x_K -x^* + s_K- s^*\|^2 \label{eq:th2_gd_convergence_rate}\\
            & + \frac{L}{K} \sum_{k=1}^{K} (x_k+ s_k -x^* -s^*)^\top \\
            & \Big(x_k +s_k - \big(x_{k-1} +s_{k-1} - \frac{\nabla f^\prime(x_{k-1}+s_{k-1})}{L}\big)\Big).
        \end{aligned}
    \end{equation*}
\end{theorem}
\begin{proof}
    See \cref{sec:th2_proof} in Appendix.
\end{proof}

The first two terms on the right-hand side of the above inequality represent the gradient descent convergence rate characterized by a step size of $1/L$. However, the third term is unbounded and could be either non-positive or positive. This suggests that there is no guaranteed global convergence in OOD situations, even with homogeneous robustness in InD scenarios.

The inequation above offers a direct approach to analyzing distinct cases of convergence. Included in the concluding line of Theorem~\ref{th:theorem1_smooth_converge_gain} is a gradient descent equation, $x_{k-1} +s_{k-1} - \nabla f^\prime(x_{k-1}+s_{k-1}) / L$. Moreover, $x_k +s_k$ represents the updated solution by the L2O model. The subtraction of the two terms reveals the discrepancy between the updates made by L2O and gradient descent on the objective variable $x_{k-1} +s_{k-1}$, thereby creating a vector directed towards $x_k+ s_k$. Similarly, $x_k+ s_k -x^* -s^*$ signifies the relative position to the optimal solution, generating another vector directed towards $x_k+ s_k$. The resulting inner product will be non-positive if the angle between these two vectors is $\pi / 2$ or more. Moreover, if the trajectory of $x_k+ s_k -x^* -s^*$ can be extrapolated from domain knowledge, a ``trust region'' surrounding $x_k+ s_k$ can be established to augment the efficacy of gradient descent. 

From Theorem~\ref{th:theorem1_smooth_converge_gain}, we develop a stringent formulation to illustrate the potential uncertainty of convergence in OOD scenarios. If we know the relative position of the optimal solution, we can fine-tune an L2O model to outperform gradient descent. Based on Theorem~\ref{th:theorem1_smooth_converge_gain}, we establish an upper bound w.r.t. $s^\prime$. This mirrors the approach in Corollary~\ref{lm:lemma2_smooth_converge_gain_ub}. 
\begin{corollary} \label{lm:lemma3_smooth_converge_rate_ub}
    Under Assumption \ref{as:assump1_ind_robust}, L2O model $d(z)$'s OOD convergence rate is upper bounded w.r.t. $\| s_{k-1}^\prime \|$ by:
    \begin{equation*}
        \begin{aligned}
            & \min_{k=1,\dots, K}F^\prime(x_k+s_k) - F^\prime(x^*+s^*) \\
            \leq & \frac{L}{2}\|x_0 +s_0 -x^* -s^*\|^2  - \frac{L}{2}\|x_K +s_K -x^* -s^*\|^2 \\
            & + \frac{1}{2K} \sum_{k=1}^{K} \big( \nabla f^\prime(x_{k-1}+s_{k-1}) - \nabla f(x_{k-1}) \big)^\top \\
            & \qquad \qquad \quad ( x_k+ s_k -x^* -s^* ) \\
            & + \frac{L}{K} \sum_{k=1}^{K}  \big(C_1 \sqrt{n} \| \nabla f^\prime(x_{k-1}+s_{k-1})\| \\
            & \qquad \qquad + C_2 \sqrt{n} \| x_k+ s_k -x^* -s^* \| \big) \|s_{k-1}^\prime\|.
        \end{aligned}
    \end{equation*}
\end{corollary}
\begin{proof}
    See \cref{sec:lm3_proof} in Appendix.
\end{proof}
Corollary~\ref{lm:lemma3_smooth_converge_rate_ub} posits that the overall convergence rate is consistently upper bounded by the magnitude of $s^\prime$. Based on Corollaries~\ref{lm:lemma2_smooth_converge_gain_ub} and \ref{lm:lemma3_smooth_converge_rate_ub}, we endeavor to reduce the magnitude of $s^\prime$ by eliminating variable, leading to the approach of a gradient-only Math-L2O framework in the next section.

\subsection{Other Three Cases} \label{sec:other_cases}
We have developed several additional theorems and lemmas for non-smooth, incremental historical modeling, and integrated smooth-non-smooth cases.. Our approach mirrors that employed in the smooth case demonstration. The backbone algorithms of math-inspired L2O fundamentally limit their convergences. For example, the Gradient Descent \cite{Ruder16} and Proximal Point \cite{Rockafellar1976} algorithms in the smooth case and the non-smooth case, respectively.

We extend the theorems and lemmas in the smooth case to derive formulas for convergence improvement of a single iteration and convergence rate across a sequence. These demonstrate the diminishing effect of OOD on convergence. Our findings conclude that constructing fewer features can mitigate this negative impact. 
More extensive demonstrations and complete proofs can be found in Appendix.

\section{Gradient-Only L2O Model}
\label{sec:design_model}
Informed by the theorems and lemmas posited in \cref{sec:design_theorem}, we introduce a gradient-only L2O model, GO-Math-L2O, which aims to enhance robustness in OOD scenarios by eliminate variable-related input features for the L2O model.

To derive the formulation of GO-Math-L2O, we employ the workflow delineated in \cite{Liu2023}. Let $T$ denote the history length. At the $k$-th iteration, suppose there exists an operator $d_k \in \mathcal{D}_C(\mathbb{R}^{3n})$, we formulate the input of our GO-Math-L2O as follows:
\begin{equation} \label{eq:d_wox}
    x_k=x_{k-1}-d_k(\nabla f(x_{k-1}), g_k, v_{k-1}),
\end{equation}
where $g_k$ denotes the implicit subgradient vector of $x_k$ to invoke the proximal gradient method \cite{Liu2023}. Moreover, we eliminate all variable-related features and define $v_k$ as the result of historical modeling \cite{Liu2023}. Different from the variable approach in \cite{Liu2023}, we propose to utilize gradient (and subgradient) to model the historical information of the optimization process since gradient sufficiently and necessarily indicates optimality in convex optimization scenarios. Such an approach reduces the magnitude of L2O's input feature (defined in \cref{sec:vv_traj}) by $1/3$, which facilitates convergence based on our proposed corollaries in \cref{sec:design_theorem}. 

Suppose there exists an operator $u_k \in \mathcal{D}_C(\mathbb{R}^{Tn})$, 
we define the following model to generate $v_k$ from the gradient and subgradient of $T$ historical iterations:
\begin{equation} \label{eq:grad_map_v_operator}
    v_k=d_k(\nabla f(x_{k-1}) + g_{k-1}, \ldots, \nabla f(x_{k-T}) + g_{k-T}).
\end{equation}
where each $g$ represents a subgradient vector. For subgradient selection, 
we should carefully choose an instance from the subgradient set of each non-smooth point since an arbitrary selection may lead to poor convergence \cite{Boyd2003}. 

We achieve a lightweight subgradient selection based on the gradient map method \cite{Vandenberghe2022} and our following model constructions. From the objective definition in \cref{sec:def_optimizee}, the non-smooth objective $r$ is trivially solvable by $\argmin$. Thus, at $k$-th iteration, we can recover an implicit subgradient vector $g_k$ of $\argmin$ by $k$-th solution $x_{k}$ and $k_{-1}$-th solution $x_{k-1}$ if the L2O operator $d_k$ in \eqref{eq:d_wox} is inversible. Next, we achieve an inversible $d_k$ based on the workflow proposed in \cite{Liu2023}.

With the above feature and component constructions, we start to define the structures and learnable parameters of our L2O operator $d_k$ in \eqref{eq:d_wox}. We formulate $d_k$ as the necessary condition of convergence \cite{Liu2023}, which means the formulation that $d_k$ should follow if convergence is achieved. 
First, denote a candidate optimal solution as $x^*$, we construct two sufficient conditions (Asymptotic \textbf{F}ixed \textbf{P}oint and \textbf{G}lobal \textbf{C}onvergence) of convergence for our L2O operator $d_k$ in \eqref{eq:d_wox}:
\begin{equation} \label{eq:l2o_model_condition_fp}
    \lim_{k \rightarrow \infty} d_k(\nabla f(x^*), -\nabla f(x^*), 0)=\mathbf{0},\tag{FP}
\end{equation}
\vspace{-6mm}
\begin{equation} \label{eq:l2o_model_condition_gc}
    \lim_{k \rightarrow \infty} x_k = x^*. \tag{GC}
\end{equation}
As discussed in \cite{Liu2023}, such two conditions are essential for optimization algorithms.

Then, we present the following Theorem \ref{lm:go_math_l2o_model} to construct $d_k$'s parameters. 
Theorem \ref{lm:go_math_l2o_model} shows that if $d_k$ converges, it should be in the form of \eqref{lm:go_math_l2o_basic}. Then, if we add a further assumption on some of the parameters, the solution on each iteration can be uniquely obtained by \eqref{lm:go_math_l2o_ppm}.
\begin{theorem} \label{lm:go_math_l2o_model}
    Suppose $T = 2$, given $f \in \mathcal{F}_L(\mathbb{R}^n)$ and $r \in \mathcal{F}(\mathbb{R}^n)$, we pick an operators from $\mathcal{D}_C(\mathbb{R}^{3n})$ and $\mathcal{D}_C(\mathbb{R}^{2n})$. If Condition~\ref{eq:l2o_model_condition_fp} and Condition~\ref{eq:l2o_model_condition_gc} hold, there exist $\mathbf{R}_k \succ 0, \mathbf{Q}_k, \mathbf{B}_k \in \mathbb{R}^{n \times n}$ and $b_{1,k}, b_{2,k} \in \mathbb{R}^n$ and satisfying:
    \vspace{-1mm}
    \begin{equation} \label{lm:go_math_l2o_basic}
        \begin{aligned}
            x_{k} =& x_{k-1}- \mathbf{R}_k\nabla f(x_{k-1}) -\mathbf{R}_k g_{k} - \mathbf{Q}_k v_{k-1}  - b_{1,k},\\
            v_{k} = & (\mathbf{I} - \mathbf{B}_k) G_k + \mathbf{B}_k G_{k-1} - b_{2,k},\\
            G_k :=& \mathbf{R}_{k}^{-1} (x_{k-1} - x_{k} - \mathbf{Q}_k v_{k-1} - b_{1,k}),
        \end{aligned} 
    \end{equation}
    where for $k=0, 1, 2, \dots$, $g_{k+1} \in \partial r(x_{k+1})$ represents implicit subgradient vector, $\mathbf{R}_k$, $\mathbf{Q}_k$, and $\mathbf{B}_k$ are bounded parameter matrices and $b_{1,k}\to 0, b_{2,k} \to 0$ as $k \to \infty$. Since $\mathbf{R}_{k}$ is symmetric positive definite, $x_{k+1}$ is uniquely determined through:
    \vspace{-2mm}
    \begin{equation} \label{lm:go_math_l2o_ppm}
        \begin{aligned}
            \argmin_{x \in \mathbb{R}^n} r(x) 
            + \frac{1}{2} \|x&-\mathbf{R}_{k}\nabla f(x_k) - \mathbf{Q}_kv_k -b_{1,k} \|_{\mathbf{R}_k^{-1}}^2,
        \end{aligned}
    \end{equation}
    where $\|\cdot\|_{\mathbf{R}_k^{-1}}$ is defined as $\|x\|_{\mathbf{R}_k^{-1}}=\sqrt{x^\top \mathbf{R}_k^{-1} x}$.
\end{theorem}
\begin{proof}
    See \cref{sec:tm2_l2o_model_proof} in Appendix.
\end{proof}

As a necessary condition for convergence, Theorem~\ref{lm:go_math_l2o_model} suggests that our gradient-only L2O model should construct parameters $\mathbf{R}$, $\mathbf{Q}$, $\mathbf{B}$, $b_1$, and $b_2$. It is worth noting that this model does not guarantee satisfaction of conditions FP and GC. The convergence is promoted by training. 

We learn to construct the parameters in  Theorem~\ref{lm:go_math_l2o_model}. First, the proof elucidates that the bias terms approach zero upon convergence. Thus, we set $b_1, b_2:=0$ and learn to construct $\mathbf{R}$, $\mathbf{Q}$, and $\mathbf{B}$. 
We take the construction in \cite{Liu2023} to implement our GO-Math-L2O model with a two-layer LSTM cell. Then, we utilize three one-layer linear neural network models with Sigmoid activation function \cite{Narayan199769} to generate $\mathbf{R}$, $\mathbf{Q}$, and $\mathbf{B}$ at each iteration, respectively, which ensures that all the matrices are bounded.
\section{Experiments}
\label{sec:eval}
We perform experiments with Python 3.9 and PyTorch 1.12 on an Ubuntu 18.04 system equipped with 128GB of memory, an Intel Xeon Gold 5320 CPU, and a pair of NVIDIA RTX 3090 GPUs. We strictly follow the experimental setup presented in \cite{Liu2023} for constructing InD evaluations. Due to the page limit, the implementation details are in \cref{sec:evaltion_detail}.

We use the Adam optimizer \cite{Kinga2015} to train our proposed model and learning-based baselines on datasets of 32,000 optimization problems with randomly sampled parameters and optimal solutions. We generate a test dataset of 1,000 iterations' objective values, averaging over 1,024 pre-generated optimization problems. We evaluate different training configurations and loss functions to select the best setting. Details are in \cref{sec:train_config}, Appendix.

% The loss is computed as the objective values summed on 100-length trajectories, then averaged over a mini-batch of 64 samples. We execute a backpropagation step every 20 iterations \cite{Liu2023}. 

\vspace{-2mm}
\paragraph{Baselines.} We compare our GD-Math-L2O (Section~\ref{sec:design_model}) against both learning-based methods and non-learning algorithms. Our main competater is the state-of-the-art (SOTA) math-inspired L2O model in  \cite{Liu2023}. Specifically, we select the best variant from this study, L2O-PA. Consistent with the outlined methodology, we also compare our approach with several hand-crafted algorithms: ISTA, FISTA \cite{Beck2009fast}, Adam \cite{Kinga2015}, and AdamHD \cite{Baydin2017}, which is Adam complemented by an adaptive learning rate. Moreover, we assess our model against two black-box L2O models, namely L2O-DM\cite{Andrychowicz2016} and L2O-RNNprop \cite{Lv2017}, and one Ada-LISTA \cite{Aberdam2021} that unrolls the gradient descent algorithm with learning.

\vspace{-2mm}
\paragraph{Optimization Objective.} 
We choose the two regression problems in \cite{Liu2023}: \textit{LASSO Regression} and \textit{Logistic Regression}, defined as follows:
\begin{equation*}
    \begin{aligned}
        \min _{x \in \mathbb{R}^n} F(x)=&\frac{1}{2}\|\mathbf{A} x - b\|^2+\lambda\|x\|_1,\\
        \min_{x \in  \mathbb{R}^n} F(x)  = & - \frac{1}{m} \sum_{i=1}^{m} \big[b_i \log (h(a_i^\top x))\\
        & +(1-b_i) \log (1-h(a_i^\top x))\big] +\lambda\|x\|_1,
    \end{aligned}
\end{equation*}
where $m:=1000$. $\mathbf{A} \in \mathbb{R}^{250\times 500}$ and $b \in \mathbb{R}^{500}$, $\{(a_i, b_i) \in \mathbb{R}^{50} \times\{0,1\}\}_{i=1}^m$ are given parameters. $h(x) := 1/(1+e^{-x})$ is sigmoid function. 
We utilize the standard normal distribution to generate samples and set $\lambda:= 0.1$ for both scenarios \cite{Liu2023}. 

We implement the Fast Iterative Shrinkage-Thresholding Algorithm (FISTA) \cite{Beck2009fast}, executing 5,000 iterations to generate labels (optimal objective values) \cite{Liu2023}. Due to page limit, we confine our presentation to \textit{LASSO Regression}. The results of \textit{Logistic Regression} are in \cref{sec:eval_logistic_results}, Appendix.

% \qy{
%     Basic Outline: 
%     Two problems: LASSO and Logistic Regression.
%     \begin{enumerate}
%         \item Q: Whether theorem and lemma need to be demonstrated?
%         \item Demonstrate InD performances of Gradient-only. With order baselines in 
%         \item Compare three OOD by two triggers. (1): Initial point shifting. (2) Objective function shifting (3) Both shifting.
%         \item Initial point shifting: Double actions. Two direction.
%         \item Objective function shifting: Double actions. Two direction.
%         \item Both shifting: Double actions. Two direction. One special case: Eliminated with initial point shifting.
%     \end{enumerate}

%     Other details:
%     \begin{enumerate}
%         \item Introduce shifted PPM solution of LASSO and Logistric Regression.
%     \end{enumerate}
% }
\vspace{-2mm}
\paragraph{OOD Scenarios.} 
% LASSO and Logistic regression in OOD scenarios are list below:
% \begin{equation}
%     \min _{x \in \mathbb{R}^n} F(x)=\frac{1}{2}\|\mathbf{A} (x+t) -b\|^2+\lambda\|x+t\|_1
% \end{equation}
We aim to quantify the effect of OOD on convergence rates. We specifically formulate two types of OOD trajectories triggered by different actions. It is crucial to note that both OOD and InD scenarios maintain an identical optimality on both objective and solution.
\begin{enumerate}
    \item[1)] $s_0 \neq 0, s_0 \in \mathbb{R}^n$. $x_0$ is altered by an adjustment factor $s_0$ that $x_0^\prime$ falls within the OOD set $\mathcal{S}_O$. Assuming the objective remains consistent, we expect $ x^\prime$ to move from the OOD $\mathcal{S}_O$ to the InD $\mathcal{S}_P$.
    \item[2)] $F^\prime(x) = F(x+t), t \in \mathbb{R}^n$. The OOD perturbation introduces a translation $t$ along the axes of the objective variable to the objective function. Thus, the optimal solution $x^{\prime *}$ diverges from that obtained under the original InD domain, even though the optimal value remains. This illustrates a scenario where the domain translates in inference. If the starting point is unchanged, $x^\prime$ is expected to move from InD domain to OOD domain.
\end{enumerate}

% , in the context of both regression problems.
We derive the non-smooth function's proximal operator for the OOD scenario, specifically for the $\ell_1$-norm. We define $r(x)$ as $\lambda |x|_1$, and define the OOD translation as $t$ on variable. The OOD proximal operator with $t$ is given by:
\vspace{-2mm}
\begin{equation*}
    \begin{aligned}
        (&\operatorname{prox}_{r, p_k}(\bar{x}))_i\\
        &:= -t + \sign(\bar{x}_i) \max (0,|\bar{x}_i|-\lambda(p_k)_i + \sign(\bar{x}_i)t).
    \end{aligned}
\end{equation*}

\subsection{InD Comparison}
\begin{figure}
    \centering
    \includegraphics[width=0.99\linewidth]{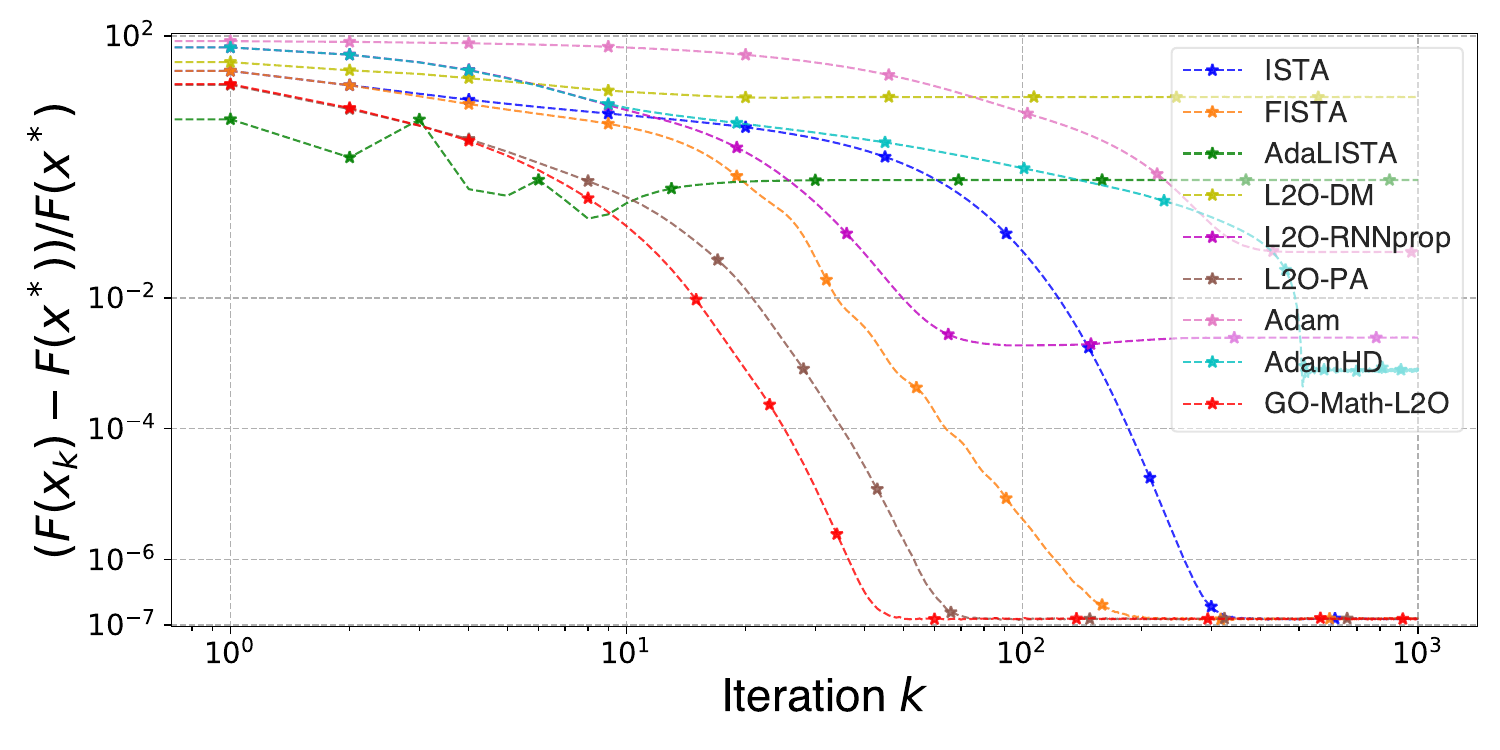}
    \vspace{-4mm}
    \caption{LASSO Regression: InD.}
    \label{fig:figure1_ind_cmp_lasso}
\end{figure}

The trajectories of solving the \textit{LASSO Regression} problems are shown in Figure~\ref{fig:figure1_ind_cmp_lasso}, where the vertical axis represents the normed objective value at a given iteration (indicated on the horizontal axis) with a label generated by FISTA \cite{Beck2009fast}. 
Our proposed method (red line) surpasses all other methods, demonstrating better optimality and quicker convergence.

Furthermore, we utilize several ablation studies on model configuration, such as gradient map recovery strategies in \cref{sec:eval_grad_map} and hyperparameter settings for learned parameter matrices in \cref{sec:eval_l_shifting}, to determine the best model configuration. The details are in the Appendix.

% \begin{figure}
%     \centering
%     \includegraphics[width=0.99\linewidth]{fig/figure1_ind_cmp_lasso.pdf}
%     \vspace{-2mm}
%     \caption{\qy{TODO our model} LASSO Regression: InD.}
%     \label{fig:lasso_ind}
% \end{figure}

\subsection{OOD Comparison}
\begin{figure}
    \centering
    \includegraphics[width=0.99\linewidth]{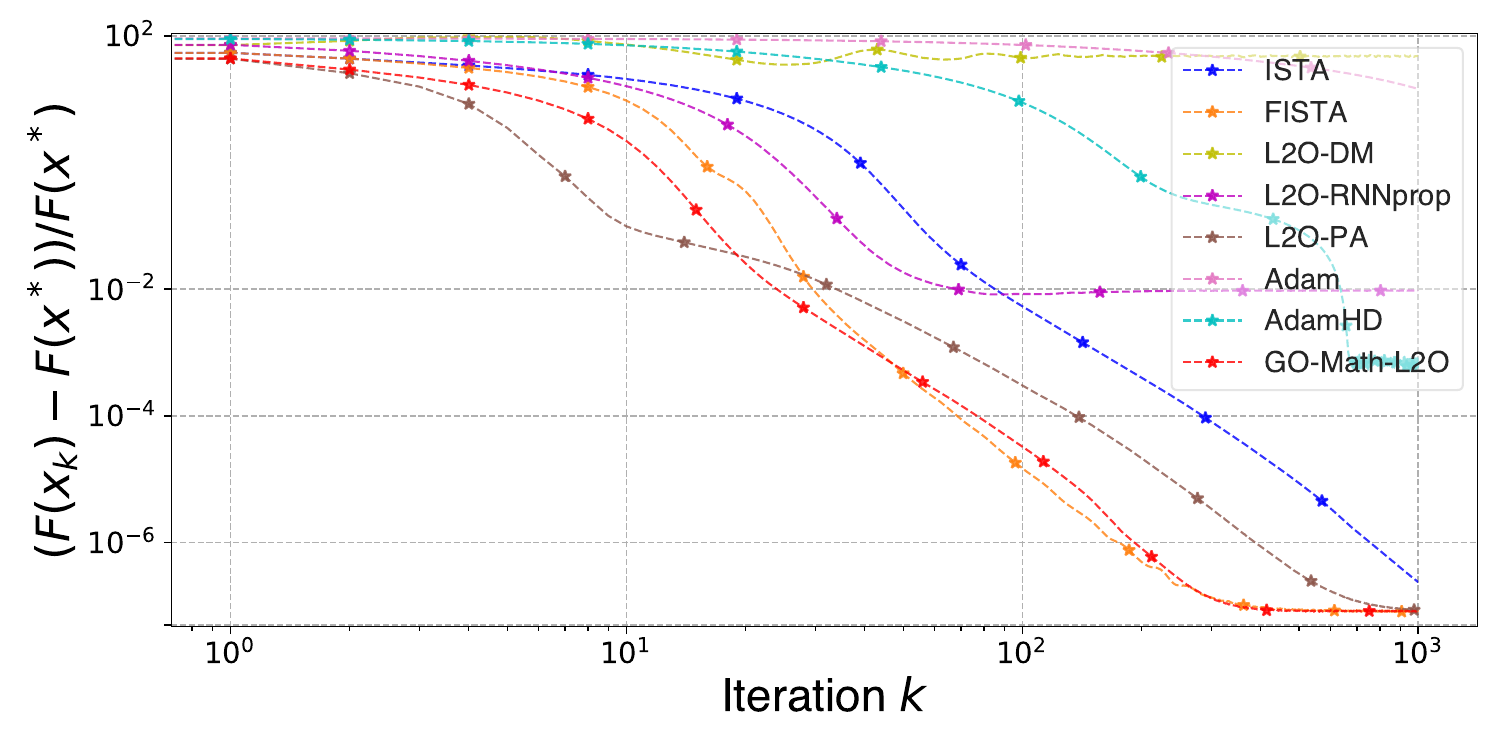}
    \vspace{-4mm}
    \caption{LASSO Regression: Real-World OOD.}
    \label{fig:figure2_ood_cmp_lasso_real}
\end{figure}

The real-world results in Firgure~\ref{fig:figure2_ood_cmp_lasso_real} show that our GO-Math-L2O (converges at 400 iterations) outperforms all other baselines (1,000 iterations). Considering the lackluster performances of other baselines in Figures~\ref{fig:figure1_ind_cmp_lasso} and \ref{fig:figure2_ood_cmp_lasso_real}, we primarily compare our GO-Math-L2O model against SOTA L2O-PA \cite{Liu2023}. We construct two synthetic OOD scenarios with the two trigger settings, where the optimal objectives align with those in Figure~\ref{fig:figure1_ind_cmp_lasso}.

\begin{figure}
    \centering
    \includegraphics[width=0.99\linewidth]{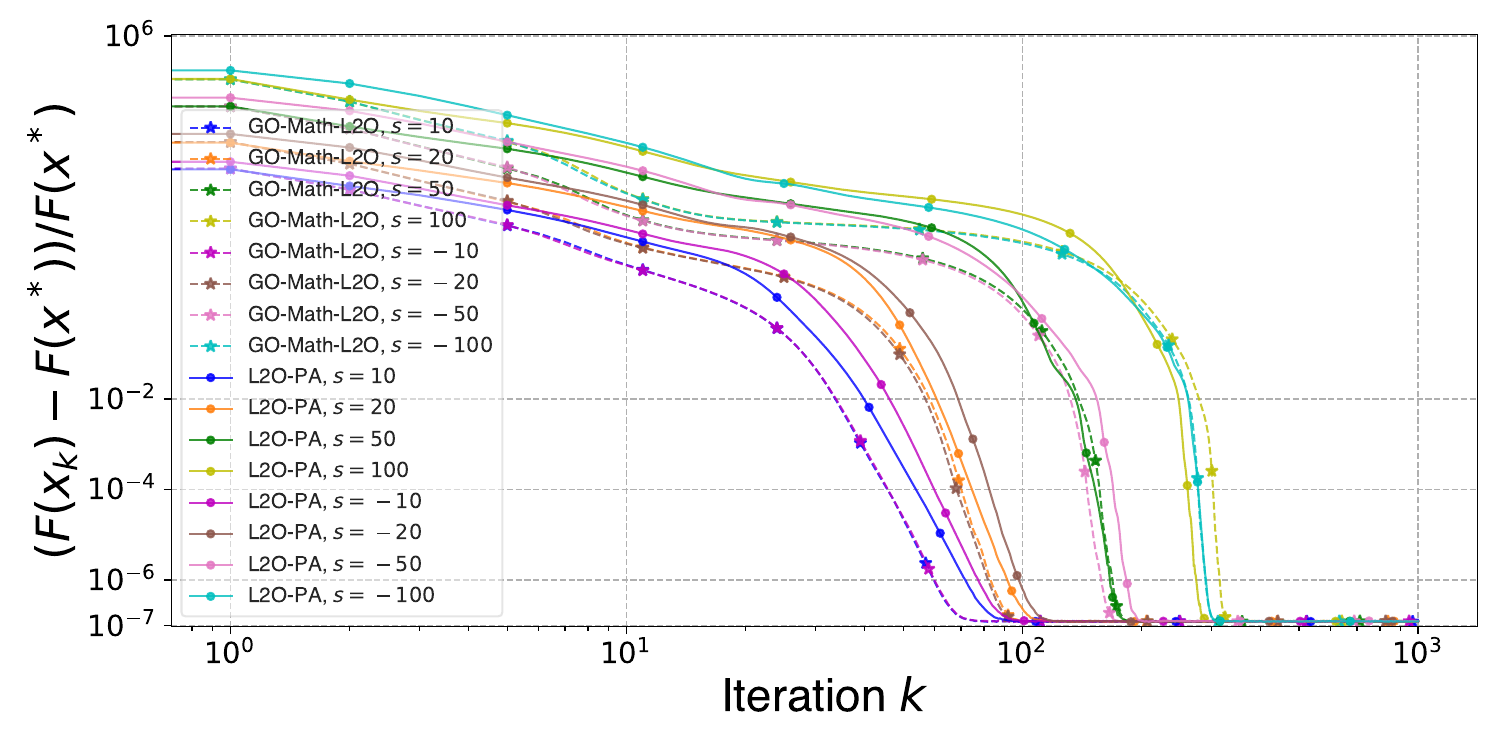}
    \vspace{-4mm}
    \caption{LASSO Regression: OOD by Trigger 1.}
    \label{fig:figure3_ood_cmp_lasso_s}
\end{figure}

Figure~\ref{fig:figure3_ood_cmp_lasso_s} portrays the scenario wherein the initial point shifts such that $s_0 \neq 0$, with the legends denoting sixteen cases. Our GO-Math-L2O model (represented by dashed lines) outshines L2O-PA (solid lines) in all instances, asserting its superior robustness. 

% Furthermore, our model exhibits similar trajectories for each pair of mirror settings, such as $s=50$ and $s=-50$. However, a nearly 300-iteration gap exists between these two settings in L2O-PA.

\begin{figure}
    \centering
    \includegraphics[width=0.99\linewidth]{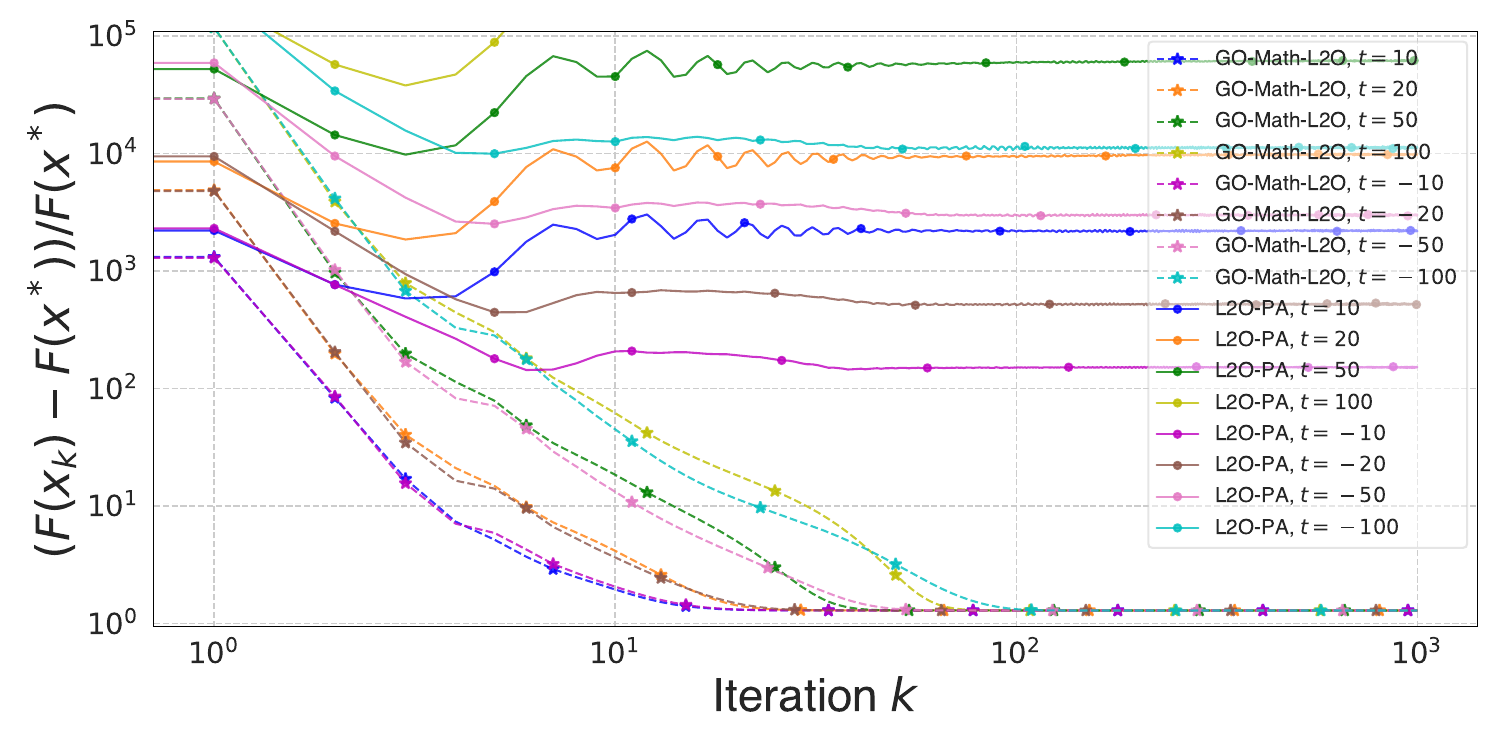}
    \vspace{-4mm}
    \caption{LASSO Regression: OOD by Trigger 2.}
    \label{fig:figure4_ood_cmp_lasso_t}
\end{figure}

The observations in Figure~\ref{fig:figure4_ood_cmp_lasso_t} for the OOD scenario involve function shifting that $F^\prime(x) = F(x+t)$. The optimal values achieved by both methods deteriorate from $10^{-7}$ (as seen in Figure~\ref{fig:figure3_ood_cmp_lasso_s}) to $10^{0}$. However, our GO-Math-L2O still outperforms L2O-PA in all cases. For example, when $t=\pm10$, our model converges at around 20 steps, but L2O-PA fails to converge. 
\vspace{-2mm}

%!TEX root = main.tex
\section{Conclusion}
\label{sec:conclusion}

This paper aims to improve the robustness of L2O in OOD scenarios. We derive a general condition to ensure robustness in InD scenarios. We propose virtual features to connect the OOD L2O's outputs with InD L2O's outputs of a whole trajectory. Based on such connections, we prove formulations to demonstrate the convergence performances in OOD scenarios. Based on the observations, we establish that the magnitude of the L2O model's input features intrinsically limits the OOD's convergence. 
Furthermore, we propose a robust L2O model with concise gradient-only features and modeling historical features with gradient and subgradient. 
Experiments show our model significantly outperforms SOTA baselines. 
% \vspace{-2mm}

% Our findings affirm that even though robustness is achieved in InD scenarios, the guarantee of convergence is compromised by OOD, potentially leading to unpredictable optimal solutions. We further establish that the magnitude of the L2O model input features intrinsically limits the convergence guarantee. 

% We assume any L2O model performs identically in all InD instances after training. Moving forward, we intend to moderate this assumption and delve into the influence of training on our present findings.
\section*{Acknowledgements}
This work is partly supported by funding from the Research Grants Council of Hong Kong (11209520, CRF C7004-22G) and CUHK (4055199).

\newpage
{
    
    \small
    \bibliographystyle{ieeenat_fullname}
    \bibliography{main}
}


\onecolumn

\clearpage
\setcounter{page}{1}
% \maketitlesupplementary

\onecolumn
{
\centering{\Large{\textbf{\thetitle}\\
\vspace{0.5em}Supplementary Material \\
\vspace{1.0em}}}
}

% \tableofcontents

\section{Proofs}
\label{sec:rationale}

\subsection{Preliminary} \label{sec:thpf_smooth_pre}

\subsubsection*{Demonstration of Equation \ref{eq:traj_mean_value_th}} \label{sec:thpf_eq3}

\begin{proof}
    Based on demonstration for Lemma 1 in \cite{Liu2023}, since $d \in \mathcal{D}_C(m)$, the outcome of$d$ is an $n$-dimensional vector. Denote the $i$-th element as $d_i (1 \leq i \leq n)$ and convert $d$ into a matrix form:
\begin{equation*}
    \begin{aligned}
        d(z^\prime) &= \big[d_1(z^\prime), \dots, d_n(z^\prime)\big]^\top, \\
        d(z) &= \big[d_1(z), \dots, d_n(z)\big]^\top.
    \end{aligned}
\end{equation*}

Regarding each $d_i(z^\prime)$ as a multi-variable function and applying the Mean Value Theorem on it, for some $\xi_i \in (0, 1)$, we can construct following equality:
\begin{equation*}
    d_i(z^\prime) - d_i(z) = \Big\langle\frac{\partial d_i}{\partial z_i}\big(\xi_i z^\prime + (1-\xi_i) z\big), z^\prime - z \Big\rangle.
\end{equation*}

Stacking all partial derivatives into one matrix yields:
\begin{equation*}
    \mathbf{J}_d = \left[\frac{\partial d_1}{\partial z}\big(\xi_1 z^\prime + (1-\xi_1) z\big), \dots, \frac{\partial d_n}{\partial z}\big(\xi_n z^\prime + (1-\xi_n) z\big) \right].
\end{equation*}
We can directly get \eqref{eq:traj_mean_value_th}. And the upper bound of $\|\mathbf{J}_d\|$ is given by:
\begin{equation*}
    \|\mathbf{J}_d\|^2 \leq \|\mathbf{J}_d\|^2_\mathrm{F} = \sum_{i=1}^{n} \left\|\frac{\partial d_i}{\partial z}\big(\xi_i z^\prime + (1-\xi_i)\right\|^2 \leq n C^2.
\end{equation*}
% Thus, $\|\mathbf{J}\| \leq  C \sqrt{n}$.
\end{proof}

\subsubsection*{A General Upper Bound from $L$-smoothness}

Suppose $z, \tilde{z}, z^\prime \in \mathcal{Z}$ are input feature vectors of the L2O model. There exists a $\xi \in [0,1]$ that $z^\prime = \xi z + (1-\xi)\tilde{z}, z^\prime \in \mathcal{Z}$. Thus, we are able to represent OOD feature $z^\prime$ with InD feature $z$. Denote the virtual Jacobian matrix of $\mathbf{N}_1(z^\prime) \in \mathcal{D}_{C_1}$ and $\mathbf{N}_2(z^\prime) \in \mathcal{D}_{C_2}$ at point $z^\prime$ as $\mathbf{J}_1$ and $\mathbf{J}_2$.
Since $\mathbf{N}_1(z)$ and $\mathbf{N}_1(z)$ are smooth, due to the Mean Value Theorem, we have the following equality between OOD and InD outputs of the L2O model:
\begin{equation*}
    \mathbf{N}_1(z) = \mathbf{N}_1(\tilde{z}) + \mathbf{J}_1(z - \tilde{z}),\quad \mathbf{N}_2(z) = \mathbf{N}_2(\tilde{z}) + \mathbf{J}_2(z - \tilde{z}).
\end{equation*}
As demonstrated above, we have $\|\mathbf{J}_1\| \leq \sqrt{n}C_1$, and $\|\mathbf{J}_2\| \leq \sqrt{n}C_2$.

As illustrated in \cref{sec:vv_traj}, we represent OOD input feature vector $z^\prime$ as a combination of InD input feature vector $z$ and virtual feature vector $s^\prime$. For a $z^\prime = z + s^\prime$, we have the following equalities:
\begin{equation} \label{thpf:mvt_l2o}
    \begin{aligned}
        &\mathbf{N}_1(z+s^\prime) = \mathbf{N}_1(z) + \mathbf{J}_1(z + s^\prime - z) = \mathbf{N}_1(z) + \mathbf{J}_1 s^\prime, \\
        &\mathbf{N}_2(z+s^\prime) = \mathbf{N}_2(z) + \mathbf{J}_2(z + s^\prime - z) = \mathbf{N}_2(z) + \mathbf{J}_2 s^\prime.
    \end{aligned}
\end{equation}

% The shifting on gradient is caused by shifting on $x$
% \begin{equation} \label{thpf:ub}
%     \begin{aligned}
%         \mathbf{N}_1(z+s^\prime) \leq \mathbf{N}_1(z)+\mathbf{J}_{1, z}(z+s^\prime-z)+\frac{C_1}{2}\|z+s^\prime-z|^2 = \mathbf{N}_1(z)+\mathbf{J}_{1, z}(s^\prime)+\frac{C_1}{2}\|s^\prime\|^2\\
%         \mathbf{N}_2(z+s^\prime) \leq \mathbf{N}_2(z)+\mathbf{J}_{2, z}(z+s^\prime-z)+\frac{C_2}{2}\|z+s^\prime-z\|^2 = \mathbf{N}_2(z)+\mathbf{J}_{2, z}(s^\prime)+\frac{C_2}{2}\|s^\prime\|^2
%     \end{aligned}
% \end{equation}

Following \cite{Liu2023}, in the smooth objective case, if we use variable and gradient to construct input features, we can further formulate $s^\prime$ as $s^\prime:= \Big[s^\top, \big(\nabla f^\prime(x+s) - \nabla f(x)\big)^\top\Big]^\top$. For $s^\prime$, we have the following inequalities:
\begin{equation} \label{thpf:lb_shift_norm}
    \begin{aligned}
        \|s^\prime\|^2 &= \|s\|^2 + \|\nabla f^\prime(x+s) - \nabla f(x)\|^2 \geq \|\nabla f^\prime(x+s) - \nabla f(x)\|^2.
    \end{aligned}
\end{equation}
The above inequation gives a theoretical lower bound of the input feature's magnitude in \cite{Liu2023}. Our following results will demonstrate that the convergence rate of learning-to-optimize will be the upper bound with respect to the magnitude of the L2O model's input feature. Based on such a lower bound, we are able to improve the convergence rate by eliminating variable features.

% \begin{equation} \label{thpf:ub_shift_norm}
%     \begin{aligned}
%         \|s^\prime\| &= \sqrt{\|s\|^2 + \|\nabla f^\prime(x+s) - \nabla f(x)\|^2}\\
%         &\leq \sqrt{\|s\|^2 + 2\|\nabla f^\prime(x+s)\| + 2\|\nabla f(x)\|^2}\\
%         &\leq \sqrt{\|s\|^2 + 2L^2 + 2L^2} \\
%         &= \sqrt{\|s\|^2 + 4L^2}.
%     \end{aligned}
% \end{equation}

% \qy{NOTE: $\mathbf{J}_1$ and $\mathbf{J}_2$ are corelated to both $z$ and $\tilde{z}$!!!}

% Due to the $L$-smoothness definition of $f(x)$ in \eqref{eq:function_space}, we have the following inequalities for gradient:
% \begin{equation} \label{thpf:up}
%     \begin{aligned}
%         \|\nabla f^\prime(x+s) - \nabla f(x)\| &\leq  \|\nabla f^\prime(x+s)\|  +  \|\nabla f(x)\| \leq 2L.
%         % ,\\\|\nabla f^\prime(x+s)\| - \|\nabla f(x)\| &\leq  L\|s\|, \\
%         % \|\nabla f^\prime(x+s)\| &\leq \|\nabla f(x)\| + L\|s\|.
%     \end{aligned}
% \end{equation}
% \begin{equation} \label{thpf:up2}
%     \|\nabla f^\prime(x+s) - \nabla f(x)\|\leq  L \|x+s-x\| =  L \|s\|.
% \end{equation}

For any $x, x^+ \in \mathbb{R}^n$, we use $x$ and $x^+$ to denote the variables before and after the update. Based on the definition of $L$-smoothness \cite{Ryan2013}, we have following upper bound on objective $F$:
\begin{equation*}
    F(x^+) \leq F(x) + \nabla f(x)^\top(x^+ - x) + \frac{L}{2}\|x^+ - x\|^2.
\end{equation*}
Note that in problem \ref{obj:ood} (\cref{sec:definition}), we define that an objective $F(x)$ has two parts: smooth part $f(x)$ and non-smooth part $r(x)$. And in the smooth case, we set $r(x):=0$.

Substituting $x^+ = x - \diag\big(\mathbf{N}_1(z)\big)^\top \nabla f(x) - \mathbf{N}_2(z)$, we have:
\begin{equation} \label{thpf:smooth_base}
    \begin{aligned}
        F(x^+) & \leq F(x) + \nabla f(x)^\top\big(x - \diag\big(\mathbf{N}_1(z)\big) \nabla f(x) - \mathbf{N}_2(z) - x\big) + \frac{1}{2}L\big\|x - \diag\big(\mathbf{N}_1(z)\big)\nabla f(x) - \mathbf{N}_2(z) - x\big\|^2, \\
        & = F(x) - \nabla f(x)^\top \diag\big(\mathbf{N}_1(z)\big) \nabla f(x) - \nabla f(x)^\top\mathbf{N}_2(z) + \frac{L}{2}\big\|\diag\big(\mathbf{N}_1(z)\big)\nabla f(x) + \mathbf{N}_2(z)\big\|^2.
    \end{aligned}
\end{equation}

If $\nabla f(x):=\mathbf{0}$, we have:
\begin{equation*}
    F(x^+) \leq F(x) + \frac{L}{2}\big\|\mathbf{N}_2(z)\big\|^2.
\end{equation*}

To ensure $F(x^+) \leq F(x)$, we should set $\big\|\mathbf{N}_2(z)\big\|:=\mathbf{0}$. Thus, $\mathbf{N}_2(z):=\mathbf{0}$. Otherwise $\nabla f(x) \neq \mathbf{0}$, we can split $\mathbf{N}_1(z)$ and $\mathbf{N}_2(z)$ in \eqref{thpf:smooth_base} by:
\begin{equation*}
    \begin{aligned}
        F(x^+) & \leq F(x) - \nabla f(x)^\top \diag\big(\mathbf{N}_1(z)\big) \nabla f(x) - \nabla f(x)^\top\mathbf{N}_2(z) + \frac{L}{2}\big\|\diag\big(\mathbf{N}_1(z)\big)\nabla f(x) + \mathbf{N}_2(z)\big\|^2, \\
        & \leq F(x) - \nabla f(x)^\top \diag\big(\mathbf{N}_1(z)\big) \nabla f(x) - \nabla f(x)^\top\mathbf{N}_2(z) + L\big\|\diag\big(\mathbf{N}_1(z)\big)\nabla f(x)\big\|^2 + L\big\|\mathbf{N}_2(z)\big\|^2, \\
        & = F(x) - \nabla f(x)^\top \diag\big(\mathbf{N}_1(z)\big) \nabla f(x) - \nabla f(x)^\top\mathbf{N}_2(z) + L\nabla f(x)^\top\diag\big(\mathbf{N}_1(z)\big)^2\nabla f(x) + L\big\|\mathbf{N}_2(z)\big\|^2, \\
        & = F(x) - \nabla f(x)^\top\left(\diag\big(\mathbf{N}_1(z)\big)-L\diag\big(\mathbf{N}_1(z)\big)^2\right)\nabla f(x) - \nabla f(x)^\top\mathbf{N}_2(z) + L\big\|\mathbf{N}_2(z)\big\|^2.
    \end{aligned}
\end{equation*}

We construct the following inequality to ensure a homogeneous decrease in objective:
\begin{equation*}
    \begin{aligned}
            - \nabla f(x)^\top\left(\diag\big(\mathbf{N}_1(z)\big)-L\diag\big(\mathbf{N}_1(z)\big)^2\right)\nabla f(x) - \nabla f(x)^\top\mathbf{N}_2(z) + L\big\|\mathbf{N}_2(z)\big\|^2 \leq 0,
    \end{aligned}
\end{equation*}
where $\big(\big)^2$ on a matrix or a vector represents the entry-wise square, respectively. We continue to use this denotation below.

We first demonstrate the convergence gain on each iteration by $- \nabla f(x)^\top\Big(\diag\big(\mathbf{N}_1(z)\big)-L\diag\big(\mathbf{N}_1(z)\big)^2\Big)\nabla f(x)$ and $- \nabla f(x)^\top\mathbf{N}_2(z) + L\big\|\mathbf{N}_2(z)\big\|^2$ respectively (Lemma \ref{lm:lemma1_ind_homogeneous_converge_gain}) and overall convergence rate over $K$ iterations. Then, we analyze the out-of-distribution effect on the convergence rate.

% \qy{TODO: Several part are for smooth case only, remove it to the proof of smooth case.}

\subsection{Proof of Lemma \ref{lm:lemma1_ind_homogeneous_converge_gain}} \label{sec:lemma1_proof}
% : InD Per-Iteration Robustness
\begin{proof}
    This proof demonstrates a sufficient condition to ensure a robust L2O model with per iteration convergence guarantee.

    Due to the proof in Section A.2. in \cite{Liu2023}, $\big\|\mathbf{N}_2(z)\big\| \rightarrow 0$ when $x^+ \rightarrow x^*$. We frist derive the conditions for $\mathbf{N}_2(z) = \mathbf{0}$ case. Based on the solutions, we derive the conditions for $\mathbf{N}_2(z)$ in $\mathbf{N}_2(z) \neq \mathbf{0}$ case.
    \paragraph{Case 1 $\mathbf{N}_2(z) = \mathbf{0}$.}
    \begin{equation} \label{thpf:smooth_case1_base}
        \begin{aligned}
             & - \nabla f(x)^\top\Big(\diag\big(\mathbf{N}_1(z)\big)-L\diag\big(\mathbf{N}_1(z)\big)^2\Big)\nabla f(x) - \nabla f(x)^\top\mathbf{N}_2(z) + L\big\|\mathbf{N}_2(z)\big\|^2 \\
             = &- \nabla f(x)^\top\Big(\diag\big(\mathbf{N}_1(z)\big)-L\diag\big(\mathbf{N}_1(z)\big)^2\Big)\nabla f(x), \\
             = &- \nabla f(x)^\top\Big(\diag(\mathbf{N}_1(z)-L\mathbf{N}_1(z)^2)\Big)\nabla f(x),
        \end{aligned}
    \end{equation}
    where $\mathbf{N}_1(z)^2$ represents coordinate-wise square over a vector. To keep $- \nabla f(x)^\top\Big(\diag\big(\mathbf{N}_1(z)-L\mathbf{N}_1(z)^2\big)\Big)\nabla f(x) \leq 0, \forall \nabla f(x) $, we should have:
    \begin{equation*}
        \begin{aligned}
            \diag\big(\mathbf{N}_1(z)-L\mathbf{N}_1(z)^2\big) & \succeq 0,\\
            \mathbf{N}_1(z)-L\mathbf{N}_1(z)^2 & \geq 0,
        \end{aligned}
    \end{equation*}
    where $L > 0$ is given by $L$-smoothness definition in \eqref{eq:function_space}. Solving the above quadratic inequality, we have the following range for $\mathbf{N}_1(z)$:
    \begin{equation} \label{thpf:smooth_N1_bound1}
        0 \leq \mathbf{N}_1(z) \leq \frac{1}{L}, \forall z \in \mathcal{Z}.
    \end{equation}
    The left-hand side $\mathbf{N}_1(z)-L\mathbf{N}_1(z)^2$ achieves maxima $\frac{1}{4L}$ at $\mathbf{N}_1(z) = \frac{1}{2L}$.

    % \qy{NOTE 1: Convergence gain of training.}
    Hence, due to the $L$-smoothness of $f(x)$, InD's convergence gain in one iteration has the following lower bound:
    \begin{equation} \label{thpf:smooth_part1_lu_bound_ind}
        -\frac{L}{4} \leq -\frac{\|\nabla f(x)\|^2}{4L}  \leq - \nabla f(x)^\top\Big(\diag\big(\mathbf{N}_1(z)\big)-L\diag\big(\mathbf{N}_1(z)\big)^2\Big)\nabla f(x).
    \end{equation}
    % The above solution demonstrate that Assumption 1 is a necessary condition for convergence. Obviously, it is a suffcient condition as well. 

    \paragraph{Case 2 $\mathbf{N}_2(z) \neq \mathbf{0}$.} 
    We first freeze $- \nabla f(x)^\top\mathbf{N}_2(z) + L\big\|\mathbf{N}_2(z)\big\|^2 $ and apply the derivation in Case 1 to keep a non-positive $- \nabla f(x)^\top\Big(\diag\big(\mathbf{N}_1(z)\big)-L\diag\big(\mathbf{N}_1(z)\big)^2\Big)\nabla f(x)$. Similar to the demonstration of $\mathbf{N}_1(z)$, we construct the following inequation for $\mathbf{N}_2(z)$:
    \begin{equation*}
        - \nabla f(x)^\top\mathbf{N}_2(z) + L\big\|\mathbf{N}_2(z)\big\|^2 \leq 0.
    \end{equation*}

    Suppose the angle between $\mathbf{N}_2(z)$ and $\nabla f(x)$ is $\theta$. The left-hand side of the above equation can be represented by:
    \begin{equation*}
        - \|\nabla f(x)\|*\big\|\mathbf{N}_2(z)\big\|\cos(\theta) + L\big\|\mathbf{N}_2(z)\big\|^2 = \big\|\mathbf{N}_2(z)\big\| \Big(- \|\nabla f(x)\|\cos(\theta) + L \big\|\mathbf{N}_2(z)\big\|\Big) \leq 0.
    \end{equation*}
    Note that $\theta$ should follow $\theta \in [0, \frac{\pi}{2}]$ to avoid inherent non-negative of left-hand side in the above inequalities. 
    Solve the above inequality, we have:
    \begin{equation} \label{thpf:smooth_N2_bound1}
        0 \leq \big\|\mathbf{N}_2(z)\big\| \leq \frac{\|\nabla f(x)\|\cos(\theta)}{L} \leq \frac{\|\nabla f(x)\|}{L}.
    \end{equation}
    Substituting it back, we get minima at $\big\|\mathbf{N}_2(z)\big\| = \frac{\|\nabla f(x)\|}{2L}$ as $-\frac{\|\nabla f(x)\|^2}{4L}$. 
    Note that if $\theta=0$, the above equation achieves maxima at $-\frac{1}{4L}\|\nabla f(x)\|^2$, which means $\mathbf{N}_2(z)$ is in the same direction with $\nabla f(x)$, i.e., $\mathbf{N}_2(z) = \frac{\nabla f(x)}{2L}$.
    % \begin{equation*}
    %     - \|\nabla f(x)\|*\frac{\|\nabla f(x)\|\cos(\theta)}{2L}\cos(\theta) + L(\frac{\|\nabla f(x)\|\cos(\theta)}{2L})^2 =-\frac{\cos(\theta)^2}{4L}\|\nabla f(x)\|^2.
    % \end{equation*}
    
    % \qy{Insight 1: This may be the another insight that we should pick it up: How to design the output of the neural network $\mathbf{N}_2(z)$? Directly set $\mathbf{N}_2(z)=\lambda \nabla f(x)$, $\lambda \in [0, \frac{1}{L}]$ in experiments.}
\end{proof}

\subsection{Proof of Corollary \ref{ob:obs_best_converge_gain}} \label{sec:obs1_proof}
% : Robust InD L2O
\begin{proof}
    From the proof of Lemma \ref{lm:lemma1_ind_homogeneous_converge_gain}, we construct separate conditions for the output of neural networks $\mathbf{N}_1$ and $\mathbf{N}_2$ by decomposing the per iteration convergence rate into two quadratic formulas with respect to the neural network models in the L2O model. 
    This proof demonstrates the most robust L2O model in the InD scenario, which achieves the per iteration convergence gain of gradient descent. 

    The quadratic formular $\mathbf{N}_1(z)-L\mathbf{N}_1(z)^2$ with respect to $\mathbf{N}_1(z)$ achieves maxima $\frac{1}{4L}$ at $\mathbf{N}_1(z) = \frac{1}{2L}$. The quadratic formular $\big\|\mathbf{N}_2(z)\big\| \Big(- \|\nabla f(x)\|\cos(\theta) + L \big\|\mathbf{N}_2(z)\big\|\Big) $  with respect to $\mathbf{N}_2(z)$ achieves minima $-\frac{\|\nabla f(x)\|^2}{4L}$ at $\big\|\mathbf{N}_2(z)\big\| = \frac{\|\nabla f(x)\|}{2L}$. We derive the best convergence rate after $K$ iterations in this part for InD and OOD cases, respectively. The convexity of $f(x)$ yields $f(x) \leq f(x^*)+\nabla f(x)^\top(x-x^*)$ \cite{Ryan2013}. 

    Due to \eqref{thpf:smooth_N1_bound1} and \eqref{thpf:smooth_N2_bound1}, when $\mathbf{N}_1(z)=\frac{1}{2L}$ and $\mathbf{N}_1(z)=\frac{\nabla f(x)}{2L}$, the update formula for one iteration is as following, which is precisely gradient descent with $\frac{1}{L}$ step size.
    \begin{equation} \label{thpf:converge_rate_ind_best}
            x^+ = x - \frac{1}{2L} \nabla f(x) - \frac{\nabla f(x)}{2L}= x - \frac{1}{L} \nabla f(x).
    \end{equation}
    We note that this corollary demonstrates that the L2O model can achieve the best upper-bound convergence rate of gradient descent. However, its lower bound is non-deterministic and relies on training. 
    % \qy{Insight 5: }

    Based on the definition of $L$-smoothness on $f$, we have:
    \begin{equation*}
        F(x^+) \leq F(x^*)+\nabla f(x)^\top(x-x^*) -\frac{\|\nabla f(x)\|^2}{2L}.
    \end{equation*}

    In $k$-th iteration, $k \geq 1$, we have:
    \begin{equation*}
        \begin{aligned}
            F(x_{k}) - F(x^*) & \leq \nabla f(x_{k-1})^\top(x_{k-1}-x^*) -\frac{\big\|\nabla f(x_{k-1})\big\|^2}{2L}  , \\
            & = \frac{1}{2L} \Big(2L \nabla f(x_{k-1})^\top(x_{k-1}-x^*) - \big\|\nabla f(x_{k-1})\big\|^2\Big) , \\
            & = \frac{1}{2L} \Big(2L \nabla f(x_{k-1})^\top(x_{k-1}-x^*) - \big\|\nabla f(x_{k-1})\big\|^2 - L^2\|x_{k-1}-x^*\|^2 + L^2\|x_{k-1}-x^*\|^2 \Big) , \\
            & = \frac{1}{2L} \Big(L^2\|x_{k-1}-x^*\|^2 - \big\|L(x_{k-1}-x^*) - \nabla f(x_{k-1})\big\|^2\Big) , \\
            & = \frac{1}{2L} \Big(L^2\|x_{k-1}-x^*\|^2 - L^2 \big\|x_{k-1} - \frac{1}{L} \nabla f(x_{k-1}) -x^*\big\|^2\Big) , \\
            (\text{Due to \eqref{thpf:converge_rate_ind_best}})& = \frac{L}{2} \Big(\big\|x_{k-1}-x^*\|^2 - \|x_{k} -x^*\big\|^2\Big).
        \end{aligned}
    \end{equation*}

    Sum over all $K$ iterations, we have:
    \begin{equation*}
        \sum_{k=1}^{K}F(x_k) - F(x^*)
        \leq \frac{L}{2} \sum_{k=1}^{K}  \big(\|x_{k-1}-x^*\|^2 - L^2 \|x_{k} -x^*\|^2\big) 
        =  \frac{L}{2} \|x_0 -x^*\|^2.
    \end{equation*}

    Since $F(x_K) - F(x^*)$ is the minimum of the left-hand side of the above, we have:
    \begin{equation*}
        F(x_K) - F(x^*) \leq \frac{L}{2K} \|x_0 -x^*\|^2.
    \end{equation*}

\end{proof}

\subsection{Proof of Theorem \ref{th:theorem1_smooth_converge_gain}} \label{sec:th1_proof}
% : OOD Convergence Gain Upper Bound Formula
% \qy{Workflow: Directly make up relationship between $\mathbf{N}_1(z + s)$ and $\mathbf{N}_1(z)$ with mean value Theorem.}

\begin{proof}
    In the proof of Lemma \ref{lm:lemma1_ind_homogeneous_converge_gain}, we have demonstrated that to ensure a robust L2O with a homogeneous convergence gain for any $x$ and $F$, we should bound the neural networks $\mathbf{N}_1(z)$ and $\mathbf{N}_2(z)$ in the L2O model into those compact sets respectively. In Corollary \ref{ob:obs_best_converge_gain}, we give one sufficient condition to ensure the best robustness for the L2O model. In this proof, upon Corollary \ref{ob:obs_best_converge_gain}, we formulate the L2O model's convergence in the OOD scenario in this proof.

    \paragraph{Convergence Gain by $\mathbf{N}_1(z)$.} 
    Due to \eqref{thpf:mvt_l2o}, we have the following equation:
    \begin{equation} \label{thpf:smooth_case1_part1_basic}
        \begin{aligned}
            & \mathbf{N}_1(z+s^\prime)-L\mathbf{N}_1(z+s^\prime)^2\\
            = & \mathbf{N}_1(z) + \mathbf{J}_1 s^\prime - L\big(\mathbf{N}_1(z) + v\big)^2  , \\
            = & \mathbf{N}_1(z) + \mathbf{J}_1 s^\prime - L\big(\mathbf{N}_1(z)^2 + (\mathbf{J}_1 s^\prime)^2 + 2\mathbf{N}_1(z)\mathbf{J}_1 s^\prime\big) , \\
            = & \underbrace{\mathbf{N}_1(z)- L\mathbf{N}_1(z)^2}_{\textcircled{1}}  +  \underbrace{\big(1 - 2 L \mathbf{N}_1(z)\big)\mathbf{J}_1 s^\prime - L (\mathbf{J}_1 s^\prime)^2}_{\textcircled{2}}.
        \end{aligned}
    \end{equation}

    Term $\textcircled{2}$ in \eqref{thpf:smooth_case1_part1_basic} is a quadratic formula of $v$ with the following range thanks to $v$:
    \begin{equation}
        \begin{aligned}
            0 \leq \textcircled{2} \leq \frac{\big(1-2L\mathbf{N}_1(z)\big)^2}{4L}, &\text{ if } 0 \leq \mathbf{J}_1 s^\prime \leq \frac{1}{L} - 2\mathbf{N}_1(z),\\
            \textcircled{2} < 0, &\text{ if } \mathbf{J}_1 s^\prime < 0 \text{ or } \mathbf{J}_1 s^\prime > \frac{1}{L} - 2\mathbf{N}_1(z).
        \end{aligned}
    \end{equation}

    % \qy{NOTE 2: Convergence gain on inference, convergence of OOD and inD are not comparable.} 
    % \qy{NOTE 3: This are Hadamard product and vector inequalities.}
    Due to \eqref{thpf:smooth_case1_part1_basic}, we have the following OOD convergence gain on shifted data $z+s^\prime$:
    \begin{equation} \label{thpf:smooth_part1_ood_rate}
        \begin{aligned}
            &- \nabla f^\prime(x+s)^\top\Big(\diag\big(\mathbf{N}_1(z+s^\prime)\big)-L\diag\big(\mathbf{N}_1(z+s^\prime)\big)^2\Big)\nabla f^\prime(x+s) \\
            = & -\nabla f^\prime(x+s)^\top \diag\Big(\mathbf{N}_1(z)- L\mathbf{N}_1(z)^2 +  \big(1 - 2 L \mathbf{N}_1(z)\big)\mathbf{J}_1 s^\prime - L (\mathbf{J}_1 s^\prime)^2\Big)\nabla f^\prime(x+s)  , \\
            = & -\nabla f^\prime(x+s)^\top \diag\big(\mathbf{N}_1(z)- L\mathbf{N}_1(z)^2\big)\nabla f^\prime(x+s) - \nabla f^\prime(x+s)^\top \diag\Big(\big(1 - 2 L \mathbf{N}_1(z)\big)\mathbf{J}_1 s^\prime - L (\mathbf{J}_1 s^\prime)^2\Big)\nabla f^\prime(x+s)  , \\ 
            = &-\nabla f^\prime(x+s)^\top \diag\big(\mathbf{N}_1(z)- L\mathbf{N}_1(z)^2\big)\nabla f^\prime(x+s) - \nabla f^\prime(x+s)^\top \diag\Big(\big(1 - 2 L \mathbf{N}_1(z)\big)\mathbf{J}_1 s^\prime - L (\mathbf{J}_1 s^\prime)^2\Big)\nabla f^\prime(x+s).
        \end{aligned}
    \end{equation}

    % \qy{NOTE 3: We'd like to derive an upper bound of gain w.r.t. $\|s\|$.} 
    % Denote the $i$-th row of $\mathbf{J}_{1}$ as $\mathbf{J}_{1 (i,:)}$ and denote the angle between $\mathbf{J}_{1 (i,:)}$ and $s^\prime$ as $\phi_i, i \in \{1,\dots, n\}$. We have 
    % \begin{equation*}
    %     \mathbf{J}_1 s^\prime = [\|\mathbf{J}_{1 (1,:)}\| \|s^\prime\| \cos(\phi_1), \|\mathbf{J}_{1 (2,:)}\| \|s^\prime\| \cos(\phi_2), \dots, \|\mathbf{J}_{1 (n,:)}\| \|s^\prime\| \cos(\phi_n)]^\top.
    % \end{equation*}

    Moreover, we derive the following equation of convergence gain with respect to the L2O model's virtual feature $s^\prime$ (the difference between OOD and InD features):
    \begin{equation} \label{thpf:smooth_part1_gain_lb}
        \begin{aligned}
            & -\nabla f^\prime(x+s)^\top \diag\big(\mathbf{N}_1(z)- L\mathbf{N}_1(z)^2\big)\nabla f^\prime(x+s) -\nabla f^\prime(x+s)^\top \diag\Big(\big(1 - 2 L \mathbf{N}_1(z)\big)\mathbf{J}_1 s^\prime - L \big(\mathbf{J}_1 s^\prime\big)^2\Big)\nabla f^\prime(x+s)\\
            = & -\nabla f^\prime(x+s)^\top \diag\big(\mathbf{N}_1(z)- L\mathbf{N}_1(z)^2\big)\nabla f^\prime(x+s) \\
            &-\nabla f^\prime(x+s)^\top \Big( \diag\big(1 - 2 L \mathbf{N}_1(z)\big)\diag(\mathbf{J}_1 s^\prime) - L \diag\big((\mathbf{J}_1 s^\prime)^\top(\mathbf{J}_1 s^\prime)\big) \Big) \nabla f^\prime(x+s).
        \end{aligned}
    \end{equation}

    The above equation is lower bounded by $-\nabla f^\prime(x+s)^\top \diag\big(\mathbf{N}_1(z)- L\mathbf{N}_1(z)^2\big)\nabla f^\prime(x+s) + \frac{1}{4L}\nabla f^\prime(x+s)^\top\diag\Big(\big(1-2L\mathbf{N}_1(z)\big)^2\Big)\nabla f^\prime(x+s)$ when $\mathbf{J}_1 s^\prime:=\frac{1}{2L} - \mathbf{N}_1(z)$. 

    Moreover, if $\mathbf{N}_1(z) := \frac{1}{2L}, \forall z \in \mathcal{Z}_P$, which means the best robutstness is achieved on the training set, the convergence gain compared with InD decreases to:
    \begin{equation} \label{thpf:smooth_part1_gain_lb_best_train}
        \begin{aligned}
            & -\nabla f^\prime(x+s)^\top \diag\left(\frac{1}{2L}- L\frac{1}{2L}^2\right)\nabla f^\prime(x+s) \\
            & -\nabla f^\prime(x+s)^\top \big( \diag(1 - 2 L \frac{1}{2L})\diag(\mathbf{J}_1 s^\prime) - L \diag(\mathbf{J}_1 s^\prime)^\top\diag(\mathbf{J}_1 s^\prime) \big) \nabla f^\prime(x+s) \\
            =& -\frac{\|\nabla f^\prime(x+s)\|^2}{4L} + L\nabla f^\prime(x+s)^\top \diag(\mathbf{J}_1 s^\prime)^\top\diag(\mathbf{J}_1 s^\prime) \nabla f^\prime(x+s)  , \\
            =& -\frac{\|\nabla f^\prime(x+s)\|^2}{4L} + L \big\|\diag(\mathbf{J}_1 s^\prime) \nabla f^\prime(x+s)\big\|^2.
        \end{aligned}
    \end{equation}
    The above result demonstrates that any non-zero $\mathbf{J}_1 s^\prime$ leads to worse convergence gain, which implies that a more well-training L2O model leads to less generalization ability. Only inadequately trained L2O models can achieve increments of convergence.

    % \qy{Insight 2 in Part 1: More well-training L2O model, less generalization ability.}

    % \qy{Insight 3 in Part 1: Only when training is not well-done, OOD is possible to leads to some potential increment of convergence.}

    \paragraph{Convergence Gain by $\mathbf{N}_2(z)$.} 

    Assume \eqref{thpf:smooth_N2_bound1} holds in training: 
    % Denote the $i$-th row of $\mathbf{J}_{2}$ as $\mathbf{J}_{2 (i,:)}$ and denote the angle between $\mathbf{J}_{2 (i,:)}$ and $s^\prime$ as $\theta_i, i \in \{1,\dots, n\}$.
    \begin{equation} \label{thpf:smooth_part2_gain_lb}
        \begin{aligned}
            & - \nabla f^\prime(x+s)^\top\mathbf{N}_2(z+s^\prime) + L\big\|\mathbf{N}_2(z+s^\prime)\big\|^2 \\
            = & -\nabla f^\prime(x+s)^\top \big(\mathbf{N}_2(z) + \mathbf{J}_2 s^\prime\big) + L \big\|\mathbf{N}_2(z) + \mathbf{J}_2 s^\prime\big\|^2  , \\
            = & -\nabla f^\prime(x+s)^\top \mathbf{N}_2(z) + L\big\|\mathbf{N}_2(z)\big\|^2 - \nabla f^\prime(x+s)^\top \mathbf{J}_2 s^\prime + L \|\mathbf{J}_2 s^\prime\|^2 + 2L \mathbf{N}_2(z)^\top \mathbf{J}_2 s^\prime , \\
            = & -\nabla f^\prime(x+s)^\top \mathbf{N}_2(z) + L\big\|\mathbf{N}_2(z)\big\|^2 - \nabla f^\prime(x+s)^\top \mathbf{J}_2 s^\prime + L \|\mathbf{J}_2 s^\prime\|^2 + 2L \mathbf{N}_2(z)^\top \mathbf{J}_2 s^\prime  , \\
            = & -\nabla f^\prime(x+s)^\top \mathbf{N}_2(z) + L\big\|\mathbf{N}_2(z)\big\|^2 - \big(\nabla f^\prime(x+s)-2L \mathbf{N}_2(z)\big)^\top \mathbf{J}_2 s^\prime + L \|\mathbf{J}_2 s^\prime\|^2.
        \end{aligned}
    \end{equation}

    If we further assume training leads to best convergence gain, i.e., $\mathbf{N}_2(z) = \frac{\nabla f(x)}{2L}, \forall z \in \mathcal{Z}_P$, which means best convergence gain achieved on a training set, we have the following per iteration convergence gain in the OOD scenario from $\mathbf{N}_2(z)$:
    \begin{equation} \label{thpf:smooth_part2_gain_lb_best_train}
        \begin{aligned}
            & -\nabla f^\prime(x+s)^\top \mathbf{N}_2(z) + L\big\|\mathbf{N}_2(z)\big\|^2 - \big(\nabla f^\prime(x+s)-2L \mathbf{N}_2(z)\big)^\top \mathbf{J}_2 s^\prime + L \|\mathbf{J}_2 s^\prime\|^2  \\
            =& - \frac{\nabla f^\prime(x+s)^\top\nabla f(x)}{2L} + \frac{\|\nabla f(x)\|^2}{4L} - \left(\nabla f^\prime(x+s)-2L \frac{\nabla f(x)}{2L} \right)^\top \mathbf{J}_2 s^\prime + L \|\mathbf{J}_2 s^\prime\|^2  , \\
            =& - \frac{\nabla f^\prime(x+s)^\top\nabla f(x)}{2L} + \frac{\|\nabla f(x)\|^2}{4L}  \\
            & + L \left(-\frac{\big(\nabla f^\prime(x+s)-\nabla f(x)\big)^\top}{L} \mathbf{J}_2 s^\prime +  \|\mathbf{J}_2 s^\prime\|^2 + \frac{\big\|\nabla f^\prime(x+s)-\nabla f(x)\big\|^2}{4L^2} - \frac{\big\|\nabla f^\prime(x+s)-\nabla f(x)\big\|^2}{4L^2} \right)  , \\
            =& - \frac{\nabla f^\prime(x+s)^\top\nabla f(x)}{2L} + \frac{\|\nabla f(x)\|^2}{4L} - \frac{\|\nabla f^\prime(x+s)-\nabla f(x)\|^2}{4L} + L \left\|\frac{\nabla f^\prime(x+s)-\nabla f(x)}{2L} - \mathbf{J}_2 s^\prime \right\|^2  , \\
            = & - \frac{\|\nabla f^\prime(x+s)\|^2}{4L} + L \left\|\frac{\nabla f^\prime(x+s)-\nabla f(x)}{2L} - \mathbf{J}_2 s^\prime \right\|^2.
        \end{aligned}
    \end{equation}

    \paragraph{Overall Convergence Gain of One Iteration.}
    Sum up \eqref{thpf:smooth_part1_gain_lb_best_train} and \eqref{thpf:smooth_part2_gain_lb_best_train}, we have the following OOD's integrated convergence gain of one iteration:
    \begin{equation}\label{thpf:smooth_integrated_gain_lb_best_train}
        -\frac{\big\|\nabla f^\prime(x+s)\big\|^2}{2L} + \underbrace{L\big\|\diag(\mathbf{J}_1 s^\prime) \nabla f^\prime(x+s)\big\|^2 + L \left\|\frac{\nabla f^\prime(x+s)-\nabla f(x)}{2L} - \mathbf{J}_2 s^\prime \right\|^2}_{\textcircled{3}}.
    \end{equation}

    $-\frac{\big\|\nabla f^\prime(x+s)\big\|^2}{2L}$ is equivalent to the convergence rate of gradient descent, which is also the most robust convergence rate that the L2O model could achieve in the InD scenario. 
    Moreover, we would like the value of the above equation to be as small as possible. However, the non-negativity of $\textcircled{3}$ shows that the convergence gain deteriorates as long as OOD happens.
\end{proof}

\subsection{Proof of Corollary \ref{lm:lemma2_smooth_converge_gain_ub}} \label{sec:lm2_proof}
% : Convergence Gain Upper Bound w.r.t. $\|s^\prime \|$
\begin{proof}
    In this proof, we formulate the upper bound of per iteration convergence gain with respect to the L2O model input feature vectors.

    Based on Triangle and Cauchy-Schwarz inequalities, we have:
    \begin{equation*}
        \begin{aligned}
            & -\frac{\big\|\nabla f^\prime(x+s)\big\|^2}{2L} + L\big\|\diag(\mathbf{J}_1 s^\prime) \nabla f^\prime(x+s)\big\|^2 + L \left\|\frac{\nabla f^\prime(x+s)-\nabla f(x)}{2L} - \mathbf{J}_2 s^\prime \right\|^2 \\
            =& -\frac{\big\|\nabla f^\prime(x+s)\big\|^2}{2L} + L\big\|\diag(\mathbf{J}_1 s^\prime) \nabla f^\prime(x+s)\big\|^2 + L \left\|\frac{\nabla f^\prime(x+s)-\nabla f(x)}{2L} - \mathbf{J}_2 s^\prime \right\|^2 , \\
            \leq & -\frac{\big\|\nabla f^\prime(x+s)\big\|^2}{2L} + L\big\|\diag(\mathbf{J}_1 s^\prime) \nabla f^\prime(x+s)\big\|^2 + 2L \left\|\frac{\nabla f^\prime(x+s)-\nabla f(x)}{2L}\|^2 + 2L \| \mathbf{J}_2 s^\prime \right\|^2  , \\
            = & -\frac{\big\|\nabla f^\prime(x+s)\big\|^2}{2L} + \frac{\|\nabla f^\prime(x+s)-\nabla f(x)\|^2}{2L} + L \big\|\diag(\mathbf{J}_1 s^\prime) \nabla f^\prime(x+s)\big\|^2 + 2L \| \mathbf{J}_2 s^\prime \|^2 , \\
            \leq & -\frac{\big\|\nabla f^\prime(x+s)\big\|^2}{2L} + \frac{\|\nabla f^\prime(x+s)-\nabla f(x)\|^2}{2L} + L \big\|\diag(\mathbf{J}_1 s^\prime) \nabla f^\prime(x+s)\big\|^2 + 2L \| \mathbf{J}_2\|^2 \|s^\prime \|^2.
        \end{aligned}
    \end{equation*}

    Due to $\|\mathbf{J}_1 \| \leq C_1\sqrt{n}$ and $\|\mathbf{J}_2 \| \leq C_2\sqrt{n}$, the above inequality is upper bounded by:
    \begin{equation} \label{thpf:smooth_integrated_gain_lb_best_train_upper_bound}
        \begin{aligned}
            & -\frac{\big\|\nabla f^\prime(x+s)\big\|^2}{2L} + \frac{\|\nabla f^\prime(x+s)-\nabla f(x)\|^2}{2L} + L \big\|\diag(\mathbf{J}_1 s^\prime) \nabla f^\prime(x+s)\big\|^2 +  2L \| \mathbf{J}_2\|^2 \|s^\prime \|^2 , \\
            \leq & -\frac{\big\|\nabla f^\prime(x+s)\big\|^2}{2L} + \frac{\|\nabla f^\prime(x+s)-\nabla f(x)\|^2}{2L} + L C_1^2 n \|s^\prime\|^2 \nabla f^\prime(x+s)^\top\mathbf{I}\nabla f^\prime(x+s) + 2L C_2^2 n \|s^\prime \|^2 , \\
            = & -\frac{\big\|\nabla f^\prime(x+s)\big\|^2}{2L} + \frac{\|\nabla f^\prime(x+s)-\nabla f(x)\|^2}{2L} + \big(L C_1^2 n \|\nabla f^\prime(x+s)\|^2  + 2L C_2^2n \big) \|s^\prime \|^2.
        \end{aligned}
    \end{equation}

    Due to the formulation of $\|s^\prime\|$ in \eqref{thpf:lb_shift_norm}, if we set $s^\prime := \nabla f^\prime(x+s) - \nabla f(x)$, which means we remove the variable feature and only use gradient as the input feature of the L2O model, we can decrease such convergence gain's upper bound from the right-hand side of inequation \ref{thpf:smooth_integrated_gain_lb_best_train} to:
    \begin{equation} \label{thpf:smooth_integrated_gain_lb_best_train_upper_bound_withoutX}
        -\frac{\big\|\nabla f^\prime(x+s)\big\|^2}{2L} + \frac{\|\nabla f^\prime(x+s)-\nabla f(x)\|^2}{2L} + \big(L C_1^2 n \|\nabla f^\prime(x+s)\|^2  + 2L C_2^2\big) \|\nabla f^\prime(x+s) - \nabla f(x)\|^2,
    \end{equation}
    where we just replace $\|s^\prime\|$ in inequation \ref{thpf:smooth_integrated_gain_lb_best_train} with $\|\nabla f^\prime(x+s) - \nabla f(x)\|$.

\end{proof}

\subsection{Proof of Theorem \ref{th:theorem2_smooth_converge_rate}} \label{sec:th2_proof}

\begin{proof}

    In \cref{sec:vv_traj}, we split the trajectory of OOD variable $x^\prime$ into a trajectory of InD variable $x$ and a trajectory of the virtual variable $s$. $x$ is updated with ``well-trained'' L2O with robustness guarantee (Corollary \ref{lm:lemma2_smooth_converge_gain_ub}), which is independent of OOD and deterministically performs as gradient descent. Thus, the uncertainty of the OOD scenario can be formulated with respect to the virtual variable $s$. This proof constructs the formulation of multi-iteration convergence rate with respect to $s$.

    First, we assume Corollary \ref{lm:lemma2_smooth_converge_gain_ub} hold, which ensures the L2O model's robust performance on InD variable $x$. Upon \eqref{thpf:mvt_l2o}, we have following equalities between the L2O model's outputs of ODD and InD scenarios:
    \begin{equation*}
        \begin{aligned}
            &\mathbf{N}_1(z+s^\prime) = \frac{1}{2L} + \mathbf{J}_1 s^\prime, \\
            &\mathbf{N}_2(z+s^\prime) = \frac{\nabla f(x)}{2L} + \mathbf{J}_2 s^\prime,
        \end{aligned}
    \end{equation*}
    where as defined in \cref{sec:vv_traj}, $z+s^\prime$ and $z$ represent L2O model's input features of OOD and InD scenarios respectively. And $s^\prime$ formulates the difference between them. 

    In $k$-th iteration, $k \geq 1$, based on the above equations, we can represent the L2O model's update formula in the OOD scenario as follows:
    \begin{equation}
        \begin{aligned}
            x_k+s_k &= x_{k-1} + s_{k-1} - \mathbf{N}_1(z_{k-1}+s^\prime_{k-1}) \nabla f^\prime(x_{k-1}+s_{k-1}) - \mathbf{N}_2(z_{k-1}+s^\prime_{k-1}),\\
            &= x_{k-1} + s_{k-1} - \left(\frac{1}{2L} - \diag(\mathbf{J}_{1,k-1} s^\prime_{k-1})\right) \nabla f^\prime(x_{k-1}+s_{k-1}) - \left(\frac{\nabla f(x_{k-1})}{2L} + \mathbf{J}_{2,k-1} s^\prime_{k-1}\right).
        \end{aligned}
    \end{equation}

    From the above equation, we want to split the InD part on the InD variable $x$ and the OOD part on the virtual variable $s$. 
    By adding an entra term $\frac{\nabla f(x_{k-1})}{L}$, we have the following reformulations:
    \begin{equation} \label{thpf:ood_update}
        \begin{aligned}
            x_k+s_k = & x_{k-1} - \frac{\nabla f(x_{k-1})}{L} + s_{k-1} - \frac{\nabla f^\prime(x_{k-1}+s_{k-1})}{2L} + \frac{\nabla f(x_{k-1})}{2L} - \diag(\mathbf{J}_{1,k-1} s_{k-1}^\prime) \nabla f^\prime(x_{k-1}+s_{k-1}) \\
            & - \mathbf{J}_{2,k-1} s_{k-1}^\prime , \\
            = & x_{k-1} - \frac{\nabla f(x_{k-1})}{L} + s_{k-1} - \frac{\nabla f^\prime(x_{k-1}+s_{k-1}) - \nabla f(x_{k-1})}{2L} - \diag(\mathbf{J}_{1,k-1} s_{k-1}^\prime) \nabla f^\prime(x_{k-1}+s_{k-1}) \\
            & - \mathbf{J}_{2,k-1} s_{k-1}^\prime .
        \end{aligned}
    \end{equation}

    Then, we can split the OOD trajectory of $x^\prime = x + s$ into two parts: InD and OOD parts on $x$ and $s$. First, we assume the InD variable $x$ is updated with the following update formula with the gradient of InD objective $f(x_{k-1})$:
    \begin{equation} \label{thpf:ood_update_indx}
        x_k = x_{k-1} - \frac{\nabla f(x_{k-1})}{L},
    \end{equation}
    For the $x$ part, with an unchanged InD initial point $x_0 \in \mathcal{S}_P$ and an InD objective $f \in \mathcal{F}_{L,P}$ is solutions given by the L2O model are always in the InD scenario. The optimal solution $x^*$ of $f$ is deterministic as well. 
    
    Moreover, removing \eqref{thpf:ood_update_indx}, the remaining terms of \eqref{thpf:ood_update} constitutes the update formula on virtual variable $s$:
    \begin{equation*}
        s_k =  s_{k-1} - \frac{\nabla f^\prime(x_{k-1}+s_{k-1}) - \nabla f(x_{k-1})}{2L} - \diag(\mathbf{J}_1 s_{k-1}^\prime) \nabla f^\prime(x_{k-1}+s_{k-1}) - \mathbf{J}_2 s_{k-1}^\prime.
    \end{equation*}

    Now is the time to derive the convergence rate. 
    Based on the definition of $L$-smoothness on $f^\prime$, in $k$-th iteration, we have the following upper bound of OOD objective $f^\prime(x_k+s_k)$:
    \begin{equation*}
        \begin{aligned}
            &F^\prime(x_k+s_k)\\
             \leq & F^\prime(x_{k-1}+s_{k-1}) + \nabla f^\prime(x_{k-1}+s_{k-1})^\top\big(x_k+s_k - (x_{k-1}+s_{k-1})\big) + \frac{L}{2}\left\|x_k+s_k - (x_{k-1}+s_{k-1})\right\|^2 , \\
            =& F^\prime(x_{k-1}+s_{k-1}) \\
            &  + \nabla f^\prime(x_{k-1}+s_{k-1})^\top\left(- \frac{\nabla f(x_{k-1})}{L} - \frac{\nabla f^\prime(x_{k-1}+s_{k-1}) - \nabla f(x_{k-1})}{2L} - \diag(\mathbf{J}_1 s_{k-1}^\prime) \nabla f^\prime(x_{k-1}+s_{k-1}) - \mathbf{J}_2 s_{k-1}^\prime\right) \\
            & + \frac{L}{2}\left\|- \frac{\nabla f(x_{k-1})}{L} - \frac{\nabla f^\prime(x_{k-1}+s_{k-1}) - \nabla f(x_{k-1})}{2L} - \diag(\mathbf{J}_1 s_{k-1}^\prime) \nabla f^\prime(x_{k-1}+s_{k-1}) - \mathbf{J}_2 s_{k-1}^\prime\right\|^2 , \\
            \leq & F^\prime(x^*+s^*) + \nabla f^\prime(x_{k-1}+s_{k-1})^\top\big(x_{k-1}+s_{k-1} - (x^*+s^*)\big) \\
            &  + \nabla f^\prime(x_{k-1}+s_{k-1})^\top\left(- \frac{\nabla f(x_{k-1})}{L} - \frac{\nabla f^\prime(x_{k-1}+s_{k-1}) - \nabla f(x_{k-1})}{2L} - \diag(\mathbf{J}_1 s_{k-1}^\prime) \nabla f^\prime(x_{k-1}+s_{k-1}) - \mathbf{J}_2 s_{k-1}^\prime\right) \\
            & + \frac{L}{2}\left\| - \frac{\nabla f(x_{k-1})}{L} - \frac{\nabla f^\prime(x_{k-1}+s_{k-1}) - \nabla f(x_{k-1})}{2L} - \diag(\mathbf{J}_1 s_{k-1}^\prime) \nabla f^\prime(x_{k-1}+s_{k-1}) - \mathbf{J}_2 s_{k-1}^\prime \right\|^2,
        \end{aligned}
    \end{equation*}
    where in the second step, we import the L2O model's in OOD update process defined in \eqref{thpf:ood_update}. In the third step, we apply the definition of convexity on $F^\prime$.

    We make the following reformulation and denote the update on virtual variable $s$ as $\Delta s_{k-1}$:
    \begin{equation*}
        \begin{aligned}
            &F^\prime(x_k+s_k) - F^\prime(x^*+s^*) \\
                \leq &  \nabla f^\prime (x_{k-1}+s_{k-1})^\top\big(x_{k-1}+s_{k-1} - (x^*+s^*)\big) \\
            &  + \nabla f^\prime (x_{k-1}+s_{k-1})^\top\left(- \frac{\nabla f(x_{k-1})}{L} - \frac{\nabla f^\prime(x_{k-1}+s_{k-1}) - \nabla f(x_{k-1})}{2L} - \diag(\mathbf{J}_1 s_{k-1}^\prime) \nabla f^\prime(x_{k-1}+s_{k-1}) - \mathbf{J}_2 s_{k-1}^\prime\right) \\
            & + \frac{L}{2}\left\| - \frac{\nabla f(x_{k-1})}{L} - \frac{\nabla f^\prime(x_{k-1}+s_{k-1}) - \nabla f(x_{k-1})}{2L} - \diag(\mathbf{J}_1 s_{k-1}^\prime) \nabla f^\prime(x_{k-1}+s_{k-1}) - \mathbf{J}_2 s_{k-1}^\prime \right\|^2, \\
                = &  \nabla f^\prime(s_{k-1}+x_{k-1})^\top(s_{k-1}+x_{k-1}-(s^*+x^*)) \\
            &  + \nabla f^\prime(s_{k-1}+x_{k-1})^\top \\
            & \quad \Bigg(x_{k-1} - \frac{\nabla f(x_{k-1})}{L} -x^* -\big((s_{k-1}+x_{k-1}) - (s^*+x^*)\big) \\
            & \qquad + s_{k-1} - \underbrace{\big(\frac{\nabla f^\prime(x_{k-1}+s_{k-1}) - \nabla f(x_{k-1})}{2L} + \diag(\mathbf{J}_1 s_{k-1}^\prime) \nabla f^\prime(x_{k-1}+s_{k-1}) + \mathbf{J}_2 s_{k-1}^\prime\big)}_{\Delta s_{k-1}} -s^* \Bigg) \\
            & + \frac{L}{2}\Bigg\| x_{k-1} - \frac{\nabla f(x_{k-1})}{L} -x^* -\big((s_{k-1}+x_{k-1}) - (s^*+x^*)\big)\\
            & \qquad + s_{k-1}- \underbrace{\frac{\nabla f^\prime(x_{k-1}+s_{k-1}) - \nabla f(x_{k-1})}{2L} + \diag(\mathbf{J}_1 s_{k-1}^\prime) \nabla f^\prime(x_{k-1}+s_{k-1}) + \mathbf{J}_2 s_{k-1}^\prime }_{\Delta s_{k-1}}-s^*  \Bigg\|^2,
        \end{aligned}
    \end{equation*}
    where we place right-hand side items according to whether they are updates for $x_{k-1}$ or $s_{k-1}$. Then, we can simplify the above inequation with $\Delta s_{k-1}$ as follows:
    \begin{equation*}
        \begin{aligned}
            &F^\prime(x_k+s_k) - F^\prime(x^*+s^*) \\
                \leq  &  \nabla f^\prime(s_{k-1}+x_{k-1})^\top(s_{k-1}+x_{k-1}-(s^*+x^*)) \\
            &  + \nabla f^\prime(s_{k-1}+x_{k-1})^\top\Big(x_{k-1} - \frac{\nabla f(x_{k-1})}{L} -x^* + s_{k-1} - \Delta s_{k-1} -s^* -\big((s_{k-1}+x_{k-1}) - (s^*+x^*)\big)\Big) \\
            & + \frac{L}{2}\Big\| x_{k-1} - \frac{\nabla f(x_{k-1})}{L} -x^*  + s_{k-1}- \Delta s_{k-1}-s^* -\big((s_{k-1}+x_{k-1}) - (s^*+x^*)\big)\Big\|^2, \\
                = & \nabla f^\prime(s_{k-1}+x_{k-1})^\top(s_{k-1}+x_{k-1}-(s^*+x^*)) - \nabla f^\prime(s_{k-1}+x_{k-1})^\top\big(s_{k-1}+x_{k-1} - (s^*+x^*)\big)\\
            &  + \nabla f^\prime(s_{k-1}+x_{k-1})^\top\Big(x_{k-1} - \frac{\nabla f(x_{k-1})}{L} -x^* + s_{k-1} - \Delta s_{k-1} -s^* \Big) \\
            & + \frac{L}{2}\Big\| x_{k-1} - \frac{\nabla f(x_{k-1})}{L} -x^*  + s_{k-1}- \Delta s_{k-1}-s^* -\big((s_{k-1}+x_{k-1}) - (s^*+x^*)\big)\Big\|^2 , \\
            = & \nabla f^\prime(s_{k-1}+x_{k-1})^\top\Big(x_{k-1} - \frac{\nabla f(x_{k-1})}{L} -x^* + s_{k-1} - \Delta s_{k-1} -s^* \Big) \\
            & + \frac{L}{2}\Big\| x_{k-1} - \frac{\nabla f(x_{k-1})}{L} -x^*  + s_{k-1}- \Delta s_{k-1}-s^* -\big((s_{k-1}+x_{k-1}) - (s^*+x^*)\big)\Big\|^2,
        \end{aligned}
    \end{equation*}
    where in the second and third steps, we combine the similar terms of the right-hand side. 

    We continue by expanding the quadratic formula of the right-hand side and making up new ones as follows:
    \begin{equation*}
        \begin{aligned}
                & \nabla f^\prime(s_{k-1}+x_{k-1})^\top\left(x_{k-1} - \frac{\nabla f(x_{k-1})}{L} -x^*\right) + \nabla f^\prime(s_{k-1}+x_{k-1})^\top(s_{k-1} -\Delta s_{k-1} -s^*) \\
            & + \frac{L}{2}\Big\| x_{k-1}-\frac{\nabla f(x_{k-1})}{L} -x^*  +  s_{k-1}- \Delta s_{k-1} -s^*  -\big((s_{k-1}+x_{k-1}) - (s^*+x^*)\big)\Big\|^2 \\
                = & \Big(\nabla f^\prime(s_{k-1}+x_{k-1}) - L\big((s_{k-1}+x_{k-1}) - (s^*+x^*)\big)\Big)^\top\left(x_{k-1} - \frac{\nabla f(x_{k-1})}{L} -x^* + s_{k-1} -\Delta s_{k-1} -s^*\right) \\
            & + \frac{L}{2}\left\| x_{k-1}-\frac{\nabla f(x_{k-1})}{L} -x^* + s_{k-1}- \Delta s_{k-1} -s^*\right\|^2 + \frac{L}{2}\|(s_{k-1}+x_{k-1}) - (s^*+x^*)\|^2 , \\
                = & \frac{L}{2}\Bigg(2\Big(\frac{\nabla f^\prime(s_{k-1}+x_{k-1})}{L} - \big((s_{k-1}+x_{k-1}) - (s^*+x^*)\big)\Big)^\top\left(x_{k-1} - \frac{\nabla f(x_{k-1})}{L} -x^* + s_{k-1} -\Delta s_{k-1} -s^*\right) \\
            & \qquad + \| x_{k-1}-\frac{\nabla f(x_{k-1})}{L} -x^* + s_{k-1}- \Delta s_{k-1} -s^*\|^2\Bigg) + \frac{L}{2}\|(s_{k-1}+x_{k-1}) - (s^*+x^*)\|^2 , \\
                = & \frac{L}{2}\left(\left\| \frac{\nabla f^\prime(s_{k-1}+x_{k-1})}{L} -\frac{\nabla f(x_{k-1})}{L}  - \Delta s_{k-1} \right\|^2 - \left\|\frac{\nabla f^\prime(s_{k-1}+x_{k-1})}{L} - \big((s_{k-1}+x_{k-1}) - (s^*+x^*)\big)\right\|^2\right) \\
            &  + \frac{L}{2}\|(s_{k-1}+x_{k-1}) - (s^*+x^*)\|^2 ,
        \end{aligned}
    \end{equation*}
    where we expand a quadratic formula in the first step and make up a new quadratic formula in the second and thrid steps.

    We add an extra term $\frac{L}{2}\|(s_k+x_k) - (s^*+x^*)\|^2 -  \frac{L}{2}\|(s_k+x_k) - (s^*+x^*)\|^2$ to expand the second quadratic formular:
    \begin{equation*}
        \begin{aligned}
                & \frac{L}{2}\left(\left\| \frac{\nabla f^\prime(s_{k-1}+x_{k-1})}{L} -\frac{\nabla f(x_{k-1})}{L}  - \Delta s_{k-1} \right\|^2 - \left\|\frac{\nabla f^\prime(s_{k-1}+x_{k-1})}{L} - \big((s_{k-1}+x_{k-1}) - (s^*+x^*)\big)\right\|^2\right) \\
            &  + \frac{L}{2}\|(s_{k-1}+x_{k-1}) - (s^*+x^*)\|^2 +  \frac{L}{2}\|(s_k+x_k) - (s^*+x^*)\|^2 -  \frac{L}{2}\|(s_k+x_k) - (s^*+x^*)\|^2\\
                = & \frac{L}{2}\Bigg(\left\| \frac{\nabla f^\prime(s_{k-1}+x_{k-1})}{L} -\frac{\nabla f(x_{k-1})}{L}  - \Delta s_{k-1} \right\|^2  \\
            & \qquad - \left\|\frac{\nabla f^\prime(s_{k-1}+x_{k-1})}{L} - \big((s_{k-1}+x_{k-1}) - (s^*+x^*)\big)\right\|^2 + \|(s_k+x_k) - (s^*+x^*)\|^2 \Bigg) \\
            &  + \frac{L}{2}\|(s_{k-1}+x_{k-1}) - (s^*+x^*)\|^2 - \frac{L}{2}\|(s_k+x_k) - (s^*+x^*)\|^2 , \\
                = & \frac{L}{2}\Bigg(\left\| \frac{\nabla f^\prime(s_{k-1}+x_{k-1})}{L} -\frac{\nabla f(x_{k-1})}{L}  - \Delta s_{k-1} \right\|^2 + \frac{L}{2}\|(s_{k-1}+x_{k-1}) - (s^*+x^*)\|^2 - \frac{L}{2}\|(s_k+x_k) - (s^*+x^*)\|^2 \\
            & + \left((s_k+x_k) - (s_{k-1}+x_{k-1}) + \frac{\nabla f^\prime(s_{k-1}+x_{k-1})}{L}\right)^\top \\
            & \quad \left((s_k+s_{k-1}+x_k+x_{k-1}) - 2(s^*+x^*) - \frac{\nabla f^\prime(s_{k-1}+x_{k-1})}{L} \right) \Bigg).
        \end{aligned}
    \end{equation*}

    Based on the definition of $x$ and $s$ updates, we can further combine the first and second terms in the above formulation after expanding the first quadratic formula:
    \begin{equation*}
        \begin{aligned}
                & \frac{L}{2}\Bigg(\left\| \frac{\nabla f^\prime(s_{k-1}+x_{k-1})}{L} -\frac{\nabla f(x_{k-1})}{L}  - \Delta s_{k-1} \right\|^2  \\
            & + \left((s_k+x_k) - (s_{k-1}+x_{k-1}) + \frac{\nabla f^\prime(s_{k-1}+x_{k-1})}{L}\right)^\top \\
            & \qquad \left((s_k+s_{k-1}+x_k+x_{k-1}) - 2(s^*+x^*) - \frac{\nabla f^\prime(s_{k-1}+x_{k-1})}{L} \right) \Bigg) \\
            &  + \frac{L}{2}\|(s_{k-1}+x_{k-1}) - (s^*+x^*)\|^2 - \frac{L}{2}\|(s_k+x_k) - (s^*+x^*)\|^2 , \\
                = & \frac{L}{2}\Bigg(  \left( \frac{\nabla f^\prime(s_{k-1}+x_{k-1})}{L} -\frac{\nabla f(x_{k-1})}{L}  - \Delta s_{k-1} \right)^\top\Big(2(s_{k-1}+x_{k-1}) - 2(s^*+x^*) \\
                & \qquad - \frac{\nabla f^\prime(s_{k-1}+x_{k-1})}{L} -\frac{\nabla f(x_{k-1})}{L}  - \Delta s_{k-1}  + \frac{\nabla f^\prime(s_{k-1}+x_{k-1})}{L} -\frac{\nabla f(x_{k-1})}{L}  - \Delta s_{k-1} \Big)\Bigg) \\
            &  + \frac{L}{2}\|(s_{k-1}+x_{k-1}) - (s^*+x^*)\|^2 - \frac{L}{2}\|(s_k+x_k) - (s^*+x^*)\|^2 , \\
                = & \frac{L}{2}  \Bigg( \left( \frac{\nabla f^\prime(s_{k-1}+x_{k-1})}{L} -\frac{\nabla f(x_{k-1})}{L}  - \Delta s_{k-1} \right)^\top(2(s_{k-1}+x_{k-1}) - 2(s^*+x^*) - 2\frac{\nabla f(x_{k-1})}{L}  - 2\Delta s_{k-1} )\Bigg) \\
            &  + \frac{L}{2}\|(s_{k-1}+x_{k-1}) - (s^*+x^*)\|^2 - \frac{L}{2}\|(s_k+x_k) - (s^*+x^*)\|^2 ,\\
                = & L\left( \frac{\nabla f^\prime(s_{k-1}+x_{k-1})}{L} + x_k - x_{k-1} + s_k - s_{k-1} \right)^\top\big( (s_k+x_k) - (s^*+x^*) \big) \\
            &  + \frac{L}{2}\|(s_{k-1}+x_{k-1}) - (s^*+x^*)\|^2 - \frac{L}{2}\|(s_k+x_k) - (s^*+x^*)\|^2.
        \end{aligned}
    \end{equation*}

    We subsitute $\Delta s_{k-1}$ back and sum over $K$ iterations to get final multi-iteration convergence rate:
    \begin{equation}
        \begin{aligned}
            & \sum_{k=1}^{K} F^\prime(x_k+s_k) - F^\prime(x^*+s^*) \\
            \leq & L \sum_{k=1}^{K} \left( \frac{\nabla f^\prime(s_{k-1}+x_{k-1})}{L} -\frac{\nabla f(x_{k-1})}{L}  - \Delta s_{k-1} \right)^\top( (s_k+x_k) - (s^*+x^*) ) \\
            &  + \frac{L}{2} \sum_{k=1}^{K} \|(s_{k-1}+x_{k-1}) - (s^*+x^*)\|^2 - \frac{L}{2}\|(s_k+x_k) - (s^*+x^*)\|^2 ,\\
            = & L \sum_{k=1}^{K} \left( \frac{\nabla f^\prime(s_{k-1}+x_{k-1})}{L} -\frac{\nabla f(x_{k-1})}{L}  - \Delta s_{k-1} \right)^\top( x_k-x^* + s_k- s^* ) \\
            & + \frac{L}{2}(\|x_0 -x^* + s_0- s^*\|^2  - \|x_K -x^* + s_K- s^*\|^2 ), \\
            = & L \sum_{k=1}^{K} \left(\frac{\nabla f^\prime(x_{k-1}+s_{k-1}) - \nabla f(x_{k-1})}{2L} - \diag(\mathbf{J}_1 s_{k-1}^\prime) \nabla f^\prime(x_{k-1}+s_{k-1}) - \mathbf{J}_2 s_{k-1}^\prime\right)^\top( x_k-x^* + s_k- s^* ) \\
            & + \frac{L}{2}(\|x_0 -x^* + s_0- s^*\|^2  - \|x_K -x^* + s_K- s^*\|^2 ), \\
            = & L \sum_{k=1}^{K} \left( \frac{\nabla f^\prime(s_{k-1}+x_{k-1})}{L} + x_k - x_{k-1} + s_k - s_{k-1} \right)^\top( x_k-x^* + s_k- s^* ) \\
            & + \frac{L}{2}(\|x_0 -x^* + s_0- s^*\|^2  - \|x_K -x^* + s_K- s^*\|^2 ).
        \end{aligned}
    \end{equation}

    Since we have demonstrated that there may be no convergence guarantees in each iteration, we cannot directly split $F^\prime(x_K+s_K) - F^\prime(x^*+s^*)$ from the left-hand side of the above inequalities, we denote that $\min_{k=1,\dots, K}F^\prime(x_k+s_k) - F^\prime(x^*+s^*)$ is minimal over $F^\prime(x_j+s_j) - F^\prime(x^*+s^*), j=1,\dots,K$, which is a scenario that leads to the best convergence rate upper bound. Without loss of generality, we can always find a $k$ that minimizes all $F^\prime(x_k+s_k) - F^\prime(x^*+s^*), k \in [1, K]$.

    After rearrangement, we have the following two equivalent expressions:
    \begin{equation} \label{thpf:theorem1_smooth_converge_rate}
        \begin{aligned}
            & \min_{k=1,\dots, K}F^\prime(x_k+s_k) - F^\prime(x^*+s^*) \\
                \leq & \frac{L}{2}\|x_0 -x^* + s_0- s^*\|^2  - \frac{L}{2}\|x_K -x^* + s_K- s^*\|^2 \\
                &+ \frac{1}{K} \sum_{k=1}^{K} \nabla f^\prime(x_{k-1}+s_{k-1})^\top( x_k+ s_k -x^* -s^* ) + \frac{L}{K} \sum_{k=1}^{K} (x_k +s_k -x_{k-1} -s_{k-1} )^\top( x_k+ s_k -x^* -s^* ), \\
            = & \frac{L}{2}\|x_0 -x^* + s_0- s^*\|^2  - \frac{L}{2}\|x_K -x^* + s_K- s^*\|^2 \\
            & + \frac{L}{K} \sum_{k=1}^{K} (x_k+ s_k -x^* -s^*)^\top (x_k +s_k - (x_{k-1} +s_{k-1} - \frac{\nabla f^\prime(x_{k-1}+s_{k-1})}{L})).
        \end{aligned}
    \end{equation}
    % \underline{Theorem~\ref{th:theorem2_smooth_converge_rate} has been proved.}
    \begin{equation}
        \begin{aligned}
            & \min_{k=1,\dots, K}F^\prime(x_k+s_k) - F^\prime(x^*+s^*) \\
                \leq & \frac{L}{2}\|x_0 -x^* + s_0- s^*\|^2  - \frac{L}{2}\|x_K -x^* + s_K- s^*\|^2 \\
                &+ \frac{1}{K} \sum_{k=1}^{K} \nabla f^\prime(x_{k-1}+s_{k-1})^\top( x_k+ s_k -x^* -s^* ) + \frac{L}{K} \sum_{k=1}^{K} (x_k +s_k -x_{k-1} -s_{k-1} )^\top( x_k+ s_k -x^* -s^* ), \\
                = & \frac{L}{2}\|x_0 +s_0 -x^* -s^*\|^2  - \frac{L}{2}\|x_K +s_K -x^* -s^*\|^2 \\
            & + \frac{1}{K} \sum_{k=1}^{K} \nabla f^\prime(x_{k-1}+s_{k-1})^\top( x_k+ s_k -x^* -s^* ) + \frac{L}{K} \sum_{k=1}^{K} \left(-\frac{\nabla f(x_{k-1})}{L} - \Delta s_{k-1}^\prime \right)^\top( x_k+ s_k -x^* -s^* ).
        \end{aligned}
    \end{equation}
\end{proof}

The above results imply that there is no convergence guarantee in OOD scenarios.

If not OOD, which means both variable and objective are from the InD scenario, i.e., $s := 0, f^\prime = f$, the convergence rate upper bound is as follows, which is precisely that of gradient-descent:
\begin{equation}
    \begin{aligned}
            & \frac{L}{2}\|x_0 -x^* + s_0- s^*\|^2  - \frac{L}{2}\|x_K -x^* + s_K- s^*\|^2 \\
            & + \frac{1}{K} \sum_{k=1}^{K} \nabla f^\prime(x_{k-1}+s_{k-1})^\top( x_k+ s_k -x^* -s^* ) + \frac{L}{K} \sum_{k=1}^{K} (x_k +s_k -x_{k-1} -s_{k-1} )^\top( x_k+ s_k -x^* -s^* ) \\
            = & \frac{L}{2}\|x_0 -x^*\|^2  - \frac{L}{2}\|x_K -x^* \|^2 + \frac{L}{K} \sum_{k=1}^{K} \frac{\nabla f(x_{k-1})}{L}^\top( x_k -x^*) + \frac{L}{K} \sum_{k=1}^{K} \left(-\frac{\nabla f(x_{k-1})}{L}\right)^\top( x_k -x^* ), \\
            = & \frac{L}{2}\|x_0 -x^*\|^2  - \frac{L}{2}\|x_K -x^* \|^2 .
    \end{aligned}
\end{equation}

Note that $s := 0$ cannot lead to the convergence rate of gradient descent since the third term in \eqref{thpf:theorem1_smooth_converge_rate} is non-zero and cannot be eliminated. 
Such an \textbf{upper} bound is trivially upper bounded by:
\begin{equation}
    \begin{aligned}
            & F^\prime(x_k+s_k) - F^\prime(x^*+s^*) \\
            \leq & \frac{L}{2}\|x_0 -x^* + s_0- s^*\|^2  - \frac{L}{2}\|x_K -x^* + s_K- s^*\|^2 + \frac{1}{K} \sum_{k=1}^{K} \nabla f^\prime(x_{k-1}+s_{k-1})^\top( x_k+ s_k -x^* -s^* )\\
        & + \frac{L}{K} \sum_{k=1}^{K} (x_k +s_k -x_{k-1} -s_{k-1} )^\top( x_k+ s_k -x^* -s^* ), \\
            \leq & \frac{L}{2}\|x_0 -x^* + s_0- s^*\|^2  - \frac{L}{2}\|x_K -x^* + s_K- s^*\|^2 + \frac{1}{K} \sum_{k=1}^{K} \| \nabla f^\prime(x_{k-1}+s_{k-1}) \| \| x_k+ s_k -x^* -s^* \| \\
        & + \frac{L}{K} \sum_{k=1}^{K} \|x_k +s_k -x_{k-1} -s_{k-1} \| \| x_k+ s_k -x^* -s^* \|.
    \end{aligned}
\end{equation}

Such an \textbf{upper} bound is trivially lower bounded as follows:
\begin{equation}
    \begin{aligned}
            & \frac{L}{2}\|x_0 -x^* + s_0- s^*\|^2  - \frac{L}{2}\|x_K -x^* + s_K- s^*\|^2 + \frac{1}{K} \sum_{k=1}^{K} \nabla f^\prime(x_{k-1}+s_{k-1})^\top( x_k+ s_k -x^* -s^* )\\
            & + \frac{L}{K} \sum_{k=1}^{K} (x_k +s_k -x_{k-1} -s_{k-1} )^\top( x_k+ s_k -x^* -s^* ) \\
        \geq & \frac{L}{2}\|x_0 -x^* + s_0- s^*\|^2  - \frac{L}{2}\|x_K -x^* + s_K- s^*\|^2  - \frac{1}{K} \sum_{k=1}^{K} \| \nabla f^\prime(x_{k-1}+s_{k-1}) \| \| x_k+ s_k -x^* -s^* \| \\
        & - \frac{L}{K} \sum_{k=1}^{K} \|x_k +s_k -x_{k-1} -s_{k-1} \| \| x_k+ s_k -x^* -s^* \|.
    \end{aligned}
\end{equation}

\subsection{Proof of Corollary \ref{lm:lemma3_smooth_converge_rate_ub}} \label{sec:lm3_proof}
% : Convergence Rate Upper Bound w.r.t. $\| s_{k-1}^\prime \|$
\begin{proof}
    % TODO: Make it more concrete: a minimal among K iterations.
    This proof derives the upper bound with respect to the magnitude of virtual input feature $\| s^\prime \|$ of the L2O model, where $s^\prime$ is proposed in \cref{sec:vv_traj} to formulate the difference between input features of the OOD and InD scenarios.

    Substituting $\Delta s_{k-1}^\prime$ yields:
    \begin{equation}
        \begin{aligned}
            & F^\prime(x_k+s_k) - F^\prime(x^*+s^*) \\
                \leq & \frac{1}{K} \sum_{k=1}^{K} \nabla f^\prime(x_{k-1}+s_{k-1})^\top( x_k+ s_k -x^* -s^* ) + \frac{L}{2}\|x_0 +s_0 -x^* -s^*\|^2  - \frac{L}{2}\|x_K +s_K -x^* -s^*\|^2\\
            & + \frac{L}{K} \sum_{k=1}^{K} \left(-\frac{\nabla f(x_{k-1})}{L} - \Delta s_{k-1}^\prime \right)^\top( x_k+ s_k -x^* -s^* )\\
                = & \frac{1}{K} \sum_{k=1}^{K} \nabla f^\prime(x_{k-1}+s_{k-1})^\top( x_k+ s_k -x^* -s^* ) + \frac{L}{2}\|x_0 +s_0 -x^* -s^*\|^2  - \frac{L}{2}\|x_K +s_K -x^* -s^*\|^2\\
            & - \frac{L}{K} \sum_{k=1}^{K} \left(\frac{\nabla f(x_{k-1})}{L} + \frac{\nabla f^\prime(x_{k-1}+s_{k-1}) - \nabla f(x_{k-1})}{2L} + \diag(\mathbf{J}_1 s_{k-1}^\prime) \nabla f^\prime(x_{k-1}+s_{k-1}) + \mathbf{J}_2 s_{k-1}^\prime \right)^\top \\
            & \qquad \quad \qquad (x_k+ s_k -x^* -s^*).
        \end{aligned}
    \end{equation}
    
    By Cauchy-Schwarz inequality and Triangle inequality, we have:
    \begin{equation}
        \begin{aligned}
            & F^\prime(x_k+s_k) - F^\prime(x^*+s^*) \\
                \leq & \frac{L}{2}\|x_0 +s_0 -x^* -s^*\|^2  - \frac{L}{2}\|x_K +s_K -x^* -s^*\|^2 \\
            & + \frac{1}{2K} \sum_{k=1}^{K} \left( \nabla f^\prime(x_{k-1}+s_{k-1}) - \nabla f(x_{k-1}) \right)^\top( x_k+ s_k -x^* -s^* ) \\
            & + \frac{L}{K} \sum_{k=1}^{K} \left(-\diag(\mathbf{J}_1 s_{k-1}^\prime) \nabla f^\prime(x_{k-1}+s_{k-1}) - \mathbf{J}_2 s_{k-1}^\prime \right)^\top (x_k+ s_k -x^* -s^*) \\
                \leq & \frac{L}{2}\|x_0 +s_0 -x^* -s^*\|^2  - \frac{L}{2}\|x_K +s_K -x^* -s^*\|^2 \\
            & + \frac{1}{2K} \sum_{k=1}^{K} \left( \nabla f^\prime(x_{k-1}+s_{k-1}) - \nabla f(x_{k-1}) \right)^\top( x_k+ s_k -x^* -s^* ) \\
            & + \frac{L}{K} \sum_{k=1}^{K} (\left\|  \diag(\mathbf{J}_1 s_{k-1}^\prime) \nabla f^\prime(x_{k-1}+s_{k-1})\right\| + \|\mathbf{J}_2 s_{k-1}^\prime \|) \| x_k+ s_k -x^* -s^* \|\\
                \leq & \frac{L}{2}\|x_0 +s_0 -x^* -s^*\|^2  - \frac{L}{2}\|x_K +s_K -x^* -s^*\|^2 \\
            & + \frac{1}{2K} \sum_{k=1}^{K} \left( \nabla f^\prime(x_{k-1}+s_{k-1}) - \nabla f(x_{k-1}) \right)^\top( x_k+ s_k -x^* -s^* ) \\
            & + \frac{L}{K} \sum_{k=1}^{K} ( C_1 \sqrt{n} \|s_{k-1}^\prime\| \| \nabla f^\prime(x_{k-1}+s_{k-1})\| + C_2\sqrt{n} \| s_{k-1}^\prime \|) \| x_k+ s_k -x^* -s^* \| \\
                = & \frac{L}{2}\|x_0 +s_0 -x^* -s^*\|^2  - \frac{L}{2}\|x_K +s_K -x^* -s^*\|^2 \\
            & + \frac{1}{2K} \sum_{k=1}^{K} \left( \nabla f^\prime(x_{k-1}+s_{k-1}) - \nabla f(x_{k-1}) \right)^\top( x_k+ s_k -x^* -s^* ) \\
            & + \frac{L}{K} \sum_{k=1}^{K}  C_1 \sqrt{n} \| \nabla f^\prime(x_{k-1}+s_{k-1})\| \|s_{k-1}^\prime\| + \frac{L}{K} \sum_{k=1}^{K} C_2 \sqrt{n} \| x_k+ s_k -x^* -s^* \| \| s_{k-1}^\prime \|.
        \end{aligned}
    \end{equation}

    % \underline{Lemma \ref{lm:lemma3_smooth_converge_rate_ub} is proved.}

\end{proof}

\subsection{Proof of Theorem \ref{lm:go_math_l2o_model}} \label{sec:tm2_l2o_model_proof}
This proof demonstrates that if the FP and GC conditions hold, the L2O model follows the structure defined in Theorem \ref{lm:go_math_l2o_model} and yields a unique solution at each iteration. We construct the proof by following the workflow proposed in \cite{Liu2023}.

First, we define our gradient-only input feature construction. Then, we apply the lemma proposed in \cite{Liu2023} to construct a few candidate parameter matrices. We propose a new formulation to achieve the two sufficient conditions, GC and FP. For the non-smooth part $r$ in the objective, we apply the proximal gradient method \cite{Michael2015} as in \cite{Liu2023} to solve a solution.

As the summation of two convex functions, $F(x)$ is convex on $x$. We have $\mathbf{0} \in \partial F(x)$ that $\mathbf{0} \in \nabla f(x^*) + \partial r(x)$. We choose $g_{x^*}$ as $\nabla f(x^*)$. When $k \to \infty$, by making the following denotation:
\begin{equation*}
    \hat{d}_k = d_k(\nabla f(x^*), -\nabla f(x^*), \mathbf{0}),
\end{equation*}
where $\nabla f(x^*)$ denotes gradient of optimal solution $x^*$. In the above definition, $\mathbf{0}$ means the results of historical modeling operator $u$ reaching zero when $k \to \infty$. Our following demonstrations will also derive the formulation of $u$ to ensure such a condition.

With $\hat{d}_k$, we can rewrite the L2O update formula of $k$-th iteration in \eqref{eq:d_wox} as:
 \begin{equation*}
    x_k=x_{k-1}-d_k(\nabla f(x_{k-1}), g_k, v_{k-1}) + d_k(\nabla f(x^*), -\nabla f(x^*), \nabla f(x^*), -\nabla f(x^*)) - \hat{d}_k. 
\end{equation*}
$g_k \in \partial r(x_k)$ represents an implicit subgradient vector at desired $x_k$, which yields the application of the proximal gradient algorithm \cite{Liu2023}.

We assume that there are following bounded parameter matrices:
\begin{equation*}
    \mathbf{J}_{j, k} \in \mathbb{R}^{n \times n}, \|\mathbf{J}_{j, k}\| \leq  C \sqrt{n}, \quad \forall j=1,2,3,
\end{equation*}
where $C$ is the upper bound on the Jacobian matrix of $d$'s function space. Without loss of generality, such an assumption is a general setting by setting a bounded activation function in machine learning \cite{Liu2023}.

Based on Lemma 1 in Section A.1. of \cite{Liu2023}, we can represent $d_k$ with the above bounded parameter matrices as follows: 
\begin{equation*}
    \begin{aligned}
        x_k=x_{k-1} & -\mathbf{J}_{1, k}(\nabla f(x_{k-1})-\nabla f(x^*))-\mathbf{J}_{2, k}(g_k+\nabla f(x^*)) -\mathbf{J}_{3, k}(v_{k-1}-\mathbf{0}) - \hat{d}_k ,\\
        =x_{k-1} & -\mathbf{J}_{1, k}(\nabla f(x_{k-1})-\nabla f(x^*))-\mathbf{J}_{2, k}(g_k+\nabla f(x^*)) -\mathbf{J}_{2, k}(\nabla f(x_{k-1})-\nabla f(x^*)) + \mathbf{J}_{2, k}(\nabla f(x_{k-1})-\nabla f(x^*))\\
        &-\mathbf{J}_{3, k}v_{k-1} - \hat{d}_k ,\\
        =x_{k-1} & -\mathbf{J}_{2, k}\nabla f(x_{k-1}) -\mathbf{J}_{2, k}g_k -\mathbf{J}_{3, k}v_{k-1} \\
        &-(\mathbf{J}_{1, k} - \mathbf{J}_{2, k})(\nabla f(x_{k-1})-\nabla f(x^*))  - \hat{d}_k. 
    \end{aligned}
\end{equation*}
In the second and third steps, we unify the parameters for smooth part and non-smooth part by the $- (\mathbf{J}_{1, k} - \mathbf{J}_{2, k})(\nabla f(x_k)-\nabla f(x^*))$ term. When $k \to \infty$, the above equality becomes:
\begin{equation*}
    \begin{aligned}
        x_k = &x_{k-1}  -\mathbf{J}_{2, k}\nabla f(x^*)-\mathbf{J}_{2, k}(-\nabla f(x^*)) -\mathbf{J}_{3, k}\mathbf{0} -(\mathbf{J}_{1, k} - \mathbf{J}_{2, k})(\nabla f(x^*)-\nabla f(x^*))  - \mathbf{0} ,\\
        =&x_{k-1},
    \end{aligned}
\end{equation*}
where we define $\lim_{k \to \infty} v_k = \mathbf{0}$. At each iteration, given a group of parameter $\mathbf{J}$ and $b$, the solution $x_k$ is uniquely constructed. Based on the method proposed in \cite{Liu2023}, we construct such parameters by learning. We define the learnable parameters as follows:
\begin{equation*}
    \begin{aligned}
        \mathbf{R}_{k}:=&\mathbf{J}_{2, k},\\
        \mathbf{Q}_{k}:=&\mathbf{J}_{3, k},\\
        % \mathbf{H}_{k}:=&\mathbf{J}_{5, k},\\
        % \mathbf{T}_{k}:=&\mathbf{J}_{6, k},\\
        b_{1,k}:=&  (\mathbf{J}_{1, k} - \mathbf{J}_{2, k})(\nabla f(x_{k-1})-\nabla f(x^*)) + \hat{d}_k.
    \end{aligned}
\end{equation*}
Thus, the update of solution is given by:
\begin{equation} \label{thpf:go_math_l2o_define}
    \begin{aligned}
    x_k = x_{k-1} & -\mathbf{R}_{k}\nabla f(x_{k-1}) -\mathbf{R}_{k}g_k -\mathbf{Q}_{k}v_{k-1} - b_{1,k}. 
    \end{aligned}
\end{equation}

As demonstrated in \cite{Liu2023}, all terms in $b_{1,k}$ reach zero as the iteration reaches $\infty$. We note that all defined parameter matrices are bounded by Lemma 1 in \cite{Liu2023}. From Triangle and Cauchy Schwarz inequalities, $b_{1,k}$ is also bounded by:
\begin{equation*}
    \|b_{1, k}\| \leq 2 \sqrt{n} C\|\nabla f(x_k)-\nabla f(x^*)\| + \|\hat{{d}}_k\|.
\end{equation*}
Here, we eliminate the requirement on an extra parameter matrix to control the boundness of $b_{1, k}$ in \cite{Liu2023}. Moreover, we note that $b_{1, k}$ can be arbitrarily defined, which means it may be either non-negative or non-positive. This observation implies that both negative and positive implementations are available. In our implementation, we following \cite{Liu2023} and use non-negative $b_{1, k}$.

Then, we derive the update formulation for $v_k$. Following \cite{Liu2023}, we set the length of historical information $T = 2$. We define the following operator to generate the historical feature vector $v_k$:
\begin{equation*}
    v_k = u_k(\nabla f(x_{k-1}) + g_{k-1}, v_{k-1}),
\end{equation*}
where we use explicit subgradient vector $g_{k-1}$. As defined in \cref{sec:design_model}, we can recover subgradient vector $g_k$ after solving $x_k$. Based on the L2O model defined in \eqref{thpf:go_math_l2o_define}, assume $\mathbf{R}_{k} \succ 0$, we have the following equation to get the summation of gradient and subgradient:
\begin{equation*}
    \nabla f(x_{k-1}) + g_{k-1} = \mathbf{R}_{k}^{-1} (x_{k-1} - x_{k} - \mathbf{Q}_k v_{k-1} - b_{1,k}).
\end{equation*}
Moreover, we take a recurrent definition of the operator $u$, which takes the output of the last iteration $v_{k-1}$ as the second input.

Suppose $\mathbf{0} := u_k(\mathbf{0}, \mathbf{0})$, which means when $k \to \infty$, the inputs of $u$ are all the gradient (and subgradient) at optimal solution and the output of $u$ converge to the gradient (and subgradient) at optimal solution as well. Suppose there are following bounded parameter matrices:
\begin{equation*}
    \mathbf{J}_{j, k} \in \mathbb{R}^{n\times n}, \|\mathbf{J}_{j, k}\| \leq \sqrt{n} C, \quad \forall j=5,6.
\end{equation*}

Denote $G_{k-1} = \nabla f(x_{k-1}) + g_{k-1}$, we have:
\begin{equation*}
    \begin{aligned}
        v_k =& u_k(G_{k-1}, v_{k-1}) - u(\mathbf{0}, \mathbf{0}) + \mathbf{0} ,\\
        =& \mathbf{J}_{5, k}(G_{k-1} - \mathbf{0}) + \mathbf{J}_{6, k}(v_{k-1} - \mathbf{0}) + \mathbf{0},\\
        =& \mathbf{J}_{5, k}G_{k-1} + \mathbf{J}_{6, k}v_{k-1} + (\mathbf{I} - \mathbf{J}_{5, k} - \mathbf{J}_{6, k})\mathbf{0} , \\
        =& \mathbf{J}_{5, k}G_{k-1} + (\mathbf{I} - \mathbf{J}_{5, k} - \mathbf{J}_{6, k})G_{k-1}+ \mathbf{J}_{6, k}v_{k-1} - (\mathbf{I} - \mathbf{J}_{5, k} - \mathbf{J}_{6, k})(G_{k-1} - \mathbf{0}) , \\
        =& (\mathbf{I}-\mathbf{J}_{6, k})G_{k-1} + \mathbf{J}_{6, k}v_{k-1} - (\mathbf{I} - \mathbf{J}_{5, k} - \mathbf{J}_{6, k})(G_{k-1} - \mathbf{0}).
    \end{aligned}
\end{equation*}
Here, we construct a reaching zero term $G_{k-1} - \mathbf{0}$ w.r.t. $G_{k-1}$, which imply that $G_{k-1}$ may not be exactly equal to $\mathbf{0}$. We define the learnable parameters as follows:
\begin{equation*}
    \begin{aligned}
        \mathbf{B}_k:=&\mathbf{J}_{6, k},\\
        b_{2,k}:=& (\mathbf{I} - \mathbf{J}_{5, k} - \mathbf{J}_{6, k})(G_{k-1} - \mathbf{0}).
    \end{aligned}
\end{equation*}

Assume $v_0:=0$, at $k$-th iteration, the historical information $v_k$ is given by the following equation:
\begin{equation} \label{thpf:his_operator_u}
    v_k = (\mathbf{I} - \mathbf{B}_k)G_{k-1} + \mathbf{B}_k v_{k-1} - b_{2,k}. 
\end{equation}
Motivated by the momentum scheme in FISTA \cite{Beck2009fast}, we set $\mathbf{B}_k$ to be negative semi-definite. Thus, $v_k$ illustrates the momentum of the gradient at $x_{k-1}$. It is worth noting that the above formulation is more like the classical momentum method, different from the Nesterov gradient method in FISTA \cite{Beck2009fast} and Math-L2O \cite{Liu2023}.

At $k-1$-th iteration, $v_{k-1}$ is yielded by: 
\begin{equation*}
    v_{k-1} = (\mathbf{I} - \mathbf{B}_{k-1})G_{k-2} + \mathbf{B}_{k-1} v_{k-2} - b_{2,k-1}. 
\end{equation*}

Substituting $v_{k-1}$ into \eqref{thpf:go_math_l2o_define} yields the following complete formulation to generate $x_k$ at $k$-th iteration:
\begin{equation} \label{thpf:go_math_l2o_basic}
    \begin{aligned}
        x_k &= x_{k-1}  -\mathbf{R}_{k}\nabla f(x_{k-1}) -\mathbf{R}_{k}g_k -\mathbf{Q}_{k}v_{k-1} - b_{1,k} \\
        &= x_{k-1} - \mathbf{R}_{k}\nabla f(x_{k-1}) - \mathbf{Q}_{k}((\mathbf{I} - \mathbf{B}_{k-1})G_{k-2} + \mathbf{B}_{k-1} v_{k-2} - b_{2,k-1}) - b_{1,k} -\mathbf{R}_{k}g_k.
    \end{aligned}
\end{equation}

% For $g_k$, we utilize a learnable diagonal matrix, denote as $\mathbf{A}_k$, $\mathbf{A}_k := \diag(a_k)$, to learn a subgradient value from the subgradient set by:
% \begin{equation*}
%     g_k = (\mathbf{I} - \mathbf{A}_k) \partial r(x_k)_{\text{lb}} + \mathbf{A}_k \partial r(x_k)_{\text{ub}}.
% \end{equation*}

We follow the method proposed in \cite{Liu2023} to derive a unique solution of $x_k$ based on the first-order derivative condition of non-smooth convex optimization. For non-smooth convex objective $r(x)$, $0 \in \partial r(x)$ is a sufficient and necessary condition for its optimality. We rewrite \eqref{thpf:go_math_l2o_basic} as:
\begin{equation*}
    x_k + \mathbf{R}_{k}g_k = x_{k-1} - \mathbf{R}_{k}\nabla f(x_{k-1}) - \mathbf{Q}_{k}((\mathbf{I} - \mathbf{B}_{k-1})G_{k-2} + \mathbf{B}_{k-1} v_{k-2} - b_{2,k-1}) - b_{1,k}.
\end{equation*}

Since $g_k \in \partial r(x_k)$, we have:
\begin{equation*}
    x_{k-1} - \mathbf{R}_{k}\nabla f(x_{k-1}) - \mathbf{Q}_{k}((\mathbf{I} - \mathbf{B}_{k-1})G_{k-2} + \mathbf{B}_{k-1} v_{k-2} - b_{2,k-1}) - b_{1,k} \in x_k + \mathbf{R}_{k}\partial r(x_k).
\end{equation*}

After rearrangement, we have:
\begin{equation*}
    0 \in \mathbf{R}_{k}\partial r(x_k) + x_k-\left(x_{k-1} - \mathbf{R}_{k}\nabla f(x_{k-1}) - \mathbf{Q}_{k}((\mathbf{I} - \mathbf{B}_{k-1})G_{k-2} + \mathbf{B}_{k-1} v_{k-2} - b_{2,k-1}) - b_{1,k}\right).
\end{equation*}

Given $\mathbf{R}_{k}$ as a symmetric positive definite matrix, we have:
\begin{equation} \label{thpf:ppm}
    0 \in \partial r(x_k) + \mathbf{R}_{k}^{-1}\left(x_k-\left(x_{k-1} - \mathbf{R}_{k}\nabla f(x_{k-1}) - \mathbf{Q}_{k}((\mathbf{I} - \mathbf{B}_{k-1})G_{k-2} + \mathbf{B}_{k-1} v_{k-2} - b_{2,k-1}) - b_{1,k}\right)\right),
\end{equation}
where $x_{k-1} - \mathbf{R}_{k}\nabla f(x_{k-1}) - \mathbf{Q}_{k}((\mathbf{I} - \mathbf{B}_{k-1})G_{k-2} + \mathbf{B}_{k-1} v_{k-2} - b_{2,k-1}) - b_{1,k}$ are exactly calculated, we denote it as $\bar{x}$.

Then, based on first-order condition, 
$x_k$ can be uniquely solved by following the proximal operator:
\begin{equation*}
    x_k= \arg \min_{x} r(x) + (1/2)(x - \bar{x})^\top \mathbf{R}_{k}^{-1}(x - \bar{x}), 
\end{equation*}
where taking $x$ as the variable, $r(x) + (1/2)(x - \bar{x})^\top \mathbf{R}_{k}^{-1}(x - \bar{x})$ is the mathematical integration of right-hand side of \eqref{thpf:ppm}.

In the experiments, we set $\mathbf{R}$, $\mathbf{Q}$, and $\mathbf{B}$ to be positive definite matrices by Sigmoid activation functions.

\section{Composite Case Results} \label{sec:smooth_with_non_smooth}
This section introduces several more theoretical findings on the composite case where the smooth and non-smooth parts in objective \ref{obj:ind} are non-degenerated. Similar to the results in the smooth case of main pages, we derive several theorems and corollaries on per iteration and multi-iteration convergence of the L2O model. 
We follow the proofs of the vanilla proximal point method and proximal gradient algorithm (PGA) in \cite{Michael2015} and \cite{Vandenberghe2022} to derive our demonstrations for theorems and corollaries, where we use a gradient map to represent the $\argmin$ operation for non-smooth optimization in PGA.

\subsection{Preliminary}
As in \eqref{obj:ind}, the objective of composite case is as below:
\begin{equation*}
    \min_{x} f(x) + r(x),
\end{equation*}
where $f(x) \in \mathcal{F}_{L}$ is a $L$-smooth and convex function and $r(x) \in \mathcal{F}$ is a proper, convex but probably non-smooth function. Notably, in the composite case, $f(x)$ and $r(x)$ are non-degenerated.

Based on the definition of $L$-smoothness on $f(x)$, $\frac{L^\prime}{2} {x}^\top x - f(x)$ is convex \cite{zhou2018fenchel}. Thus, for any points $y,x \in \mathbb{R}^n$, the convexity of $f$ yields the following upper bound of $f(y)$:
\begin{equation} \label{thpf:composite_lsmooth}
    f(y) \leq f(x) + \nabla f(x)^\top (y - x) + \frac{L}{2} \|y - x\|^2.
\end{equation}

We take the definition of the $k$-th iteration update formulation given by the L2O model in \cite{Liu2023} as below:
\begin{equation} \label{eq:def_l2o_composite}
    x_{k} = x_{k-1} - \diag\left(\mathbf{N}_{1}(z_{k-1})\right)\big(\nabla f(x_{k-1}) + g_{k}\big) - \mathbf{N}_{2}(z_{k-1}),
\end{equation}
where we set $\diag\left(\mathbf{N}_{1}(z_{k-1})\right) \succeq 0$ as a symmetric positive definition matrix and $g_{k} \in \partial r(x)$ as an implicit subgradient value at $x_{k}$ \cite{Liu2023}. We have the following reformulation of the above equation:
\begin{equation*}
    \diag\left(\mathbf{N}_{1}(z_{k-1})\right)^{-1}\Big(x_{k} - \big( x_{k-1} - \diag\left(\mathbf{N}_1(z_{k-1})\right)\nabla f(x_{k-1})  - \mathbf{N}_{2}(z_{k-1})\big)\Big) + g_{k} = \mathbf{0}.
\end{equation*}

Since $g_{k} \in \partial r(x)$, we can represent the above equation with the following relationship:
\begin{equation*}
    \mathbf{0} \in  \partial r(x_{k}) + \diag\left(\mathbf{N}_{1}(z_{k-1})\right)^{-1}\Big(x_{k} - \big( x_{k-1} - \diag\left(\mathbf{N}_1(z_{k-1})\right)\nabla f(x_{k-1})  - \mathbf{N}_{2}(z_{k-1})\big)\Big). 
\end{equation*}
Due to the first-order condition for convex optimization, as in \cite{Liu2023}, appling the proximal gradient method in \cite{Rockafellar1976}, we can use the following proximal operator to solve a $x_k$:
\begin{equation} \label{eq:l2o_composite}
    \begin{aligned}
        &\operatorname{prox}_{\diag\left(\mathbf{N}_{1}(z_{k-1})\right)^{-1}}\Big(x_{k} - \big( x_{k-1} - \diag\left(\mathbf{N}_1(z_{k-1})\right)\nabla f(x_{k-1})  - \mathbf{N}_{2}(z_{k-1})\big)\Big) \\
        =&\argmin_{x_{k}} r(x_{k}) + \frac{1}{2} \left\|x_{k} - \big( x_{k-1} - \diag\left(\mathbf{N}_1(z_{k-1})\right)\nabla f(x_{k-1})  - \mathbf{N}_{2}(z_{k-1})\big) \right\|^2_{\diag\left(\mathbf{N}_{1}(z_{k-1})\right)^{-1}},
    \end{aligned}
\end{equation}
where the norm $\|\cdot\|_{\mathbf{P}_k^{-1}}$ is defined as $\|x\|_{\mathbf{P}_k^{-1}}:=\sqrt{x^{\top} \mathbf{P}_k^{-1} x}$ \cite{Liu2023}. 
From our definition in \cref{sec:def_optimizee}, the non-smooth function $r$ is solvable. The $\argmin$ operation in the above operator will explicitly generate an optimal solution.

The unknown implicit process in $\argmin$ makes it hard to explicitly analyze the update from $x_{k-1}$ to $x_k$. However, there are precisely two parts in the above operator, i.e., the $\argmin$ operation on non-smooth function $r$ and gradient descent operation $\big( x_{k-1} - \diag\left(\mathbf{N}_1(z_{k-1})\right)\nabla f(x_{k-1})  - \mathbf{N}_{2}(z_{k-1})\big)$ on the smooth function $f$. We can regard the non-smooth function $r$ update as an implicit subgradient descent and combine the two parts into one proximal gradient descent with both smooth and non-smooth gradients \cite{Vandenberghe2022}. As in \cite{Vandenberghe2022}, the proximal gradient is named the gradient map.

Different from \cite{Vandenberghe2022}, we define the gradient map for L2O model in this work, denote as $G_{\mathbf{N}_1(z)}(x_{k-1})$. To represent the update from $x_{k-1}$ to the $x_k$ given by the L2O model defined in \eqref{eq:def_l2o_composite}, we define the it as following operations:
\begin{equation*}
    \begin{aligned}
        &G_{\diag\left(\mathbf{N}_1(z_{k-1})\right)^{-1}}(x_{k-1}) \\
        :=& \diag\left(\mathbf{N}_1(z_{k-1})\right)^{-1} \Big(x_{k-1} - \operatorname{prox}_{\diag\left(\mathbf{N}_1(z_{k-1})\right)^{-1}}\big( x_{k-1} - \diag\left(\mathbf{N}_1(z_{k-1})\right)\nabla f(x_{k-1})  - \mathbf{N}_{2}(z_{k-1})\big) -\mathbf{N}_2(z_{k-1})\Big).
    \end{aligned}
\end{equation*}
Then, $G_{\mathbf{N}_1(z)}(x_{k-1})$ yields the following update formulation from $x_{k-1}$ to $x_k$, which is similar to the L2O model in the smooth case.
\begin{equation} \label{thpf:ind_GM_composite}
    \begin{aligned}
        x_{k} &= \operatorname{prox}_{\diag\left(\mathbf{N}_1(z_{k-1})\right)^{-1}}\Big( x_{k-1} - \diag\left(\mathbf{N}_1(z_{k-1})\right)\nabla f(x_{k-1})  - \mathbf{N}_{2}(z_{k-1})\Big), \\
        &= x_{k-1} - \diag\left(\mathbf{N}_1(z_{k-1})\right)G_{\diag\left(\mathbf{N}_1(z_{k-1})\right)^{-1}}(x_{k-1}) -\mathbf{N}_2(z_{k-1}).
    \end{aligned}
\end{equation}
Substitute the above $x_{k}$'s representation with gradient map into the $L$-smoothness inequation in \eqref{thpf:composite_lsmooth}, we have the following upper bound of $f(x_{k})$ from $L$-smoothness:
\begin{equation} \label{thpf:composite_ub_pre}
    \begin{aligned}
        f(x_{k}) \leq f(x_{k-1}) &+ \nabla f(x_{k-1})^\top (x_{k
        } - x_{k-1}) + \frac{L}{2} \|x_{k
        } - x_{k-1}\|^2 , \\
        \leq f(x_{k-1}) &- \nabla f(x_{k-1})^\top (\diag\left(\mathbf{N}_1(z_{k-1})\right)G_{\diag\left(\mathbf{N}_1(z_{k-1})\right)^{-1}}(x_{k-1}) +\mathbf{N}_2(z_{k-1})), \\
        & + \frac{L}{2} \left\| \diag\left(\mathbf{N}_1(z_{k-1})\right)G_{\diag\left(\mathbf{N}_1(z_{k-1})\right)^{-1}}(x_{k-1}) +\mathbf{N}_2(z_{k-1}) \right\|^2.
    \end{aligned}
\end{equation}

Moreover, we would like to construct the representation of $\nabla f(x_{k-1}) + g_{k}$ by the gradient map. 
From the gradient map definition in \eqref{thpf:ind_GM_composite} and the L2O model definition in \eqref{eq:def_l2o_composite}, we directly have the following equality of $G_{\diag\left(\mathbf{N}_1(z_{k-1})\right)^{-1}}(x_{k-1})$:
\begin{equation} \label{thpf:G_fg_equation_composite}
    G_{\diag\left(\mathbf{N}_1(z_{k-1})\right)^{-1}}(x_{k-1}) =  \nabla f(x_{k-1}) + g_{k},
\end{equation}
where $g_{k} \in \partial r(x_{k})$ is the virtual subgradient of the non-smooth part $r$ of objective. Thus, we have $G_{\diag\left(\mathbf{N}_1(z_{k-1})\right)^{-1}}(x_{k-1}) - \nabla f(x_{k-1}) \in \partial r(x_{k})$, 
along with the definition of the convexity of $r$, for any $x,t \in \mathbb{R}^n$, we have the following inequality between $r(t)$ and $r(x)$:
\begin{equation*} 
    r(t) \geq r(x_{k}) + \big(G_{\diag\left(\mathbf{N}_1(z_{k-1})\right)^{-1}}(x_{k-1}) - \nabla f(x_{k-1})\big)^\top (t - x_{k}).
\end{equation*}
After rearrangement, we have the following upper bound of $r(x_{k})$ with any arbitrary $t\in \mathbb{R}^n$:
\begin{equation} \label{thpf:composite_ub_pre2}
    r(x_{k}) \leq r(t) - \big(G_{\diag\left(\mathbf{N}_1(z_{k-1})\right)^{-1}}(x_{k-1}) - \nabla f(x_{k-1})\big)^\top (t - x_{k}).
\end{equation}

Finally, we present the following lemma to construct a general relationship between the objectives of any arbitrary two points:
\begin{lemma} \label{thpf:lm_GM_basic_bound_composite}
    $\forall x_{k}, t \in \mathbf{R}^n$:
    \begin{equation*}
        \begin{aligned}
            &F(x_{k}) \\
            \leq&  F(t) + G_{\diag\left(\mathbf{N}_1(z_{k-1})\right)^{-1}}(x_{k-1})^\top (x_{k-1} - t) \\
            & + \frac{L}{2} \Bigg(\left\| \diag\left(\mathbf{N}_1(z_{k-1})\right)G_{\diag\left(\mathbf{N}_1(z_{k-1})\right)^{-1}}(x_{k-1}) +\mathbf{N}_2(z_{k-1}) - \frac{G_{\diag\left(\mathbf{N}_1(z_{k-1})\right)^{-1}}(x_{k-1})}{L}\right\|^2 \\
            & \quad \qquad - \left\|\frac{G_{\diag\left(\mathbf{N}_1(z_{k-1})\right)^{-1}}(x_{k-1})}{L}\right\|^2\Bigg).
        \end{aligned}
    \end{equation*}
\end{lemma}
The proof is as follows.

\begin{proof}
    First, based on the definition that $\forall x \in \mathbb{R}^n, F(x) = f(x) + r(x)$, adding $r(x_{k})$ into inequality \ref{thpf:composite_ub_pre} yields a full representation of objective $F$:
    \begin{equation*}
        \begin{aligned}
            & F(x_{k}) \\
            \leq & f(x_{k-1}) - \nabla f(x_{k-1})^\top \big(\diag\left(\mathbf{N}_1(z_{k-1})\right)G_{\diag\left(\mathbf{N}_1(z_{k-1})\right)^{-1}}(x_{k-1}) +\mathbf{N}_2(z_{k-1}) \big) \\
            & + \frac{L}{2} \left\| \diag\left(\mathbf{N}_1(z_{k-1})\right)G_{\diag\left(\mathbf{N}_1(z_{k-1})\right)^{-1}}(x_{k-1}) +\mathbf{N}_2(z_{k-1}) \right\|^2 + r(x_{k}).\\
        \end{aligned}
    \end{equation*}

    Since $f$ is convex and differentiable, $\forall x_{k-1}, t \in \mathbb{R}^n$, we have $f(x_{k-1}) \leq f(t) - \nabla f(x_{k-1})^\top(t-x_{k-1})$, adding it into the above inequation yields:
    \begin{equation*}
        \begin{aligned}
            & F(x_{k}) \\
            \leq & f(t) - \nabla f(x_{k-1})^\top(t-x_{k-1}) - \nabla f(x_{k-1})^\top \big(\diag\left(\mathbf{N}_1(z_{k-1})\right)G_{\diag\left(\mathbf{N}_1(z_{k-1})\right)^{-1}}(x_{k-1}) +\mathbf{N}_2(z_{k-1})\big) \\
            & + \frac{L}{2} \left\| \diag\left(\mathbf{N}_1(z_{k-1})\right)G_{\diag\left(\mathbf{N}_1(z_{k-1})\right)^{-1}}(x_{k-1}) +\mathbf{N}_2(z_{k-1})\right\|^2 + r(x_{k}).
        \end{aligned}
    \end{equation*}
    Moreover, adding the upper bound of $r(x_k)$ in inequation \ref{thpf:composite_ub_pre2} yields:
    \begin{equation*}
        \begin{aligned}
            & F(x_{k}) \\
            \leq & f(t) - \nabla f(x_{k-1})^\top(t-x_{k-1}) - \nabla f(x_{k-1})^\top \big(\diag\left(\mathbf{N}_1(z_{k-1})\right)G_{\diag\left(\mathbf{N}_1(z_{k-1})\right)^{-1}}(x_{k-1}) +\mathbf{N}_2(z_{k-1})\big) \\
            & + \frac{L}{2} \left\| \diag\left(\mathbf{N}_1(z_{k-1})\right)G_{\diag\left(\mathbf{N}_1(z_{k-1})\right)^{-1}}(x_{k-1}) +\mathbf{N}_2(z_{k-1})\right\|^2 \\
            & + r(t) - \big(G_{\diag\left(\mathbf{N}_1(z_{k-1})\right)^{-1}}(x_{k-1}) - \nabla f(x_{k-1})\big)^\top \Big(t - \big(x_{k-1} - \diag\left(\mathbf{N}_1(z_{k-1})\right)G_{\diag\left(\mathbf{N}_1(z_{k-1})\right)^{-1}}(x_{k-1}) -\mathbf{N}_2(z_{k-1})\big)\Big).
        \end{aligned}
    \end{equation*}

    Then, we make the following rearrangement on the right-hand side of the above inequation:
    \begin{equation*}
        \begin{aligned}
            & F(x_{k}) \\
            \leq & f(t) - \nabla f(x_{k-1})^\top(t-x_{k-1}) - \nabla f(x_{k-1})^\top \big(\diag\left(\mathbf{N}_1(z_{k-1})\right)G_{\diag\left(\mathbf{N}_1(z_{k-1})\right)^{-1}}(x_{k-1}) +\mathbf{N}_2(z_{k-1})\big) \\
            & + \frac{L}{2} \left\| \diag\left(\mathbf{N}_1(z_{k-1})\right)G_{\diag\left(\mathbf{N}_1(z_{k-1})\right)^{-1}}(x_{k-1}) +\mathbf{N}_2(z_{k-1})\right\|^2 \\
            & + r(t) - \big(G_{\diag\left(\mathbf{N}_1(z_{k-1})\right)^{-1}}(x_{k-1}) - \nabla f(x_{k-1})\big)^\top \Big(t - \big(x_{k-1} - \diag\left(\mathbf{N}_1(z_{k-1})\right)G_{\diag\left(\mathbf{N}_1(z_{k-1})\right)^{-1}}(x_{k-1}) -\mathbf{N}_2(z_{k-1})\big)\Big), \\
            = & f(t) + r(t) + \frac{L}{2} \left\| \diag\left(\mathbf{N}_1(z_{k-1})\right)G_{\diag\left(\mathbf{N}_1(z_{k-1})\right)^{-1}}(x_{k-1}) +\mathbf{N}_2(z_{k-1})\right\|^2 \\
            & - G_{\diag\left(\mathbf{N}_1(z_{k-1})\right)^{-1}}(x_{k-1})^\top (t - (x_{k-1} - \diag\left(\mathbf{N}_1(z_{k-1})\right)G_{\diag\left(\mathbf{N}_1(z_{k-1})\right)^{-1}}(x_{k-1}) -\mathbf{N}_2(z_{k-1}))), \\
            = & F(t) + G_{\diag\left(\mathbf{N}_1(z_{k-1})\right)^{-1}}(x_{k-1})^\top (x_{k-1} - t) \\
            & + \frac{L}{2} \left\| \diag\left(\mathbf{N}_1(z_{k-1})\right)G_{\diag\left(\mathbf{N}_1(z_{k-1})\right)^{-1}}(x_{k-1}) +\mathbf{N}_2(z_{k-1})\right\|^2 \\
            & - G_{\diag\left(\mathbf{N}_1(z_{k-1})\right)^{-1}}(x_{k-1})^\top \big(\diag\left(\mathbf{N}_1(z_{k-1})\right)G_{\diag\left(\mathbf{N}_1(z_{k-1})\right)^{-1}}(x_{k-1}) + \mathbf{N}_2(z_{k-1})\big),
        \end{aligned}
    \end{equation*}
    where we put $f(t)$ and $r(t)$ together and combine a similar term to achieve the simplification in the second step. In the third step, we combine $f(t)+r(t)$ as $F(t)$ based on the objective definition.

    Finally, making up a perfect square between the last two terms finishes the proof:
    \begin{equation*}
        \begin{aligned}
            & F(x_{k}) \\
            = & F(t) + G_{\diag\left(\mathbf{N}_1(z_{k-1})\right)^{-1}}(x_{k-1})^\top (x_{k-1} - t) \\
            & + \frac{L}{2} \left\| \diag\left(\mathbf{N}_1(z_{k-1})\right)G_{\diag\left(\mathbf{N}_1(z_{k-1})\right)^{-1}}(x_{k-1}) +\mathbf{N}_2(z_{k-1})\right\|^2 \\
            & - G_{\diag\left(\mathbf{N}_1(z_{k-1})\right)^{-1}}(x_{k-1})^\top \big(\diag\left(\mathbf{N}_1(z_{k-1})\right)G_{\diag\left(\mathbf{N}_1(z_{k-1})\right)^{-1}}(x_{k-1}) + \mathbf{N}_2(z_{k-1})\big), \\
            = & F(t) + G_{\diag\left(\mathbf{N}_1(z_{k-1})\right)^{-1}}(x_{k-1})^\top (x_{k-1} - t) \\
            & + \frac{L}{2} \Big(\| \diag\left(\mathbf{N}_1(z_{k-1})\right)G_{\diag\left(\mathbf{N}_1(z_{k-1})\right)^{-1}}(x_{k-1}) +\mathbf{N}_2(z_{k-1})\|^2 \\
            & \qquad - \frac{2}{L} G_{\diag\left(\mathbf{N}_1(z_{k-1})\right)^{-1}}(x_{k-1})^\top \big(\diag\left(\mathbf{N}_1(z_{k-1})\right)G_{\diag\left(\mathbf{N}_1(z_{k-1})\right)^{-1}}(x_{k-1}) + \mathbf{N}_2(z_{k-1})\big)\Big) \\
            = & F(t) + G_{\diag\left(\mathbf{N}_1(z_{k-1})\right)^{-1}}(x_{k-1})^\top (x_{k-1} - t) \\
            & + \frac{L}{2} \Bigg(\left\| \diag\left(\mathbf{N}_1(z_{k-1})\right)G_{\diag\left(\mathbf{N}_1(z_{k-1})\right)^{-1}}(x_{k-1}) +\mathbf{N}_2(z_{k-1}) - \frac{G_{\diag\left(\mathbf{N}_1(z_{k-1})\right)^{-1}}(x_{k-1})}{L}\right\|^2 \\
            & \quad \qquad - \left\|\frac{G_{\diag\left(\mathbf{N}_1(z_{k-1})\right)^{-1}}(x_{k-1})}{L}\right\|^2\Bigg).
        \end{aligned}
    \end{equation*}
\end{proof}

We are ready to derive convergence analysis based on Lemma~\ref{thpf:lm_GM_basic_bound_composite}. We will iteratively apply Lemma~\ref{thpf:lm_GM_basic_bound_composite} to construct the difference in objective between one and last iteration and between one iteration and the optimum.

\subsection{InD Convergence Upper Bound}
Similar to Lemma \ref{lm:lemma1_ind_homogeneous_converge_gain} for the smooth case, for the composite case, we propose the following lemma for per iteration convergence gain to ensure the L2O model is robust in the InD scenario.
\begin{lemma} \label{lm:ind_converge_gain_composite}
    For $\forall z_{k-1} \in \mathcal{Z}_P, \forall x_{k-1} \in \mathcal{S}_P$, 
    if $\mathbf{N}_1(z_{k-1})$ and $\mathbf{N}_2(z_{k-1})$ are bounded by following compact sets:
    \begin{equation*}
        \begin{aligned}
            &\mathbf{N}_1(z_{k-1}) \in \left[\mathbf{0}, \frac{2}{L} \mathbf{1} \right],\\
            & \left\| \diag\left(\mathbf{N}_1(z_{k-1})\right)\big(\nabla f(x_{k-1}) + g_{k}\big) +\mathbf{N}_2(z_{k-1}) - \frac{\nabla f(x_{k-1}) + g_{k}}{L}\right\| \leq \left\|\frac{\nabla f(x_{k-1}) + g_{k}}{L} \right\|, \forall \mathbf{N}_1(z_{k-1}) \in \left[\mathbf{0}, \frac{2}{L} \mathbf{1} \right],
        \end{aligned}
    \end{equation*}
    where $g_{k} \in \partial r(x_{k})$, for any $x_k$ generated by L2O model in \eqref{eq:l2o_composite}, we have the following homogeneous derease on objective:
    \begin{equation*}
        F(x_k) - F(x_{k-1}) \leq 0.
    \end{equation*}
\end{lemma}
\begin{proof}
    Based on Lemma~\ref{thpf:lm_GM_basic_bound_composite}, set $t := x_{k-1}$, we have the following inequation between two objectives: 
    \begin{equation} \label{thpf:converge_gain_composite_ineq}
        \begin{aligned}
            &F(x_{k}) - F(x_{k-1})\\
            \leq& \frac{L}{2} \Bigg(\left\| \diag\left(\mathbf{N}_1(z_{k-1})\right)G_{\diag\left(\mathbf{N}_1(z_{k-1})\right)^{-1}}(x_{k-1}) +\mathbf{N}_2(z_{k-1}) - \frac{G_{\diag\left(\mathbf{N}_1(z_{k-1})\right)^{-1}}(x_{k-1})}{L}\right\|^2 \\
            & \quad - \left\|\frac{G_{\diag\left(\mathbf{N}_1(z_{k-1})\right)^{-1}}(x_{k-1})}{L}\right\|^2\Bigg).
        \end{aligned}
    \end{equation}

    To ensure $F(x_{k-1}) \leq F(x_{k-1})$, we should keep right-hand side non-positive. Thus, we have the following inequality:
    \begin{equation} \label{thpf:ind_converge_gain_composite_ineq}
        \begin{aligned}
            & \left\| \diag\left(\mathbf{N}_1(z_{k-1})\right)G_{\diag\left(\mathbf{N}_1(z_{k-1})\right)^{-1}}(x_{k-1}) +\mathbf{N}_2(z_{k-1}) - \frac{G_{\diag\left(\mathbf{N}_1(z_{k-1})\right)^{-1}}(x_{k-1})}{L}\right\| \leq \frac{\left\|G_{\diag\left(\mathbf{N}_1(z_{k-1})\right)^{-1}}(x_{k-1})\right\|}{L}.
        \end{aligned}
    \end{equation}

    Similarly, we first freeze $\mathbf{N}_2(z_{k-1})$ and discuss $\mathbf{N}_1(z_{k-1})$-only terms, which yields:
    \begin{equation*}
        \begin{aligned}
            & \left\| \diag\left(\mathbf{N}_1(z_{k-1})\right)G_{\diag\left(\mathbf{N}_1(z_{k-1})\right)^{-1}}(x_{k-1}) - \frac{G_{\diag\left(\mathbf{N}_1(z_{k-1})\right)^{-1}}(x_{k-1})}{L}\right\| \leq \frac{\left\|G_{\diag\left(\mathbf{N}_1(z_{k-1})\right)^{-1}}(x_{k-1})\right\|}{L}, \\
            & \left\| \left(L\diag\left(\mathbf{N}_1(z_{k-1})\right) - \mathbf{I}\right)\frac{G_{\diag\left(\mathbf{N}_1(z_{k-1})\right)^{-1}}(x_{k-1})}{L} \right\| \leq \frac{\left\|G_{\diag\left(\mathbf{N}_1(z_{k-1})\right)^{-1}}(x_{k-1})\right\|}{L} .
        \end{aligned}
    \end{equation*}

    Solve the above inequation, we have the following upper bound of $\mathbf{N}_1(z_{k-1})$:
    \begin{equation*}
        \mathbf{N}_1(z_{k-1}) \in \left[\mathbf{0},  \frac{2}{L}\mathbf{1} \right].
    \end{equation*}
    
    Furthermore, each choice of $\mathbf{N}_1(z_{k-1})$ yields a range of $\mathbf{N}_2(z_{k-1})$. For example, $\mathbf{N}_1(z_{k-1}) := 0$ yields the following inequality for $\mathbf{N}_1(z_{k-1})$:
    \begin{equation*}
        \begin{aligned}
            & \left\| \mathbf{N}_2(z_{k-1}) - \frac{G_{\diag\left(\mathbf{N}_1(z_{k-1})\right)^{-1}}(x_{k-1})}{L}\right\| \leq \frac{\left\|G_{\diag\left(\mathbf{N}_1(z_{k-1})\right)^{-1}}(x_{k-1})\right\|}{L}.
        \end{aligned}
    \end{equation*}
    Solve the above inequation, $\mathbf{N}_2(z_{k-1})$ is bounded as follows:
    \begin{equation*}
        \mathbf{N}_2(z_{k-1}) \in \left[\mathbf{0},  \frac{2}{L}|G_{\diag\left(\mathbf{N}_1(z_{k-1})\right)^{-1}}(x_{k-1})| \right].
    \end{equation*}
    For example, $\mathbf{N}_1(z_{k-1}) := \frac{2}{L}\mathbf{1}$ yields:
    \begin{equation*}
        \mathbf{N}_2(z_{k-1}) \in \left[-\frac{2}{L}|G_{\diag\left(\mathbf{N}_1(z_{k-1})\right)^{-1}}(x_{k-1})|, \mathbf{0}  \right].
    \end{equation*}

    Replacing the $G_{\diag\left(\mathbf{N}_1(z_{k-1})\right)^{-1}}(x_{k-1})$ in inequation \ref{thpf:ind_converge_gain_composite_ineq} with $G_{\diag\left(\mathbf{N}_1(z_{k-1})\right)^{-1}}(x_{k-1}) = \nabla f(x_{k-1}) + g_{k}$ in \eqref{thpf:G_fg_equation_composite} yields:
    \begin{equation*}
        \begin{aligned}
            & \left\| \diag\left(\mathbf{N}_1(z_{k-1})\right)\big(\nabla f(x_{k-1}) + g_{k}\big) +\mathbf{N}_2(z_{k-1}) - \frac{\nabla f(x_{k-1}) + g_{k}}{L} \right\| \leq \left\|\frac{\nabla f(x_{k-1}) + g_{k}}{L} \right\|.
        \end{aligned}
    \end{equation*}
\end{proof}

Similar to Corollary \ref{ob:obs_best_converge_gain} for the smooth case, for the composite case, we propose the following corollary to achieve the best robust L2O model with the largest per iteration convergence gain.
\begin{corollary} \label{ob:composite_ind_best_converge_gain}
    For any $z_{k-1} \in \mathcal{Z}_P$, we let: 
    \begin{equation*}
        \mathbf{N}_1(z_{k-1}) := \frac{1}{L}\mathbf{1}, \mathbf{N}_2(z_{k-1}) := \mathbf{0}, 
    \end{equation*}
    the Math-L2O model in \eqref{eq:l2o_smooth} is exactly gradient descent update with convergence rate:
    \begin{equation*}
        F(x_K) - F(x^*) \leq \frac{L}{2K} \|x_0 -x^*\|^2.
    \end{equation*}
\end{corollary}
\begin{proof}
    In the last term of inequality \ref{thpf:converge_gain_composite_ineq}, the best convergence gain yields:
    \begin{equation} \label{thpf:ind_GM_composite_best_l2ogd}
        \left\| \diag\left(\mathbf{N}_1(z_{k-1})\right)G_{\diag\left(\mathbf{N}_1(z_{k-1})\right)^{-1}}(x_{k-1}) +\mathbf{N}_2(z_{k-1}) - \frac{G_{\diag\left(\mathbf{N}_1(z_{k-1})\right)^{-1}}(x_{k-1})}{L}\right\| := 0.
    \end{equation}
    $\mathbf{N}_1(z_{k-1}) = \frac{1}{L}\mathbf{1}, \mathbf{N}_2(z_{k-1}) = \mathbf{0}$ is a feasible solution.

    Given $\mathbf{N}_1(z_{k-1}) = \frac{1}{L}\mathbf{1}, \mathbf{N}_2(z_{k-1}) = \mathbf{0}$, the update formula in \eqref{thpf:ind_GM_composite} is:
    \begin{equation} \label{thpf:ind_GM_composite_best}
        \begin{aligned}
            x_{k} = x_{k-1} - \frac{1}{L}G_{\diag\left(\mathbf{N}_1(z_{k-1})\right)^{-1}}(x_{k-1}).
        \end{aligned}
    \end{equation}

    Based on Lemma~\ref{thpf:lm_GM_basic_bound_composite}, set $t := x^*$, we have the following inequality between the objective at $k$-th iteration and the optimum: 
    \begin{equation*}
        \begin{aligned}
            &F(x_{k}) - F(x^*)\\
            \leq& G_{\diag\left(\mathbf{N}_1(z_{k-1})\right)^{-1}}(x_{k-1})^\top (x_{k-1} - x^*)\\
            &+\frac{L}{2} \Bigg(\left\| \diag\left(\mathbf{N}_1(z_{k-1})\right)G_{\diag\left(\mathbf{N}_1(z_{k-1})\right)^{-1}}(x_{k-1}) +\mathbf{N}_2(z_{k-1}) - \frac{G_{\diag\left(\mathbf{N}_1(z_{k-1})\right)^{-1}}(x_{k-1})}{L}\right\|^2 \\
            & \quad \qquad - \left\|\frac{G_{\diag\left(\mathbf{N}_1(z_{k-1})\right)^{-1}}(x_{k-1})}{L}\right\|^2\Bigg), \\
            =& G_{\diag\left(\mathbf{N}_1(z_{k-1})\right)^{-1}}(x_{k-1})^\top (x_{k-1} - x^*) - \frac{L}{2} \left\|\frac{G_{\diag\left(\mathbf{N}_1(z_{k-1})\right)^{-1}}(x_{k-1})}{L}\right\|^2, \\
            =& \frac{L}{2} \left(\frac{2}{L} G_{\diag\left(\mathbf{N}_1(z_{k-1})\right)^{-1}}(x_{k-1})^\top (x_{k-1} - x^*) -  \left\|\frac{G_{\diag\left(\mathbf{N}_1(z_{k-1})\right)^{-1}}(x_{k-1})}{L}\right\|^2\right), \\
            =& \frac{L}{2} \left(\|x_{k-1} - x^*\|^2-  \left\|x_{k-1} - x^* - \frac{G_{\diag\left(\mathbf{N}_1(z_{k-1})\right)^{-1}}(x_{k-1})}{L}\right\|^2\right), \\
            =& \frac{L}{2} (\|x_{k-1} - x^*\|^2-  \|x_k - x^*\|^2), \\
        \end{aligned}
    \end{equation*}
    where in the second step, we apply the equality in \eqref{thpf:ind_GM_composite_best_l2ogd} and remove the degenerated terms. In the 4th step, we make up a perfect square. In the 5th step, we apply the L2O model's update formula in \eqref{thpf:ind_GM_composite_best}.

    Sum over $K$ iterations yields:
    \begin{equation} \label{thpf:ind_converge_rate_composite_ineq}
        \begin{aligned}
            F(x_{K}) - F(x^*) \leq \frac{L}{2K} (\|x_{0} - x^*\|^2-  \|x_{K} - x^*\|^2) \leq \frac{L}{2K} \|x_{0} - x^*\|^2.
        \end{aligned}
    \end{equation}
\end{proof}

\subsubsection*{OOD Definitions}
We first derive some preliminary formulations before the convergence analysis for OOD scenarios.

We make the following assumptions identical to those in the smooth case. 
Suppose $z, \tilde{z}, z^\prime \in \mathcal{Z}$ are input feature vectors of the L2O model. There exists a vector $\alpha \in [0,1]$ that $z^\prime := \alpha z + (1-\alpha)\tilde{z}, z^\prime \in \mathcal{Z}$. Denote virtual Jacobian matrix of $\mathbf{N}_1(z^\prime)$ and $\mathbf{N}_2(z^\prime)$ at point $z^\prime$ as $\mathbf{J}_1$ and $\mathbf{J}_2$.

Since $\mathbf{N}_1(z)$ and $\mathbf{N}_2(z)$ are smooth, due to the Mean Value Theorem, we have the following equalities:
\begin{equation*}
    \mathbf{N}_1(z) = \mathbf{N}_1(\tilde{z}) + \mathbf{J}_1(z - \tilde{z}),\quad \mathbf{N}_2(z) = \mathbf{N}_2(\tilde{z}) + \mathbf{J}_2(z - \tilde{z}).
\end{equation*}
As demonstrated in the preliminary of the smooth case, we have $\|\mathbf{J}_1\| \leq \sqrt{n}C_1$, and $\|\mathbf{J}_2\| \leq \sqrt{n}C_2$. 

Given a virtual variable $s \in \mathbb{R}^n$ to represent the OOD shifting on variable $x$, define the virtual feature (difference of L2O model's input feature between OOD and InD scenarios) as $s^\prime = [s^\top, (\nabla f^\prime(x+s) - \nabla f(x))^\top, (g^\prime - g)^\top]^\top$, where $g^\prime \in \partial r^\prime(x+s), g \in \partial r(x)$ are subgradient instances of OOD and InD scenarios respectively. We have the following equations to formulate the L2O model's behaviors in OOD and InD scenarios:
\begin{equation} \label{thpf:lb_composite}
    \begin{aligned}
        &\mathbf{N}_1(z+s^\prime) = \mathbf{N}_1(z) + \mathbf{J}_1(z + s^\prime - z) = \mathbf{N}_1(z) + \mathbf{J}_1 s^\prime\\
        &\mathbf{N}_2(z+s^\prime) = \mathbf{N}_2(z) + \mathbf{J}_2(z + s^\prime - z) = \mathbf{N}_2(z) + \mathbf{J}_2 s^\prime.
    \end{aligned}
\end{equation}

% \begin{equation} \label{thpf:ub_composite}
%     \begin{aligned}
%         \mathbf{N}_1(z+s^\prime) \leq \mathbf{N}_1(z)+\mathbf{J}_{1, z}(z+s^\prime-z)+\frac{C_1}{2}\|z+s^\prime-z\|^2 = \mathbf{N}_1(z)+\mathbf{J}_{1, z}(s^\prime)+\frac{C_1}{2}\|s^\prime\|^2\\
%         \mathbf{N}_2(z+s^\prime) \leq \mathbf{N}_2(z)+\mathbf{J}_{2, z}(z+s^\prime-z)+\frac{C_2}{2}\|z+s^\prime-z\|^2 = \mathbf{N}_2(z)+\mathbf{J}_{2, z}(s^\prime)+\frac{C_2}{2}\|s^\prime\|^2
%     \end{aligned}
% \end{equation}

Based on Lemma~\ref{thpf:lm_GM_basic_bound_composite}, $\forall x_k \in \mathcal{S}_p, s_k \in \mathbb{R}^n$, OOD yields the following inequality between any two values of objective:
\begin{equation} \label{thpf:composite_ood_ub_pre1}
    \begin{aligned}
        &F^\prime(x_{k} + s_{k}) \\
        \leq&  F^\prime(t) + G_{\diag(\mathbf{N}_1(z_{k-1} + s^\prime_{k-1}))^{-1}}(x_{k-1} +s_{k-1})^\top (x_{k-1} +s_{k-1} - t) \\
        & + \frac{L}{2} \Bigg\| \diag(\mathbf{N}_1(z_{k-1} + s^\prime_{k-1}))G_{\diag(\mathbf{N}_1(z_{k-1} + s^\prime_{k-1}))^{-1}}(x_{k-1} +s_{k-1}) +\mathbf{N}_2(z_{k-1} + s^\prime_{k-1}) \\
        & \qquad - \frac{G_{\diag(\mathbf{N}_1(z_{k-1} + s^\prime_{k-1}))^{-1}}(x_{k-1} +s_{k-1})}{L}\Bigg\|^2 - \frac{L}{2}\left\|\frac{G_{\diag(\mathbf{N}_1(z_{k-1} + s^\prime_{k-1}))^{-1}}(x_{k-1} +s_{k-1})}{L}\right\|^2.
    \end{aligned}
\end{equation}

For gradient mapping, OOD yields:
\begin{equation*}
    \begin{aligned}
        &-\diag(\mathbf{N}_{1}(z_{k-1}+s^\prime_{k-1}))^{-1}\left(x_{k} +s_{k}- \left( x_{k-1}+s_{k-1} - \diag(\mathbf{N}_{1}(z_{k-1}+s^\prime_{k-1}))\nabla f^\prime(x_{k-1}+s_{k-1}) - \mathbf{N}_{2}(z_{k-1}+s^\prime_{k-1})\right)\right) \\
        =&  -\diag(\mathbf{N}_{1}(z_{k-1}+s^\prime_{k-1}))^{-1} \\
        & \quad \Big(x_{k-1}+s_{k-1} - \diag(\mathbf{N}_1(z_{k-1}+s^\prime_{k-1}))G^\prime_{\diag(\mathbf{N}_1(z_{k-1}+s^\prime_{k-1}))^{-1}}(x_{k-1}+s_{k-1}) -\mathbf{N}_2(z_{k-1}+s^\prime_{k-1}) \\
        & \qquad - \left( x_{k-1}+s_{k-1} - \diag(\mathbf{N}_1(z_{k-1}+s^\prime_{k-1}))\nabla f^\prime(x_{k-1}+s_{k-1}) - \mathbf{N}_{2}(z_{k-1}+s^\prime_{k-1})\right) \Big), \\
        =&  G^\prime_{\diag(\mathbf{N}_1(z_{k-1}+s^\prime_{k-1}))^{-1}}(x_{k-1}+s_{k-1}) - \nabla f^\prime(x_{k-1}+s_{k-1}),
    \end{aligned}
\end{equation*}
where we use $G^\prime_{\diag(\mathbf{N}_1(z_{k-1}+s^\prime_{k-1}))^{-1}}$ to represent the gradient map in the OOD scenario.

Moreover, similar to \eqref{thpf:G_fg_equation_composite}, we have the following formulation of using gradient map to represent gradient and subgradient in the OOD scenario:
\begin{equation} \label{thpf:G_fg_equation_ood_composite}
    G^\prime_{\diag(\mathbf{N}_1(z_{k-1} + s^\prime_{k-1}))^{-1}}(x_{k-1} +s_{k-1}) =  \nabla f^\prime(x_{k-1} +s_{k-1}) + g^\prime_{k},
\end{equation}
where $g^\prime_{k} \in \partial r^\prime(x_{k} +s_{k})$.

Similar to Assumption~\ref{as:assump1_ind_robust} for the smooth case, we derive the following assumption to ensure the most robust L2O model for the composite case.
\begin{assumption} \label{as:assump_ind_robust_composite}
    After training, $\forall x_{k-1} \in \mathcal{S}_P, \forall z_{k-1} \in \mathcal{Z}_P$, $\mathbf{N}_1(z_{k-1}):= \frac{1}{L}\mathbf{1}$ and $\mathbf{N}_2(z_{k-1}) := \mathbf{0}$.
\end{assumption}

\subsection{OOD Per-Iteration Convergence Gain}
Based on the Lemma~\ref{lm:ind_converge_gain_composite} and Corollary~\ref{ob:composite_ind_best_converge_gain}, Assumption \ref{as:assump_ind_robust_composite} leads to an L2O model with best robustness on all InD instances. In the following theorem, we quantify the diminution in convergence rate instigated by the virtual feature $s^\prime$ defined in \cref{sec:vv_traj}. 

\begin{theorem} \label{lm:ood_converge_gain_composite}
    Under Assumption \ref{as:assump_ind_robust_composite}, there exists virtual Jacobian matrices $\mathbf{J}_{1,k-1}, \mathbf{J}_{2,k-1}, k= 1,2, \dots, K$ that OOD's convergence improvement of one iteration is upper bounded by following inequality:
    \begin{equation*}
        \begin{aligned}
            & F^\prime(x_k + s_k) - F^\prime(x_{k-1} + s_{k-1}) \\
             \leq &-\frac{\|\nabla f^\prime(x_{k-1} +s_{k-1}) + g^\prime_{k}\|^2}{2L}   + \frac{L}{2} \| \diag(\mathbf{J}_1 s^\prime_{k-1})(\nabla f^\prime(x_{k-1} +s_{k-1}) + g^\prime_{k}) + \mathbf{J}_2 s^\prime_{k-1} \|^2,
        \end{aligned}
    \end{equation*}
    where $g^\prime_{k} \in \partial r^\prime(x_{k} +s_{k})$.
\end{theorem}

\begin{proof}
    From \eqref{thpf:lb_composite}, we have the following reformulations of $\mathbf{N}_1(z_{k-1} + s^\prime_{k-1})$ and $\mathbf{N}_2(z_{k-1} + s^\prime_{k-1})$:
    \begin{equation*}
        \begin{aligned}
            &\mathbf{N}_1(z_{k-1} + s^\prime_{k-1})  = \mathbf{N}_1(z_{k-1}) + \mathbf{J}_1 s^\prime_{k-1},\\
            &\mathbf{N}_2(z_{k-1} + s^\prime_{k-1}) = \mathbf{N}_2(z_{k-1}) + \mathbf{J}_2 s^\prime_{k-1}.
        \end{aligned}
    \end{equation*}

    Substituting the definitions of $\mathbf{N}_1(z_{k-1})$ and $\mathbf{N}_2(z_{k-1})$ in Assumption \ref{as:assump_ind_robust_composite} yields:
    \begin{equation} \label{thpf:ood_N1N2_composite}
        \begin{aligned}
            &\mathbf{N}_1(z_{k-1} + s^\prime_{k-1})  = \frac{1}{L}\mathbf{1} + \mathbf{J}_1 s^\prime_{k-1}\\
            &\mathbf{N}_2(z_{k-1} + s^\prime_{k-1}) = \mathbf{J}_2 s^\prime_{k-1}.
        \end{aligned}
    \end{equation}

    We then apply construct inequality between objectives of two adjacent iterations. Substituting $t := x_{k-1} +s_{k-1}$ into inequality \ref{thpf:composite_ood_ub_pre1} yields:
    \begin{equation*}
        \begin{aligned}
            &F^\prime(x_{k} + s_{k}) - F^\prime(x_{k-1} +s_{k-1})\\
            \leq&  \frac{L}{2} \Bigg\| \diag(\mathbf{N}_1(z_{k-1} + s^\prime_{k-1}))G^\prime_{\diag(\mathbf{N}_1(z_{k-1} + s^\prime_{k-1}))^{-1}}(x_{k-1} +s_{k-1}) +\mathbf{N}_2(z_{k-1} + s^\prime_{k-1}) \\
            & \quad - \frac{G^\prime_{\diag(\mathbf{N}_1(z_{k-1} + s^\prime_{k-1}))^{-1}}(x_{k-1} +s_{k-1})}{L}\Bigg\|^2 - \frac{L}{2}\Bigg\|\frac{G^\prime_{\diag(\mathbf{N}_1(z_{k-1} + s^\prime_{k-1}))^{-1}}(x_{k-1} +s_{k-1})}{L}\Bigg\|^2.
        \end{aligned}
    \end{equation*}

    Substituting \eqref{thpf:ood_N1N2_composite} into above inequality yields:
    \begin{equation*}
        \begin{aligned}
            &F^\prime(x_{k} + s_{k}) - F^\prime(x_{k-1} +s_{k-1})\\
            \leq&  \frac{L}{2} \| \diag(\mathbf{J}_1 s^\prime_{k-1})G^\prime_{\diag(\mathbf{N}_1(z_{k-1} + s^\prime_{k-1}))^{-1}}(x_{k-1} +s_{k-1}) + \mathbf{J}_2 s^\prime_{k-1} \|^2  - \frac{L}{2}\left\|\frac{G^\prime_{\diag(\mathbf{N}_1(z_{k-1} + s^\prime_{k-1}))^{-1}}(x_{k-1} +s_{k-1})}{L} \right\|^2.
        \end{aligned}
    \end{equation*}

    Based on \eqref{thpf:ood_N1N2_composite}, we recover gradient and subgradient from gradient map:
    \begin{equation*}
        \begin{aligned}
            &F^\prime(x_{k} + s_{k}) - F^\prime(x_{k-1} +s_{k-1})\\
            \leq&   - \frac{L}{2}\left\|\frac{\nabla f^\prime(x_{k-1} +s_{k-1}) + g^\prime_{k}}{L}\right\|^2 + \frac{L}{2} \| \diag(\mathbf{J}_1 s^\prime_{k-1})(\nabla f^\prime(x_{k-1} +s_{k-1}) + g^\prime_{k}) + \mathbf{J}_2 s^\prime_{k-1} \|^2 , \\
            =& - \frac{\|\nabla f^\prime(x_{k-1} +s_{k-1}) + g^\prime_{k}\|^2}{2L} + \frac{L}{2} \| \diag(\mathbf{J}_1 s^\prime_{k-1})(\nabla f^\prime(x_{k-1} +s_{k-1}) + g^\prime_{k}) + \mathbf{J}_2 s^\prime_{k-1} \|^2.
        \end{aligned}
    \end{equation*}
\end{proof}

Moreover, we derive the upper bound of per iteration convergence gain in the following Corollary~\ref{lm:ood_converge_gain_ub_composite}.
\begin{corollary} \label{lm:ood_converge_gain_ub_composite}
    Under Assumption \ref{as:assump_ind_robust_composite}, the convergence improvement for one iteration of the OOD scenario can be upper bounded w.r.t. $\| s_{k-1}^\prime \|$ by:
    \begin{equation*} 
        \begin{aligned}
            & F^\prime(x_k + s_k) - F^\prime(x_{k-1} + s_{k-1})  \\
            \leq & -\frac{\|\nabla f^\prime(x_{k-1} +s_{k-1}) + g^\prime_{k}\|^2}{2L} +  ( Ln^2C_1^2 \| \nabla f^\prime(x_{k-1} +s_{k-1}) + g^\prime_{k} \|^2  +  Ln^2C_2^2) \| s^\prime_{k-1} \|^2,
        \end{aligned}
    \end{equation*}
    where $g^\prime_{k} \in \partial r^\prime(x_{k} +s_{k})$.
\end{corollary}
\begin{proof}
    Based on Triangle and Cauchy-Schwarz inequalities, we have:
    \begin{equation*}
        \begin{aligned}
            & F^\prime(x_k + s_k) - F^\prime(x_{k-1} + s_{k-1}) \\
            \leq &-\frac{\|\nabla f^\prime(x_{k-1} +s_{k-1}) + g^\prime_{k}\|^2}{2L}   + \frac{L}{2} \| \diag(\mathbf{J}_1 s^\prime_{k-1})(\nabla f^\prime(x_{k-1} +s_{k-1}) + g^\prime_{k}) + \mathbf{J}_2 s^\prime_{k-1} \|^2 , \\
            \leq &-\frac{\|\nabla f^\prime(x_{k-1} +s_{k-1}) + g^\prime_{k}\|^2}{2L}   + L \| \diag(\mathbf{J}_1 s^\prime_{k-1})(\nabla f^\prime(x_{k-1} +s_{k-1}) + g^\prime_{k}) \|^2 +  L \|\mathbf{J}_2 s^\prime_{k-1} \|^2 , \\
            \leq &-\frac{\|\nabla f^\prime(x_{k-1} +s_{k-1}) + g^\prime_{k}\|^2}{2L}   + L \| \mathbf{J}_1 s^\prime_{k-1} \|^2 \| \nabla f^\prime(x_{k-1} +s_{k-1}) + g^\prime_{k} \|^2 +  L \|\mathbf{J}_2 s^\prime_{k-1} \|^2 , \\
            \leq &-\frac{\|\nabla f^\prime(x_{k-1} +s_{k-1}) + g^\prime_{k}\|^2}{2L}   + L n^2C_1^2   \| \nabla f^\prime(x_{k-1} +s_{k-1}) + g^\prime_{k} \|^2 \| s^\prime_{k-1} \|^2 +  L n^2C_2^2 \| s^\prime_{k-1} \|^2 , \\
            = &-\frac{\|\nabla f^\prime(x_{k-1} +s_{k-1}) + g^\prime_{k}\|^2}{2L} +  ( Ln^2C_1^2 \| \nabla f^\prime(x_{k-1} +s_{k-1}) + g^\prime_{k} \|^2  +  Ln^2C_2^2) \| s^\prime_{k-1} \|^2,
        \end{aligned}
    \end{equation*}
    where $g^\prime_{k} \in \partial r^\prime(x_{k} +s_{k})$.
\end{proof}

\subsection{OOD Multi-Iteration Convergence Rate}
\begin{theorem} \label{th:ood_converge_rate_composite}
    Under Assumption \ref{as:assump_ind_robust_composite}, OOD's convergence rate of $K$ iterations is upper bounded by:
    \begin{equation*} 
        \begin{aligned}
            &\min_{k=1,\dots, K}F^\prime(x_k+s_k) - F^\prime(x^* +s^*)\\
            \leq & \frac{L}{2K} \|x_{0} +s_{0} - x^*-s^*\|^2 - \frac{L}{2K}\|x_{K} +s_{K} - x^*-s^* \|^2 \\
            & + \frac{L}{K} \sum_{k=1}^{K} \left( x_k +s_k -x_{k-1}-s_{k-1} +\frac{\nabla f^\prime(x_{k-1} +s_{k-1}) + g^\prime_{k}}{L}\right)^\top (x_{k} +s_{k} - x^*-s^* ).
        \end{aligned}
    \end{equation*}
\end{theorem}

\begin{proof}
    We construct the relationship between $k$-th iteration's and optimal objectives by substituting $t := x^* +s^*$ into inequality \ref{thpf:composite_ood_ub_pre1} yields:
    \begin{equation*}
        \begin{aligned}
            &F^\prime(x_{k} + s_{k}) - F^\prime(x^* +s^*)\\
            \leq& G^\prime_{\diag(\mathbf{N}_1(z_{k-1} + s^\prime_{k-1}))^{-1}}(x_{k-1} +s_{k-1})^\top (x_{k-1} +s_{k-1} - x^*-s^*)\\
            &+ \frac{L}{2} \Bigg\| \diag(\mathbf{N}_1(z_{k-1} + s^\prime_{k-1}))G^\prime_{\diag(\mathbf{N}_1(z_{k-1} + s^\prime_{k-1}))^{-1}}(x_{k-1} +s_{k-1}) +\mathbf{N}_2(z_{k-1} + s^\prime_{k-1}) \\
            & \qquad \quad - \frac{G^\prime_{\diag(\mathbf{N}_1(z_{k-1} + s^\prime_{k-1}))^{-1}}(x_{k-1} +s_{k-1})}{L}\Bigg\|^2 - \frac{L}{2}\Bigg\|\frac{G^\prime_{\diag(\mathbf{N}_1(z_{k-1} + s^\prime_{k-1}))^{-1}}(x_{k-1} +s_{k-1})}{L}\Bigg\|^2.
        \end{aligned}
    \end{equation*}

    We eliminate InD terms by substituting \eqref{thpf:ood_N1N2_composite} into above inequality yields:
    \begin{equation*}
        \begin{aligned}
            &F^\prime(x_{k} + s_{k}) - F^\prime(x_{k-1} +s_{k-1})\\
            \leq&  G^\prime_{\diag(\mathbf{N}_1(z_{k-1} + s^\prime_{k-1}))^{-1}}(x_{k-1} +s_{k-1})^\top (x_{k-1} +s_{k-1} - x^*-s^*) \\
            & + \frac{L}{2} \left\| \diag(\mathbf{J}_1 s^\prime_{k-1})G^\prime_{\diag(\mathbf{N}_1(z_{k-1} + s^\prime_{k-1}))^{-1}}(x_{k-1} +s_{k-1}) + \mathbf{J}_2 s^\prime_{k-1} \right\|^2 - \frac{L}{2}\left\|\frac{G^\prime_{\diag(\mathbf{N}_1(z_{k-1} + s^\prime_{k-1}))^{-1}}(x_{k-1} +s_{k-1})}{L} \right\|^2.
        \end{aligned}
    \end{equation*}
    
    Then, we recover the gradient and subgradient from the gradient map by substituting \eqref{thpf:G_fg_equation_ood_composite} into the above inequality yields:
    \begin{equation} \label{thpf:ood_converge_rate_pre_composite}
        \begin{aligned}
            &F^\prime(x_{k} + s_{k}) - F^\prime(x_{k-1} +s_{k-1})\\
            \leq&  (\nabla f^\prime(x_{k-1} +s_{k-1}) + g^\prime_{k})^\top (x_{k-1} +s_{k-1} - x^*-s^*) - \frac{L}{2}\left\|\frac{\nabla f^\prime(x_{k-1} +s_{k-1}) + g^\prime_{k}}{L}\right\|^2 \\
            & + \frac{L}{2} \left\| \diag(\mathbf{J}_1 s^\prime_{k-1})(\nabla f^\prime(x_{k-1} +s_{k-1}) + g^\prime_{k}) + \mathbf{J}_2 s^\prime_{k-1} \right\|^2 , \\
            =& \frac{L}{2} \left(2 \frac{\nabla f^\prime(x_{k-1} +s_{k-1}) + g^\prime_{k}}{L}^\top (x_{k-1} +s_{k-1} - x^*-s^*) - \left\|\frac{\nabla f^\prime(x_{k-1} +s_{k-1}) + g^\prime_{k}}{L}\right\|^2\right) \\
            & + \frac{L}{2} \| \diag(\mathbf{J}_1 s^\prime_{k-1})(\nabla f^\prime(x_{k-1} +s_{k-1}) + g^\prime_{k}) + \mathbf{J}_2 s^\prime_{k-1} \|^2.
        \end{aligned}
    \end{equation}

    By the definition of $G^\prime_{\diag(\mathbf{N}_1(z_{k}))^{-1}}(x_{k-1} + s_{k-1})$ in \eqref{thpf:G_fg_equation_ood_composite}, we can represent $x_{k} +s_{k}$ by the following equation:
    \begin{equation} \label{thpf:ood_l2o_composite}
        \begin{aligned}
            x_{k} + s_{k}  &= \operatorname{prox}_{\diag(\mathbf{N}_1(z_{k-1} + s^\prime_{k-1}))^{-1}}(x_{k-1} + s_{k-1} - \diag(\mathbf{N}_1(z_{k-1} + s^\prime_{k-1}))\nabla f(x_{k-1} + s_{k-1}) - \mathbf{N}_2(z_{k-1} + s^\prime_{k-1})), \\
            &= x_{k-1} + s_{k-1} - \diag(\mathbf{N}_1(z_{k-1} + s^\prime_{k-1}))G^\prime_{\diag(\mathbf{N}_1(z_{k-1} + s^\prime_{k-1}))^{-1}}(x_{k-1} + s_{k-1}) -\mathbf{N}_2(z_{k-1} + s^\prime_{k-1}), \\
            &= x_{k-1} + s_{k-1} - \diag(\frac{1}{L}\mathbf{1} + \mathbf{J}_1 s^\prime_{k-1})(\nabla f^\prime(x_{k-1} +s_{k-1}) + g^\prime_{k}) -\mathbf{J}_2 s^\prime_{k-1}, \\
            &= x_{k-1} + s_{k-1} - \frac{\nabla f^\prime(x_{k-1} +s_{k-1}) + g^\prime_{k}}{L} - \diag(\mathbf{J}_1 s^\prime_{k-1})(\nabla f^\prime(x_{k-1} +s_{k-1}) + g^\prime_{k}) -\mathbf{J}_2 s^\prime_{k-1},
        \end{aligned}
    \end{equation}
    where $g^\prime_{k} \in \partial r^\prime(x_{k} +s_{k})$ is a subgradient vector.

    Similarly, we aim to make up a perfect square in \eqref{thpf:ood_converge_rate_pre_composite} with the above formulation of the update given by the L2O model in the OOD scenario. The demonstration is as follows.

    First, in order to apply \eqref{thpf:ood_l2o_composite}, we would like to make up several terms of \eqref{thpf:ood_l2o_composite}:
    \begin{equation*}
        \begin{aligned}
            &F^\prime(x_{k} + s_{k}) - F^\prime(x_{k-1} +s_{k-1})\\
            \leq& \frac{L}{2} (2 \frac{\nabla f^\prime(x_{k-1} +s_{k-1}) + g^\prime_{k}}{L}^\top (x_{k-1} +s_{k-1} - x^*-s^*) - \|\frac{\nabla f^\prime(x_{k-1} +s_{k-1}) + g^\prime_{k}}{L}\|^2) \\
            & + \frac{L}{2} \| \diag(\mathbf{J}_1 s^\prime_{k-1})(\nabla f^\prime(x_{k-1} +s_{k-1}) + g^\prime_{k}) + \mathbf{J}_2 s^\prime_{k-1} \|^2 , \\
            =& \frac{L}{2} \Big(2 \big(\diag(\frac{1}{L}\mathbf{1} + \mathbf{J}_1 s^\prime_{k-1})(\nabla f^\prime(x_{k-1} +s_{k-1}) + g^\prime_{k}) - \diag(\mathbf{J}_1 s^\prime_{k-1})(\nabla f^\prime(x_{k-1} +s_{k-1}) + g^\prime_{k})\big)^\top (x_{k-1} +s_{k-1} - x^*-s^*) \\
            & - \|\diag(\frac{1}{L}\mathbf{1} + \mathbf{J}_1 s^\prime_{k-1})(\nabla f^\prime(x_{k-1} +s_{k-1}) + g^\prime_{k}) - \diag(\mathbf{J}_1 s^\prime_{k-1})(\nabla f^\prime(x_{k-1} +s_{k-1}) + g^\prime_{k})\|^2 \Big) \\
            & + \frac{L}{2} \| \diag(\mathbf{J}_1 s^\prime_{k-1})(\nabla f^\prime(x_{k-1} +s_{k-1}) + g^\prime_{k}) + \mathbf{J}_2 s^\prime_{k-1} \|^2 , \\
            =& \frac{L}{2} \Big(2 \big(\diag(\frac{1}{L}\mathbf{1} + \mathbf{J}_1 s^\prime_{k-1})(\nabla f^\prime(x_{k-1} +s_{k-1}) + g^\prime_{k})+\mathbf{J}_2 s^\prime_{k-1} \\
            & \qquad - \diag(\mathbf{J}_1 s^\prime_{k-1})(\nabla f^\prime(x_{k-1} +s_{k-1}) + g^\prime_{k})-\mathbf{J}_2 s^\prime_{k-1} \big)^\top (x_{k-1} +s_{k-1} - x^*-s^*) \\
            & \qquad - \|\diag(\frac{1}{L}\mathbf{1} + \mathbf{J}_1 s^\prime_{k-1})(\nabla f^\prime(x_{k-1} +s_{k-1}) + g^\prime_{k})+\mathbf{J}_2 s^\prime_{k-1} - \diag(\mathbf{J}_1 s^\prime_{k-1})(\nabla f^\prime(x_{k-1} +s_{k-1}) + g^\prime_{k})-\mathbf{J}_2 s^\prime_{k-1} \|^2 \Big) \\
            & + \frac{L}{2} \| \diag(\mathbf{J}_1 s^\prime_{k-1})(\nabla f^\prime(x_{k-1} +s_{k-1}) + g^\prime_{k}) + \mathbf{J}_2 s^\prime_{k-1} \|^2,
        \end{aligned}
    \end{equation*}
    where in the first step, we makeup the $\diag(\frac{1}{L}\mathbf{1} + \mathbf{J}_1 s^\prime_{k-1})(\nabla f^\prime(x_{k-1} +s_{k-1}) + g^\prime_{k})$ of \eqref{thpf:ood_l2o_composite}. In the second step, we makeup the $\mathbf{J}_2 s^\prime_{k-1}$ of \eqref{thpf:ood_l2o_composite}. 
    
    Then, we expand the quadratic term in the second line of above inequation's right-hand side and merge similar terms:
    \begin{equation*}
        \begin{aligned}
            &F^\prime(x_{k} + s_{k}) - F^\prime(x_{k-1} +s_{k-1})\\
            \leq& \frac{L}{2} \Big(2 \big( \diag(\frac{1}{L}\mathbf{1} + \mathbf{J}_1 s^\prime_{k-1})(\nabla f^\prime(x_{k-1} +s_{k-1}) + g^\prime_{k})+\mathbf{J}_2 s^\prime_{k-1}\big)^\top (x_{k-1} +s_{k-1} - x^*-s^*) \\
            & \quad - \|\diag(\frac{1}{L}\mathbf{1} + \mathbf{J}_1 s^\prime_{k-1})(\nabla f^\prime(x_{k-1} +s_{k-1}) + g^\prime_{k})+\mathbf{J}_2 s^\prime_{k-1}\|^2 \Big) \\
            &- 2\frac{L}{2} \big( \diag(\mathbf{J}_1 s^\prime_{k-1})(\nabla f^\prime(x_{k-1} +s_{k-1}) + g^\prime_{k})+\mathbf{J}_2 s^\prime_{k-1} \big)^\top (x_{k-1} +s_{k-1} - x^*-s^*) \\
            & + 2\frac{L}{2} \big(\diag(\frac{1}{L}\mathbf{1} + \mathbf{J}_1 s^\prime_{k-1})(\nabla f^\prime(x_{k-1} +s_{k-1}) + g^\prime_{k}) +\mathbf{J}_2 s^\prime_{k-1}\big)^\top \big(\diag(\mathbf{J}_1 s^\prime_{k-1})(\nabla f^\prime(x_{k-1} +s_{k-1}) + g^\prime_{k}) + \mathbf{J}_2 s^\prime_{k-1}\big)\\
            & - \frac{L}{2} \| \diag(\mathbf{J}_1 s^\prime_{k-1})(\nabla f^\prime(x_{k-1} +s_{k-1}) + g^\prime_{k})+\mathbf{J}_2 s^\prime_{k-1} \|^2  + \frac{L}{2} \| \diag(\mathbf{J}_1 s^\prime_{k-1})(\nabla f^\prime(x_{k-1} +s_{k-1}) + g^\prime_{k}) + \mathbf{J}_2 s^\prime_{k-1} \|^2.
        \end{aligned}
    \end{equation*}

    Finially, we are able to make up a perfect square on the first two lines of the right-hand side:
    \begin{equation*}
        \begin{aligned}
            &F^\prime(x_{k} + s_{k}) - F^\prime(x_{k-1} +s_{k-1})\\
            \leq& \frac{L}{2} (\|x_{k-1} +s_{k-1} - x^*-s^*\|^2 - \|x_{k-1} +s_{k-1} - x^*-s^* - \diag(\frac{1}{L}\mathbf{1} + \mathbf{J}_1 s^\prime_{k-1})(\nabla f^\prime(x_{k-1} +s_{k-1}) + g^\prime_{k}) -\mathbf{J}_2 s^\prime_{k-1} \|^2) \\
            &- 2\frac{L}{2} \big( \diag(\mathbf{J}_1 s^\prime_{k-1})(\nabla f^\prime(x_{k-1} +s_{k-1}) + g^\prime_{k})+\mathbf{J}_2 s^\prime_{k-1} \big)^\top (x_{k-1} +s_{k-1} - x^*-s^*) \\
            & + 2\frac{L}{2} \big(\diag(\frac{1}{L}\mathbf{1} + \mathbf{J}_1 s^\prime_{k-1})(\nabla f^\prime(x_{k-1} +s_{k-1}) + g^\prime_{k}) +\mathbf{J}_2 s^\prime_{k-1}\big)^\top \big(\diag(\mathbf{J}_1 s^\prime_{k-1})(\nabla f^\prime(x_{k-1} +s_{k-1}) + g^\prime_{k}) + \mathbf{J}_2 s^\prime_{k-1}\big).
        \end{aligned}
    \end{equation*}

    Moreover, we can apply the update formula in \eqref{thpf:ood_l2o_composite} to simplify the above inequation as follows:
    \begin{equation} \label{thpf:ood_convergence_gain_ub_pre_composite}
        \begin{aligned}
            &F^\prime(x_{k} + s_{k}) - F^\prime(x_{k-1} +s_{k-1})\\
            \leq& \frac{L}{2} (\|x_{k-1} +s_{k-1} - x^*-s^*\|^2 - \|x_{k} +s_{k} - x^*-s^* \|^2) \\
            &- L \big( \diag(\mathbf{J}_1 s^\prime_{k-1})(\nabla f^\prime(x_{k-1} +s_{k-1}) + g^\prime_{k})+\mathbf{J}_2 s^\prime_{k-1} \big)^\top \\
            & \qquad \Big(x_{k-1} +s_{k-1} - x^*-s^* - \diag(\frac{1}{L}\mathbf{1} + \mathbf{J}_1 s^\prime_{k-1})(\nabla f^\prime(x_{k-1} +s_{k-1}) + g^\prime_{k}) -\mathbf{J}_2 s^\prime_{k-1}\Big),
        \end{aligned}
    \end{equation}
    where we replace the update on $x_{k-1} + s_{k-1}$ with $x_k + s_k$ in the first line.
    
    Similarly, we propose to maintain the InD update formula on $x_{k-1}$ as $x_{k} = x_{k-1} - \frac{\nabla f(x_{k-1} +s_{k-1}) + g_{k}}{L}$, which further yields:
    \begin{equation*}
        \begin{aligned}
            x_{k} + s_{k} =& x_{k-1} + s_{k-1} - \frac{\nabla f^\prime(x_{k-1} +s_{k-1}) + g^\prime_{k}}{L} - \diag(\mathbf{J}_1 s^\prime_{k-1})(\nabla f^\prime(x_{k-1} +s_{k-1}) + g^\prime_{k}) -\mathbf{J}_2 s^\prime_{k-1}, \\
            =& x_{k-1} - \frac{\nabla f(x_{k-1} +s_{k-1}) + g_{k}}{L} \\
            & + s_{k-1} - \frac{\nabla f^\prime(x_{k-1} +s_{k-1}) + g^\prime_{k}-\nabla f(x_{k-1} +s_{k-1})-g_{k}}{L} - \diag(\mathbf{J}_1 s^\prime_{k-1})(\nabla f^\prime(x_{k-1} +s_{k-1}) + g^\prime_{k}) -\mathbf{J}_2 s^\prime_{k-1},
        \end{aligned}
    \end{equation*}
    where, we construct $s_{k}$ by following equation:
    \begin{equation*}
        \begin{aligned}
            s_{k} = s_{k-1} - \frac{\nabla f^\prime(x_{k-1} +s_{k-1}) + g^\prime_{k}-\nabla f(x_{k-1} +s_{k-1})-g_{k}}{L} - \diag(\mathbf{J}_1 s^\prime_{k-1})(\nabla f^\prime(x_{k-1} +s_{k-1}) + g^\prime_{k}) -\mathbf{J}_2 s^\prime_{k-1}.
        \end{aligned}
    \end{equation*}

    Substituting above equation into right-hand side of inequality \ref{thpf:ood_convergence_gain_ub_pre_composite} yields:
    \begin{equation}
        \begin{aligned}
            &F^\prime(x_{k} + s_{k}) - F^\prime(x_{k-1} +s_{k-1})\\
            \leq & \frac{L}{2} (\|x_{k-1} +s_{k-1} - x^*-s^*\|^2 - \|x_{k} +s_{k} - x^*-s^* \|^2) \\
            &- L \big( \diag(\mathbf{J}_1 s^\prime_{k-1})(\nabla f^\prime(x_{k-1} +s_{k-1}) + g^\prime_{k})+\mathbf{J}_2 s^\prime_{k-1} \big)^\top  \\
            & \qquad \Big(x_{k-1} +s_{k-1} - x^*-s^* - \diag(\frac{1}{L}\mathbf{1} + \mathbf{J}_1 s^\prime_{k-1})(\nabla f^\prime(x_{k-1} +s_{k-1}) + g^\prime_{k}) -\mathbf{J}_2 s^\prime_{k-1}\Big) , \\
            = & \frac{L}{2} (\|x_{k-1} +s_{k-1} - x^*-s^*\|^2 - \|x_{k} +s_{k} - x^*-s^* \|^2) \\
            &+ L \left( x_k +s_k -x_{k-1}-s_{k-1} +\frac{\nabla f^\prime(x_{k-1} +s_{k-1}) + g^\prime_{k}}{L}\right)^\top \Bigg(x_{k-1} - \frac{\nabla f(x_{k-1} +s_{k-1}) + g_{k}}{L} - x^*\\
            & +s_{k-1}  - \frac{\nabla f^\prime(x_{k-1} +s_{k-1}) + g^\prime_{k}-\nabla f(x_{k-1} +s_{k-1})-g_{k}}{L} - \diag(\mathbf{J}_1 s^\prime_{k-1})(\nabla f^\prime(x_{k-1} +s_{k-1}) + g^\prime_{k}) -\mathbf{J}_2 s^\prime_{k-1} -s^* \Bigg) , \\
            = & \frac{L}{2} (\|x_{k-1} +s_{k-1} - x^*-s^*\|^2 - \|x_{k} +s_{k} - x^*-s^* \|^2) \\
            & + L ( x_k +s_k -x_{k-1}-s_{k-1} +\frac{\nabla f^\prime(x_{k-1} +s_{k-1}) + g^\prime_{k}}{L})^\top (x_{k} +s_{k} - x^*-s^* ).
        \end{aligned}
    \end{equation}
    % TODO: Make it more concrete: a minimal among K iterations.
    Over $K$-iterations, we have:
    \begin{equation*}
        \begin{aligned}
            &\min_{k=1,\dots, K}F^\prime(x_k+s_k) - F^\prime(x^* +s^*)\\
            \leq & \frac{L}{2K} \|x_{0} +s_{0} - x^*-s^*\|^2 - \frac{L}{2K}\|x_{K} +s_{K} - x^*-s^* \|^2 \\
            & + \frac{L}{K} \sum_{k=1}^{K} ( x_k +s_k -x_{k-1}-s_{k-1} +\frac{\nabla f^\prime(x_{k-1} +s_{k-1}) + g^\prime_{k}}{L})^\top (x_{k} +s_{k} - x^*-s^* ).
        \end{aligned}
    \end{equation*}

\end{proof}

Moreover, we derive the upper bound of multi-iteration convergence rate in the following Corollary~\ref{lm:corollary9_composite_converge_rate_ub}.
\begin{corollary} \label{lm:corollary9_composite_converge_rate_ub}
    Under Assumption \ref{as:assump_ind_robust_composite}, L2O model $d$'s (\eqref{eq:l2o_composite}) convergence rate is upper bounded by w.r.t. $\| s_{k-1}^\prime \|$ by:
    \begin{equation*}
        \begin{aligned}
            &\min_{k=1,\dots, K}F^\prime(x_k+s_k) - F^\prime(x^* +s^*)\\
            \leq & \frac{L}{2K} \|x_{0} +s_{0} - x^*-s^*\|^2 - \frac{L}{2K}\|x_{K} +s_{K} - x^*-s^* \|^2 - \frac{1}{K} \sum_{k=1}^{K} (\nabla f^\prime(x_{k-1} +s_{k-1}))^\top (x_{k} +s_{k} - x^*-s^* ) \\
            &+ \frac{L}{K} \sum_{k=1}^{K} ( \sqrt{n}C_1  \| \nabla f^\prime(x_{k-1}+s_{k-1})\| + \sqrt{n}C_2 ) \| x_k+ s_k -x^* -s^* \| \|s_{k-1}^\prime\|.
        \end{aligned}
    \end{equation*}
\end{corollary}
\begin{proof}
    First, we rewrite the convergence rate upper bound as follows:
    \begin{equation*}
        \begin{aligned}
            &\min_{k=1,\dots, K}F^\prime(x_k+s_k) - F^\prime(x^* +s^*)\\
            \leq & \frac{L}{2K} \|x_{0} +s_{0} - x^*-s^*\|^2 - \frac{L}{2K}\|x_{K} +s_{K} - x^*-s^* \|^2 + \frac{L}{K} \sum_{k=1}^{K} ( x_k +s_k -x_{k-1}-s_{k-1})^\top (x_{k} +s_{k} - x^*-s^* ), \\
            = & \frac{L}{2K} \|x_{0} +s_{0} - x^*-s^*\|^2 - \frac{L}{2K}\|x_{K} +s_{K} - x^*-s^* \|^2 - \frac{1}{K} \sum_{k=1}^{K} (\nabla f^\prime(x_{k-1} +s_{k-1}))^\top (x_{k} +s_{k} - x^*-s^* ) \\
            & + \frac{L}{K} \sum_{k=1}^{K} ( -\diag(\mathbf{J}_1 s^\prime_{k-1})(\nabla f^\prime(x_{k-1} +s_{k-1}) + g^\prime_{k}) - \mathbf{J}_2 s^\prime_{k-1} )^\top (x_{k} +s_{k} - x^*-s^* ).
        \end{aligned}
    \end{equation*}

    Next, we derive its upper bound w.r.t. $\| s_{k-1}^\prime \|$. Cauchy-Schwarz inequality and Triangle inequality yield:
    \begin{equation*}
        \begin{aligned}
            &\min_{k=1,\dots, K}F^\prime(x_k+s_k) - F^\prime(x^* +s^*)\\
            \leq & \frac{L}{2K} \|x_{0} +s_{0} - x^*-s^*\|^2 - \frac{L}{2K}\|x_{K} +s_{K} - x^*-s^* \|^2 - \frac{1}{K} \sum_{k=1}^{K} (\nabla f^\prime(x_{k-1} +s_{k-1}))^\top (x_{k} +s_{k} - x^*-s^* ) \\
            & + \frac{L}{K} \sum_{k=1}^{K} ( -\diag(\mathbf{J}_1 s^\prime_{k-1})(\nabla f^\prime(x_{k-1} +s_{k-1}) + g^\prime_{k}) - \mathbf{J}_2 s^\prime_{k-1} )^\top (x_{k} +s_{k} - x^*-s^* ) , \\
            \leq & \frac{L}{2K} \|x_{0} +s_{0} - x^*-s^*\|^2 - \frac{L}{2K}\|x_{K} +s_{K} - x^*-s^* \|^2 - \frac{1}{K} \sum_{k=1}^{K} (\nabla f^\prime(x_{k-1} +s_{k-1}))^\top (x_{k} +s_{k} - x^*-s^* ) \\
            & + \frac{L}{K} \sum_{k=1}^{K} ( \|\diag(\mathbf{J}_1 s^\prime_{k-1})(\nabla f^\prime(x_{k-1} +s_{k-1}) + g^\prime_{k})\| + \|\mathbf{J}_2 s^\prime_{k-1}\| ) \| x_{k} +s_{k} - x^*-s^* \| , \\
            \leq & \frac{L}{2K} \|x_{0} +s_{0} - x^*-s^*\|^2 - \frac{L}{2K}\|x_{K} +s_{K} - x^*-s^* \|^2 - \frac{1}{K} \sum_{k=1}^{K} (\nabla f^\prime(x_{k-1} +s_{k-1}))^\top (x_{k} +s_{k} - x^*-s^* ) \\
            &+ \frac{L}{K} \sum_{k=1}^{K} ( \sqrt{n}C_1 \|s_{k-1}^\prime\| \| \nabla f^\prime(x_{k-1}+s_{k-1})\| + \sqrt{n}C_2\| s_{k-1}^\prime \|) \| x_k+ s_k -x^* -s^* \| , \\
            = & \frac{L}{2K} \|x_{0} +s_{0} - x^*-s^*\|^2 - \frac{L}{2K}\|x_{K} +s_{K} - x^*-s^* \|^2 - \frac{1}{K} \sum_{k=1}^{K} (\nabla f^\prime(x_{k-1} +s_{k-1}))^\top (x_{k} +s_{k} - x^*-s^* ) \\
            &+ \frac{L}{K} \sum_{k=1}^{K} \big( \sqrt{n}C_1  \| \nabla f^\prime(x_{k-1}+s_{k-1})\| + \sqrt{n}C_2 \big) \| x_k+ s_k -x^* -s^* \| \|s_{k-1}^\prime\|.
        \end{aligned}
    \end{equation*}

\end{proof}

\section{Non-Smooth Case Results} \label{sec:thpf_non_smooth}
% Inductively, we anaylze the non-smooth case from composite case. 
% Based on the definition in \cref{sec:def_optimizee}, 

% \subsection{Preliminary} \label{sec:thpf_non_smooth_pre}

For the non-smooth case, we set the smooth part in the objective of problem~\ref{obj:ind} to be zero $f(x):=0$, and the objective becomes:
\begin{equation} \label{obj:ind_non_smooth}
    \min_{x} r(x), \tag{P}
\end{equation}
where $x \in \mathcal{S}_P$ and $r \in \mathcal{F}_P$. Based on the definition, $r(x)$ is proper and convex, where the ``proper'' means $r(x)$ is trivially solvable for any $x$.

In the OOD scenario, the optimization problem becomes:
\begin{equation} \label{obj:ood_non_smooth}
    \min_{x^\prime} r^\prime(x^\prime), \tag{O}
\end{equation}
where $x^\prime \in \mathcal{S}_O$ and $r \in \mathcal{F}_O$. 

Based on the definition, $r^\prime(x)$ is still proper and convex. We can directly get the solution from:
\begin{equation*}
    {x^\prime}^* = \argmin_{x^\prime} r^\prime(x^\prime).
\end{equation*}

Thus, constructing a L2O model is unnecessary for the smooth case. We eliminate the demonstrations for this case.

\section{Longer Horizon Case Results} \label{sec:longer_horizon}
In the smooth and composite cases, we have demonstrated convergence analysis per iteration and multi-iteration convergence analysis for L2O. Modern algorithms utilize historical information to accelerate convergence, such as Nesterov momentum in FISTA algorithm \cite{Beck2009fast} and long short-term memory in LSTM-based unrolling algorithms \cite{Lv2017, Liu2023}. This case establishes convergence analysis with historical modeling in L2O. We take the SOTA Math-L2O framework \cite{Liu2023} to define the historical modeling part, a general and problem-independent approach that ensures our proposed theorems are also general.

We first establish that the input features of neural networks should be consistent with the definition of L2O model $d$. Suppose there exist a $d \in \mathcal{D}_C(\mathbb{R}^{m\times n}) \to \mathbb{R}^n$. Due to Lemma 1 in \cite{Liu2023}, for any $x_1, y_1, x_2, y_2, \dots, x_m, y_m \in \mathbb{R}^n$, there exists matrices $\mathbf{J}_1, \mathbf{J}_2, \dots, \mathbf{J}_m$ that:
\begin{equation*}
    d(x_1, x_2, \dots, x_m) = d(x_1^*, x_2^*, \dots, x_m^*) + \sum_{j=1}^{m} \mathbf{J}_j(x_j - x_j^*),
\end{equation*}
where $\mathbf{J}_j$ is $j$-th block of $d$'s Jacobian matrix at a interior point between $[x_1^\top, x_2^\top, \dots, x_m^\top]^\top$ and $[{x_1^*}^\top, {x_2^*}^\top, \dots, {x_m^*}^\top]^\top$.

The L2O is constructing such $\mathbf{J}$s by learning \cite{Liu2023}. Denote a NN as $\mathbf{N}_j$ and its input feature vector as $s$. We propose the following lemma to formulate $s$ to at least include all variables of $d$.
\begin{lemma}
    For any feature vector $s$ such that $\mathbf{J}_j = \mathbf{N}_j(s), j = 1,2, \dots, m$, $s$ should follows: 
    \begin{equation*}
        \{x_1^\top, x_2^\top, \dots, x_m^\top\} \subseteq s.
    \end{equation*}
\end{lemma}

\begin{proof}
    We prove the above lemma by contradiction. Suppose not, which means there exists a $s^\prime$ that $\exists x_i \notin s^\prime$. Then for $x_j, j = 1,2, \dots, m$, we have $\mathbf{J}_j = \mathbf{N}_j(s^\prime)$ by the definition. 
    First, suppose $j \neq i$. 
    Since $d$ is arbitrary, $x_j$ is not guaranteed linear with $x_i$ in $d$. Hence, $x_i$ should be one of the input features of $\mathbf{N}_j$. 
    Moreover, suppose $j = i$. $d$ is not guaranteed to always be less than the first order on $x_i$. Hence, $x_i$ should be one of the input features of $\mathbf{N}_j$. 

    The above scenarios cause contradictions with the assumption that $\exists x_i \notin s^\prime$, leading to the lemma's conclusion.
\end{proof}

\subsection{Preliminary}
Similar to the composite case in \cref{sec:smooth_with_non_smooth}. We make the following preliminary constructions. The objective is as follows:
\begin{equation*}
    \min_{x} f(x) + r(x),
\end{equation*}
where $f(x) \in \mathcal{F}_{L}$ is a $L$-smooth and convex function and $r(x) \in \mathcal{F}$ is a proper and convex function.

The definition of $L$-smoothness yields following upper bound of $f(y)$ for $\forall x, y \in \mathbb{R}^n$:
\begin{equation} \label{thpf:longer_horizon_lsmooth}
    f(y) \leq f(x) + \nabla f(x)^\top (y - x) + \frac{L}{2} \|y - x\|^2.
\end{equation}

For $k$-th iteration, we use $y_{k-1} \in \mathbb{R}^n$ to represent the historical information and use $z_{k-1}$ to represent the input feature vector for the L2O model. The L2O model is defined as follows:
\begin{equation*}
    x_k=x_{k-1} - d (z_{k-1}),
\end{equation*}
where $z_{k-1}$ is defined as $z_{k-1} := [x_{k-1}^\top, \nabla f(x_{k-1})^\top, g_{k}^\top, y_{k-1}^\top], g_{k} \in \partial r(x_k)$ \cite{Liu2023}. Without loss of generality, $y_{k-1}$ represents the result of any historical modeling methods. For example, we can use neural network models to achieve momentum-like modeling \cite{Liu2023}.

Utilizing $T$ to denote the number of iterations in historical modeling, following \cite{Liu2023}, we set $T:=1$. Inductively, we define $y_{k-1}$ as:
\begin{equation*}
    y_{k-1} = \Big(\mathbf{I} - \diag \big(\mathbf{N}_{4}([v_{k-1}^\top, v_{k-2}^\top]^\top)\big) \Big) v_{k-1} + \mathbf{N}_{4}([v_{k-1}^\top, v_{k-2}^\top]^\top) v_{k-2} + \mathbf{N}_{5}([v_{k-1}^\top, v_{k-2}^\top]^\top), 
\end{equation*}
where $\mathbf{N}_{4} \in \mathcal{D}_{C_4}(\mathbb{R}^{2n})$ and $\mathbf{N}_{5} \in \mathcal{D}_{C_5}(\mathbb{R}^{2n})$ are two neural network operators of the L2O model. $v \in \mathbb{R}^n$ denotes an input vector. Without loss of generality, such a definition covers any historical modeling methods with any feature selection. For example, $v$ can be a variable $x$ \cite{Liu2023}, a gradient $\nabla f(x)$ or a subgradient $g \in \partial r(x)$.

We are ready to demonstrate convergence analysis for longer-horizon cases. 
We focus on two kinds of historical feature selections: the horizon of variable $x$'s sequence and \cite{Liu2023} the horizon of gradient $\nabla f(x)$'s (and subgradient $\partial r(x)$'s) sequence(s).

\subsubsection*{Variable Method}
The variable method is from \cite{Liu2023}, where variable features are utilized to model the historical information. 
First, the neural network models' input vector $z_{k-1}$ is defined by:
\begin{equation} \label{thpf:longer_horizon_z_variable}
    z_{k-1} = [x_{k-1}^\top, \nabla f(x_{k-1})^\top, g_{k}^\top, y_{k-1}^\top]^\top,
\end{equation}
where $g_{k} \in \partial r(x_{k})$ is a subgradient vector. And $y_{k-1}$ denotes the feature from historical modeling. 
Then, the update given by the L2O model $d$ is defined as:
\begin{equation*}
    \begin{aligned}
    x_{k} = x_{k-1} &- \diag\left(\mathbf{N}_{1}(z_{k-1})\right)\nabla f(x_{k-1}) - \diag\left(\mathbf{N}_{1}(z_{k-1})\right)g_{k} - \diag(\mathbf{N}_{3}(z_{k-1}))(y_{k-1} - x_{k-1}) - \mathbf{N}_{2}(z_{k-1}).
    \end{aligned}
\end{equation*}

In this case, we use variable $x$ to construct the input vector $v$ for historical model $\mathbf{N}_{4}$ and denote $u_{k-1} := [x_{k-1}^\top, x_{k-2}^\top]^\top$. Based on Lemma 1 in Section A.1. of \cite{Liu2023}, we define historical modeling result $y_{k-1}$ as a linear-like combination of $x_{k-1}$ and $x_{k-2}$:
\begin{equation*}
    y_{k-1} = (\mathbf{I} - \diag(\mathbf{N}_{4}(u_{k-1}))) x_{k-1} + \diag(\mathbf{N}_{4}(u_{k-1}))x_{k-2},
\end{equation*}
where we eliminate the reaching zero bias term in \cite{Liu2023}. 

Based on \cite{Liu2023}, we define the L2O model as:
\begin{equation*}
    \begin{aligned}
        x_{k} = x_{k-1} &- \diag\left(\mathbf{N}_{1}(z_{k-1})\right)\big(\nabla f(x_{k-1}) + g_{k}\big) - \mathbf{N}_{2}(z_{k-1}) \\
        & -\diag(\mathbf{N}_{3}(z_{k-1}))\diag(\mathbf{N}_{4}(u_{k-1}))(-x_{k-1} + x_{k-2}),
    \end{aligned}
\end{equation*}
where we set $\diag\left(\mathbf{N}_{1}(z_{k-1})\right) \succ 0$ and $g_{k}$ is an implicit subgradient value at $x_{k}$. Moreover, we omit all bias terms since they are demonstrated to vanish along iteration \cite{Liu2023}.

Assume $\diag\left(\mathbf{N}_{1}(z_{k-1})\right)$ is a symmetric positive definite, similar to those in the composite case, we have the following necessary and sufficient conditions from the definition of the convex function $r$:
\begin{equation}
    \begin{aligned}
        &\diag\left(\mathbf{N}_{1}(z_{k-1})\right)^{-1}\Big(x_{k} - \big( x_{k-1} - \diag\left(\mathbf{N}_1(z_{k-1})\right)\nabla f(x_{k-1})  - \mathbf{N}_{2}(z_{k-1}) \\
        & \qquad -\diag(\mathbf{N}_{3}(z_{k-1}))\diag(\mathbf{N}_{4}(u_{k-1}))(-x_{k-1} + x_{k-2}) \big)\Big) + g_{k} = 0, \\
        & 0 \in  \partial r(x_{k}) + \diag\left(\mathbf{N}_{1}(z_{k-1})\right)^{-1} \Big(x_{k} - \big( x_{k-1} - \diag\left(\mathbf{N}_{1}(z_{k-1})\right)\nabla f(x_{k-1}) - \mathbf{N}_{2}(z_{k-1}) \\
        & \qquad -\diag(\mathbf{N}_{3}(z_{k-1}))\diag(\mathbf{N}_{4}(u_{k-1}))(-x_{k-1} + x_{k-2}) \big) \Big),
    \end{aligned}
\end{equation}
which yields the following proximal operator:
\begin{equation} \label{thpf:l2o_longer_horizon_variable}
    \begin{aligned}
        &\operatorname{prox}_{\diag\left(\mathbf{N}_{1}(z_{k-1})\right)^{-1}} \big(x_{k-1} - \diag\left(\mathbf{N}_{1}(z_{k-1})\right)\nabla f(x_{k-1}) - \mathbf{N}_{2}(z_{k-1})-\diag(\mathbf{N}_{3}(z_{k-1}))\diag(\mathbf{N}_{4}(u_{k-1}))(-x_{k-1} + x_{k-2}) \big)  \\
        =&\argmin_{x_{k}} r(x_{k}) + \frac{1}{2} \Big\|x_{k} - \big( x_{k-1} - \diag\left(\mathbf{N}_{1}(z_{k-1})\right)\nabla f(x_{k-1}) - \mathbf{N}_{2}(z_{k-1}) \\
        & \qquad \qquad \qquad \qquad -\diag(\mathbf{N}_{3}(z_{k-1}))\diag(\mathbf{N}_{4}(u_{k-1}))(-x_{k-1} + x_{k-2})  \big) \Big\|^2_{\diag\left(\mathbf{N}_{1}(z_{k-1})\right)^{-1}}.
    \end{aligned}
\end{equation}

\subsubsection*{Gradient (and Subgradient) Method}
This method utilizes gradient-related features to achieve historical modeling. 
First, the neural network models' input vector $z_{k-1}$ is defined by:
\begin{equation} \label{thpf:longer_horizon_z_gradient}
    z_{k-1} = [\nabla f(x_{k-1})^\top, g_{k}^\top, y_{k-1}^\top]^\top,
\end{equation}
where $g_{k} \in \partial r(x_{k})$. And $y_{k-1}$ denotes the feature from historical modeling. 
Compared with variable method in \cite{Liu2023}, we remove variable $x_{k-1}$ from \eqref{thpf:longer_horizon_z_variable}. Thus, compared with the variable method, $y_{k-1}$ represents a different feature based on the historical modeling method without the variable.

Similarly, in this case, we use gradient and subgradient to construct the input vector $v$ for historical model $\mathbf{N}_{4}$ and denote $w_{k-1} := [(\nabla f(x_{k-1})+g_{k-1})^\top, (\nabla f(x_{k-2})+g_{k-2})^\top]^\top$. 
For any $x \in \mathbb{R}^n$, we denote the lower bound and the upper bound of $\partial r(x)$ as $\partial r(x)_{\text{lb}}$ and $\partial r(x)_{\text{ub}}$ respectively. 
Based on Lemma 1 in Section A.1. of \cite{Liu2023}, using explicit subgradient, we define $y_{k-1}$ by:
\begin{equation*}
    \begin{aligned}
        y_{k-1} =& (\mathbf{I}- \diag(\mathbf{N}_{4}(w_{k-1})) (\nabla f(x_{k-1})+g_{k-1})) + \diag(\mathbf{N}_{4}(w_{k-1}))(\nabla f(x_{k-2})+g_{k-2}), \\
        g_{k-1} =& (\mathbf{I} - \diag(\mathbf{N}_{5}(r_{k-1})))\partial r(x_{k-1})_{\text{lb}} + \diag(\mathbf{N}_{5}(r_{k-1}))\partial r(x_{k-1})_{\text{ub}}\\
        g_{k-2} =& (\mathbf{I} - \diag(\mathbf{N}_{5}(r_{k-2})))\partial r(x_{k-2})_{\text{lb}} + \diag(\mathbf{N}_{5}(r_{k-2}))\partial r(x_{k-2})_{\text{ub}},\\
        x_{k} =& x_{k-1} - \diag\left(\mathbf{N}_{1}(z_{k-1})\right)\nabla f(x_{k-1}) - \diag\left(\mathbf{N}_{1}(z_{k-1})\right)g_{k} - \diag(\mathbf{N}_{3}(z_{k-1}))(y_{k-1} - (\nabla f(x_{k-1})+g_{k-1})) \\
        &- \mathbf{N}_{2}(z_{k-1}),
    \end{aligned}
\end{equation*}
where $g_{k} \in \partial r(x_{k})$, $g_{k-1} \in \partial r(x_{k-1})$, and $g_{k-2} \in \partial r(x_{k-2})$, $r_{k-1} := [\partial r(x_{k-1})_{\text{lb}}^\top, \partial r(x_{k-1})_{\text{ub}}^\top]^\top$. In the second and third equations, we apply two extra neural network models, denoted as $\mathbf{N}_{5}$ and $\mathbf{N}_{6}$ to learn subgradient vectors.

Based on our proposed L2O model in \eqref{thpf:go_math_l2o_basic}, the L2O model of gradient method is given by:
\begin{equation*}
    \begin{aligned}
        x_{k} = &x_{k-1} - \diag\left(\mathbf{N}_{1}(z_{k-1})\right)\big(\nabla f(x_{k-1}) + g_{k}\big) - \mathbf{N}_{2}(z_{k-1}) \\
        & -\diag(\mathbf{N}_{3}(z_{k-1}))\diag(\mathbf{N}_{4}(w_{k-1})) \Big(- \big(\nabla f(x_{k-1})+\big(\big(\mathbf{I} - \diag(\mathbf{N}_{5}(r_{k-1}))\big)\partial r(x_{k-1})_{\text{lb}} \\
        & + \diag(\mathbf{N}_{5}(r_{k-1}))\partial r(x_{k-1})_{\text{ub}}\big) \big) + \nabla f(x_{k-2})+\big((\mathbf{I} - \diag(\mathbf{N}_{5}(r_{k-2})))\partial r(x_{k-2})_{\text{lb}} + \diag(\mathbf{N}_{5}(r_{k-2}))\partial r(x_{k-2})_{\text{ub}}\big)\Big),
    \end{aligned}
\end{equation*}
where we set $\diag\left(\mathbf{N}_{1}(z_{k-1})\right) \succ 0$ and $g_{k}$ is an implicit subgradient value at $x_{k}$. Notably, in this definition, without loss of generality, we make a simpification by setting $\mathbf{Q} = \mathbf{H}$ and $\mathbf{B} = \mathbf{C}$ in \eqref{thpf:go_math_l2o_basic}, which are defined as $\diag(\mathbf{N}_{3}(z_{k-1}))$ and $\diag(\mathbf{N}_{4}(w_{k-1}))$ respectively. Moreover, we take explicit subgradient longer horizon modeling of two subgradient values of iteation $k-1$ and $k-2$ by $(\mathbf{I} - \diag(\mathbf{N}_{5}(r_{k-1})))\partial r(x_{k-1})_{\text{lb}} + \diag(\mathbf{N}_{5}(r_{k-1}))$ and $(\mathbf{I} - \diag(\mathbf{N}_{5}(r_{k-2})))\partial r(x_{k-2})_{\text{lb}} + \diag(\mathbf{N}_{5}(r_{k-2}))$ respectively. We also omit all bias terms since they are demonstrated to vanish along iteration in \cref{sec:tm2_l2o_model_proof}.

Assume $\diag\left(\mathbf{N}_{1}(z_{k-1})\right)$ is symmetric positive definite, the necessary and sufficient conditions from convexity definition are:
\begin{equation}
    \begin{aligned}
        &\diag\left(\mathbf{N}_{1}(z_{k-1})\right)^{-1}\Big(x_{k} - \big( x_{k-1} - \diag\left(\mathbf{N}_1(z_{k-1})\right)\nabla f(x_{k-1})  - \mathbf{N}_{2}(z_{k-1}) \\
        & \qquad -\diag(\mathbf{N}_{3}(z_{k-1}))\diag(\mathbf{N}_{4}(w_{k-1})) (- (\nabla f(x_{k-1})+g_{k-1}) + \nabla f(x_{k-2})+g_{k-2}) \big)\Big) + g_{k} = 0, \\
        & 0 \in  \partial r(x_{k}) + \diag\left(\mathbf{N}_{1}(z_{k-1})\right)^{-1} \Big(x_{k} - \big( x_{k-1} - \diag\left(\mathbf{N}_{1}(z_{k-1})\right)\nabla f(x_{k-1}) - \mathbf{N}_{2}(z_{k-1}) \\
        & \qquad -\diag(\mathbf{N}_{3}(z_{k-1}))\diag(\mathbf{N}_{4}(w_{k-1})) (- (\nabla f(x_{k-1})+g_{k-1}) + \nabla f(x_{k-2})+g_{k-2}) \big) \Big),
    \end{aligned}
\end{equation}
which yields the following proximal operator:
\begin{equation} \label{thpf:l2o_longer_horizon_gradient}
    \begin{aligned}
        &\operatorname{prox}_{\diag\left(\mathbf{N}_{1}(z_{k-1})\right)^{-1}}\big(x_{k-1} - \diag\left(\mathbf{N}_{1}(z_{k-1})\right)\nabla f(x_{k-1}) - \mathbf{N}_{2}(z_{k-1}) \\
        & \qquad -\diag(\mathbf{N}_{3}(z_{k-1}))\diag(\mathbf{N}_{4}(w_{k-1})) (- (\nabla f(x_{k-1})+g_{k-1}) + \nabla f(x_{k-2})+g_{k-2}) \big)  \\
        =&\argmin_{x_{k}} r(x_{k}) + \frac{1}{2} \|x_{k} - ( x_{k-1} - \diag\left(\mathbf{N}_{1}(z_{k-1})\right)\nabla f(x_{k-1}) - \mathbf{N}_{2}(z_{k-1}) \\
        & \qquad -\diag(\mathbf{N}_{3}(z_{k-1}))\diag(\mathbf{N}_{4}(w_{k-1})) (- (\nabla f(x_{k-1})+g_{k-1}) + \nabla f(x_{k-2})+g_{k-2}) ) \|^2_{\diag\left(\mathbf{N}_{1}(z_{k-1})\right)^{-1}}.
    \end{aligned}
\end{equation}

\subsubsection*{Gradient Map}
As introduced in the composite case, we still apply the gradient map method to facilitate convergence analysis. 
We note that both cases share a similar definition of gradient map. Utilizing a denotation $ P_{k-1}$ to represent the historical modeling results in both cases, we can represent the update of the L2O model in both cases with the following equation:
\begin{equation} \label{thpf:l2o_longer_horizon_p}
    \begin{aligned}
        x_{k} = x_{k-1} &- \diag\left(\mathbf{N}_{1}(z_{k-1})\right)\big(\nabla f(x_{k-1}) + g_{k}\big) - \mathbf{N}_{2}(z_{k-1}) -\diag(\mathbf{N}_{3}(z_{k-1})) P_{k-1},
    \end{aligned}
\end{equation}
where, in the variable method, $P_{k-1}$ is conducted by:
\begin{equation} \label{thpf:longer_horizon_variable_case_p}
     P_{k-1}:= \diag(\mathbf{N}_{4}(u_{k-1}))(-x_{k-1} + x_{k-2}).
\end{equation}
In the gradient method, $P_{k-1}$ is conducted by:
\begin{equation} \label{thpf:longer_horizon_gradient_case_p}\
    \begin{aligned}
         P_{k-1}:= \diag(\mathbf{N}_{4}(w_{k-1}))\Big(&- \big(\nabla f(x_{k-1})+(\mathbf{I} - \diag(\mathbf{N}_{5}(r_{k-1})))\partial r(x_{k-1})_{\text{lb}} + \diag(\mathbf{N}_{5}(r_{k-1}))\partial r(x_{k-1})_{\text{ub}} \big) \\
        & + \nabla f(x_{k-2})+(\mathbf{I} - \diag(\mathbf{N}_{5}(r_{k-2})))\partial r(x_{k-2})_{\text{lb}} + \diag(\mathbf{N}_{5}(r_{k-2}))\partial r(x_{k-2})_{\text{ub}}\Big).
    \end{aligned}
\end{equation}

Then, we define a gradient map $G_{\mathbf{N}_1(z)}(x_{k-1})$ that:
\begin{equation*}
    \begin{aligned}
        &G_{\diag\left(\mathbf{N}_1(z_{k-1})\right)^{-1}}(x_{k-1}) \\
        =& \diag\left(\mathbf{N}_1(z_{k-1})\right)^{-1} \Big(x_{k-1} - \operatorname{prox}_{\diag\left(\mathbf{N}_1(z_{k-1})\right)^{-1}}\big(x_{k-1} - \diag\left(\mathbf{N}_1(z_{k-1})\right)\nabla f(x_{k-1}) - \mathbf{N}_{2}(z_{k-1}) -\diag(\mathbf{N}_{3}(z_{k-1})) P_{k-1} \big) \\
        & \qquad \qquad \qquad \qquad \quad-\mathbf{N}_2(z_{k-1}) -\diag(\mathbf{N}_{3}(z_{k-1})) P_{k-1}\Big).
    \end{aligned}
\end{equation*}
And we can represent $x_{k}$ with $G_{\mathbf{N}_1(z)}(x_{k-1})$ by:
\begin{equation} \label{thpf:ind_GM_longer_horizon_p}
    \begin{aligned}
        x_{k} =& \operatorname{prox}_{\diag\left(\mathbf{N}_1(z_{k-1})\right)^{-1}}\Big( x_{k-1} - \diag\left(\mathbf{N}_1(z_{k-1})\right)\nabla f(x_{k-1})  - \mathbf{N}_{2}(z_{k-1})\Big) -\diag(\mathbf{N}_{3}(z_{k-1})) P_{k-1}) , \\
        =& x_{k-1} - \diag\left(\mathbf{N}_1(z_{k-1})\right)G_{\diag\left(\mathbf{N}_1(z_{k-1})\right)^{-1}}(x_{k-1}) -\mathbf{N}_2(z_{k-1}) -\diag(\mathbf{N}_{3}(z_{k-1})) P_{k-1}.
    \end{aligned}
\end{equation}
Substitute the above $x_{k}$'s representation into \eqref{thpf:longer_horizon_lsmooth}, we have the following upper bound of $f(x_{k})$:
\begin{equation} \label{thpf:l2o_longer_ub_pre}
    \begin{aligned}
        f(x_{k}) \leq &f(x_{k-1}) + \nabla f(x_{k-1})^\top (x_{k} - x_{k-1}) + \frac{L}{2} \|x_{k} - x_{k-1}\|^2 , \\
        \leq &f(x_{k-1}) - \nabla f(x_{k-1})^\top (\diag\left(\mathbf{N}_1(z_{k-1})\right)G_{\diag\left(\mathbf{N}_1(z_{k-1})\right)^{-1}}(x_{k-1}) +\mathbf{N}_2(z_{k-1}) +\diag(\mathbf{N}_{3}(z_{k-1})) P_{k-1}) \\
        & + \frac{L}{2} \| \diag\left(\mathbf{N}_1(z_{k-1})\right)G_{\diag\left(\mathbf{N}_1(z_{k-1})\right)^{-1}}(x_{k-1}) +\mathbf{N}_2(z_{k-1}) +\diag(\mathbf{N}_{3}(z_{k-1})) P_{k-1} \|^2.
    \end{aligned}
\end{equation}

Similar to \eqref{thpf:G_fg_equation_composite} in the composite case, we still have the following representation of gradient and subgradient:
\begin{equation} \label{thpf:G_fg_equation_l2o_longer}
    G_{\diag\left(\mathbf{N}_1(z_{k-1})\right)^{-1}}(x_{k-1}) =  \nabla f(x_{k-1}) + g_{k},
\end{equation}
where $g_{k} \in \partial r(x_{k})$.

Similar to Lemma~\ref{thpf:lm_GM_basic_bound_composite} in the composite case, the general relationship between the objectives of any arbitrary two points in the longer horizon case is as follows:
\begin{lemma} \label{thpf:lm_GM_basic_bound_long}
    \begin{equation*}
        \begin{aligned}
            &F(x_{k}) \\
            \leq&  F(t) + G_{\diag\left(\mathbf{N}_1(z_{k-1})\right)^{-1}}(x_{k-1})^\top (x_{k-1} - t) \\
            & + \frac{L}{2} \Bigg(\left\| \diag\left(\mathbf{N}_1(z_{k-1})\right)G_{\diag\left(\mathbf{N}_1(z_{k-1})\right)^{-1}}(x_{k-1}) +\mathbf{N}_2(z_{k-1}) +\diag(\mathbf{N}_{3}(z_{k-1})) P_{k-1} - \frac{G_{\diag\left(\mathbf{N}_1(z_{k-1})\right)^{-1}}(x_{k-1})}{L}\right\|^2 \\
            & \qquad \quad - \left\|\frac{G_{\diag\left(\mathbf{N}_1(z_{k-1})\right)^{-1}}(x_{k-1})}{L}\right\|^2\Bigg).
        \end{aligned}
    \end{equation*}
\end{lemma}
The above inequation differs from that of Lemma~\ref{thpf:lm_GM_basic_bound_composite} on the right-hand side. An extra term on historical modeling result $P_{k-1}$ exists. There are two different modeling methods to construct $P_{k-1}$, i.e. variable method in \eqref{thpf:longer_horizon_variable_case_p} and gradient method in \eqref{thpf:longer_horizon_gradient_case_p}.

\begin{proof}
    Workflow of the proof is identical to that of Lemma~\ref{thpf:lm_GM_basic_bound_composite} with a additional but stable term $\diag(\mathbf{N}_{3}(z_{k-1})) P_{k-1}$.

    First, we make objective $F$ and apply an upper bound from the convexity definition and gradient map.
    \begin{equation*}
        \begin{aligned}
            & F(x_{k}) \\
            \leq & f(x_{k-1}) - \nabla f(x_{k-1})^\top \big(\diag\left(\mathbf{N}_1(z_{k-1})\right)G_{\diag\left(\mathbf{N}_1(z_{k-1})\right)^{-1}}(x_{k-1}) +\mathbf{N}_2(z_{k-1}) +\diag(\mathbf{N}_{3}(z_{k-1})) P_{k-1}\big) \\
            & + \frac{L}{2} \| \diag\left(\mathbf{N}_1(z_{k-1})\right)G_{\diag\left(\mathbf{N}_1(z_{k-1})\right)^{-1}}(x_{k-1}) +\mathbf{N}_2(z_{k-1}) +\diag(\mathbf{N}_{3}(z_{k-1})) P_{k-1}\|^2 + r(x_{k}) , \\
            \leq & f(t) - \nabla f(x_{k-1})^\top(t-x_{k-1}) \\
            & - \nabla f(x_{k-1})^\top (\diag\left(\mathbf{N}_1(z_{k-1})\right)G_{\diag\left(\mathbf{N}_1(z_{k-1})\right)^{-1}}(x_{k-1}) +\mathbf{N}_2(z_{k-1}) +\diag(\mathbf{N}_{3}(z_{k-1})) P_{k-1}) \\
            & + \frac{L}{2} \| \diag\left(\mathbf{N}_1(z_{k-1})\right)G_{\diag\left(\mathbf{N}_1(z_{k-1})\right)^{-1}}(x_{k-1}) +\mathbf{N}_2(z_{k-1}) +\diag(\mathbf{N}_{3}(z_{k-1})) P_{k-1} \|^2 + r(x_{k}) , \\
            \leq & f(t) - \nabla f(x_{k-1})^\top(t-x_{k-1}) \\
            & - \nabla f(x_{k-1})^\top (\diag\left(\mathbf{N}_1(z_{k-1})\right)G_{\diag\left(\mathbf{N}_1(z_{k-1})\right)^{-1}}(x_{k-1}) +\mathbf{N}_2(z_{k-1}) +\diag(\mathbf{N}_{3}(z_{k-1})) P_{k-1}) \\
            & + \frac{L}{2} \| \diag\left(\mathbf{N}_1(z_{k-1})\right)G_{\diag\left(\mathbf{N}_1(z_{k-1})\right)^{-1}}(x_{k-1}) +\mathbf{N}_2(z_{k-1}) +\diag(\mathbf{N}_{3}(z_{k-1})) P_{k-1} \|^2 \\
            & + r(t) - \Big(G_{\diag\left(\mathbf{N}_1(z_{k-1})\right)^{-1}}(x_{k-1}) - \nabla f(x_{k-1})\Big)^\top \\
            & \qquad \qquad \Big(t - \big(x_{k-1} - \diag\left(\mathbf{N}_1(z_{k-1})\right)G_{\diag\left(\mathbf{N}_1(z_{k-1})\right)^{-1}}(x_{k-1}) -\mathbf{N}_2(z_{k-1}) +\diag(\mathbf{N}_{3}(z_{k-1})) P_{k-1}\big)\Big).
        \end{aligned}
    \end{equation*}
    In first step, we add $r(x_{k})$ to inequality \ref{thpf:l2o_longer_ub_pre}. In the second step, we substitute the first-order condition of convex $f$ on $x_{k-1}$. In the third step, we substitute the gradient map representation of the first-order condition of convex $r$ on $x_{k-1}$. 
    
    Then, we make up the first perfect square:
    \begin{equation*}
        \begin{aligned}
            & F(x_{k}) \\
            \leq & f(t) - \nabla f(x_{k-1})^\top(t-x_{k-1}) \\
            & - \nabla f(x_{k-1})^\top \big(\diag\left(\mathbf{N}_1(z_{k-1})\right)G_{\diag\left(\mathbf{N}_1(z_{k-1})\right)^{-1}}(x_{k-1}) +\mathbf{N}_2(z_{k-1}) +\diag(\mathbf{N}_{3}(z_{k-1})) P_{k-1}\big) \\
            & + \frac{L}{2} \| \diag\left(\mathbf{N}_1(z_{k-1})\right)G_{\diag\left(\mathbf{N}_1(z_{k-1})\right)^{-1}}(x_{k-1}) +\mathbf{N}_2(z_{k-1}) +\diag(\mathbf{N}_{3}(z_{k-1})) P_{k-1}\|^2 \\
            & + r(t) - \left(G_{\diag\left(\mathbf{N}_1(z_{k-1})\right)^{-1}}(x_{k-1}) - \nabla f(x_{k-1})\right)^\top \\
            & \qquad \qquad\qquad \Big(t - \big(x_{k-1} - \diag\left(\mathbf{N}_1(z_{k-1})\right)G_{\diag\left(\mathbf{N}_1(z_{k-1})\right)^{-1}}(x_{k-1}) -\mathbf{N}_2(z_{k-1}) -\diag(\mathbf{N}_{3}(z_{k-1})) P_{k-1}\big)\Big) , \\
            = & f(t) + r(t) + \frac{L}{2} \| \diag\left(\mathbf{N}_1(z_{k-1})\right)G_{\diag\left(\mathbf{N}_1(z_{k-1})\right)^{-1}}(x_{k-1}) +\mathbf{N}_2(z_{k-1} +\diag(\mathbf{N}_{3}(z_{k-1})) P_{k-1})\|^2 \\
            & - G_{\diag\left(\mathbf{N}_1(z_{k-1})\right)^{-1}}(x_{k-1})^\top \\
            & \qquad \Big(t - \big(x_{k-1} - \diag\left(\mathbf{N}_1(z_{k-1})\right)G_{\diag\left(\mathbf{N}_1(z_{k-1})\right)^{-1}}(x_{k-1}) -\mathbf{N}_2(z_{k-1}) -\diag(\mathbf{N}_{3}(z_{k-1})) P_{k-1}\big)\Big) , \\
            = & F(t) + G_{\diag\left(\mathbf{N}_1(z_{k-1})\right)^{-1}}(x_{k-1})^\top (x_{k-1} - t) \\
            & + \frac{L}{2} \| \diag\left(\mathbf{N}_1(z_{k-1})\right)G_{\diag\left(\mathbf{N}_1(z_{k-1})\right)^{-1}}(x_{k-1}) +\mathbf{N}_2(z_{k-1}) +\diag(\mathbf{N}_{3}(z_{k-1})) P_{k-1}\|^2 \\
            &- G_{\diag\left(\mathbf{N}_1(z_{k-1})\right)^{-1}}(x_{k-1})^\top (\diag\left(\mathbf{N}_1(z_{k-1})\right)G_{\diag\left(\mathbf{N}_1(z_{k-1})\right)^{-1}}(x_{k-1}) + \mathbf{N}_2(z_{k-1}) +\diag(\mathbf{N}_{3}(z_{k-1})) P_{k-1}).
        \end{aligned}
    \end{equation*}

    Second perfect square:
    \begin{equation}
        \begin{aligned}
            & F(x_{k}) \\
            \leq & F(t) + G_{\diag\left(\mathbf{N}_1(z_{k-1})\right)^{-1}}(x_{k-1})^\top (x_{k-1} - t) \\
            & + \frac{L}{2} \Big(\| \diag\left(\mathbf{N}_1(z_{k-1})\right)G_{\diag\left(\mathbf{N}_1(z_{k-1})\right)^{-1}}(x_{k-1}) +\mathbf{N}_2(z_{k-1}) +\diag(\mathbf{N}_{3}(z_{k-1})) P_{k-1}\|^2 \\
            & - \frac{2}{L} G_{\diag\left(\mathbf{N}_1(z_{k-1})\right)^{-1}}(x_{k-1})^\top \big(\diag\left(\mathbf{N}_1(z_{k-1})\right)G_{\diag\left(\mathbf{N}_1(z_{k-1})\right)^{-1}}(x_{k-1}) + \mathbf{N}_2(z_{k-1}) +\diag(\mathbf{N}_{3}(z_{k-1})) P_{k-1} \big) \Big) , \\
            = & F(t) + G_{\diag\left(\mathbf{N}_1(z_{k-1})\right)^{-1}}(x_{k-1})^\top (x_{k-1} - t) \\
            & + \frac{L}{2} \Bigg(\left\| \diag\left(\mathbf{N}_1(z_{k-1})\right)G_{\diag\left(\mathbf{N}_1(z_{k-1})\right)^{-1}}(x_{k-1}) +\mathbf{N}_2(z_{k-1}) +\diag(\mathbf{N}_{3}(z_{k-1})) P_{k-1} - \frac{G_{\diag\left(\mathbf{N}_1(z_{k-1})\right)^{-1}}(x_{k-1})}{L}\right\|^2 \\
            & \qquad \quad - \left\|\frac{G_{\diag\left(\mathbf{N}_1(z_{k-1})\right)^{-1}}(x_{k-1})}{L}\right\|^2 \Bigg).
        \end{aligned}
    \end{equation}
\end{proof}

Similarly, we can derive convergence analysis by iteratively applying Lemma~\ref{thpf:lm_GM_basic_bound_long}.

\subsection{InD Convergence Upper Bound}
Similar to Lemma~\ref{lm:ind_converge_gain_composite} in the composite case, we use the following lemma to ensure an InD robust L2O model in the longer horizon case.
\begin{lemma} \label{lm:ind_converge_gain_longer_horizon}
    For $\forall z_{k-1} \in \mathcal{Z}_P, \forall x_{k-1} \in \mathcal{S}_P$, 
    if $\mathbf{N}_1(z_{k-1}), \mathbf{N}_2(z_{k-1}), \mathbf{N}_3(z_{k-1})$ are bounded by following compact sets:
    \begin{equation*}
        \begin{aligned}
            &\mathbf{N}_1(z_{k-1}) \in \left[\mathbf{0}, \frac{2}{L} \mathbf{1} \right],\\
            &\left\| \diag\left(\mathbf{N}_1(z_{k-1})\right)\big(\nabla f(x_{k-1}) + g_{k}\big) +\mathbf{N}_2(z_{k-1}) +\diag(\mathbf{N}_{3}(z_{k-1})) P_{k-1} - \frac{\nabla f(x_{k-1}) + g_{k}}{L}\right\| \leq  \left\|\frac{\nabla f(x_{k-1}) + g_{k}}{L} \right\|, \\
            & \forall \mathbf{N}_1(z_{k-1}) \in \left[\mathbf{0}, \frac{2}{L} \mathbf{1} \right],
        \end{aligned}
    \end{equation*}
    where $g_{k} \in \partial r(x_{k})$. 
    
    Then, for any $x_k$ generated by L2O model in \eqref{thpf:l2o_longer_horizon_variable}, we have the following homogeneous derease on objective:
    \begin{equation*}
        F(x_k) - F(x_{k-1}) \leq 0.
    \end{equation*}
\end{lemma}
\begin{proof}
    The proof is similar that of Lemma~\ref{lm:ind_converge_gain_composite} in the composite case with an extra $\diag(\mathbf{N}_{3}(z_{k-1})) P_{k-1}$ term. 

    We first freeze operator $\mathbf{N}_{2}$ and $\mathbf{N}_{3}$ and derive the bound for $\mathbf{N}_{1}$. Then, each given $\mathbf{N}_{1}$ yields a bound for $\mathbf{N}_{2}$ and $\mathbf{N}_{3}$.

    Based on the Lemma~\ref{thpf:lm_GM_basic_bound_long}, substituting $t := x_{k-1}$ yields the following upper bound of the objective decrease: 
    \begin{equation} \label{thpf:converge_gain_longer_horizon_ineq}
        \begin{aligned}
            &F(x_{k}) - F(x_{k-1})\\
            \leq& \frac{L}{2} \Bigg(\left\| \diag\left(\mathbf{N}_1(z_{k-1})\right)G_{\diag\left(\mathbf{N}_1(z_{k-1})\right)^{-1}}(x_{k-1}) +\mathbf{N}_2(z_{k-1}) +\diag(\mathbf{N}_{3}(z_{k-1})) P_{k-1}- \frac{G_{\diag\left(\mathbf{N}_1(z_{k-1})\right)^{-1}}(x_{k-1})}{L}\right\|^2 \\
            &\quad - \left\|\frac{G_{\diag\left(\mathbf{N}_1(z_{k-1})\right)^{-1}}(x_{k-1})}{L}\right\|^2\Bigg).
        \end{aligned}
    \end{equation}

    To ensure $F(x_k) \leq F(x_{k-1})$, we should keep right-hand side non-positive, which yields:
    \begin{equation*}
        \begin{aligned}
            & \frac{L}{2} \Bigg(\left\| \diag\left(\mathbf{N}_1(z_{k-1})\right)G_{\diag\left(\mathbf{N}_1(z_{k-1})\right)^{-1}}(x_{k-1}) +\mathbf{N}_2(z_{k-1}) +\diag(\mathbf{N}_{3}(z_{k-1})) P_{k-1} - \frac{G_{\diag\left(\mathbf{N}_1(z_{k-1})\right)^{-1}}(x_{k-1})}{L}\right\|^2 \\
            & \qquad - \left\|\frac{G_{\diag\left(\mathbf{N}_1(z_{k-1})\right)^{-1}}(x_{k-1})}{L}\right\|^2\Bigg) \leq 0.
        \end{aligned}
    \end{equation*}
    
    After rearrangement, we have the following inequality:
    \begin{equation} \label{thpf:ind_converge_gain_longer_horizon_ineq}
        \begin{aligned}
            & \left\| \diag\left(\mathbf{N}_1(z_{k-1})\right)G_{\diag\left(\mathbf{N}_1(z_{k-1})\right)^{-1}}(x_{k-1}) +\mathbf{N}_2(z_{k-1}) +\diag(\mathbf{N}_{3}(z_{k-1})) P_{k-1} - \frac{G_{\diag\left(\mathbf{N}_1(z_{k-1})\right)^{-1}}(x_{k-1})}{L}\right\| \\
            & \leq \frac{\left\|G_{\diag\left(\mathbf{N}_1(z_{k-1})\right)^{-1}}(x_{k-1})\right\|}{L}.
        \end{aligned}
    \end{equation}

    Similarly, we first freeze $\mathbf{N}_2(z_{k-1}), \diag(\mathbf{N}_{3}(z_{k-1})) P_{k-1}$ and discuss $\mathbf{N}_1(z_{k-1})$-only terms, which yields:
    \begin{equation*}
        \begin{aligned}
            & \left\| \diag\left(\mathbf{N}_1(z_{k-1})\right)G_{\diag\left(\mathbf{N}_1(z_{k-1})\right)^{-1}}(x_{k-1}) - \frac{G_{\diag\left(\mathbf{N}_1(z_{k-1})\right)^{-1}}(x_{k-1})}{L}\right\| \leq \frac{\left\|G_{\diag\left(\mathbf{N}_1(z_{k-1})\right)^{-1}}(x_{k-1})\right\|}{L},\\
            & \| (L\diag\left(\mathbf{N}_1(z_{k-1})\right) - \mathbf{I})\frac{G_{\diag\left(\mathbf{N}_1(z_{k-1})\right)^{-1}}(x_{k-1})}{L} \| \leq \frac{\left\|G_{\diag\left(\mathbf{N}_1(z_{k-1})\right)^{-1}}(x_{k-1})\right\|}{L}.
        \end{aligned}
    \end{equation*}

    Solving the inequation, we have the following boundness for $\mathbf{N}_1(z_{k-1})$:
    \begin{equation*}
        \mathbf{N}_1(z_{k-1}) \in \left[\mathbf{0},  \frac{2}{L}\mathbf{1} \right].
    \end{equation*}

    Similarly, each choice of $\mathbf{N}_1(z_{k-1})$ yields a pair of bounds for $\mathbf{N}_2(z_{k-1})$ and $\diag(\mathbf{N}_{3}(z_{k-1})) P_{k-1}$. 
    For example, $\mathbf{N}_1(z_{k-1}) := 0$ yields:
    \begin{equation*}
        \begin{aligned}
            & \left\| \mathbf{N}_2(z_{k-1}) +\diag(\mathbf{N}_{3}(z_{k-1})) P_{k-1} -\frac{G_{\diag\left(\mathbf{N}_1(z_{k-1})\right)^{-1}}(x_{k-1})}{L}\right\| \leq \frac{\left\|G_{\diag\left(\mathbf{N}_1(z_{k-1})\right)^{-1}}(x_{k-1})\right\|}{L}.
        \end{aligned}
    \end{equation*}

    Inductively, freezing $\diag(\mathbf{N}_{3}(z_{k-1})) P_{k-1}$ yields:
    \begin{equation*}
        \begin{aligned}
            & \left\| \mathbf{N}_2(z_{k-1}) -\frac{G_{\diag\left(\mathbf{N}_1(z_{k-1})\right)^{-1}}(x_{k-1})}{L}\right\| \leq \frac{\left\|G_{\diag\left(\mathbf{N}_1(z_{k-1})\right)^{-1}}(x_{k-1})\right\|}{L}.
        \end{aligned}
    \end{equation*}

    Solving the above inequation, we have the following boundness on $\mathbf{N}_2(z_{k-1})$:
    \begin{equation*}
        \mathbf{N}_2(z_{k-1}) \in \left[\mathbf{0},  \frac{2}{L}|G_{\diag\left(\mathbf{N}_1(z_{k-1})\right)^{-1}}(x_{k-1})| \right].
    \end{equation*}

    Then, if $\mathbf{N}_2(z_{k-1}) = 0$, we can construct following inequality for $\mathbf{N}_3(z_{k-1})$:
    \begin{equation*}
        \begin{aligned}
            & \left\| \diag(\mathbf{N}_{3}(z_{k-1})) P_{k-1} -\frac{G_{\diag\left(\mathbf{N}_1(z_{k-1})\right)^{-1}}(x_{k-1})}{L}\right\| \leq \frac{\left\|G_{\diag\left(\mathbf{N}_1(z_{k-1})\right)^{-1}}(x_{k-1})\right\|}{L},
        \end{aligned}
    \end{equation*}
    which yields:
    \begin{equation*}
        \diag(\mathbf{N}_{3}(z_{k-1})) P_{k-1} \in \left[\mathbf{0},  \frac{2}{L}|G_{\diag\left(\mathbf{N}_1(z_{k-1})\right)^{-1}}(x_{k-1})| \right].
    \end{equation*}
    If $\mathbf{N}_2(z_{k-1}) = \frac{1}{L}G_{\diag\left(\mathbf{N}_1(z_{k-1})\right)^{-1}}(x_{k-1})$, the inequality is:
    \begin{equation*}
        \begin{aligned}
            & \left\| \diag(\mathbf{N}_{3}(z_{k-1})) P_{k-1} \right\| \leq \frac{\left\|G_{\diag\left(\mathbf{N}_1(z_{k-1})\right)^{-1}}(x_{k-1})\right\|}{L},
        \end{aligned}
    \end{equation*}
    which yields:
    \begin{equation*}
        \diag(\mathbf{N}_{3}(z_{k-1})) P_{k-1} \in \left[-\frac{1}{L}|G_{\diag\left(\mathbf{N}_1(z_{k-1})\right)^{-1}}(x_{k-1})|,  \frac{1}{L}|G_{\diag\left(\mathbf{N}_1(z_{k-1})\right)^{-1}}(x_{k-1})| \right].
    \end{equation*}

    Recovering the $G_{\diag\left(\mathbf{N}_1(z_{k-1})\right)^{-1}}(x_{k-1})$ in inequation \ref{thpf:ind_converge_gain_longer_horizon_ineq} with $G_{\diag\left(\mathbf{N}_1(z_{k-1})\right)^{-1}}(x_{k-1}) = \nabla f(x_{k-1}) + g_{k}$ in \eqref{thpf:G_fg_equation_l2o_longer} yields:
    \begin{equation*}
        \begin{aligned}
            & \| \diag\left(\mathbf{N}_1(z_{k-1})\right)\big(\nabla f(x_{k-1}) + g_{k}\big) +\mathbf{N}_2(z_{k-1}) +\diag(\mathbf{N}_{3}(z_{k-1})) P_{k-1} - \frac{\nabla f(x_{k-1}) + g_{k}}{L}\| \leq \|\frac{\nabla f(x_{k-1}) + g_{k}}{L} \|,
        \end{aligned}
    \end{equation*}
    where $g_{k} \in \partial r(x_{k})$.
\end{proof}

Similar to Corollary~\ref{ob:composite_ind_best_converge_gain} in the composite case, we present the following corollary to ensure a robust L2O model in the InD scenario.
\begin{corollary} \label{ob:longer_horizon_ind_best_converge_gain}
    For any $z_{k-1} \in \mathcal{Z}_P$, we let: 
    \begin{equation*}
        \mathbf{N}_1(z_{k-1}) := \frac{1}{L}\mathbf{1}, \mathbf{N}_2(z_{k-1}) := \mathbf{0}, \mathbf{N}_{3}(z_{k-1}) := \mathbf{0},  P_{k-1} := \mathbf{0},
    \end{equation*}
    the Math-L2O model in \eqref{thpf:l2o_longer_horizon_p} is exactly gradient descent update with convergence rate:
    \begin{equation*}
        F(x_K) - F(x^*) \leq \frac{L}{2K} \|x_0 -x^*\|^2.
    \end{equation*}
\end{corollary}
\begin{proof}
    In the last term of inequation \ref{thpf:converge_gain_longer_horizon_ineq}, the best convergence gain yields:
    \begin{equation*}
        \left\| \diag\left(\mathbf{N}_1(z_{k-1})\right)G_{\diag\left(\mathbf{N}_1(z_{k-1})\right)^{-1}}(x_{k-1}) +\mathbf{N}_2(z_{k-1}) +\diag(\mathbf{N}_{3}(z_{k-1})) P_{k-1} - \frac{G_{\diag\left(\mathbf{N}_1(z_{k-1})\right)^{-1}}(x_{k-1})}{L}\right\| := 0.
    \end{equation*}
    $\mathbf{N}_1(z_{k-1}) = \frac{1}{L}\mathbf{1}, \mathbf{N}_2(z_{k-1}) = \mathbf{0}, \mathbf{N}_{3}(z_{k-1}) = \mathbf{0},  P_{k-1} = \mathbf{0}$ is a feasible solution.

    Given $\mathbf{N}_1(z_{k-1}) = \frac{1}{L}\mathbf{1}, \mathbf{N}_2(z_{k-1}) = \mathbf{0}, \mathbf{N}_{3}(z_{k-1}) = \mathbf{0},  P_{k-1} = \mathbf{0}$, the update formula in \eqref{thpf:ind_GM_longer_horizon_p} is:
    \begin{equation} \label{thpf:ind_GM_longer_horizon_best}
        \begin{aligned}
            x_{k} = x_{k-1} - \frac{1}{L}G_{\diag\left(\mathbf{N}_1(z_{k-1})\right)^{-1}}(x_{k-1}).
        \end{aligned}
    \end{equation}

    Based on Lemma~\ref{thpf:lm_GM_basic_bound_long}, when $t := x^*$, we have the following inequality to evaluate per iteration convergence gain: 
    \begin{equation*}
        \begin{aligned}
            &F(x_{k}) - F(x^*)\\
            \leq& G_{\diag\left(\mathbf{N}_1(z_{k-1})\right)^{-1}}(x_{k-1})^\top (x_{k-1} - x^*)\\
            &+\frac{L}{2} \Bigg(\left\| \diag\left(\mathbf{N}_1(z_{k-1})\right)G_{\diag\left(\mathbf{N}_1(z_{k-1})\right)^{-1}}(x_{k-1}) +\mathbf{N}_2(z_{k-1}) +\diag(\mathbf{N}_{3}(z_{k-1})) P_{k-1} - \frac{G_{\diag\left(\mathbf{N}_1(z_{k-1})\right)^{-1}}(x_{k-1})}{L}\right\|^2 \\
            & \qquad \quad - \left\|\frac{G_{\diag\left(\mathbf{N}_1(z_{k-1})\right)^{-1}}(x_{k-1})}{L}\right\|^2 \Bigg) , \\
            =& G_{\diag\left(\mathbf{N}_1(z_{k-1})\right)^{-1}}(x_{k-1})^\top (x_{k-1} - x^*) - \frac{L}{2} \left\|\frac{G_{\diag\left(\mathbf{N}_1(z_{k-1})\right)^{-1}}(x_{k-1})}{L}\right\|^2 , \\
            =& \frac{L}{2} \Bigg(\frac{2}{L} G_{\diag\left(\mathbf{N}_1(z_{k-1})\right)^{-1}}(x_{k-1})^\top (x_{k-1} - x^*) -  \left\|\frac{G_{\diag\left(\mathbf{N}_1(z_{k-1})\right)^{-1}}(x_{k-1})}{L}\right\|^2\Bigg) , \\
            =& \frac{L}{2} \Bigg(\|x_{k-1} - x^*\|^2-  \left\|x_{k-1} - x^* - \frac{G_{\diag\left(\mathbf{N}_1(z_{k-1})\right)^{-1}}(x_{k-1})}{L}\right\|^2 \Bigg) , \\
            =& \frac{L}{2} (\|x_{k-1} - x^*\|^2-  \|x_{k-1} - x^*\|^2).
        \end{aligned}
    \end{equation*}

    Sum over $K$ iterations, we have the following InD multi-iteration convergence rate:
    \begin{equation} \label{thpf:ind_converge_rate_longer_horizon_ineq}
        \begin{aligned}
            F(x_{K}) - F(x^*) \leq \frac{L}{2K} (\|x_{0} - x^*\|^2-  \|x_{K} - x^*\|^2) \leq \frac{L}{2K} \|x_{0} - x^*\|^2.
        \end{aligned}
    \end{equation}
\end{proof}

Based on Corollary~\ref{ob:longer_horizon_ind_best_converge_gain}, we further analyze the difference between such a constraint in the variable and gradient methods. The definition in \eqref{thpf:longer_horizon_variable_case_p} and \eqref{thpf:longer_horizon_gradient_case_p} yields the following two different conditions for two methods, respectively.
\paragraph{Variable Method:}
\begin{equation*}
    \diag(\mathbf{N}_{3}(z_{k-1}))\diag(\mathbf{N}_{4}(u_{k-1}))(-x_{k-1} + x_{k-2}) = 0,
\end{equation*}
where $u_{k-1} = [x_{k-1}^\top, x_{k-2}^\top]^\top$ is the feature constructed with variable.

\paragraph{Gradient Method:}
\begin{equation*}
    \begin{aligned}
        \diag(\mathbf{N}_{3}(z_{k-1}))\diag(\mathbf{N}_{4}(w_{k-1}))\left(- \left(\nabla f(x_{k-1})+g_{k-1}\right) + \nabla f(x_{k-2}) +g_{k-2}\right) = 0,
    \end{aligned}
\end{equation*}
where $w_{k-1} = [(\nabla f(x_{k-1})+g_{k-1})^\top, (\nabla f(x_{k-2})+g_{k-2})^\top]^\top$ is the feature from gradient and subgradient. Moreover, there are two extra neural network operators, $\mathbf{N}_{5}$ and $\mathbf{N}_{6}$, to construct the subgradient vectors $g_{k-1}$ and $g_{k-2}$, respectively:
\begin{equation*}
    \begin{aligned}
        r_{k-1} &= [\partial r(x_{k-1})_{\text{lb}}^\top, \partial r(x_{k-1})_{\text{ub}}^\top]^\top, \\
        r_{k-2} &= [\partial r(x_{k-2})_{\text{lb}}^\top, \partial r(x_{k-2})_{\text{ub}}^\top]^\top,\\
        g_{k-1} &= (\mathbf{I} - \diag(\mathbf{N}_{5}(r_{k-1})))\partial r(x_{k-1})_{\text{lb}} + \diag(\mathbf{N}_{5}(r_{k-1}))\partial r(x_{k-1})_{\text{ub}},\\
        g_{k-2} &= (\mathbf{I} - \diag(\mathbf{N}_{5}(r_{k-2})))\partial r(x_{k-2})_{\text{lb}} + \diag(\mathbf{N}_{5}(r_{k-2}))\partial r(x_{k-2})_{\text{ub}}.
    \end{aligned}
\end{equation*}

We denote $\mathcal{U}_P$ and $\mathcal{W}_P$ as feature spaces upon InD variable space $\mathcal{S}_P$, which are similarly defined as $\mathcal{Z}_P$. 
Inductively, we define that $P_{k-1} := 0$ is given by $\mathbf{N}_{4}(u_{k-1}) := 0$ and $\mathbf{N}_{4}(w_{k-1}) := 0$ in the variable and gradient methods respectively, $\forall u_{k-1} \in \mathcal{U}_P$ and $\forall w_{k-1} \in \mathcal{W}_P$.

We assume both the variable method and gradient methods when modeling longer horizon modeling achieve robustness after training, as in the following assumption:
\begin{assumption} \label{as:assump_ind_robust_longer_horizon}
    After training, $\forall x_{k-1} \in \mathcal{S}_P, \forall z_{k-1} \in \mathcal{Z}_P$, $\forall u_{k-1} \in \mathcal{U}_P$, $\forall w_{k-1} \in \mathcal{W}_P$,  $\mathbf{N}_1(z_{k-1}):= \frac{1}{L}\mathbf{1}$, $\mathbf{N}_2(z_{k-1}) := \mathbf{0}$, $\mathbf{N}_{3}(z_{k-1}) := \mathbf{0}$, $\mathbf{N}_{4}(u_{k-1}) := \mathbf{0}$, and $\mathbf{N}_{4}(w_{k-1}) := \mathbf{0}$.
\end{assumption}

\subsubsection*{OOD Definitions}
Similar to the composite case, we first derive some preliminary formulations for OOD scenarios before the demonstrations. The following definitions follow the same workflow for those in the composite case.

Suppose $z, \tilde{z}, z^\prime \in \mathcal{Z}$, there exists a $\alpha \in [0,1]$ that $z^\prime := \alpha z + (1-\alpha)\tilde{z}, z^\prime \in \mathcal{Z}$. Denote the virtual Jacobian matrix of $\mathbf{N}_1(z^\prime)$, $\mathbf{N}_2(z^\prime)$, $\mathbf{N}_3(z^\prime)$ at point $z^\prime$ as $\mathbf{J}_1$, $\mathbf{J}_2$, $\mathbf{J}_3$, respectively, and $\|\mathbf{J}_1\| \leq \sqrt{n} C_1$, $\|\mathbf{J}_2\| \leq \sqrt{n}C_2$, and $\|\mathbf{J}_3\| \leq \sqrt{n}C_3$. 

Since $\mathbf{N}_1(z)$, $\mathbf{N}_2(z)$, $\mathbf{N}_3(z)$ are smooth, due to the Mean Value Theorem, we have the following equalities:
\begin{equation*}
    \mathbf{N}_1(z) = \mathbf{N}_1(\tilde{z}) + \mathbf{J}_1(z - \tilde{z}),\quad \mathbf{N}_2(z) = \mathbf{N}_2(\tilde{z}) + \mathbf{J}_2(z - \tilde{z}),\quad \mathbf{N}_3(z) = \mathbf{N}_3(\tilde{z}) + \mathbf{J}_3(z - \tilde{z}).
\end{equation*}

Given an OOD virtual variable $s \in \mathbb{R}^n$ (difference in variables between OOD and InD scenarios), we denote the virtual feature (difference in L2O model's input features between OOD and InD scenarios) as $s^\prime$. For $z + s^\prime$, based on Assumption~\ref{as:assump_ind_robust_longer_horizon}, we have the following representations of the L2O model's outputs in the OOD scenario by those in the InD scenario:
\begin{equation} \label{thpf:lb_N1N2N3_longer_horizon}
    \begin{aligned}
        &\mathbf{N}_1(z+s^\prime) = \mathbf{N}_1(z) + \mathbf{J}_1(z + s^\prime - z) = \mathbf{N}_1(z) +\mathbf{J}_1 s^\prime, \\
        &\mathbf{N}_2(z+s^\prime) = \mathbf{N}_2(z) + \mathbf{J}_2(z + s^\prime - z) = \mathbf{N}_2(z) +\mathbf{J}_2 s^\prime, \\
        &\mathbf{N}_3(z+s^\prime) = \mathbf{N}_3(z) + \mathbf{J}_3(z + s^\prime - z) = \mathbf{N}_3(z) +\mathbf{J}_3 s^\prime.
    \end{aligned}
\end{equation}

% \begin{equation} \label{thpf:ub_N1N2N3_longer_horizon}
%     \begin{aligned}
%         \mathbf{N}_1(z+s^\prime) \leq \mathbf{N}_1(z)+\mathbf{J}_{1, z}(z+s^\prime-z)+\frac{C_1}{2}\|z+s^\prime-z|^2 = \mathbf{N}_1(z)+\mathbf{J}_{1, z}(s^\prime)+\frac{C_1}{2}\|s^\prime\|^2\\
%         \mathbf{N}_2(z+s^\prime) \leq \mathbf{N}_2(z)+\mathbf{J}_{2, z}(z+s^\prime-z)+\frac{C_2}{2}\|z+s^\prime-z\|^2 = \mathbf{N}_2(z)+\mathbf{J}_{2, z}(s^\prime)+\frac{C_2}{2}\|s^\prime\|^2\\
%         \mathbf{N}_3(z+s^\prime) \leq \mathbf{N}_3(z)+\mathbf{J}_{3, z}(z+s^\prime-z)+\frac{C_3}{2}\|z+s^\prime-z\|^2 = \mathbf{N}_3(z)+\mathbf{J}_{3, z}(s^\prime)+\frac{C_3}{2}\|s^\prime\|^2.
%     \end{aligned}
% \end{equation}

Further, for historical modeling operator $\mathbf{N}_4$, we have the following two definitions for variable and gradient methods since different modeling methods have different input feature selections.

\subsubsection*{Variable Method}
We take a similar construction to represent OOD output with InD output for $\mathbf{N}_4$. 
Suppose $u, \tilde{u}, u^\prime \in \mathcal{U}$ where $\mathcal{U}$ denotes variable space of operator $\mathbf{N}_4$ for the gradient method. There exists a $\alpha \in [0,1]$ that $u^\prime := \alpha u + (1-\alpha)\tilde{u}, u^\prime \in \mathcal{U}$. Denote the virtual Jacobian matrix of $\mathbf{N}_4(u^\prime)$ at point $u^\prime$ as $\mathbf{J}_4$, $\|\mathbf{J}_4\| \leq \sqrt{n} C_4$, $\mathbf{N}_4(u^\prime)$ follows:
\begin{equation*} 
    \mathbf{N}_4(u) = \mathbf{N}_4(\tilde{u}) + \mathbf{J}_4(u - \tilde{u}).
\end{equation*}
Given two virtual variables $s_{1}, s_{2} \in \mathbb{R}^n$ (variable difference), the difference of neural network $\mathbf{N}_4$ between OOD and InD scenarios $u^\prime$ is defined as:
\begin{equation} \label{thpf:longer_horizon_u_def}
    u^\prime=[s_{1}^\top, s_{2}^\top]^\top.
\end{equation}

For $u + u^\prime$, based on Assumption~\ref{as:assump_ind_robust_longer_horizon}, $\mathbf{N}_4(u)=0$, we have the following representation of OOD output with InD output:
\begin{equation} \label{thpf:longer_horizon_lb_N4_variable}
    \mathbf{N}_4(u+u^\prime) = \mathbf{N}_4(u) + \mathbf{J}_4(u + u^\prime - u) = \mathbf{N}_4(u) + \mathbf{J}_4 u^\prime = \mathbf{J}_4 u^\prime.
\end{equation}
% \begin{equation} \label{thpf:longer_horizon_ub_N4_variable}
%     \mathbf{N}_4(u+u^\prime) \leq \mathbf{N}_4(u)+\mathbf{J}_{4, u}(u+u^\prime-u)+\frac{C_4}{2}\|u+u^\prime-u\|^2 = \mathbf{N}_4(u)+\mathbf{J}_{4, u}(u^\prime)+\frac{C_4}{2}\|u^\prime\|^2.
% \end{equation}

The OOD historical modeling result $y^\prime_{k-1}$ of the variable method is given by:
\begin{equation*}
    \begin{aligned}
        y^\prime_{k-1} &= -\mathbf{N}_{4}(u^\prime_{k-1}) x^\prime_{k-1} + \mathbf{N}_{4}(u^\prime_{k-1})x^\prime_{k-2} , \\
        &= - \mathbf{N}_{4}(u^\prime_{k-1})(x_{k-1}+s_{k-1}) + \mathbf{N}_{4}(u^\prime_{k-1})(x_{k-2}+s_{k-2}) , \\
        & = - \diag(\mathbf{J}_4 u^\prime)(x_{k-1}+s_{k-1}) + \diag(\mathbf{J}_4 u^\prime)(x_{k-2}+s_{k-2}).
    \end{aligned}
\end{equation*}

The InD historical modeling result $y_{k-1}$ of the variable method is given by:
\begin{equation*}
    y_{k-1} = -\mathbf{N}_{4}(u_{k-1}) x_{k-1} + \mathbf{N}_{4}(u_{k-1})x_{k-2} = 0, 
\end{equation*}

Their difference between OOD and InD scenarios is given by:
\begin{equation*}
    \begin{aligned}
        y^\prime_{k-1} - y_{k-1} &= - \diag(\mathbf{J}_4 u^\prime)(x_{k-1}+s_{k-1}) + \diag(\mathbf{J}_4 u^\prime)(x_{k-2}+s_{k-2}) - 0 \\
        &= -\diag(\mathbf{J}_4 u^\prime)(x_{k-1}-x_{k-2} + s_{k-1}-s_{k-2}).
    \end{aligned}
\end{equation*}

Based on the above definitions, at $k$-th iteration, virtual feature $s^\prime$ (difference of features between OOD and InD scenarios) of the variable method is defined by:
\begin{equation} \label{thpf:longer_horizon_s_prime_variable}
    \begin{aligned}
        s^\prime_{k-1} & = \left[s_{k-1}^\top, (\nabla f^\prime(x_{k-1}+s_{k-1}) - \nabla f(x))^\top, (g^\prime_{k} - g_{k})^\top, (y^\prime_{k-1} - y_{k-1})^\top\right]^\top , \\
        & = \left[s_{k-1}^\top, (\nabla f^\prime(x_{k-1}+s_{k-1}) - \nabla f(x))^\top, (g^\prime_{k} - g_{k})^\top, \big(-\diag(\mathbf{J}_4 u^\prime)(x_{k-1}-x_{k-2} + s_{k-1}-s_{k-2})\big)^\top\right]^\top, 
    \end{aligned}
\end{equation}
where $g^\prime_{k} \in \partial r^\prime(x_{k}+s_{k})$ and $g_{k} \in \partial r(x_{k})$.

\subsubsection*{Gradient Method}
Similarly, suppose $w, \tilde{w}, w^\prime \in \mathcal{W}$, where $\mathcal{W}$ denotes variable space of operator $\mathbf{N}_4$ for the gradient method. There exists a $\alpha \in [0,1]$ that $w^\prime := \alpha w + (1-\alpha)\tilde{w}, w^\prime \in \mathcal{W}$. Denote virtual Jacobian matrix of $\mathbf{N}_4(w^\prime)$ at point $w^\prime$ as $\mathbf{J}_4$, $\|\mathbf{J}_4\| \leq \sqrt{n}C_4$, $\mathbf{N}_4(w^\prime)$ follows:
\begin{equation} 
    \mathbf{N}_4(w) = \mathbf{N}_4(\tilde{w}) + \mathbf{J}_4(w - \tilde{w}).
\end{equation}
Given two virtual variable $s_{1}, s_{2} \in \mathbb{R}^n$ (difference of variables between OOD and InD scenarios), the difference of neural network $\mathbf{N}_4$ between OOD and InD scenarios $w^\prime$ is defined as: 
\begin{equation} \label{thpf:longer_horizon_w_def}
    w^\prime=[(\nabla f(x_{1}+s_{1})+g^\prime_{1})^\top - (\nabla f(x_{1})+g_{1})^\top, (\nabla f(x_{2}+s_{2})+g^\prime_{2})^\top - (\nabla f(x_{2})+g_{2})^\top]^\top.
\end{equation}
For $w + w^\prime$, based on Assumption~\ref{as:assump_ind_robust_longer_horizon}, $\mathbf{N}_4(w)=0$,, we have the following equalities:
\begin{equation} \label{thpf:longer_horizon_lb_N4_gradient}
    \mathbf{N}_4(w+w^\prime) = \mathbf{N}_4(w) + \mathbf{J}_4(w + w^\prime - W) = \mathbf{J}_4 w^\prime.
\end{equation}
% \begin{equation} \label{thpf:longer_horizon_ub_N4_gradient}
%     \mathbf{N}_4(w+w^\prime) \leq \mathbf{N}_4(w)+\mathbf{J}_{4, w}(w+w^\prime-w)+\frac{C_4}{2}\|w+w^\prime-w\|^2 = \mathbf{N}_4(w)+\mathbf{J}_{4, w}(w^\prime)+\frac{C_4}{2}\|w^\prime\|^2.
% \end{equation}

We eliminate the definition for $\mathbf{N}_5$ since we have defined it to be a diagonal matrix whose diagonal entries $\in [0, 1]$.

The OOD historical modeling result $y^\prime_{k-1}$ of the gradient method is given by:
\begin{equation*}
    \begin{aligned}
        y^\prime_{k-1} & = - \mathbf{N}_{4}(w^\prime_{k-1}) (\nabla f^\prime(x^\prime_{k-1})+g^\prime_{k-1}) + \mathbf{N}_{4}(w^\prime_{k-1})(\nabla f^\prime(x^\prime_{k-2})+g^\prime_{k-2})\\
        & = - \mathbf{N}_{4}(w^\prime_{k-1}) (\nabla f^\prime(x_{k-1}+s_{k-1})+g^\prime_{k-1}) + \mathbf{N}_{4}(w^\prime_{k-1})(\nabla f^\prime(x_{k-2}+s_{k-2})+g^\prime_{k-2}) \\
        & =  -\diag(\mathbf{J}_4 w^\prime) (\nabla f^\prime(x_{k-1}+s_{k-1})+g^\prime_{k-1}) + \diag(\mathbf{J}_4 w^\prime)(\nabla f^\prime(x_{k-2}+s_{k-2})+g^\prime_{k-2})
    \end{aligned}
\end{equation*}

The InD historical modeling result $y_{k-1}$ of the variable method is given by:
\begin{equation*}
    y_{k-1}  = - \mathbf{N}_{4}(w_{k-1}) (\nabla f(x_{k-1})+g_{k-1}) + \mathbf{N}_{4}(w_{k-1})(\nabla f^\prime(x_{k-2})+g^\prime_{k-2}) = 0.
\end{equation*}

Their difference between OOD and InD scenarios is given by:
\begin{equation*}
    \begin{aligned}
        y^\prime_{k-1} - y_{k-1} =& - \diag(\mathbf{J}_4 w^\prime) (\nabla f^\prime(x_{k-1}+s_{k-1})+g^\prime_{k-1}) + \diag(\mathbf{J}_4 w^\prime)(\nabla f^\prime(x_{k-2}+s_{k-2})+g^\prime_{k-2}) - 0 \\
        =& -\diag(\mathbf{J}_4 w^\prime) \big(\nabla f^\prime(x_{k-1}+s_{k-1})-\nabla f^\prime(x_{k-2}+s_{k-2}) + g^\prime_{k-1}-g^\prime_{k-2}\big).
    \end{aligned}
\end{equation*}

Based on above definitions, at $k$-th iteration, virtual feature $s^\prime$ (difference between OOD and InD scenarios) of the gradient method is defined by:
\begin{equation} \label{thpf:longer_horizon_s_prime_gradient}
    \begin{aligned}
        & s^\prime_{k-1} \\
        = &  \left[(\nabla f^\prime(x_{k-1}+s_{k-1}) - \nabla f(x))^\top, (g^\prime_{k} - g_{k})^\top, (y^\prime_{k-1} - y_{k-1})^\top\right]^\top \\
        =&  \Bigg[(\nabla f^\prime(x_{k-1}+s_{k-1}) - \nabla f(x))^\top, (g^\prime_{k} - g_{k})^\top, \\
        & \Big(-\diag(\mathbf{J}_4 w^\prime) \big(\nabla f^\prime(x_{k-1}+s_{k-1})-\nabla f^\prime(x_{k-2}+s_{k-2}) + g^\prime_{k-1}-g^\prime_{k-2}\big) \Big)^\top\Bigg]^\top ,
    \end{aligned}
\end{equation}
where $g^\prime_{k} \in \partial r^\prime(x_{k}+s_{k})$ and $g_{k} \in \partial r(x_{k})$.

\paragraph{OOD Update Formulation}
Similar to that in the composite case, 
based on Lemma~\ref{thpf:lm_GM_basic_bound_long}, $\forall x_k \in \mathcal{S}_p, s_k \in \mathbb{R}^n$, OOD yields the following inequality between any two values of objective:
\begin{equation} \label{thpf:longer_horizon_ood_ub_pre1}
    \begin{aligned}
        &F^\prime(x_{k} + s_{k}) \\
        \leq&  F^\prime(t) + G_{\diag(\mathbf{N}_1(z_{k-1} + s^\prime_{k-1}))^{-1}}(x_{k-1} +s_{k-1})^\top (x_{k-1} +s_{k-1} - t) \\
        & + \frac{L}{2} \Bigg\| \diag(\mathbf{N}_1(z_{k-1} + s^\prime_{k-1}))G_{\diag(\mathbf{N}_1(z_{k-1} + s^\prime_{k-1}))^{-1}}(x_{k-1} +s_{k-1}) +\mathbf{N}_2(z_{k-1} + s^\prime_{k-1}) + \diag(\mathbf{N}_{3}(z_{k-1} + s^\prime_{k-1}))P^\prime_{k-1} \\
        & \qquad \quad - \frac{G_{\diag(\mathbf{N}_1(z_{k-1} + s^\prime_{k-1}))^{-1}}(x_{k-1} +s_{k-1})}{L}\Bigg\|^2  - \frac{L}{2}\Bigg\|\frac{G_{\diag(\mathbf{N}_1(z_{k-1} + s^\prime_{k-1}))^{-1}}(x_{k-1} +s_{k-1})}{L}\Bigg\|^2.
    \end{aligned}
\end{equation}

Similar to the composite case, we directly get the following formulation for the OOD gradient map:
\begin{equation} \label{thpf:G_fg_equation_ood_longer_horizon}
    G^\prime_{\diag(\mathbf{N}_1(z_{k-1} + s^\prime_{k-1}))^{-1}}(x_{k-1} +s_{k-1}) =  \nabla f^\prime(x_{k-1} +s_{k-1}) + g^\prime_{k},
\end{equation}
where $g^\prime_{k} \in \partial r^\prime(x_{k} +s_{k})$.

\subsection{OOD Per-Iteration Convergence Gain} \label{sec:OOD_analysis}
Based on the Lemma~\ref{lm:ind_converge_gain_longer_horizon} and Corollary~\ref{ob:longer_horizon_ind_best_converge_gain}, Assumption \ref{as:assump_ind_robust_longer_horizon} leads to an L2O model with best robustness on all InD instances.

Based on Assumption \ref{as:assump_ind_robust_longer_horizon}, in the following theorem, we quantify the diminution in convergence rate instigated by the virtual feature $s^\prime$ defined in \cref{sec:vv_traj}. 

\begin{theorem} \label{lm:ood_converge_gain_longer_horizon}
    Under Assumption \ref{as:assump_ind_robust_longer_horizon}, there exists virtual Jacobian matrices $\mathbf{J}_{1,k-1}, \mathbf{J}_{2,k-1}, \mathbf{J}_{3,k-1}, k= 1,2, \dots, K$ that OOD's convergence improvement of one iteration is upper bounded by following inequality:
    \begin{equation*}
        \begin{aligned}
            & F^\prime(x_k + s_k) - F^\prime(x_{k-1} + s_{k-1}) \\
             \leq &-\frac{\|\nabla f^\prime(x_{k-1} +s_{k-1}) + g^\prime_{k}\|^2}{2L}   + \frac{L}{2} \| \diag(\mathbf{J}_1 s^\prime_{k-1})(\nabla f^\prime(x_{k-1} +s_{k-1}) + g^\prime_{k}) + \mathbf{J}_2 s^\prime_{k-1} + \diag(\mathbf{J}_3 s^\prime_{k-1})P^\prime_{k-1} \|^2,
        \end{aligned}
    \end{equation*}
    where $g^\prime_{k} \in \partial r^\prime(x_{k} +s_{k})$ and $P^\prime_{k-1}$ represents historical modeling result.
\end{theorem}

\begin{proof}
    From \eqref{thpf:lb_N1N2N3_longer_horizon}, for operators $\mathbf{N}_1$, $\mathbf{N}_2$, and $\mathbf{N}_3$, we have the following representations of OOD outputs by their InD outputs:
    \begin{equation*}
        \begin{aligned}
            &\mathbf{N}_1(z_{k-1} + s^\prime_{k-1})  = \mathbf{N}_1(z_{k-1}) + \mathbf{J}_1 s^\prime_{k-1} , \\
            &\mathbf{N}_2(z_{k-1} + s^\prime_{k-1}) = \mathbf{N}_2(z_{k-1}) + \mathbf{J}_2 s^\prime_{k-1} , \\
            &\mathbf{N}_3(z_{k-1}+s^\prime_{k-1}) = \mathbf{N}_3(z_{k-1}) + \mathbf{J}_3 s^\prime_{k-1}.
        \end{aligned}
    \end{equation*}

    Substituting the definitions of $\mathbf{N}_1(z_{k-1})$, $\mathbf{N}_2(z_{k-1})$, and $\mathbf{N}_3(z_{k-1})$ in Assumption \ref{as:assump_ind_robust_longer_horizon} yields:
    \begin{equation} \label{thpf:ood_N1N2_longer_horizon}
        \begin{aligned}
            &\mathbf{N}_1(z_{k-1} + s^\prime_{k-1})  = \frac{1}{L}\mathbf{1} + \mathbf{J}_1 s^\prime_{k-1}\\
            &\mathbf{N}_2(z_{k-1} + s^\prime_{k-1}) = \mathbf{J}_2 s^\prime_{k-1} \\
            &\mathbf{N}_3(z_{k-1} + s^\prime_{k-1}) = \mathbf{J}_3 s^\prime_{k-1}.
        \end{aligned}
    \end{equation}

    We substitut $t := x_{k-1} +s_{k-1}$ into inequation \ref{thpf:longer_horizon_ood_ub_pre1} to construct the objective difference between two iterations:
    \begin{equation*}
        \begin{aligned}
            &F^\prime(x_{k} + s_{k}) - F^\prime(x_{k-1} +s_{k-1})\\
            \leq&  \frac{L}{2} \Bigg\| \diag(\mathbf{N}_1(z_{k-1} + s^\prime_{k-1}))G^\prime_{\diag(\mathbf{N}_1(z_{k-1} + s^\prime_{k-1}))^{-1}}(x_{k-1} +s_{k-1}) +\mathbf{N}_2(z_{k-1} + s^\prime_{k-1}) + \diag(\mathbf{N}_{3}(z_{k-1} + s^\prime_{k-1}))P^\prime_{k-1} \\
            & \qquad - \frac{G^\prime_{\diag(\mathbf{N}_1(z_{k-1} + s^\prime_{k-1}))^{-1}}(x_{k-1} +s_{k-1})}{L}\Bigg\|^2 - \frac{L}{2}\Bigg\|\frac{G^\prime_{\diag(\mathbf{N}_1(z_{k-1} + s^\prime_{k-1}))^{-1}}(x_{k-1} +s_{k-1})}{L}\Bigg\|^2.
        \end{aligned}
    \end{equation*}

    By \eqref{thpf:ood_N1N2_longer_horizon}, we can represent the OOD outputs and achive the following reformulation:
    \begin{equation*}
        \begin{aligned}
            &F^\prime(x_{k} + s_{k}) - F^\prime(x_{k-1} +s_{k-1})\\
            \leq&  \frac{L}{2} \| \diag(\mathbf{J}_1 s^\prime_{k-1})G^\prime_{\diag(\mathbf{N}_1(z_{k-1} + s^\prime_{k-1}))^{-1}}(x_{k-1} +s_{k-1}) + \mathbf{J}_2 s^\prime_{k-1} +\diag(\mathbf{J}_3 s^\prime_{k-1})P^\prime_{k-1} \|^2 \\
            & - \frac{L}{2}\left\|\frac{G^\prime_{\diag(\mathbf{N}_1(z_{k-1} + s^\prime_{k-1}))^{-1}}(x_{k-1} +s_{k-1})}{L}\right\|^2.
        \end{aligned}
    \end{equation*}

    Then, based on \eqref{thpf:G_fg_equation_ood_longer_horizon}, we recover gradient and subgradient from the gradient map:
    \begin{equation*}
        \begin{aligned}
            &F^\prime(x_{k} + s_{k}) - F^\prime(x_{k-1} +s_{k-1})\\
            \leq&   - \frac{L}{2}\left\|\frac{\nabla f^\prime(x_{k-1} +s_{k-1}) + g^\prime_{k}}{L}\right\|^2 + \frac{L}{2} \| \diag(\mathbf{J}_1 s^\prime_{k-1})(\nabla f^\prime(x_{k-1} +s_{k-1}) + g^\prime_{k}) + \mathbf{J}_2 s^\prime_{k-1} + \diag(\mathbf{J}_3 s^\prime_{k-1})P^\prime_{k-1} \|^2 , \\
            =& - \frac{\|\nabla f^\prime(x_{k-1} +s_{k-1}) + g^\prime_{k}\|^2}{2L} + \frac{L}{2} \| \diag(\mathbf{J}_1 s^\prime_{k-1})(\nabla f^\prime(x_{k-1} +s_{k-1}) + g^\prime_{k}) + \mathbf{J}_2 s^\prime_{k-1}  + \diag(\mathbf{J}_3 s^\prime_{k-1})P^\prime_{k-1}\|^2.
        \end{aligned}
    \end{equation*}
\end{proof}

For variable and gradient methods, based on Theorem~\ref{lm:ood_converge_gain_longer_horizon}, we have the following different theorems of per iteration convergence gain.

\subsubsection*{Variable Method} 
From \eqref{thpf:longer_horizon_lb_N4_variable} and definition in Assumption \ref{as:assump_ind_robust_longer_horizon}, we have the following representation of $\mathbf{N}_4(u+u^\prime)$:
\begin{equation*}
    \mathbf{N}_4(u+u^\prime) = \mathbf{J}_4 u^\prime.
\end{equation*}

For the variable method, Theorem~\ref{lm:ood_converge_gain_longer_horizon} yields the following theorem of the per iteration convergence gain:
\begin{theorem} \label{th:ood_converge_gain_longer_horizon_variable}
    Under Assumption \ref{as:assump_ind_robust_longer_horizon}, there exists virtual Jacobian matrices $\mathbf{J}_{1,k-1}, \mathbf{J}_{2,k-1}, \mathbf{J}_{3,k-1}, \mathbf{J}_{4,k-1}, k= 1,2, \dots, K$ that OOD's convergence improvement of one iteration is upper bounded by following inequality:
    \begin{equation*}
        \begin{aligned}
            & F^\prime(x_k + s_k) - F^\prime(x_{k-1} + s_{k-1}) \\
             \leq &-\frac{\|\nabla f^\prime(x_{k-1} +s_{k-1}) + g^\prime_{k}\|^2}{2L}   + \frac{L}{2} \| \diag(\mathbf{J}_1 s^\prime_{k-1})(\nabla f^\prime(x_{k-1} +s_{k-1}) + g^\prime_{k}) + \mathbf{J}_2 s^\prime_{k-1} \\
             & + \diag(\mathbf{J}_3 s^\prime_{k-1}) \diag(\mathbf{J}_4 u^\prime_{k-1})(-(x_{k-1} + s_{k-1}) + x_{k-2} + s_{k-2}) \|^2,
        \end{aligned}
    \end{equation*}
    where $g^\prime_{k} \in \partial r^\prime(x_{k} +s_{k})$.
\end{theorem}

Theorem~\ref{th:ood_converge_gain_longer_horizon_variable} yields following corollary for its upper bound:
\begin{corollary} \label{lm:ood_converge_gain_ub_longer_horizon_variable}
    Under Assumption \ref{as:assump_ind_robust_longer_horizon}, the convergence improvement for one iteration of the OOD scenario can be upper bounded w.r.t. $s_{k-1}$ and $s_{k-2}$ by:
    \begin{equation*}
        \begin{aligned}
            & F^\prime(x_k + s_k) - F^\prime(x_{k-1} + s_{k-1}) \\
            \leq &-\frac{\|\nabla f^\prime(x_{k-1} +s_{k-1}) + g^\prime_{k}\|^2}{2L}  \\
            & + \Big(L C_1^2   \| \nabla f^\prime(x_{k-1} +s_{k-1}) + g^\prime_{k} \|^2 +L C_2^2 +L C_3^2 C_4^2 (\|s_{k-1}\|^2 + \|s_{k-2}\|^2) \|x_{k-1}-x_{k-2} + s_{k-1}-s_{k-2}\|^2 \Big) \\
            & \quad \times \Big(\|s_{k-1}\|^2 + \|\nabla f^\prime(x_{k-1} + s_{k-1}) - \nabla f(x_{k-1})\|^2 + \|g^\prime_{k} - g_{k}\|^2 \\
            & \qquad \quad + C_4^2 (\|s_{k-1}\|^2 + \|s_{k-2}\|^2) \|x_{k-1}-x_{k-2} + s_{k-1}-s_{k-2}\|^2\Big),
        \end{aligned}
    \end{equation*}
    where $g^\prime_{k} \in \partial r^\prime(x_{k} +s_{k})$ and $g_{k} \in \partial r(x_{k})$.
\end{corollary}
\begin{proof}
    We iteratively apply Triangle and Cauchy Schwarz inequalities:
    \begin{equation} \label{thpf:ood_converge_gain_ub_longer_horizon_variable_pre}
        \begin{aligned}
            & F^\prime(x_k + s_k) - F^\prime(x_{k-1} + s_{k-1}) \\
            \leq &-\frac{\|\nabla f^\prime(x_{k-1} +s_{k-1}) + g^\prime_{k}\|^2}{2L}   + \frac{L}{2} \| \diag(\mathbf{J}_1 s^\prime_{k-1})(\nabla f^\prime(x_{k-1} +s_{k-1}) + g^\prime_{k}) + \mathbf{J}_2 s^\prime_{k-1} \\
            & + \diag(\mathbf{J}_3 s^\prime_{k-1}) \diag(\mathbf{J}_4 u^\prime_{k-1})(-(x_{k-1} + s_{k-1}) + x_{k-2} + s_{k-2}) \|^2 , \\
            \leq &-\frac{\|\nabla f^\prime(x_{k-1} +s_{k-1}) + g^\prime_{k}\|^2}{2L}   + L \| \diag(\mathbf{J}_1 s^\prime_{k-1})(\nabla f^\prime(x_{k-1} +s_{k-1}) + g^\prime_{k}) \|^2 +  L \|\mathbf{J}_2 s^\prime_{k-1} \|^2 \\
            & + L\|\diag(\mathbf{J}_3 s^\prime_{k-1}) \diag(\mathbf{J}_4 u^\prime_{k-1})(-(x_{k-1} + s_{k-1}) + x_{k-2} + s_{k-2}) \|^2 , \\
            \leq &-\frac{\|\nabla f^\prime(x_{k-1} +s_{k-1}) + g^\prime_{k}\|^2}{2L}   + L \| \mathbf{J}_1 s^\prime_{k-1} \|^2 \| \nabla f^\prime(x_{k-1} +s_{k-1}) + g^\prime_{k} \|^2 +  L \|\mathbf{J}_2 s^\prime_{k-1} \|^2 \\
            & + L\|\mathbf{J}_3 s^\prime_{k-1}\|^2 \|\mathbf{J}_4 u^\prime_{k-1}\|^2 \|-(x_{k-1} + s_{k-1}) + x_{k-2} + s_{k-2}\|^2 , \\
            \leq &-\frac{\|\nabla f^\prime(x_{k-1} +s_{k-1}) + g^\prime_{k}\|^2}{2L}   + L C_1^2   \| \nabla f^\prime(x_{k-1} +s_{k-1}) + g^\prime_{k} \|^2 \| s^\prime_{k-1} \|^2 +  L C_2^2 \| s^\prime_{k-1} \|^2\\
            & + L C_3^2 C_4^2 \|s^\prime_{k-1}\|^2 \|u^\prime_{k-1}\|^2 \|x_{k-1}-x_{k-2} + s_{k-1}-s_{k-2}\|^2 , \\
            = &-\frac{\|\nabla f^\prime(x_{k-1} +s_{k-1}) + g^\prime_{k}\|^2}{2L} \\
            &  + \Big(L C_1^2   \| \nabla f^\prime(x_{k-1} +s_{k-1}) + g^\prime_{k} \|^2 +  L C_2^2 + L C_3^2 C_4^2  \|u^\prime_{k-1}\|^2 \|x_{k-1}-x_{k-2} + s_{k-1}-s_{k-2}\|^2 \Big) \| s^\prime_{k-1} \|^2 ,
        \end{aligned}
    \end{equation}
    where $g^\prime_{k} \in \partial r^\prime(x_{k} +s_{k})$.

    Based on the definition of $u^\prime_{k-1}$ in \eqref{thpf:longer_horizon_u_def}, we calculate its vector-norm by:
    \begin{equation*}
        \|u^\prime_{k-1}\|^2 = \|[s_{k-1}, s_{k-2}]\|^2 = \|s_{k-1}\|^2 + \|s_{k-2}\|^2.
    \end{equation*}
    Substituting it into inequation \ref{thpf:ood_converge_gain_ub_longer_horizon_variable_pre} yields the following upper bound:
    \begin{equation} \label{thpf:ood_converge_gain_ub_longer_horizon_variable_step1}
        \begin{aligned}
            & F^\prime(x_k + s_k) - F^\prime(x_{k-1} + s_{k-1}) \\
            \leq &-\frac{\|\nabla f^\prime(x_{k-1} +s_{k-1}) + g^\prime_{k}\|^2}{2L}  \\
            & + \Big(L C_1^2   \| \nabla f^\prime(x_{k-1} +s_{k-1}) + g^\prime_{k} \|^2 +L C_2^2 +L C_3^2 C_4^2 (\|s_{k-1}\|^2 + \|s_{k-2}\|^2) \|x_{k-1}-x_{k-2} + s_{k-1}-s_{k-2}\|^2 \Big) \| s^\prime_{k-1} \|^2.
        \end{aligned}
    \end{equation}

    Moreover, the definition of $s^\prime_{k-1}$ with variable method in \eqref{thpf:longer_horizon_s_prime_variable} yields:
    \begin{equation*}
        \begin{aligned}
            \| s^\prime_{k-1} \|^2 =& \|s_{k-1}\|^2 + \|\nabla f^\prime(x_{k-1} + s_{k-1}) - \nabla f(x_{k-1})\|^2 + \|g^\prime_{k} - g_{k}\|^2 + \|-\diag(\mathbf{J}_4 u^\prime)(x_{k-1}-x_{k-2} + s_{k-1}-s_{k-2})\|^2,  \\
            \leq& \|s_{k-1}\|^2 + \|\nabla f^\prime(x_{k-1} + s_{k-1}) - \nabla f(x_{k-1})\|^2  + \|g^\prime_{k} - g_{k}\|^2 + \|-\diag(\mathbf{J}_4 u^\prime)(x_{k-1}-x_{k-2} + s_{k-1}-s_{k-2})\|^2 , \\
            \leq& \|s_{k-1}\|^2 + \|\nabla f^\prime(x_{k-1} + s_{k-1}) - \nabla f(x_{k-1})\|^2 + \|g^\prime_{k} - g_{k}\|^2 + \|\mathbf{J}_4 u^\prime\|^2 \|x_{k-1}-x_{k-2} + s_{k-1}-s_{k-2}\|^2 , \\
            =& \|s_{k-1}\|^2 + \|\nabla f^\prime(x_{k-1} + s_{k-1}) - \nabla f(x_{k-1})\|^2 + \|g^\prime_{k} - g_{k}\|^2 + \|\mathbf{J}_4 u^\prime\|^2 \|x_{k-1}-x_{k-2} + s_{k-1}-s_{k-2}\|^2 , \\
            \leq & \|s_{k-1}\|^2 + \|\nabla f^\prime(x_{k-1} + s_{k-1}) - \nabla f(x_{k-1})\|^2 + \|g^\prime_{k} - g_{k}\|^2 + C_4^2 \|u^\prime\|^2 \|x_{k-1}-x_{k-2} + s_{k-1}-s_{k-2}\|^2 , \\
            = & \|s_{k-1}\|^2 + \|\nabla f^\prime(x_{k-1} + s_{k-1}) - \nabla f(x_{k-1})\|^2 + \|g^\prime_{k} - g_{k}\|^2 \\
            &+ C_4^2 \left(\|s_{k-1}\|^2 + \|s_{k-2}\|^2\right) \|x_{k-1}-x_{k-2} + s_{k-1}-s_{k-2}\|^2,
        \end{aligned}
    \end{equation*}
    where $g^\prime_{k} \in \partial r^\prime(x_{k} +s_{k})$, $g_{k} \in \partial r(x_{k})$, and the third step is based on Cauchy-Schwarz inequality.

    Substituting the above vector norm into the inequation \ref{thpf:ood_converge_gain_ub_longer_horizon_variable_step1} yields the final upper bound:
    \begin{equation*}
        \begin{aligned}
            & F^\prime(x_k + s_k) - F^\prime(x_{k-1} + s_{k-1}) \\
            \leq &-\frac{\|\nabla f^\prime(x_{k-1} +s_{k-1}) + g^\prime_{k}\|^2}{2L}  \\
            & + \Big(L C_1^2   \| \nabla f^\prime(x_{k-1} +s_{k-1}) + g^\prime_{k} \|^2 +L C_2^2 +L C_3^2 C_4^2 (\|s_{k-1}\|^2 + \|s_{k-2}\|^2) \|x_{k-1}-x_{k-2} + s_{k-1}-s_{k-2}\|^2 \Big) \\
            & \quad \times \Big(\|s_{k-1}\|^2 + \|\nabla f^\prime(x_{k-1} + s_{k-1}) - \nabla f(x_{k-1})\|^2 + \|g^\prime_{k} - g_{k}\|^2 \\
            & \quad \qquad + C_4^2 (\|s_{k-1}\|^2 + \|s_{k-2}\|^2) \|x_{k-1}-x_{k-2} + s_{k-1}-s_{k-2}\|^2\Big),
        \end{aligned}
    \end{equation*}
    where $g^\prime_{k} \in \partial r^\prime(x_{k} +s_{k})$ and $g_{k} \in \partial r(x_{k})$.
\end{proof}

\subsubsection*{Gradient Method}
From \eqref{thpf:longer_horizon_lb_N4_gradient} and definition in Assumption \ref{as:assump_ind_robust_longer_horizon}, we have the following representation of $\mathbf{N}_4(u+u^\prime)$:
\begin{equation*}
    \mathbf{N}_4(w+w^\prime) = \mathbf{J}_4 w^\prime.
\end{equation*}

For the gradient method, Theorem~\ref{lm:ood_converge_gain_longer_horizon} yields the following theorem of the per iteration convergence gain:
\begin{theorem}\label{th:ood_converge_gain_longer_horizon_gradient}
    Under Assumption \ref{as:assump_ind_robust_longer_horizon}, there exists virtual Jacobian matrices $\mathbf{J}_{1,k-1}, \mathbf{J}_{2,k-1}, \mathbf{J}_{3,k-1}, \mathbf{J}_{4,k-1}, k= 1,2, \dots, K$ that OOD's convergence improvement of one iteration is upper bounded by following inequality:
    \begin{equation*}
        \begin{aligned}
            & F^\prime(x_k + s_k) - F^\prime(x_{k-1} + s_{k-1}) \\
            \leq &-\frac{\|\nabla f^\prime(x_{k-1} +s_{k-1}) + g^\prime_{k}\|^2}{2L}   + \frac{L}{2} \| \diag(\mathbf{J}_1 s^\prime_{k-1})(\nabla f^\prime(x_{k-1} +s_{k-1}) + g^\prime_{k}) + \mathbf{J}_2 s^\prime_{k-1} \\
            & + \diag(\mathbf{J}_3 s^\prime_{k-1}) \diag(\mathbf{J}_4 w^\prime_{k-1})\big(-(\nabla f^\prime(x_{k-1}+s_{k-1})+g^\prime_{k-1}) + \nabla f^\prime(x_{k-2}+s_{k-2})+g^\prime_{k-2}\big) \|^2,
        \end{aligned}
    \end{equation*}
    where $g^\prime_{k} \in \partial r^\prime(x_{k} +s_{k})$ and $g^\prime_{k-1}, g^\prime_{k-2}$ follow:
    \begin{equation*}
        \begin{aligned}
            g^\prime_{k-1} &\in [\partial r^\prime(x_{k-1}+s_{k-1})_{\text{lb}}, \partial r^\prime(x_{k-1}+s_{k-1})_{\text{ub}}], \\
            g^\prime_{k-2} &\in [\partial r^\prime(x_{k-2}+s_{k-2})_{\text{lb}}, \partial r^\prime(x_{k-2}+s_{k-2})_{\text{ub}}].
        \end{aligned}
    \end{equation*}
\end{theorem}

Theorem~\ref{th:ood_converge_gain_longer_horizon_gradient} yields the following corollary for its upper bound:
\begin{corollary} \label{lm:ood_converge_gain_ub_longer_horizon_gradient}
    Under Assumption \ref{as:assump_ind_robust_longer_horizon}, the convergence improvement for one iteration of the OOD scenario can be upper bounded w.r.t. $s_{k-1}$ and $s_{k-2}$ by:
    \begin{equation*}
        \begin{aligned}
            & F^\prime(x_k + s_k) - F^\prime(x_{k-1} + s_{k-1}) \\
            \leq &-\frac{\|\nabla f^\prime(x_{k-1} +s_{k-1}) + g^\prime_{k}\|^2}{2L} \\
            & + \Big( Ln^2C_1^2 \| \nabla f^\prime(x_{k-1} +s_{k-1}) + g^\prime_{k} \|^2  +  Ln^2C_2^2 \\
            & \qquad + L n^4C_3^2 C_4^2 \big(L^2(\|s_{k-1}\|^2 + \|s_{k-2}\|^2) + \|g^\prime_{k-1}-g_{k-1}\|^2 + \|g^\prime_{k-2}-g_{k-2}\|^2\big) \\
            & \qquad \qquad \times\big(L^2\|x_{k-2}+s_{k-2}-x_{k-1}-s_{k-1}\|^2 + \|g^\prime_{k-2}-g^\prime_{k-1}\|^2 \big) \Big) \\
            & \quad \times \Big( \|\nabla f^\prime(x_{k-1} + s_{k-1}) - \nabla f(x_{k-1})\|^2 + \|g^\prime_{k} - g_{k}\|^2 \\
            & \qquad \quad + n^2C_4^2  \big(L^2(\|s_{k-1}\|^2 + \|s_{k-2}\|^2) + \|g^\prime_{k-1}-g_{k-1}\|^2 + \|g^\prime_{k-2}-g_{k-2}\|^2\big) \\
            & \qquad \qquad \times \big(L^2\|x_{k-1}+s_{k-1}-x_{k-2}-s_{k-2}\|^2 + \|g^\prime_{k-1}-g^\prime_{k-2}\|^2\big) \Big),
        \end{aligned}
    \end{equation*}
    where $g^\prime_{k} \in \partial r^\prime(x_{k} +s_{k})$, $g^\prime_{k-1} \in \partial r^\prime(x_{k-1} +s_{k-1})$, and $g^\prime_{k-2} \in \partial r^\prime(x_{k-2} +s_{k-2})$.
\end{corollary}
\begin{proof}
    \begin{equation} \label{thpf:ood_converge_gain_ub_longer_horizon_gradient_pre}
        \begin{aligned}
            & F^\prime(x_k + s_k) - F^\prime(x_{k-1} + s_{k-1}) \\
            \leq &-\frac{\|\nabla f^\prime(x_{k-1} +s_{k-1}) + g^\prime_{k}\|^2}{2L}   + \frac{L}{2} \| \diag(\mathbf{J}_1 s^\prime_{k-1})(\nabla f^\prime(x_{k-1} +s_{k-1}) + g^\prime_{k}) + \mathbf{J}_2 s^\prime_{k-1} \\
            & + \diag(\mathbf{J}_3 s^\prime_{k-1}) \diag(\mathbf{J}_4 w^\prime_{k-1})\big(-(\nabla f^\prime(x_{k-1}+s_{k-1})+g^\prime_{k-1}) + \nabla f^\prime(x_{k-2}+s_{k-2})+g^\prime_{k-2}\big) \|^2 , \\
            \leq &-\frac{\|\nabla f^\prime(x_{k-1} +s_{k-1}) + g^\prime_{k}\|^2}{2L}   + L \| \diag(\mathbf{J}_1 s^\prime_{k-1})(\nabla f^\prime(x_{k-1} +s_{k-1}) + g^\prime_{k}) \|^2 +  L \|\mathbf{J}_2 s^\prime_{k-1} \|^2 \\
            & + L\|\diag(\mathbf{J}_3 s^\prime_{k-1}) \diag(\mathbf{J}_4 w^\prime_{k-1}) \big(-(\nabla f^\prime(x_{k-1}+s_{k-1})+g^\prime_{k-1}) + \nabla f^\prime(x_{k-2}+s_{k-2})+g^\prime_{k-2}\big) \|^2 , \\
            \leq &-\frac{\|\nabla f^\prime(x_{k-1} +s_{k-1}) + g^\prime_{k}\|^2}{2L}   + L \| \mathbf{J}_1 s^\prime_{k-1} \|^2 \| \nabla f^\prime(x_{k-1} +s_{k-1}) + g^\prime_{k} \|^2 +  L \|\mathbf{J}_2 s^\prime_{k-1} \|^2 \\
            & + L\|\mathbf{J}_3 s^\prime_{k-1}\|^2 \|\mathbf{J}_4 w^\prime_{k-1}\|^2 \|\nabla f^\prime(x_{k-2}+s_{k-2})-\nabla f^\prime(x_{k-1}+s_{k-1})+ g^\prime_{k-2}-g^\prime_{k-1}\|^2 , \\
            \leq &-\frac{\|\nabla f^\prime(x_{k-1} +s_{k-1}) + g^\prime_{k}\|^2}{2L}   + L n^2C_1^2   \| \nabla f^\prime(x_{k-1} +s_{k-1}) + g^\prime_{k} \|^2 \| s^\prime_{k-1} \|^2 +  L n^2C_2^2 \| s^\prime_{k-1} \|^2\\
            & + Ln^2 C_3^2 n^2C_4^2 \|s^\prime_{k-1}\|^2 \|w^\prime_{k-1}\|^2 \|\nabla f^\prime(x_{k-2}+s_{k-2})-\nabla f^\prime(x_{k-1}+s_{k-1})+ g^\prime_{k-2}-g^\prime_{k-1}\|^2 , \\
            \leq &-\frac{\|\nabla f^\prime(x_{k-1} +s_{k-1}) + g^\prime_{k}\|^2}{2L}   + L n^2C_1^2   \| \nabla f^\prime(x_{k-1} +s_{k-1}) + g^\prime_{k} \|^2 \| s^\prime_{k-1} \|^2 +  L n^2C_2^2 \| s^\prime_{k-1} \|^2\\
            & + L n^4C_3^2 C_4^2 \|s^\prime_{k-1}\|^2 \|w^\prime_{k-1}\|^2 \big(\|\nabla f^\prime(x_{k-2}+s_{k-2})-\nabla f^\prime(x_{k-1}+s_{k-1})\|^2+ \|g^\prime_{k-2}-g^\prime_{k-1}\|^2\big) , \\
            \leq &-\frac{\|\nabla f^\prime(x_{k-1} +s_{k-1}) + g^\prime_{k}\|^2}{2L} +  \big( Ln^2C_1^2 \| \nabla f^\prime(x_{k-1} +s_{k-1}) + g^\prime_{k} \|^2  +  Ln^2C_2^2 \big) \| s^\prime_{k-1} \|^2 \\
            & + L n^4C_3^2 C_4^2  \|w^\prime_{k-1}\|^2 \big(L^2\|x_{k-2}+s_{k-2}-x_{k-1}-s_{k-1}\|^2 + \|g^\prime_{k-2}-g^\prime_{k-1}\|^2 \big) \|s^\prime_{k-1}\|^2,\\
            = &-\frac{\|\nabla f^\prime(x_{k-1} +s_{k-1}) + g^\prime_{k}\|^2}{2L} \\
            & + \Big( Ln^2C_1^2 \| \nabla f^\prime(x_{k-1} +s_{k-1}) + g^\prime_{k} \|^2  +  Ln^2C_2^2 \\
            & \qquad + L n^4 C_3^2 C_4^2  \|w^\prime_{k-1}\|^2 \big(L^2\|x_{k-2}+s_{k-2}-x_{k-1}-s_{k-1}\|^2 + \|g^\prime_{k-2}-g^\prime_{k-1}\|^2 \big) \Big) \| s^\prime_{k-1} \|^2,
        \end{aligned}
    \end{equation}
    where $g^\prime_{k} \in \partial r^\prime(x_{k} +s_{k})$, $g^\prime_{k-1} \in \partial r^\prime(x_{k-1} +s_{k-1})$, and $g^\prime_{k-2} \in \partial r^\prime(x_{k-2} +s_{k-2})$. The last step is based on the definition of $L$-smoothness on $f^\prime$.

    Based on the definition of $w^\prime_{k-1}$ in \eqref{thpf:longer_horizon_w_def}, we calculate its vector-norm by:
    \begin{equation*}
        w^\prime_{k-1} = \left[(\nabla f(x_{k-1}+s_{k-1})+g^\prime_{k-1} - \nabla f(x_{k-1})-g_{k-1})^\top, (\nabla f(x_{k-2}+s_{k-2})+g^\prime_{k-2} - \nabla f(x_{k-2}) -g_{k-2})^\top\right]^\top.
    \end{equation*}

    Thus, we have:
    \begin{equation*}
        \begin{aligned}
            &\|w^\prime_{k-1}\|^2\\
            =&\|\nabla f(x_{k-1}+s_{k-1})+g^\prime_{k-1} - \nabla f(x_{k-1})-g_{k-1}, \nabla f(x_{k-2}+s_{k-2})+g^\prime_{k-2} - \nabla f(x_{k-2}) -g_{k-2}\|^2 , \\
            =& \|\nabla f(x_{k-1}+s_{k-1})-\nabla f(x_{k-1}) +g^\prime_{k-1}-g_{k-1}, \nabla f(x_{k-2}+s_{k-2})-\nabla f(x_{k-2}) +g^\prime_{k-2}-g_{k-2}\|^2 , \\
            = & \|\nabla f(x_{k-1}+s_{k-1})-\nabla f(x_{k-1}) +g^\prime_{k-1}-g_{k-1}\|^2 + \|\nabla f(x_{k-2}+s_{k-2})-\nabla f(x_{k-2}) +g^\prime_{k-2}-g_{k-2}\|^2 , \\
            \leq & \|\nabla f(x_{k-1}+s_{k-1})-\nabla f(x_{k-1})\|^2 +\|g^\prime_{k-1}-g_{k-1}\|^2 + \|\nabla f(x_{k-2}+s_{k-2})-\nabla f(x_{k-2})\|^2 + \|g^\prime_{k-2}-g_{k-2}\|^2 , \\
            \leq & L^2\|x_{k-1}+s_{k-1}-x_{k-1}\|^2 +\|g^\prime_{k-1}-g_{k-1}\|^2 + L^2\|x_{k-2}+s_{k-2}-x_{k-2}\|^2 + \|g^\prime_{k-2}-g_{k-2}\|^2 , \\
            = & L^2(\|s_{k-1}\|^2 + \|s_{k-2}\|^2) + \|g^\prime_{k-1}-g_{k-1}\|^2 + \|g^\prime_{k-2}-g_{k-2}\|^2.
        \end{aligned}
    \end{equation*}
    In steps 1-3, we rearrange items. The 4th step is based on Triangle inequality. The 4th step is based on Cauchy-Schwarz inequality. The 5th step is based on the definition of $L$-smoothness on $f$.

    Substituting $\|w^\prime_{k-1}\|^2$'s upper bound into above inequality \ref{thpf:ood_converge_gain_ub_longer_horizon_gradient_pre} yields:
    \begin{equation*}
        \begin{aligned}
            & F^\prime(x_k + s_k) - F^\prime(x_{k-1} + s_{k-1}) \\
            \leq &-\frac{\|\nabla f^\prime(x_{k-1} +s_{k-1}) + g^\prime_{k}\|^2}{2L} \\
            & + \Big( Ln^2C_1^2 \| \nabla f^\prime(x_{k-1} +s_{k-1}) + g^\prime_{k} \|^2  +  Ln^2C_2^2  \\
            & \quad + L n^4 C_3^2 C_4^2 \big(L^2(\|s_{k-1}\|^2 + \|s_{k-2}\|^2) + \|g^\prime_{k-1}-g_{k-1}\|^2 + \|g^\prime_{k-2}-g_{k-2}\|^2\big) \\
            & \qquad \times\big(L^2\|x_{k-2}+s_{k-2}-x_{k-1}-s_{k-1}\|^2 + \|g^\prime_{k-2}-g^\prime_{k-1}\|^2 \big) \Big) \| s^\prime_{k-1} \|^2,
        \end{aligned}
    \end{equation*}
    where $g^\prime_{k} \in \partial r^\prime(x_{k} +s_{k})$, $g_{k} \in \partial r(x_{k})$, $g^\prime_{k-1} \in \partial r^\prime(x_{k-1} +s_{k-1})$, and $g^\prime_{k-2} \in \partial r^\prime(x_{k-2} +s_{k-2})$.

    Moreover, the definition of $s^\prime_{k-1}$ with variable method in \eqref{thpf:longer_horizon_s_prime_gradient} yields:
    \begin{equation*}
        \begin{aligned}
            &\| s^\prime_{k-1} \|^2 \\
           = & \|\nabla f^\prime(x_{k-1} + s_{k-1}) - \nabla f(x_{k-1})\|^2 + \|g^\prime_{k} - g_{k}\|^2 \\
           & + \|-\diag(\mathbf{J}_4 w^\prime) \big(\nabla f^\prime(x_{k-1}+s_{k-1})-\nabla f^\prime(x_{k-2}+s_{k-2}) + g^\prime_{k-1}-g^\prime_{k-2}\big) \|^2 , \\
        %    \leq & \|\nabla f^\prime(x_{k-1} + s_{k-1}) - \nabla f(x_{k-1})\|^2 + \|g^\prime_{k} - g_{k}\|^2 + \|\nabla f^\prime(x_{k-1}+s_{k-1})-\nabla f(x_{k-1}) + g^\prime_{k-1}-g_{k-1}\|^2 \\
        %    & + \|-\diag(\mathbf{J}_4 w^\prime) \big(\nabla f^\prime(x_{k-1}+s_{k-1})-\nabla f^\prime(x_{k-2}+s_{k-2}) + g^\prime_{k-1}-g^\prime_{k-2}\big)\|^2 \\
           \leq & \|\nabla f^\prime(x_{k-1} + s_{k-1}) - \nabla f(x_{k-1})\|^2 + \|g^\prime_{k} - g_{k}\|^2  + \|\mathbf{J}_4 w^\prime\|^2 \|\nabla f^\prime(x_{k-1}+s_{k-1})-\nabla f^\prime(x_{k-2}+s_{k-2}) + g^\prime_{k-1}-g^\prime_{k-2}\|^2 , \\
           \leq & \|\nabla f^\prime(x_{k-1} + s_{k-1}) - \nabla f(x_{k-1})\|^2 + \|g^\prime_{k} - g_{k}\|^2  \\
           & + \|\mathbf{J}_4 w^\prime\|^2 \|\nabla f^\prime(x_{k-1}+s_{k-1})-\nabla f^\prime(x_{k-2}+s_{k-2})\|^2 + \|\mathbf{J}_4 w^\prime\|^2 \|g^\prime_{k-1}-g^\prime_{k-2}\|^2 , \\
           = & \|\nabla f^\prime(x_{k-1} + s_{k-1}) - \nabla f(x_{k-1})\|^2 + \|g^\prime_{k} - g_{k}\|^2  \\
           & + \|\mathbf{J}_4 w^\prime\|^2 \big(\|\nabla f^\prime(x_{k-1}+s_{k-1})-\nabla f^\prime(x_{k-2}+s_{k-2})\|^2 +  \|g^\prime_{k-1}-g^\prime_{k-2}\|^2\big) , \\
           \leq & \|\nabla f^\prime(x_{k-1} + s_{k-1}) - \nabla f(x_{k-1})\|^2 + \|g^\prime_{k} - g_{k}\|^2 \\
           & + \|\mathbf{J}_4 w^\prime\|^2 \big(L^2 \|x_{k-1}+s_{k-1}-x_{k-2}-s_{k-2}\|^2 + \|g^\prime_{k-1}-g^\prime_{k-2}\|^2\big) , \\
           \leq & \|\nabla f^\prime(x_{k-1} + s_{k-1}) - \nabla f(x_{k-1})\|^2 + \|g^\prime_{k} - g_{k}\|^2 \\
           & + n^2C_4^2 \|w^\prime\|^2 \big(L^2 \|x_{k-1}+s_{k-1}-x_{k-2}-s_{k-2}\|^2 + \|g^\prime_{k-1}-g^\prime_{k-2}\|^2 \big) , \\
           \leq & \|\nabla f^\prime(x_{k-1} + s_{k-1}) - \nabla f(x_{k-1})\|^2 + \|g^\prime_{k} - g_{k}\|^2 \\
           & + n^2C_4^2  \big(L^2(\|s_{k-1}\|^2 + \|s_{k-2}\|^2) + \|g^\prime_{k-1}-g_{k-1}\|^2 + \|g^\prime_{k-2}-g_{k-2}\|^2\big) \\
           & \qquad \quad \times \big(L^2\|x_{k-1}+s_{k-1}-x_{k-2}-s_{k-2}\|^2 + \|g^\prime_{k-1}-g^\prime_{k-2}\|^2\big),
        \end{aligned}
    \end{equation*}
    where $g^\prime_{k} \in \partial r^\prime(x_{k}+s_{k})$ and $g_{k} \in \partial r(x_{k})$. The second and third steps are based on Cauchy-Schwarz inequality and Triangle inequality. The 5th step is based on the definition of $L$-smoothness on $f^\prime$.

    Substituting the above formulation into the above inequality yields:
    \begin{equation*}
        \begin{aligned}
            & F^\prime(x_k + s_k) - F^\prime(x_{k-1} + s_{k-1}) \\
            \leq &-\frac{\|\nabla f^\prime(x_{k-1} +s_{k-1}) + g^\prime_{k}\|^2}{2L} \\
            & + \Big( Ln^2C_1^2 \| \nabla f^\prime(x_{k-1} +s_{k-1}) + g^\prime_{k} \|^2  +  Ln^2C_2^2 \\
            & \qquad  + L n^4C_3^2 C_4^2 \big(L^2(\|s_{k-1}\|^2 + \|s_{k-2}\|^2) + \|g^\prime_{k-1}-g_{k-1}\|^2 + \|g^\prime_{k-2}-g_{k-2}\|^2\big) \\
            & \qquad \qquad \times\big(L^2\|x_{k-2}+s_{k-2}-x_{k-1}-s_{k-1}\|^2 + \|g^\prime_{k-2}-g^\prime_{k-1}\|^2 \big) \Big) \\
            & \quad \times \Big( \|\nabla f^\prime(x_{k-1} + s_{k-1}) - \nabla f(x_{k-1})\|^2 + \|g^\prime_{k} - g_{k}\|^2 \\
            & \qquad + n^2C_4^2  \big(L^2(\|s_{k-1}\|^2 + \|s_{k-2}\|^2) + \|g^\prime_{k-1}-g_{k-1}\|^2 + \|g^\prime_{k-2}-g_{k-2}\|^2\big) \\
            & \qquad \qquad  \times \big(L^2\|x_{k-1}+s_{k-1}-x_{k-2}-s_{k-2}\|^2 + \|g^\prime_{k-1}-g^\prime_{k-2}\|^2\big) \Big),
        \end{aligned}
    \end{equation*}
    where $g^\prime_{k} \in \partial r^\prime(x_{k} +s_{k})$, $g^\prime_{k-1} \in \partial r^\prime(x_{k-1} +s_{k-1})$, and $g^\prime_{k-2} \in \partial r^\prime(x_{k-2} +s_{k-2})$.
    
    % \qy{NOTE: Variable method:}
    % \begin{equation*}
    %     \begin{aligned}
    %         & F^\prime(x_k + s_k) - F^\prime(x_{k-1} + s_{k-1}) \\
    %         \leq &-\frac{\|\nabla f^\prime(x_{k-1} +s_{k-1}) + g^\prime_{k}\|^2}{2L}  \\
    %         & + \Big(L C_1^2   \| \nabla f^\prime(x_{k-1} +s_{k-1}) + g^\prime_{k} \|^2 +L C_2^2 +L C_3^2 C_4^2 (\|s_{k-1}\|^2 + \|s_{k-2}\|^2) \|x_{k-1}-x_{k-2} + s_{k-1}-s_{k-2}\|^2 \Big) \\
    %         & \quad \times \Big(2\|s_{k-1}\|^2 + \|\nabla f^\prime(x_{k-1} + s_{k-1}) - \nabla f(x_{k-1})\|^2 + \|g^\prime_{k} - g_{k}\|^2 + C_4^2 (\|s_{k-1}\|^2 + \|s_{k-2}\|^2) \|x_{k-1}-x_{k-2} + s_{k-1}-s_{k-2}\|^2\Big),
    %     \end{aligned}
    % \end{equation*}
\end{proof}

\subsubsection*{Comparison between Variable Method and Gradient Method} \label{sec:OOD_variable_gradient_cmp}

As introduced in corollaries \ref{lm:ood_converge_gain_ub_longer_horizon_variable} and \ref{lm:ood_converge_gain_ub_longer_horizon_gradient}, we have demonstrated the per iteration convergence gain of the variable and the gradient methods, respectively. 
We are ready to compare the variable and gradient methods regarding such bounds. We categorically derive the analysis with and without the non-smooth part in the objective. We focus on the case without non-smooth parts based on the assumption that the non-smooth function in the objective is trivially solvable. Such an assumption is achievable in real-world scenarios. For example, in the blurring task for computer vision \cite{Beck2009fast}, the non-smooth function is $L_1$-norm and serves as a regularization term for the smooth objective.

\paragraph{Without Subgradient Case}
In this case, we remove all subgradient in historical modeling, which yields:
\begin{equation*}
    g^\prime_{k-1} :=0,  g^\prime_{k-2} := 0, g_{k-1} :=0,  g_{k-2} := 0.
\end{equation*}

Thus, Corollary~\ref{lm:ood_converge_gain_ub_longer_horizon_gradient} is simplified into:
\begin{equation*}
    \begin{aligned}
        & F^\prime(x_k + s_k) - F^\prime(x_{k-1} + s_{k-1}) \\
        \leq &-\frac{\|\nabla f^\prime(x_{k-1} +s_{k-1}) + g^\prime_{k}\|^2}{2L} \\
        & + \Big( Ln^2C_1^2 \| \nabla f^\prime(x_{k-1} +s_{k-1}) + g^\prime_{k} \|^2  +  Ln^2C_2^2  \\
        & \qquad+ L n^4C_3^2 C_4^{2,g} \big(L^2(\|s_{k-1}\|^2 + \|s_{k-2}\|^2) \big) \big(L^2\|x_{k-2}+s_{k-2}-x_{k-1}-s_{k-1}\|^2  \big) \Big) \\
        & \quad \times \Big( \|\nabla f^\prime(x_{k-1} + s_{k-1}) - \nabla f(x_{k-1})\|^2 + \|g^\prime_{k} - g_{k}\|^2  \\
        & \qquad \quad + n^2C_4^{2,g}  \big(L^2(\|s_{k-1}\|^2 + \|s_{k-2}\|^2) \big)  \big(L^2\|x_{k-1}+s_{k-1}-x_{k-2}-s_{k-2}\|^2 \big) \Big),
    \end{aligned}
\end{equation*}
where $g^\prime_{k} \in \partial r^\prime(x_{k} +s_{k})$ and we use the superscript $^g$ to represent gradient method's $C_4$.

If we further assume $f^\prime(x_{k-1} + s_{k-1}) := f(x_{k-1} + s_{k-1} + t), t \in \mathbb{R}^n$, which means the OOD on objective is a shifting on variable, we can further get following upper bound for the gradient method's per iteration convergence gain:
\begin{equation} \label{thpf:longer_horizon_converge_gain_ub_gradient_nosubgrad}
    \begin{aligned}
        & F^\prime(x_k + s_k) - F^\prime(x_{k-1} + s_{k-1}) \\
        \leq &-\frac{\|\nabla f^\prime(x_{k-1} +s_{k-1}) + g^\prime_{k}\|^2}{2L} \\
        & + \Big( Ln^2C_1^2 \| \nabla f^\prime(x_{k-1} +s_{k-1}) + g^\prime_{k} \|^2  +  n^2C_2^2  \\
        & \qquad + L n^4 C_3^2 C_4^{2, g} \big(L^2(\|s_{k-1}\|^2 + \|s_{k-2}\|^2) \big) \big(L^2\|x_{k-2}+s_{k-2}-x_{k-1}-s_{k-1}\|^2  \big) \Big) \\
        & \quad \times \Big( \|\nabla f(x_{k-1} + s_{k-1} + t) - \nabla f(x_{k-1})\|^2 + \|g^\prime_{k} - g_{k}\|^2  \\
        & \qquad \quad + n^2C_4^{2, g}  \big(L^2(\|s_{k-1}\|^2 + \|s_{k-2}\|^2) \big) \big(L^2\|x_{k-1}+s_{k-1}-x_{k-2}-s_{k-2}\|^2 \big) \Big) , \\
        \leq &-\frac{\|\nabla f^\prime(x_{k-1} +s_{k-1}) + g^\prime_{k}\|^2}{2L} \\
        & + \Big( Ln^2C_1^2 \| \nabla f^\prime(x_{k-1} +s_{k-1}) + g^\prime_{k} \|^2  +  Ln^2C_2^2  \\
        & \qquad + L n^4C_3^2 C_4^{2, g} \big(L^2(\|s_{k-1}\|^2 + \|s_{k-2}\|^2) \big) \big(L^2\|x_{k-2}+s_{k-2}-x_{k-1}-s_{k-1}\|^2  \big) \Big) \\
        & \quad \times \Big( L^2 \|s_{k-1} + t\|^2 + \|g^\prime_{k} - g_{k}\|^2  + n^2C_4^{2, g}  \big(L^2(\|s_{k-1}\|^2 + \|s_{k-2}\|^2) \big) \big(L^2\|x_{k-1}+s_{k-1}-x_{k-2}-s_{k-2}\|^2 \big) \Big) , \\
        = &-\frac{\|\nabla f^\prime(x_{k-1} +s_{k-1}) + g^\prime_{k}\|^2}{2L} \\
        & + \Big( Ln^2C_1^2 \| \nabla f^\prime(x_{k-1} +s_{k-1}) + g^\prime_{k} \|^2  +  Ln^2C_2^2  \\
        & \qquad + L n^4C_3^2 C_4^{2, g} L^4 (\|s_{k-1}\|^2 + \|s_{k-2}\|^2) (\|x_{k-2}+s_{k-2}-x_{k-1}-s_{k-1}\|^2  ) \Big) \\
        & \quad \times \Big( L^2 \|s_{k-1} + t\|^2 + \|g^\prime_{k} - g_{k}\|^2  + n^2C_4^{2, g} L^4 (\|s_{k-1}\|^2 + \|s_{k-2}\|^2) (\|x_{k-1}+s_{k-1}-x_{k-2}-s_{k-2}\|^2 ) \Big).
    \end{aligned}
\end{equation}

Similarly, we get the bound of the variable method by:
\begin{equation}\label{thpf:longer_horizon_converge_gain_ub_variable_nosubgrad}
    \begin{aligned}
        & F^\prime(x_k + s_k) - F^\prime(x_{k-1} + s_{k-1}) \\
        \leq &-\frac{\|\nabla f^\prime(x_{k-1} +s_{k-1}) + g^\prime_{k}\|^2}{2L}  \\
        & + \Big(L n^2C_1^2   \| \nabla f^\prime(x_{k-1} +s_{k-1}) + g^\prime_{k} \|^2 +L n^2C_2^2 +L n^4 C_3^2 C_4^{2, v} (\|s_{k-1}\|^2 + \|s_{k-2}\|^2) \|x_{k-1}-x_{k-2} + s_{k-1}-s_{k-2}\|^2 \Big) \\
        & \quad \times \Big(\|s_{k-1}\|^2 + L^2 \|s_{k-1} + t\|^2  + \|g^\prime_{k} - g_{k}\|^2 + n^2C_4^{2, v} (\|s_{k-1}\|^2 + \|s_{k-2}\|^2) \|x_{k-1}-x_{k-2} + s_{k-1}-s_{k-2}\|^2\Big),
    \end{aligned}
\end{equation}
where we use the superscript $^v$ to represent variable method's $C_4$.

If $L \leq 1$, which means the objective is smooth, the upper bound yielded by the gradient method in inequality \ref{thpf:longer_horizon_converge_gain_ub_gradient_nosubgrad} is intrinsically smaller than that in inequality \ref{thpf:longer_horizon_converge_gain_ub_variable_nosubgrad}, which means that gradient-based longer horizon modeling methods are better for functions with small gradients. We note that $L \leq 1$ is general in real-world scenarios. For example, $L \leq 1$ in logistic regression tasks are inherently achieved by average among features.

Otherwise, if $L > 1$, to keep an identical boundness in inequalities \ref{thpf:longer_horizon_converge_gain_ub_gradient_nosubgrad} and \ref{thpf:longer_horizon_converge_gain_ub_variable_nosubgrad}, we can also achieve a lower convergence bound by shrinking the output of operator $\mathbf{N}_4$ in the  gradient method, such as setting $C_4^{g} = C_4^{v} / (L^2)$ in \ref{thpf:longer_horizon_converge_gain_ub_gradient_nosubgrad}, which yields:
\begin{equation}
    \begin{aligned}
        & F^\prime(x_k + s_k) - F^\prime(x_{k-1} + s_{k-1}) \\
        \leq &-\frac{\|\nabla f^\prime(x_{k-1} +s_{k-1}) + g^\prime_{k}\|^2}{2L} \\
        & + \Big( Ln^2C_1^2 \| \nabla f^\prime(x_{k-1} +s_{k-1}) + g^\prime_{k} \|^2  +  Ln^2C_2^2  + L n^4C_3^2 C_4^{2,v} (\|s_{k-1}\|^2 + \|s_{k-2}\|^2) (\|x_{k-2}+s_{k-2}-x_{k-1}-s_{k-1}\|^2  ) \Big) \\
        & \quad \times \Big( L^2 \|s_{k-1} + t\|^2 + \|g^\prime_{k} - g_{k}\|^2  + n^2C_4^{2,v} (\|s_{k-1}\|^2 + \|s_{k-2}\|^2) (\|x_{k-1}+s_{k-1}-x_{k-2}-s_{k-2}\|^2 ) \Big).
    \end{aligned}
\end{equation}

Moreover, the remaining difference between such two upper bounds are $L^2 \|s_{k-1} + t\|^2$ in the gradient method and $\|s_{k-1}\|^2 + L^2 \|s_{k-1} + t\|^2$ in the variable method. Hence, the gradient method yields a smaller convergence gain upper bound.

% Thus, $t:=0$ yields the following inequality to keep better robustness for the gradient method:
% \begin{equation*}
%     2L^2 \|s_{k-1}\|^2 \leq 2\|s_{k-1}\|^2 + L^2 \|s_{k-1}\|^2.
% \end{equation*}
% Given $s_{k-1} \neq 0$, solve the inequality, we get $L \in [1, \sqrt{2}]$.

To sum up, we have the following conclusions:
\begin{enumerate}
    \item[1)] $L \in [0, 1]$. The gradient-based longer horizon modeling method is more robust in OOD scenarios.
    \item[2)] $L \in (1, \infty]$. By setting $C_4^{g} \leq C_4^{v} / (L^2)$, the gradient-based longer horizon modeling method is more robust in OOD scenarios.
    % \item[3)] Otherwise, the variable-based longer horizon modeling method is robust in OOD scenarios.
\end{enumerate}

\paragraph{With Subgradient Case}
We eliminate this case since we assume $r(x)$ is a proper function that can be trivially solved.

% They are bounded by:
% \begin{equation} \label{thpf:longer_horizon_ood_subgradient_bound}
%     \begin{aligned}
%         \|g^\prime_{k-1}\| &\leq \max(\|\partial r(x_{k-1}+s_{k-1})_{\text{lb}}\|, \|\partial r(x_{k-1}+s_{k-1})_{\text{ub}}\|),  \\
%         \|g^\prime_{k-2}\| &\leq \max(\|\partial r(x_{k-2}+s_{k-2})_{\text{lb}}\|, \|\partial r(x_{k-2}+s_{k-2})_{\text{ub}}\|).
%     \end{aligned}
% \end{equation}

% \qy{TODO: Discuss the different construction for variable method and gradient method to keep a similar robustness.} 

\subsection{OOD Multi-Iteration Convergence Rate}
\begin{theorem} \label{th:ood_converge_rate_longer_horizon}
    Under Assumption \ref{as:assump_ind_robust_longer_horizon}, OOD's convergence rate of $K$ iterations is upper bounded by:
    \begin{equation*} 
        \begin{aligned}
            &\min_{k=1,\dots, K}F^\prime(x_k+s_k) - F^\prime(x^* +s^*)\\
            \leq & \frac{L}{2K} \|x_{0} +s_{0} - x^*-s^*\|^2 - \frac{L}{2K}\|x_{K} +s_{K} - x^*-s^* \|^2 \\
            & + \frac{L}{K} \sum_{k=1}^{K} ( x_k +s_k -x_{k-1}-s_{k-1} +\frac{\nabla f^\prime(x_{k-1} +s_{k-1}) + g^\prime_{k}}{L})^\top (x_{k} +s_{k} - x^*-s^* ).
        \end{aligned}
    \end{equation*}
\end{theorem}

\begin{proof}
    Same as the demonstration for Theorem~\ref{th:ood_converge_rate_composite}.
\end{proof}

\begin{corollary}
    Under Assumption \ref{as:assump_ind_robust_longer_horizon}, L2O model $d$'s (\eqref{thpf:l2o_longer_horizon_p}) convergence rate is upper bounded by w.r.t. $\| s_{k-1}^\prime \|$ by:
    \begin{equation*}
        \begin{aligned}
            &\min_{k=1,\dots, K}F^\prime(x_k+s_k) - F^\prime(x^* +s^*)\\
            \leq & \frac{L}{2K} \|x_{0} +s_{0} - x^*-s^*\|^2 - \frac{L}{2K}\|x_{K} +s_{K} - x^*-s^* \|^2 - \frac{1}{K} \sum_{k=1}^{K} (\nabla f^\prime(x_{k-1} +s_{k-1}))^\top (x_{k} +s_{k} - x^*-s^* ) \\
            &+ \frac{L}{K} \sum_{k=1}^{K} \big(( \sqrt{n}C_1  \| \nabla f^\prime(x_{k-1}+s_{k-1})\| + \sqrt{n}C_2 )  \|s_{k-1}^\prime\| + \sqrt{n}C_3\| P^\prime_{k-1} \|\big) \| x_k+ s_k -x^* -s^* \|.
        \end{aligned}
    \end{equation*}
\end{corollary}

\begin{proof}
    First, we rewrite the convergence rate upper bound as the following inequalities:
    \begin{equation*}
        \begin{aligned}
            &\min_{k=1,\dots, K}F^\prime(x_k+s_k) - F^\prime(x^* +s^*)\\
            \leq & \frac{L}{2K} \|x_{0} +s_{0} - x^*-s^*\|^2 - \frac{L}{2K}\|x_{K} +s_{K} - x^*-s^* \|^2 + \frac{L}{K} \sum_{k=1}^{K} ( x_k +s_k -x_{k-1}-s_{k-1})^\top (x_{k} +s_{k} - x^*-s^* )\\
            = & \frac{L}{2K} \|x_{0} +s_{0} - x^*-s^*\|^2 - \frac{L}{2K}\|x_{K} +s_{K} - x^*-s^* \|^2 - \frac{1}{K} \sum_{k=1}^{K} (\nabla f^\prime(x_{k-1} +s_{k-1}))^\top (x_{k} +s_{k} - x^*-s^* ) \\
            & + \frac{L}{K} \sum_{k=1}^{K} ( -\diag(\mathbf{J}_1 s^\prime_{k-1})(\nabla f^\prime(x_{k-1} +s_{k-1}) + g^\prime_{k}) - \mathbf{J}_2 s^\prime_{k-1} - \mathbf{J}_3 P^\prime_{k-1})^\top (x_{k} +s_{k} - x^*-s^* ).
        \end{aligned}
    \end{equation*}

    Next, we derive its upper bound w.r.t. $\| s_{k-1}^\prime \|$. Cauchy-Schwarz inequality and Triangle inequality yield:
    \begin{equation*}
        \begin{aligned}
            &\min_{k=1,\dots, K}F^\prime(x_k+s_k) - F^\prime(x^* +s^*)\\
            \leq & \frac{L}{2K} \|x_{0} +s_{0} - x^*-s^*\|^2 - \frac{L}{2K}\|x_{K} +s_{K} - x^*-s^* \|^2 - \frac{1}{K} \sum_{k=1}^{K} (\nabla f^\prime(x_{k-1} +s_{k-1}))^\top (x_{k} +s_{k} - x^*-s^* ) \\
            & + \frac{L}{K} \sum_{k=1}^{K} ( -\diag(\mathbf{J}_1 s^\prime_{k-1})(\nabla f^\prime(x_{k-1} +s_{k-1}) + g^\prime_{k}) - \mathbf{J}_2 s^\prime_{k-1} - \mathbf{J}_3 P^\prime_{k-1})^\top (x_{k} +s_{k} - x^*-s^* ) , \\
            \leq & \frac{L}{2K} \|x_{0} +s_{0} - x^*-s^*\|^2 - \frac{L}{2K}\|x_{K} +s_{K} - x^*-s^* \|^2 - \frac{1}{K} \sum_{k=1}^{K} (\nabla f^\prime(x_{k-1} +s_{k-1}))^\top (x_{k} +s_{k} - x^*-s^* ) \\
            & + \frac{L}{K} \sum_{k=1}^{K} ( \|\diag(\mathbf{J}_1 s^\prime_{k-1})(\nabla f^\prime(x_{k-1} +s_{k-1}) + g^\prime_{k})\| + \|\mathbf{J}_2 s^\prime_{k-1}\| + \|\mathbf{J}_3 P^\prime_{k-1}\|) \| x_{k} +s_{k} - x^*-s^* \| , \\
            \leq & \frac{L}{2K} \|x_{0} +s_{0} - x^*-s^*\|^2 - \frac{L}{2K}\|x_{K} +s_{K} - x^*-s^* \|^2 - \frac{1}{K} \sum_{k=1}^{K} (\nabla f^\prime(x_{k-1} +s_{k-1}))^\top (x_{k} +s_{k} - x^*-s^* ) \\
            &+ \frac{L}{K} \sum_{k=1}^{K} ( \sqrt{n}C_1 \|s_{k-1}^\prime\| \| \nabla f^\prime(x_{k-1}+s_{k-1})\| + \sqrt{n}C_2\| s_{k-1}^\prime \| + \sqrt{n}C_3\| P^\prime_{k-1} \|) \| x_k+ s_k -x^* -s^* \| , \\
            = & \frac{L}{2K} \|x_{0} +s_{0} - x^*-s^*\|^2 - \frac{L}{2K}\|x_{K} +s_{K} - x^*-s^* \|^2 - \frac{1}{K} \sum_{k=1}^{K} (\nabla f^\prime(x_{k-1} +s_{k-1}))^\top (x_{k} +s_{k} - x^*-s^* ) \\
            &+ \frac{L}{K} \sum_{k=1}^{K} \left(( \sqrt{n}C_1  \| \nabla f^\prime(x_{k-1}+s_{k-1})\| + \sqrt{n}C_2 )  \|s_{k-1}^\prime\| + \sqrt{n}C_3\| P^\prime_{k-1} \|\right) \| x_k+ s_k -x^* -s^* \|.
        \end{aligned}
    \end{equation*}
\end{proof}

\section{Details of Experiments} \label{sec:evaltion_detail}

\subsection{Implementation Details}
Our implementation is conducted with PyTorch based on the open-source code provided by the official implementation of \cite{Liu2023} in \textit{https://github.com/xhchrn/MS4L2O}. 
We follow the settings in \cite{Liu2023} to implement our GO-Math-L2O model. We construct a coordinate-wise model where our model takes gradient features according to a variable as an input and generates the update for that coordinate independently on all coordinates. 

We implement the learnable parameter matrices $\mathbf{R}$, $\mathbf{Q}$, and $\mathbf{B}$ in Theorem~\ref{lm:go_math_l2o_model} as diagonal matrices. We use neural networks to generate vectors with an identical shape to variable $x$ and use them to conduct the diagonal entries of $\mathbf{R}$, $\mathbf{Q}$, and $\mathbf{B}$. 
We use different models for $\mathbf{R}$, $\mathbf{Q}$, and $\mathbf{B}$, respectively, both of which take the same inner feature of the former layer. 
Following L2O-PA in \cite{Liu2023}, we add an inner linear model between the linear models and the LSTM cell. The complete forward data flow is: the LSTM cell $\to$ the inner linear model $\to$ linear models of $\mathbf{R}$, $\mathbf{Q}$, and $\mathbf{B}$.

For LSTM's input features in our GO-Math-L2O, we set it to be smooth gradient $\nabla f(x)$ and two boundaries of non-smooth gradient set, denoted as $\partial r(x)_{\text{lb}}$ and $\partial r(x)_{\text{ub}}$. Thus, the input feature is the concatenation of such three features, $[\nabla f(x)^\top, \partial r(x)_{\text{lb}}^\top, \partial r(x)_{\text{ub}}^\top]^\top$. Moreover, for each block, we normalize input features with the vector norm of the initial point's input feature, i.e., $\|\nabla f(x_0)\|$, $\|\partial r(x_0)_{\text{lb}}\|$, and $\|\partial r(x_0)_{\text{ub}}\|$. At $k$-th iteration, the input feature is $\left[\frac{\nabla f(x_k)}{\|\nabla f(x_0)\|}^\top,  \frac{\partial r(x_k)_{\text{lb}}}{\|\partial r(x_0)_{\text{lb}}\|}^\top, \frac{\partial r(x_k)_{\text{ub}}}{\partial r(x_0)_{\text{ub}}}^\top\right]^\top$. 

For L2O-PA \cite{Liu2023}, the input feature is the concatenation of variable and gradient vectors $[x^\top, \nabla f(x)^\top]^\top$. 

For other baseline methods introduced in \cref{sec:eval}, we use the implementations provided in the official implementation of \cite{Liu2023}.

We randomly sample the initial points for all methods. We use deterministic seeds in samplings to ensure reproducibility. Thus, even for non-learning methods, our experimental results are different from those in \cite{Liu2023}. However, compared with the origin point set in \cite{Liu2023}, the random initial point setting is better for robustness evaluation.

% Notably, we improve the L2O-PA \cite{Liu2023} by adding two such two boundaries to the input of their LSTM. Thus, the only difference between our GO-Math-L2O and baselines L2O-PA is that L2O-PA has an extra feature of variable $x$. 

\subsection{Output Activation}
As in Theorem~\ref{lm:go_math_l2o_model}, at each iteration, we achieve a symmetric positive definite $\mathbf{R}_k$ by Sigmoid function. The output of the Sigmoid function is in $(0, 1)$. Thus, Frobenius-norm of $\mathbf{R}_k$ is bounded by $\sqrt{n}$, where $n$ is the dimension of variable $x$. In \cite{Liu2023}, a larger range is achieved by a direct multiplication with a given constant. 
We set the constants for $\mathbf{R}_k$, $\mathbf{Q}_k$, and $\mathbf{B}_k$ to be $2$, $2$, and $1$, respectively. Thus, $\|\mathbf{R}_k\| \leq 2\sqrt{n}$,  $\|\mathbf{Q}_k\| \leq 2\sqrt{n}$, and $\|\mathbf{B}_k\| \leq \sqrt{n}$.

We follow the activation function setting for LASSO and Logistic regressions in \cite{Liu2023}: Sigmoid for LASSO regression and Softplus for Logistic regression. The Sigmoid function is doubled to get $(0,2)$ range outputs \cite{Liu2023}. The activation function is applied on all parameters in Theorem~\ref{lm:go_math_l2o_model}.

Following \cite{Liu2023}, we utilize the objective's smoothness scalar $L$ to shrink the parameters multiplied before gradient, i.e., parameter matrix $\mathbf{R}$ (and $\mathbf{Q}$ for the first two variants in the following section). We set $L$ as the maximal eigenvalue of the Hessian matrix on optimization problem. For \textit{LASSO Regerssion}, $L$ is given by:
\begin{equation*}
    \|\mathbf{A}\|_2,
\end{equation*}
where $\mathbf{A}$ is the given parameter matrix in objectives defined in \cref{sec:eval}.

For \textit{Logistic Regerssion}, $L$ is given by:
\begin{equation*}
    \left\|\frac{1}{m}\sum_{i=1}^m a_i a_i^T h\left(a_i^\top x\right)\left(1-h\left(a_i^\top x\right)\right)\right\|_2,
\end{equation*}
where $h$ is the sigmoid function and each $a_i$ is a given parameter farture vector defined in \cref{sec:eval}. Moreover, since $h\left(a_i^\top x\right)\left(1-h\left(a_i^\top x\right)\right) \leq 1$, we construct the following upper bound of the above formulation to get a $x$-unrelated $L$:
\begin{equation*}
    \left\|\frac{1}{m}\sum_{i=1}^m a_i a_i^T \right\|_2.
\end{equation*}

\subsection{Evaluation Metric} 
Following \cite{Liu2023}, we use a classical algorithm, named FISTA \cite{Beck2009fast}, to generate optimal solutions for both \textit{LASSO Regression} and \textit{Logistic Regression} problems. Based on \cite{Liu2023}, we run FISTA for 5,000 iterations to ensure the accuracy. Then, all the evaluation solutions are normalized with the optimal solutions by the following equation:
\begin{equation*}
    \frac{F(x)-F(x^*)}{F(x^*)}.
\end{equation*}

\subsection{Gradient Map Ablation} \label{sec:eval_grad_map}
We construct the following three gradient map implementations and select the one with best InD performance. 
At $k$-th iteration, the fist one is the standard gradient map (denoted as \textbf{STD}) as follows:
\begin{equation*}
    G_{k-1} = \mathbf{R}_{k}^{-1} (x_{k-1} - x_{k} - \mathbf{Q}_k v_{k-1} - b_{1,k}),
\end{equation*}
where $G_{k-1}$ is equivalent to $\nabla f(x_{k-1}) + g_{k-1}$.

Then, we eliminate the minus term for historical information $v_{k-1}$ to let $G_{k-1}$ cover the historical information (denoted as \textbf{LH}):
\begin{equation*}
    G_{k-1} = \mathbf{R}_{k}^{-1} (x_{k-1} - x_{k} - b_{1,k}).
\end{equation*}

Moreover, we eliminate $\mathbf{R}_{k}$ inversion to improve numerical stability (denoted as \textbf{LHNoR}):
\begin{equation*}
    G_{k-1} = x_{k-1} - x_{k} - b_{1,k}.
\end{equation*}
It is worth noting that such an implementation differs from Math-L2O in \cite{Liu2023}, where we follow a classical momentum scheme by applying momentum posteriorly. However, Math-L2O use the Nesterov momentum method by adding momentum to the approximation point before the gradient calculation.

\begin{figure}
    \centering
    \includegraphics[width=0.6\linewidth]{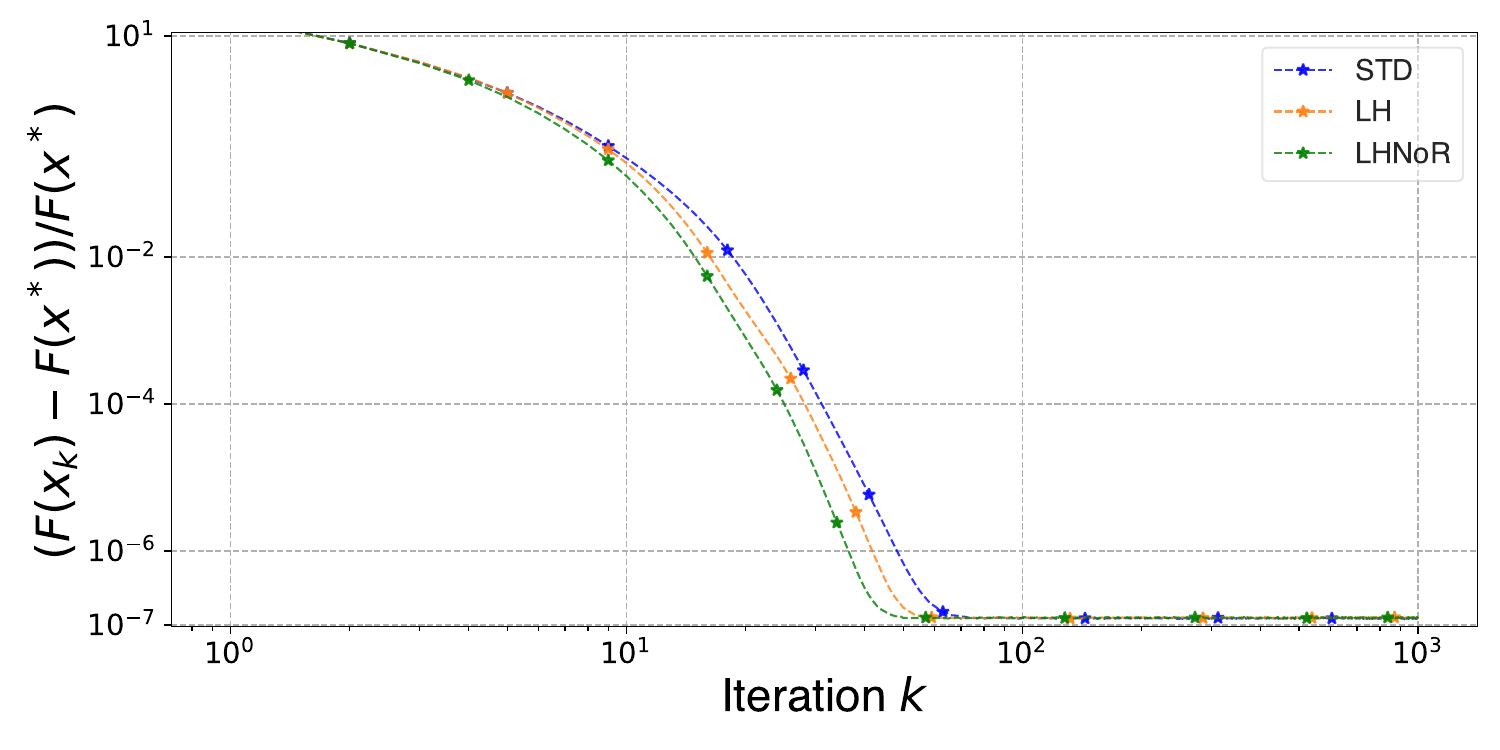}
    \vspace{-4mm}
    \caption{LASSO Regression: Ablation Study on Gradient Map Configurations.}
    \label{fig:gradmap_ablation_lasso_ind}
\end{figure}

The InD results are shown in Figure~\ref{fig:gradmap_ablation_lasso_ind}, where \textbf{LHNoR} outperforms the other two methods. We use the \textbf{LHNoR} version in the following experiments.

\subsection{Training Configurations} \label{sec:train_config}
We use Adam \cite{Kinga2015} as the optimizer to train our model and learning-based baselines. We set the weight decay to zero for all models. For our proposed model,  we clamp the norm of the gradient vector to one. However, we do not apply this setting to other baselines since Math-L2O \cite{Liu2023} fails to converge at all OOD scenarios. 
Following \cite{Liu2023}, we evaluate the in-training model with the evaluation set every 20 iterations.

We evaluate the InD performance of LASSO regression with different training settings in Table~\ref{table:train_config} to choose the best setting. The ``BP Frequency'' represents the iterations utilized to achieve one backpropagation and total iterations. For example, 20/100 means backpropagating every 20 iterations in 100 iterations. For training epochs and learning rates, we consider two candidate settings. One epoch and $0.01$ is the setting in \cite{Liu2023}. We conduct another one with three epochs and a decayed learning rate starting from $0.01$. Since our proposed model has more parameters than Math-L2O in \cite{Liu2023}, we test two different mini-batch settings, 64 and 128, where the 128 mini-batch case has a double training sample that the  64 mini-batch case. Furthermore, we consider alleviating the imbalance problem in a sequence of objective values and design a weighted-sum loss function by the indices of iterations. For a BP length $T$, given a objective value sequence $x_1, x_2, \dots, x_T$, the weighted-sum loss is given by:
\begin{equation*}
    \sum_{i=1}^{T} \frac{i}{\sum_{i=1}^{T} i} F(x_i).
\end{equation*}
In contrast, the mean loss used in \cite{Liu2023} is given by:
\begin{equation*}
    \sum_{i=1}^{T} \frac{1}{T} F(x_i).
\end{equation*}

\begin{table}[t]
	\centering
	\caption{Training Settings}
	\vspace{-2mm}
	\begin{tabular}{cccccc}
		\hline
		\textbf{Index} &\textbf{BP Frequency} &\textbf{Epochs} &\textbf{Learning Rate} &\textbf{Batch Size} &\textbf{Loss Function} \\ \hline
		1 &20/100 &1 &$0.01$ &64 &Mean\\ 
		2 &20/100 &1 &$0.01$ &64 &Weighted-Sum\\ 
		3 &20/100 &1 &$0.01$ &128 &Mean\\ 
		4 &20/100 &1 &$0.01$ &128 &Weighted-Sum\\ 
		5 &20/100 &3 &$0.01$, Decay to 10\% Per-Epoch &64 &Mean\\ 
		6 &20/100 &3 &$0.01$, Decay to 10\% Per-Epoch &64 &Weighted-Sum\\ 
		7 &20/100 &3 &$0.01$, Decay to 10\% Per-Epoch &128 &Mean\\ 
		8 &20/100 &3 &$0.01$, Decay to 10\% Per-Epoch &128 &Weighted-Sum\\ \hline
        9 &50/100 &1 &$0.01$ &64 &Mean\\ 
		10 &50/100 &1 &$0.01$ &64 &Weighted-Sum\\ 
		11 &50/100 &1 &$0.01$ &128 &Mean\\ 
		12 &50/100 &1 &$0.01$ &128 &Weighted-Sum\\ 
		13 &50/100 &3 &$0.01$, Decay to 10\% Per-Epoch &64 &Mean\\ 
		14 &50/100 &3 &$0.01$, Decay to 10\% Per-Epoch &64 &Weighted-Sum\\ 
		15 &50/100 &3 &$0.01$, Decay to 10\% Per-Epoch &128 &Mean\\ 
		16 &50/100 &3 &$0.01$, Decay to 10\% Per-Epoch &128 &Weighted-Sum\\ \hline
        17 &100/100 &1 &$0.01$ &64 &Mean\\
        18 &100/100 &1 &$0.01$ &64 &Weighted-Sum\\ 
		19 &100/100 &1 &$0.01$ &128 &Mean\\ 
		20 &100/100 &1 &$0.01$ &128 &Weighted-Sum\\ 
		21 &100/100 &3 &$0.01$, Decay to 10\% Per-Epoch &64 &Mean\\ 
		22 &100/100 &3 &$0.01$, Decay to 10\% Per-Epoch &64 &Weighted-Sum\\ 
		23 &100/100 &3 &$0.01$, Decay to 10\% Per-Epoch &128 &Mean\\ 
		24 &100/100 &3 &$0.01$, Decay to 10\% Per-Epoch &128 &Weighted-Sum\\ \hline
	\end{tabular}
	\label{table:train_config}
\end{table}

% \begin{table}[t]
% 	\centering
% 	\caption{Candidates of Training Settings}
% 	\vspace{-2mm}
% 	\begin{tabular}{cccccc}
% 		\hline
% 		\textbf{Index} &\textbf{Batch Size} &\textbf{BP Frequency} &\textbf{Epochs} &\textbf{Learning Rate}  &\textbf{Loss Function} \\ \hline
% 		1 &64 &10/100 &1 &$0.01$  &Mean\\ 
% 		2 &64 &10/100 &1 &$0.01$  &Weighted-Sum\\ 
% 		3 &64 &10/100 &3 &$0.01$, Decay to 10\% Per-Epoch  &Mean\\ 
% 		4 &64 &10/100 &3 &$0.01$, Decay to 10\% Per-Epoch  &Weighted-Sum\\ \hline
%         5 &128 &10/100 &1 &$0.01$  &Mean\\ 
% 		6 &128 &10/100 &1 &$0.01$  &Weighted-Sum\\ 
% 		7 &128 &10/100 &3 &$0.01$, Decay to 10\% Per-Epoch  &Mean\\ 
% 		8 &128 &10/100 &3 &$0.01$, Decay to 10\% Per-Epoch  &Weighted-Sum\\ \hline
%         9 &256 &10/100 &1 &$0.01$  &Mean\\ 
% 		10 &256 &10/100 &1 &$0.01$  &Weighted-Sum\\ 
% 		11 &256 &10/100 &3 &$0.01$, Decay to 10\% Per-Epoch  &Mean\\ 
% 		12 &256 &10/100 &3 &$0.01$, Decay to 10\% Per-Epoch  &Weighted-Sum\\ \hline
% 	\end{tabular}
% 	\label{table:train_config}
% \end{table}

\begin{figure}
    \centering
    \includegraphics[width=0.6\linewidth]{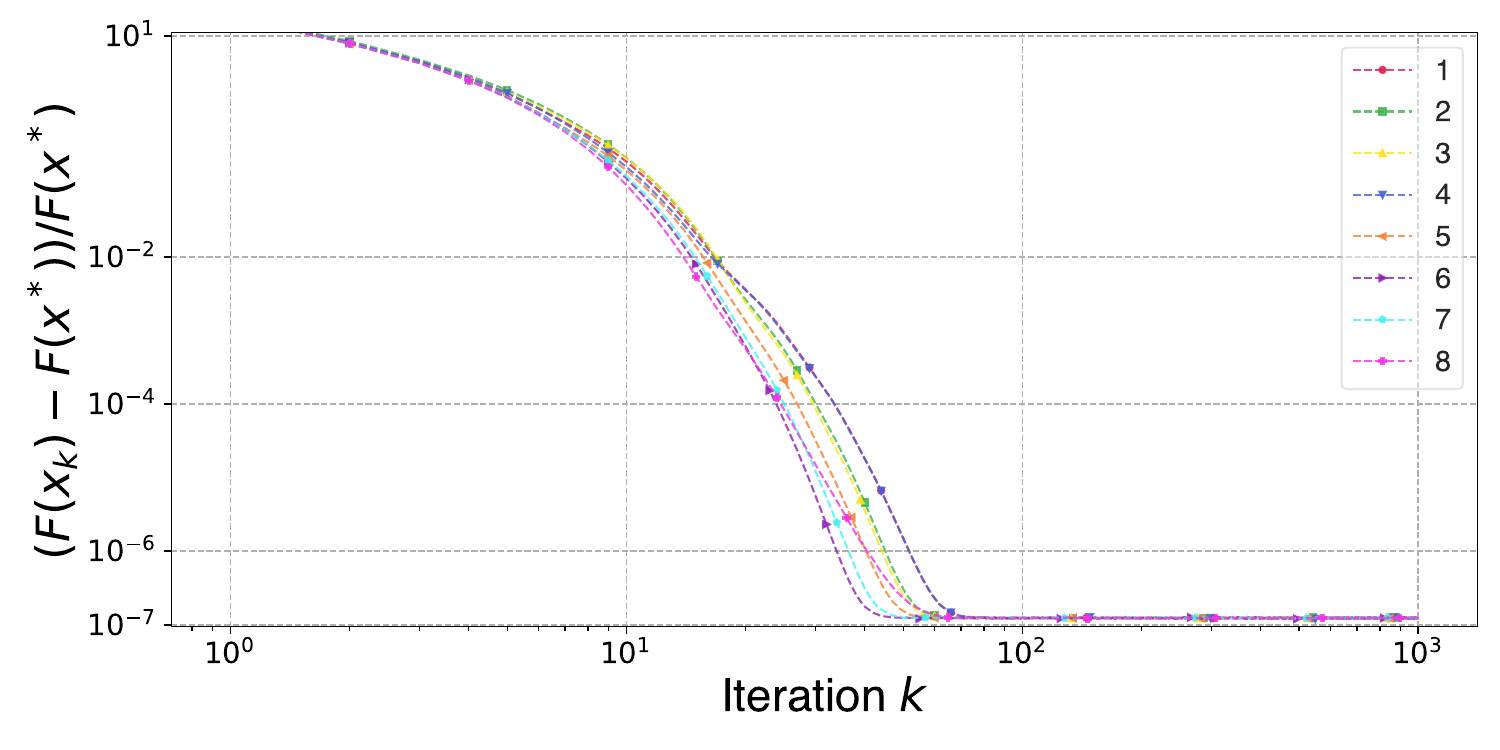}
    \vspace{-4mm}
    \caption{LASSO Regression: Ablation Study on Training Settings, 20/100 BP Frequency.}
    \label{fig:training_config_20100}
\end{figure}

% TODO: To be updated, following three figures.
The results of settings 1 to 8 are shown in Figure~\ref{fig:training_config_20100}. The experimental results demonstrate that the best settings for ``20/100'' BP frequency are settings 7 and 8.

\begin{figure}
    \centering
    \includegraphics[width=0.6\linewidth]{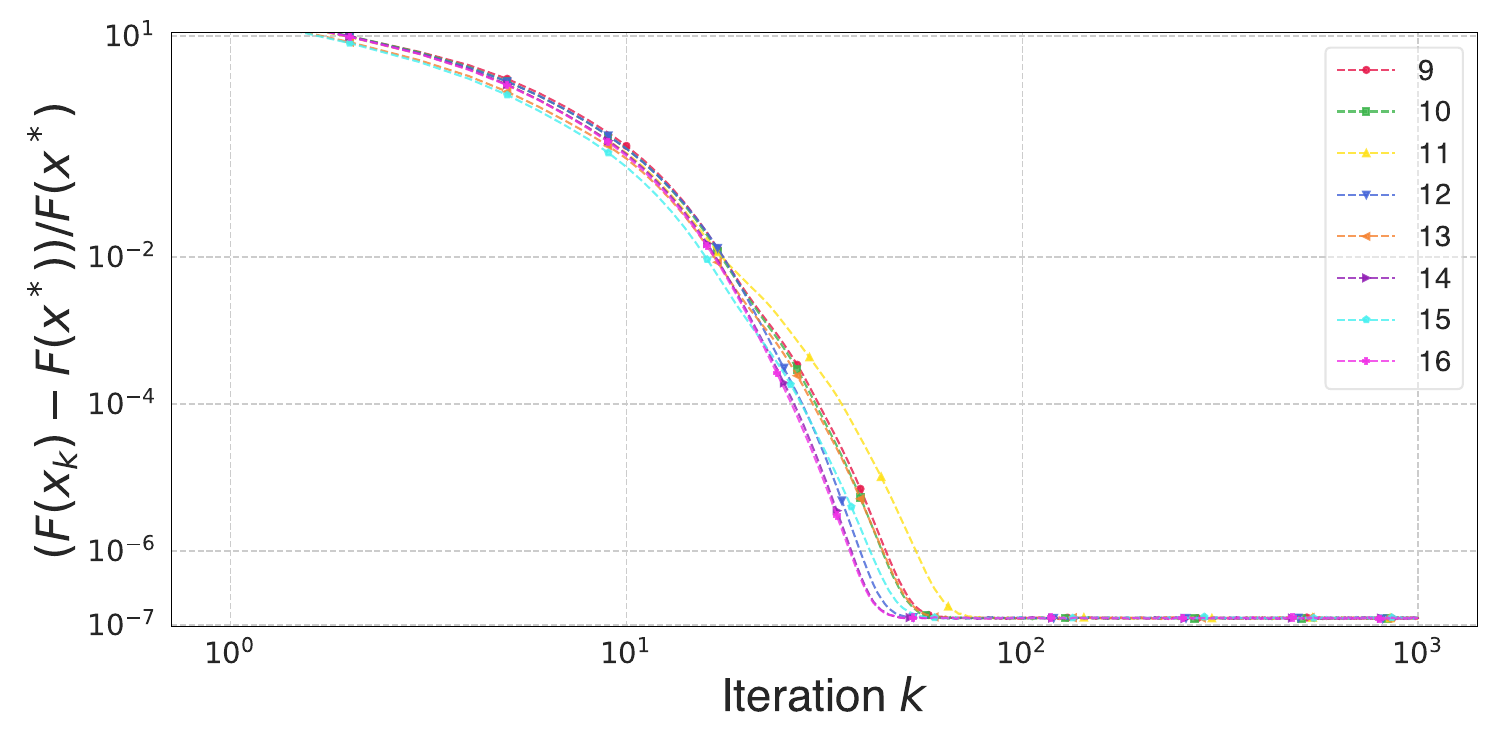}
    \vspace{-4mm}
    \caption{LASSO Regression: Ablation Study on Training Settings, 50/100 BP Frequency.}
    \label{fig:training_config_50100}
\end{figure}

The results of settings 9 to 16 are shown in Figure~\ref{fig:training_config_50100}. The experimental results demonstrate that the best settings for ``50/100'' BP frequency are settings 14 and 16.

\begin{figure}
    \centering
    \includegraphics[width=0.6\linewidth]{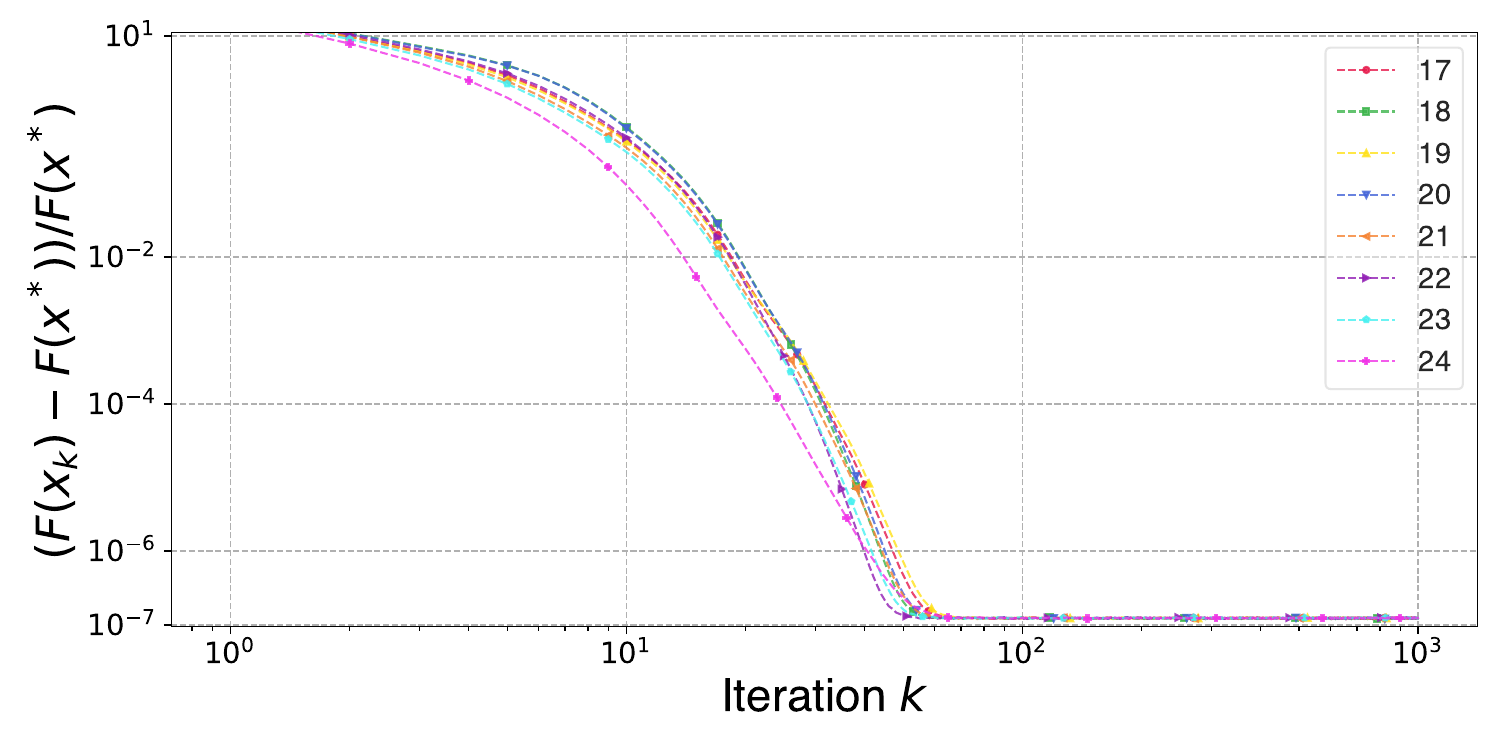}
    \vspace{-4mm}
    \caption{LASSO Regression: Ablation Study on Training Settings, 100/100 BP Frequency.}
    \label{fig:training_config_100100}
\end{figure}

The results of settings 17 to 24 are shown in Figure~\ref{fig:training_config_100100}. The experimental results demonstrate that the best setting for ``100/100'' BP frequency is setting 24.

\begin{figure}
    \centering
    \includegraphics[width=0.6\linewidth]{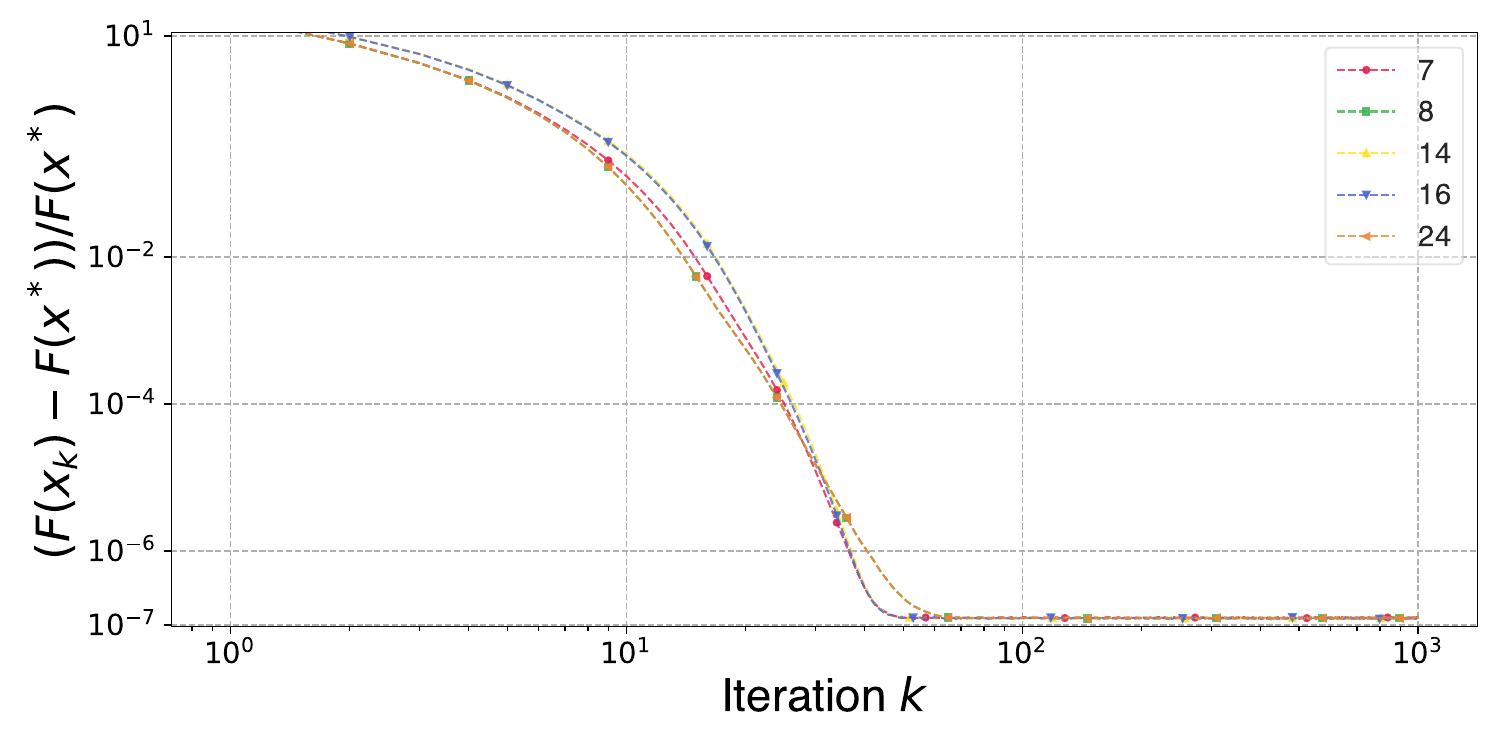}
    \vspace{-4mm}
    \caption{LASSO Regression: Ablation Study on Training Settings, Best.}
    \label{fig:training_config_best}
\end{figure}

A further comparison between settings 7, 8, 14, 16, and 24 is illustrated in Figure~\ref{fig:training_config_best}. Based on the result, we conclude that training settings do not dominate the InD performance of our proposed Go-Math-L2O model. We choose setting 7 as our training configuration since we observe that the baseline Math-L2O method fails to converge at all OOD scenarios if we increase the BP frequency to ``50/100'' or ``100/100''.

\subsection{OOD Improvement Configurations} \label{sec:eval_l_shifting}
Based on our analysis in \cref{sec:OOD_analysis}, by shrinking the range of the parameters in optimizing steep objectives, our proposed model can achieve better robustness than variable-based historical modeling in SOTA Math-L2O \cite{Liu2023}. This section evaluates the performance with different extents of parameter shrinking. Specifically, we set the shrinkings on $\mathbf{Q}$ by dividing the smoothness parameter $L$ of smooth objective. Different settings are listed in Table~\ref{table:Q_shrinking}.
\begin{table}[t]
	\centering
	\caption{$\mathbf{Q}$-Shrinking Settings}
	\vspace{-2mm}
	\begin{tabular}{cc}
		\hline
		\textbf{Index} &$\mathbf{Q}$ \textbf{Settings}  \\ \hline
		1 &$\mathbf{Q}$ \\ 
		2 &$\mathbf{Q}/\sqrt{L}$  \\ 
        3 &$\mathbf{Q}/L$  \\ 
        4 &$\mathbf{Q}/L^2$  \\ \hline
	\end{tabular}
	\label{table:Q_shrinking}
\end{table}

\begin{figure}
    \centering
    \includegraphics[width=0.6\linewidth]{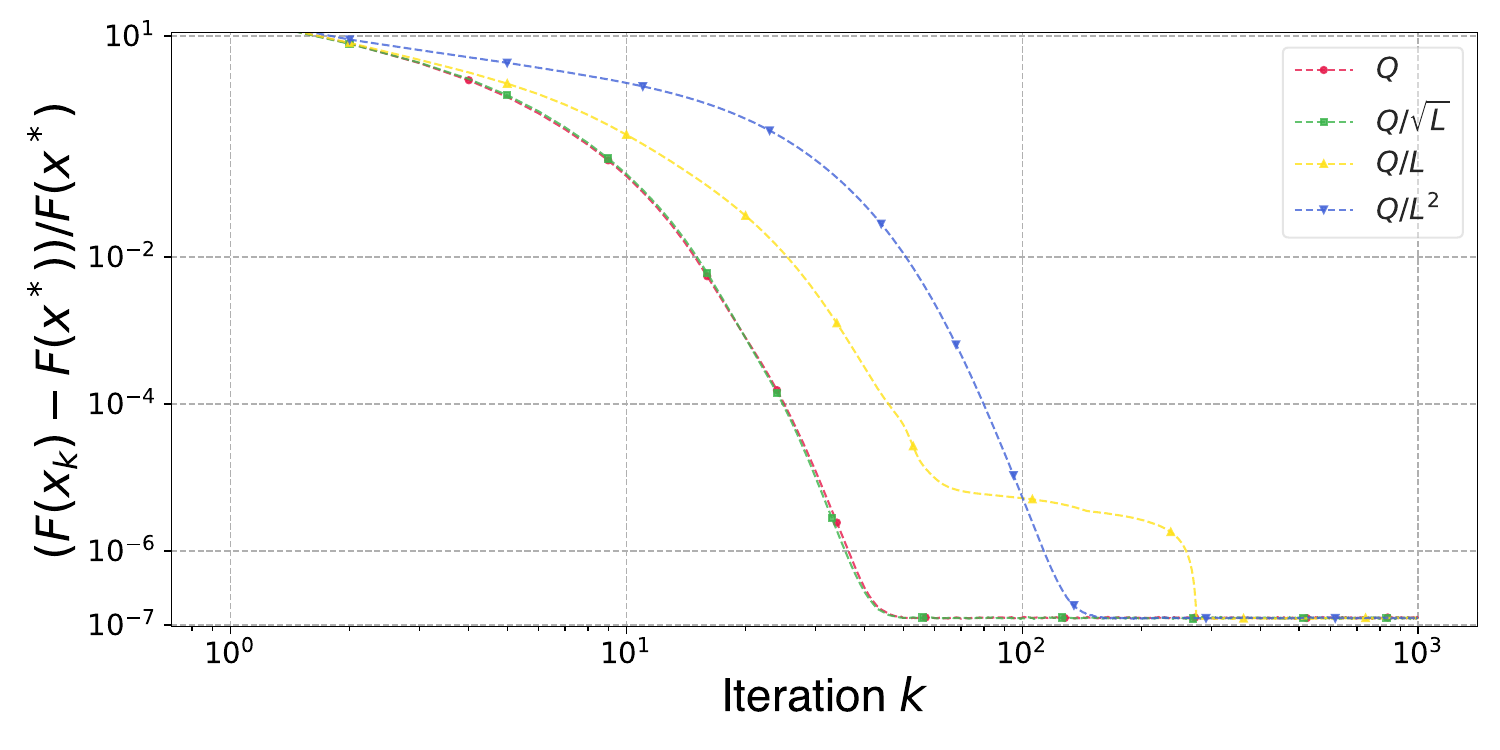}
    \vspace{-4mm}
    \caption{LASSO Regression: Ablation Study on $\mathbf{Q}$ Settings, InD scenario.}
    \label{fig:q_config_ind}
\end{figure}

Since $L$ is calculated individually for each instance by its largest eigenvalue of the Hessian matrix of the objective. Compared with the $\mathbf{Q}$ only version, adding $L$ changes $\mathbf{Q}$'s distribution. Thus, we separately train each setting within Table~\ref{table:Q_shrinking}. 
The InD results of the settings in Table~\ref{table:Q_shrinking} are shown in Figure~\ref{fig:q_config_ind}. The illustrated results show that $\mathbf{Q}/L$ and $\mathbf{Q}/L^2$ cause poor InD performance. $\mathbf{Q}/\sqrt{L}$ has a similar InD convergence to $\mathbf{Q}$.

\begin{figure}
    \centering
    \includegraphics[width=0.6\linewidth]{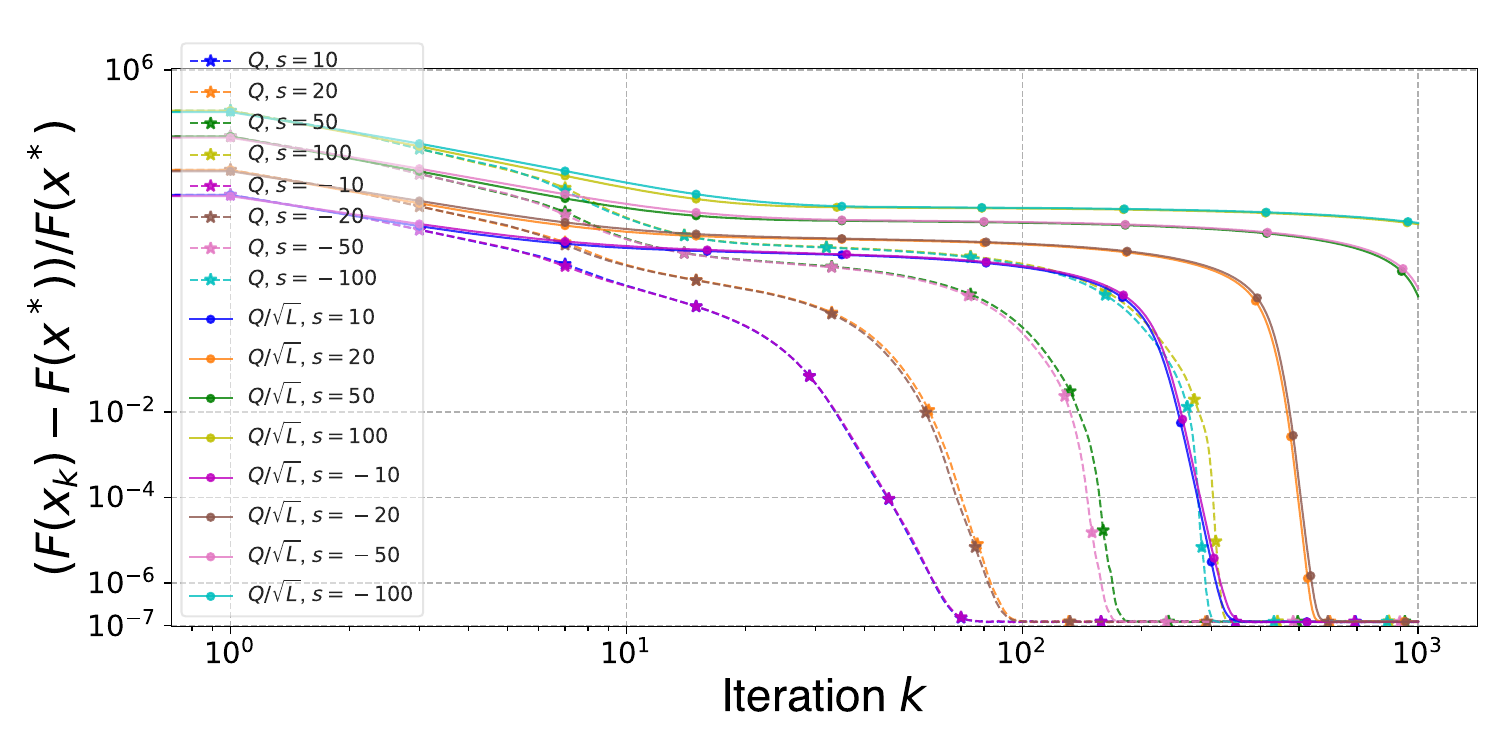}
    \vspace{-4mm}
    \caption{LASSO Regression: Ablation Study on $\mathbf{Q}$ Settings, OOD by Trigger 1.}
    \label{fig:q_config_ood_s}
\end{figure}

\begin{figure}
    \centering
    \includegraphics[width=0.6\linewidth]{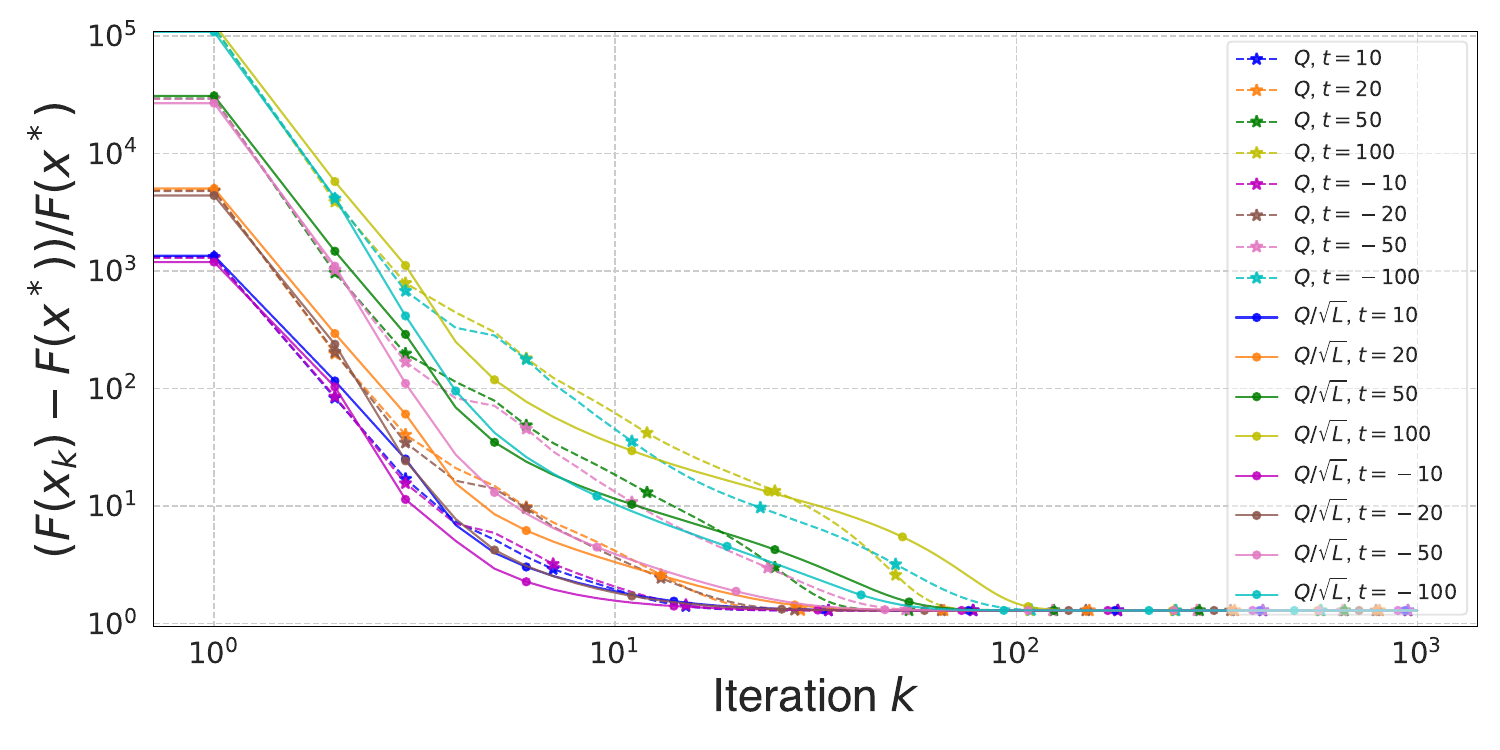}
    \vspace{-4mm}
    \caption{LASSO Regression: Ablation Study on $\mathbf{Q}$ Settings, OOD by Trigger 2.}
    \label{fig:q_config_ood_t}
\end{figure}

Furthermore, we add an extra experiment to compare their OOD performances, shown in Figures~\ref{fig:q_config_ood_s} and \ref{fig:q_config_ood_t}. The results show that the $\mathbf{Q}$ setting outperforms $\mathbf{Q}/\sqrt{L}$ in all initial point OOD scenarios (Figure~\ref{fig:q_config_ood_s}) and achieves better outperformance with larger OOD shiftings. Both methods perform similarly in objective OOD scenarios (Figure~\ref{fig:q_config_ood_t}).

It is worth noting that this result does not violate our theoretical comparison result in \cref{sec:OOD_variable_gradient_cmp}, where our gradient-only method needs further parameter shrinking strategies to address the deficiency of weaker robustness by larger magnitude in sharp objective cases. Our normalization method on input gradient-only features and our recurrent gradient map setting that eliminates $\mathbf{R}$ inversion have achieved a similar input magnitude to the variable method in \cite{Liu2023}. Moreover, in Figure~\ref{fig:q_config_ood_t}, $\mathbf{Q}/\sqrt{L}$ setting performs similarly to the $\mathbf{Q}$.

\subsection{Real-World Evaluation}
We further evaluate our model on real-world optimization problems. We follow the methodology proposed in \cite{Liu2023} to construct the following real-world datasets:
\begin{enumerate}
    \item[1)] LASSO Regression. 1,000 patches are chosen from the BSDS500 dataset. $\mathbf{A}$ are calculated with K-SVD method and $\lambda$ is set to be $0.5$.
    \item[2)] Logistic Regression. Ionoshpere dataset contains 4,601 $a_i, b_i \in \mathbb{R}^{34}$ for each sample. Spambase dataset contains 4,601 $a_i, b_i \in \mathbb{R}^{57}$ for each sample.
\end{enumerate}

\subsection{Logistic Regression Results} \label{sec:eval_logistic_results}
\begin{figure}
    \centering
    \includegraphics[width=0.6\linewidth]{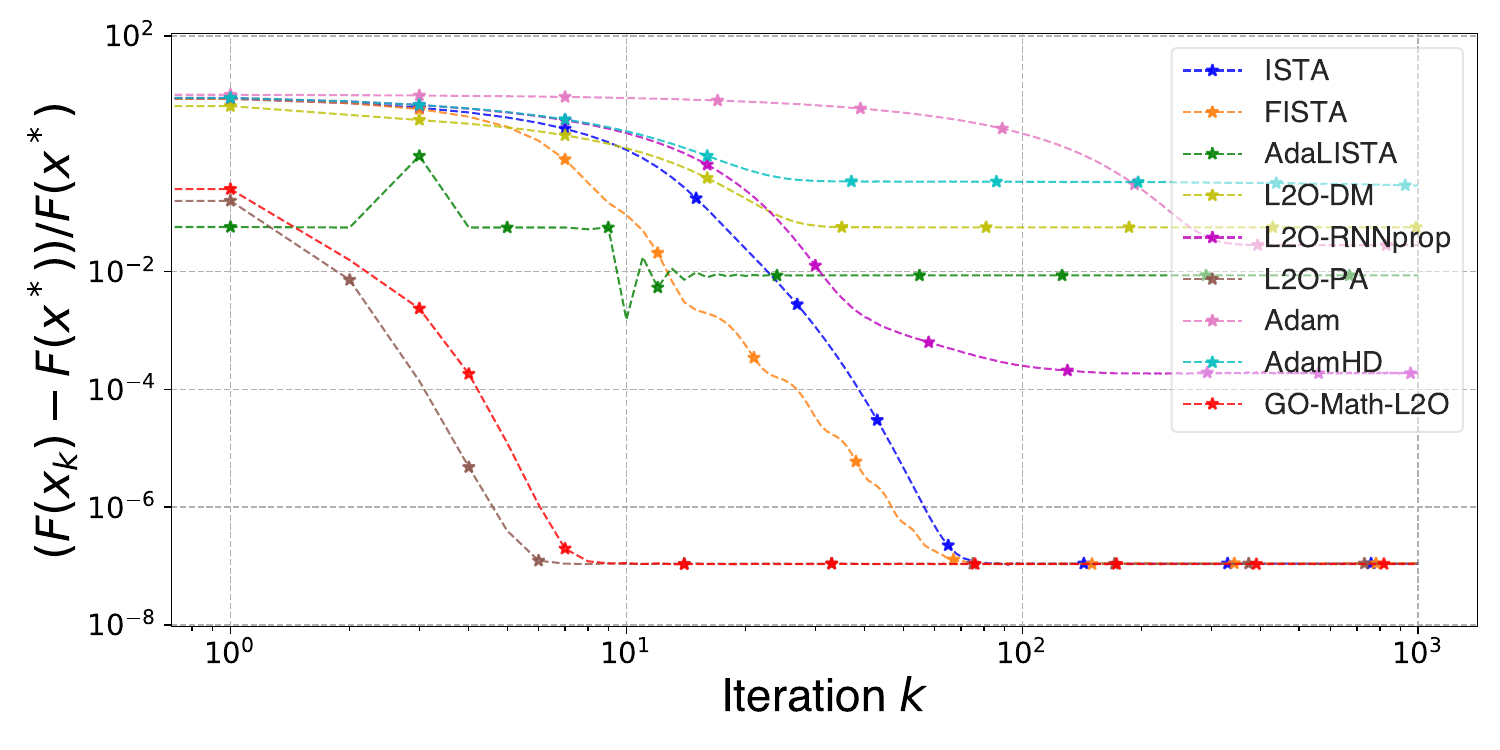}
    \vspace{-4mm}
    \caption{Logistic Regression: InD.}
    \label{fig:figure2_ind_cmp_logisticl1}
\end{figure}

InD comparison is shown in Figure~\ref{fig:figure2_ind_cmp_logisticl1}. Our proposed Go-Math-L2O performs similarly to L2O-PA and outperforms other baselines.

\begin{figure}
    \centering
    \includegraphics[width=0.6\linewidth]{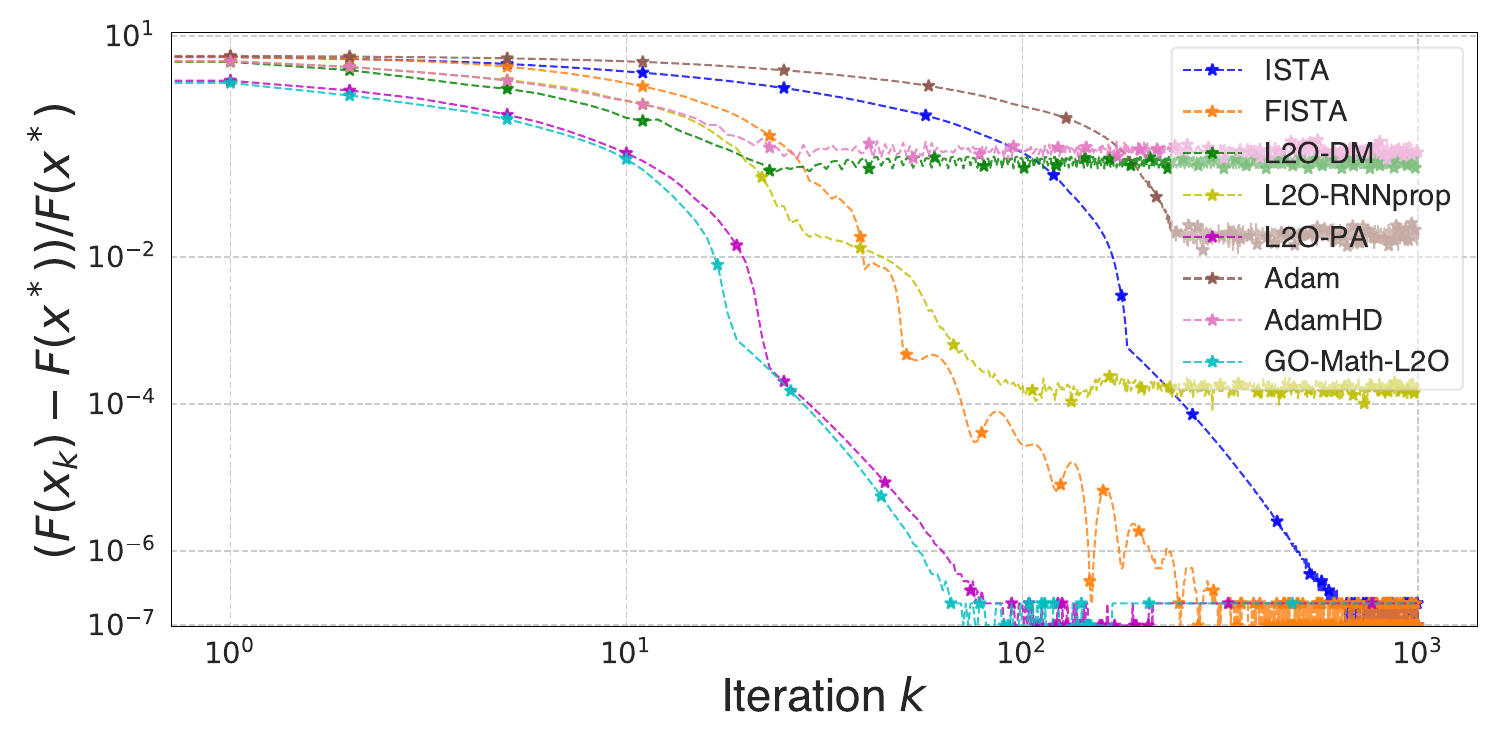}
    \vspace{-4mm}
    \caption{Logistic Regression: Real-World Ionoshpere Dataset.}
    \label{fig:logistic_ood_ionoshpere}
\end{figure}

\begin{figure}
    \centering
    \includegraphics[width=0.6\linewidth]{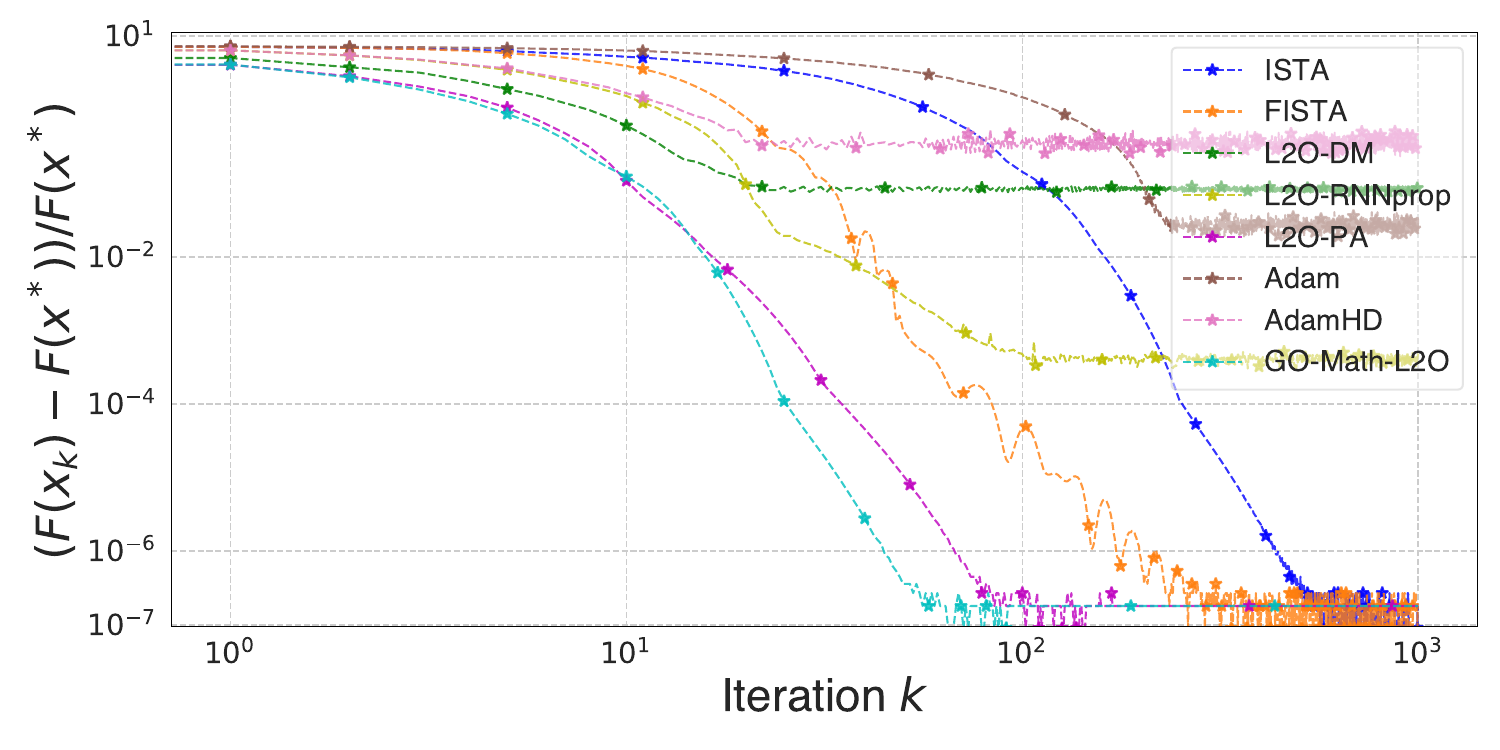}
    \vspace{-4mm}
    \caption{Logistic Regression: Real-World Spambase Dataset.}
    \label{fig:logistic_ood_spambase}
\end{figure}

The OOD comparison on two real-world datasets, Ionoshpere and Spambase, are shown in Figures~\ref{fig:logistic_ood_ionoshpere} and \ref{fig:logistic_ood_spambase}. Our GO-Math-L2O model outperforms all other baselines.

\begin{figure}
    \centering
    \includegraphics[width=0.6\linewidth]{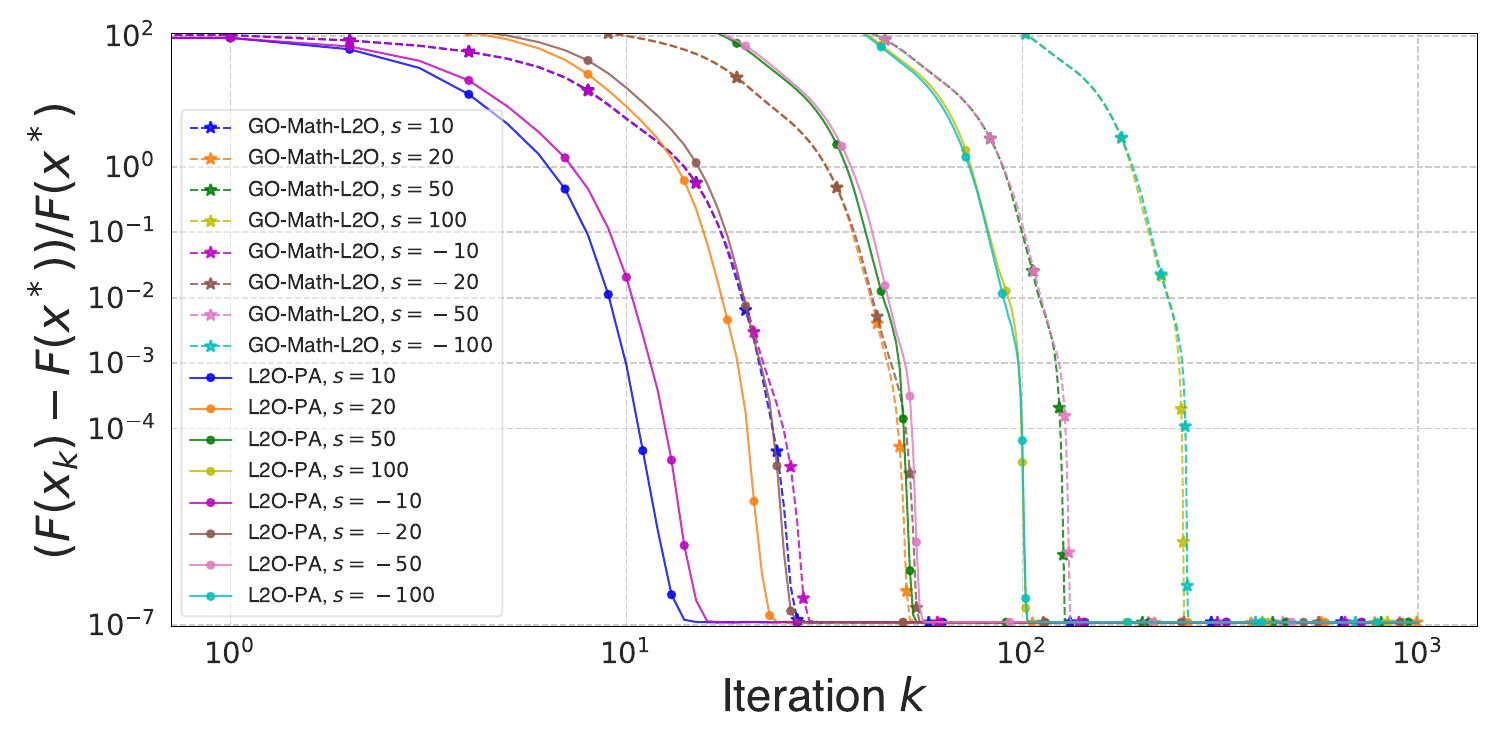}
    \vspace{-2mm}
    \caption{Logistic Regression: OOD by Trigger 1.}
    \label{fig:figure3_ood_cmp_logistic_s}
\end{figure}

Figure~\ref{fig:figure3_ood_cmp_logistic_s} depicts the OOD scenarios in Logistic regression where the initial point deviates from around zero, i.e., the OOD initial point is $x_0+s$, where $s$ denotes the extent of the initial point shifting. Under these conditions, our proposed GO-Math-L2O model performs similarly to L2O-PA \cite{Liu2023}.

% when initial point shifting with $s=\pm10, \pm20, \pm50$.
% For the large initial point shifting cases ($S=\pm100$), our model has similar convergence rates to the L2O-PA.

% The L2O-PA method \cite{Liu2023} cannot converge in all these instances. 

% Our proposed GO-Math-L2O achieves convergence comparable to InD scenarios for $s=\mp10, \mp20$. Nonetheless, substantial shifts in the initial point present challenges to our model.

\begin{figure}
    \centering
    \includegraphics[width=0.6\linewidth]{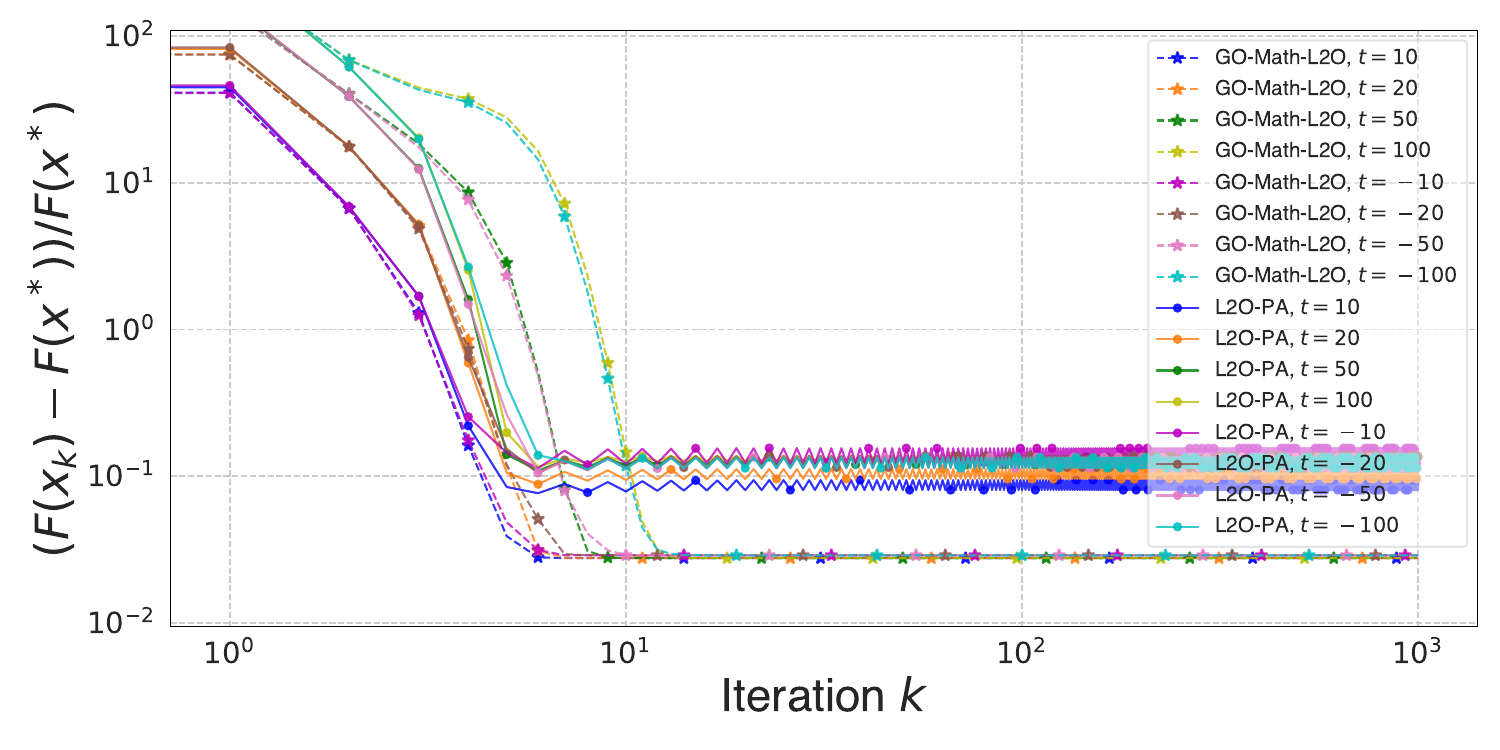}
    \vspace{-2mm}
    \caption{Logistic Regression: OOD by Trigger 2.}
    \label{fig:figure4_ood_cmp_logistic_t}
\end{figure}

Figure~\ref{fig:figure4_ood_cmp_logistic_t} presents the results for the OOD scenarios of objective shifting, i.e., the OOD objective is $F^\prime(x) = F(x+t)$, where $s$ denotes the extent of initial point shifting. Results in $t=\pm10, \pm20$ cases demonstrate that our proposed GO-Math-L2O method converges significantly faster than L2O-PA \cite{Liu2023}. For $t=\pm50, \pm100$ cases, our model can also converge to better optimums after oscillations. Moreover, the results also show that L2O-PA fails to converge when objective shifts.

% \subsection{Other Results of Logistric Regression}

% \newpage
% {
%     \small
%     \bibliographystyle{ieeenat_fullname}
%     \bibliography{appendix.bib}
% }

% % 
% Having the supplementary compiled together with the main paper means that:
% % 
% \begin{itemize}
% \item The supplementary can back-reference sections of the main paper, for example, we can refer to \cref{sec:intro};
% \item The main paper can forward reference sub-sections within the supplementary explicitly (e.g. referring to a particular experiment); 
% \item When submitted to arXiv, the supplementary will already included at the end of the paper.
% \end{itemize}
% 
% To split the supplementary pages from the main paper, you can use \href{https://support.apple.com/en-ca/guide/preview/prvw11793/mac#:~:text=Delete%20a%20page%20from%20a,or%20choose%20Edit%20%3E%20Delete).}{Preview (on macOS)}, \href{https://www.adobe.com/acrobat/how-to/delete-pages-from-pdf.html#:~:text=Choose%20%E2%80%9CTools%E2%80%9D%20%3E%20%E2%80%9COrganize,or%20pages%20from%20the%20file.}{Adobe Acrobat} (on all OSs), as well as \href{https://superuser.com/questions/517986/is-it-possible-to-delete-some-pages-of-a-pdf-document}{command line tools}.

\end{document}